\documentclass{article} 
\usepackage[final]{colm2025_conference}
\usepackage{graphicx}
\usepackage{wrapfig}
\usepackage{subcaption}
\usepackage{microtype}
\usepackage{hyperref}
\usepackage{url}
\usepackage{booktabs}
\usepackage{amsmath}
\usepackage{lineno}
\usepackage{booktabs}
\usepackage{multirow}
\usepackage{algorithm}
\usepackage{algorithmic}
\definecolor{darkblue}{rgb}{0, 0, 0.5}
\hypersetup{colorlinks=true, citecolor=darkblue, linkcolor=darkblue, urlcolor=darkblue}

\title{\textit{KVSink}: Understanding and Enhancing the Preservation of Attention Sinks in KV Cache Quantization for LLMs }
\author{Zunhai Su \\
Shenzhen International Graduate School, Tsinghua University\\
\texttt{zh-su23@mails.tsinghua.edu.cn} 
\AND
Kehong Yuan\thanks{Corresponding author: Kehong Yuan} \\
Shenzhen International Graduate School, Tsinghua University\\
\texttt{yuankh@sz.tsinghua.edu.cn} 
}
\begin{document}

\ifcolmsubmission
\linenumbers
\fi

\maketitle

\begin{abstract}
Key-Value (KV) cache quantization has become a widely adopted optimization technique for efficient large language models (LLMs) inference by reducing KV cache memory usage and mitigating memory-bound constraints.
Recent studies have emphasized the importance of preserving the original precision of KVs for the first few tokens to ensure the protection of attention sinks.
While this approach has proven effective in mitigating performance degradation, its underlying principles remain insufficiently understood.
Moreover, it fails to address the recent discovery that attention sinks can emerge beyond the initial token positions.
In this work, we elucidate the underlying mechanisms of attention sinks during inference by examining their role in the cross-layer evolution of extreme activation outliers.
Additionally, we provide a comprehensive analysis of the interplay between attention sinks and KV cache quantization. 
Based on our enhanced understanding, we introduce \textit{\textbf{KVSink}}, a plug-and-play method that effectively predicts sink tokens with negligible overhead, enabling more thorough preservation. 
Extensive experiments demonstrate that KVSink outperforms the existing Preserve-First-N (PFN) strategy, offering more effective preservation of attention sinks during KV cache quantization.
Moreover, when applied to the well-established KVQuant method, KVSink further improves perplexity (PPL) and reduces reliance on 16-bit numerical outliers.
\end{abstract}
\section{Introduction}
Transformer-based \citep{vaswani2017attention} large language models (LLMs), including GPT \citep{achiam2023gpt}, LLaMA \citep{dubey2024llama}, and DeepSeek \citep{liu2024deepseek,guo2025deepseek}, have revolutionized various domains of artificial intelligence research, including natural language processing \citep{hadi2023survey,zhao2023survey}, computer vision \citep{zhang2024vision}, and multimodal understanding \citep{liang2024survey}.
However, the impressive capabilities of LLMs come with significant challenges due to their extensive size and computational demands \citep{zhu2024survey}, along with the substantial Key-Value (KV) cache generated during inference \citep{li2024survey}, all of which hinder their deployment and practical application.
KV cache facilitates LLMs inference by avoiding recomputation of past KVs. 
However, as the batch size and context length increase, the oversized KV caches become a significant memory bottleneck \citep{liu2024kivi}.
KV cache compression has emerged as a promising direction to mitigate this challenge \citep{shi2024keep,li2024survey}, encompassing a broad array of techniques—including quantization \citep{hooper2025kvquant,su2025rotatekv}, pruning \citep{xiao2023efficient,zhang2023h2o}, fusion \citep{liu2025minicache,wan2024look}, budget-aware allocation \citep{xiao2024duoattention,cai2024pyramidkv}, and low-rank decomposition \citep{chang2024palu,saxena2024eigen}.
\begin{figure}[t!]
    \centering
    \vspace{-5mm}
    \begin{subfigure}{0.3\textwidth} 
        \centering
    \includegraphics[width=\linewidth]{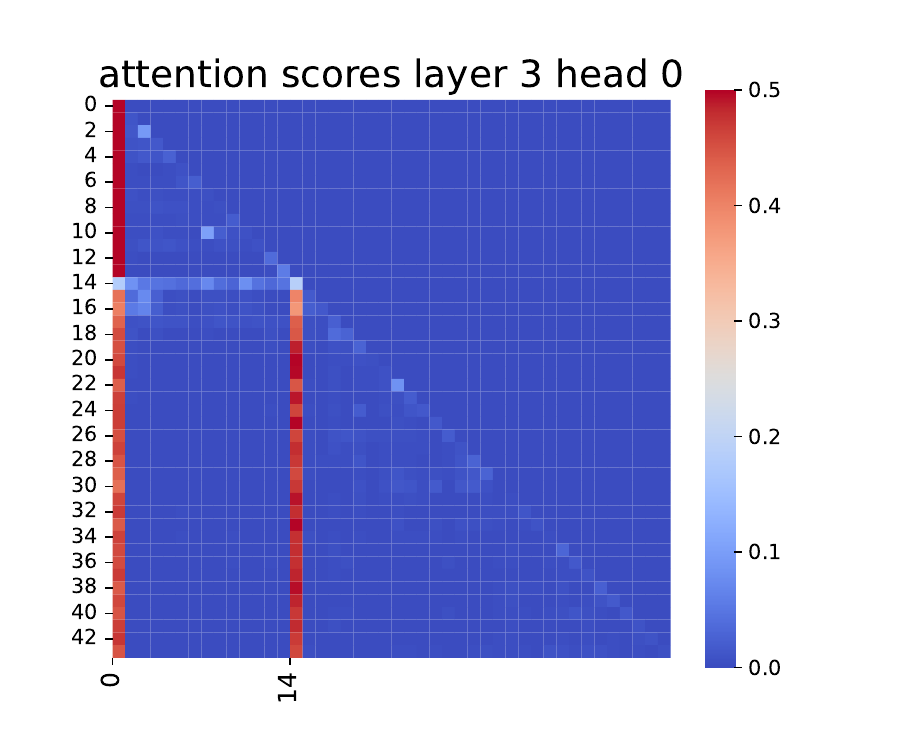}
        \vspace{-5mm}
        \caption{Attention sinks}
        \label{attn_sink}
    \end{subfigure}
    \begin{subfigure}{0.3\textwidth}
        \centering
    \includegraphics[width=\linewidth]{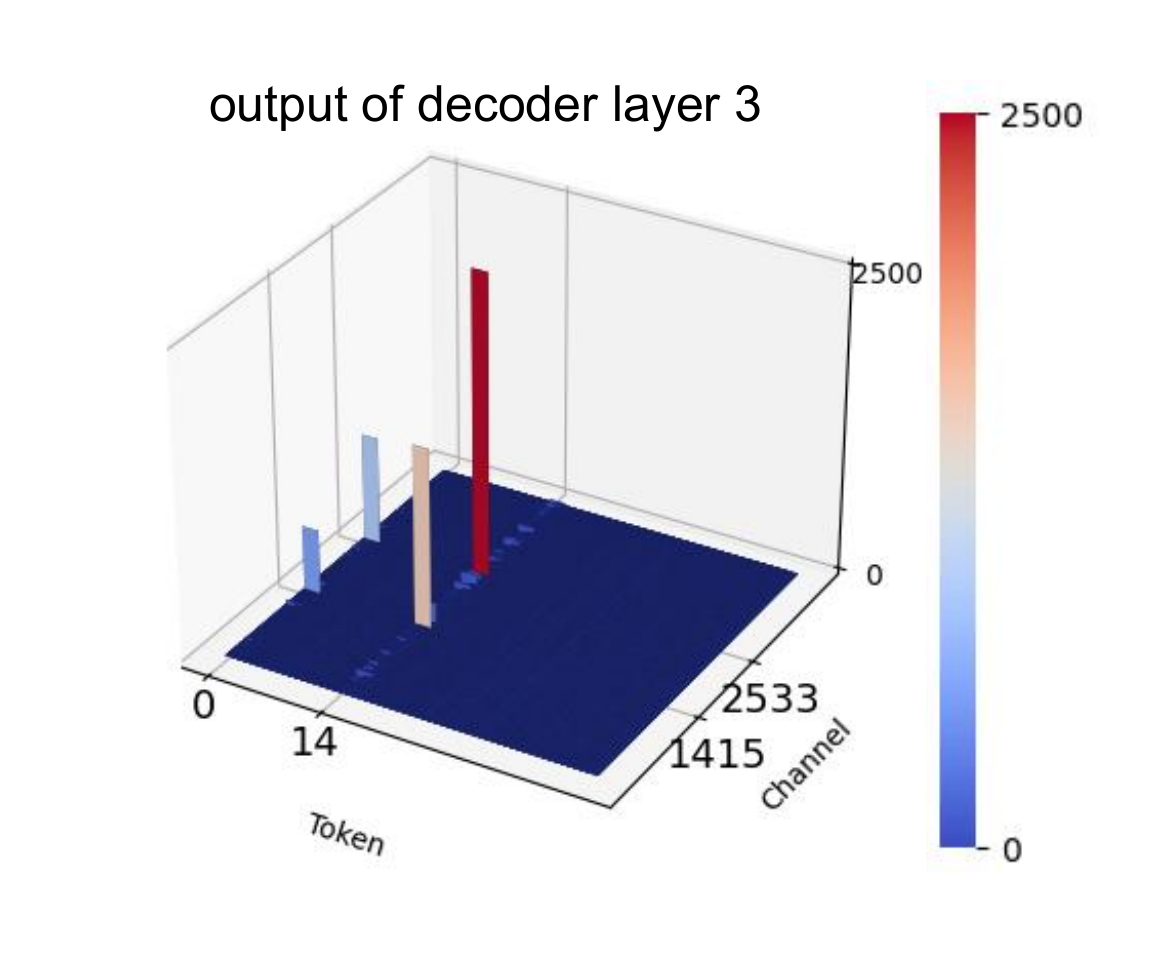}
        \vspace{-5mm}
        \caption{Activation outliers}
        \label{hidden_atates_1}
    \end{subfigure}
    \begin{subfigure}{0.3\textwidth}
        \centering
    \includegraphics[width=\linewidth]{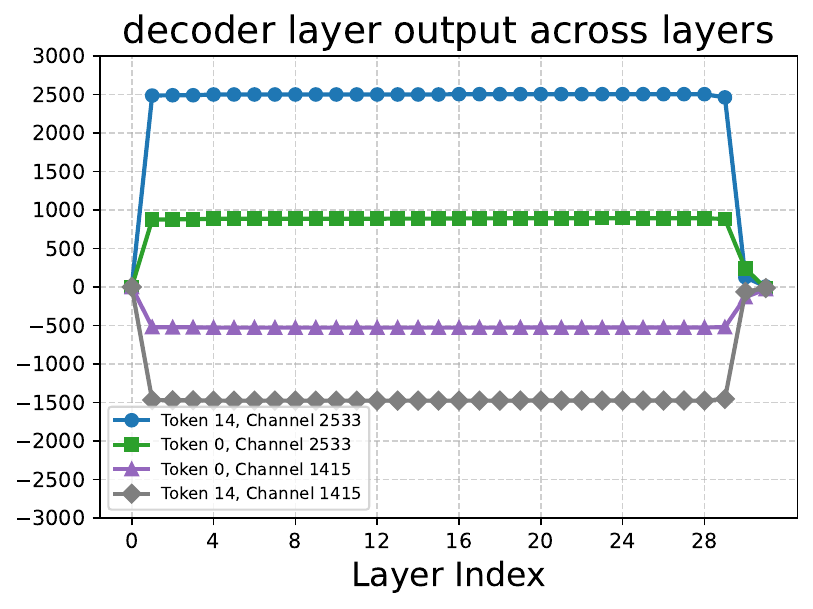}
        \vspace{-5mm}
        \caption{Stable outliers}
        \label{hidden_atates_2}
    \end{subfigure}
    \vspace{-2mm}
    \caption{Visualizations of attention sinks and extreme activation outliers in LLaMA2-7B.  
    (\ref{attn_sink}) illustrates the presence of attention sinks at tokens 0 and 14.  
    (\ref{hidden_atates_1}) shows extreme activation outliers in the output of the decoder layer 3, which emerge at attention sink tokens.
    (\ref{hidden_atates_2}) shows the distribution of stable outliers across decoder layers.    
    Unless otherwise specified, all visualizations use the following input from MMLU \citep{hendryckstest2021}:  
    "The following are multiple-choice questions (with answers) about machine learning. \textbackslash n\textbackslash n..." 
    }
\label{attn_sink_and_outlier}    
\end{figure}
\begin{figure}[t!]
    \vspace{-3mm}
    \centering    
    \begin{subfigure}{0.32\textwidth}
        \centering
    \includegraphics[width=\linewidth]{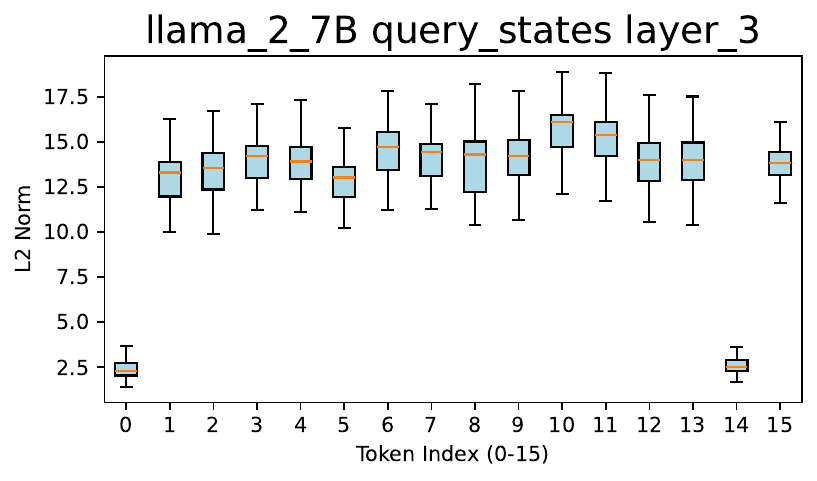}
    \end{subfigure}
    \begin{subfigure}{0.32\textwidth}
        \centering
    \includegraphics[width=\linewidth]{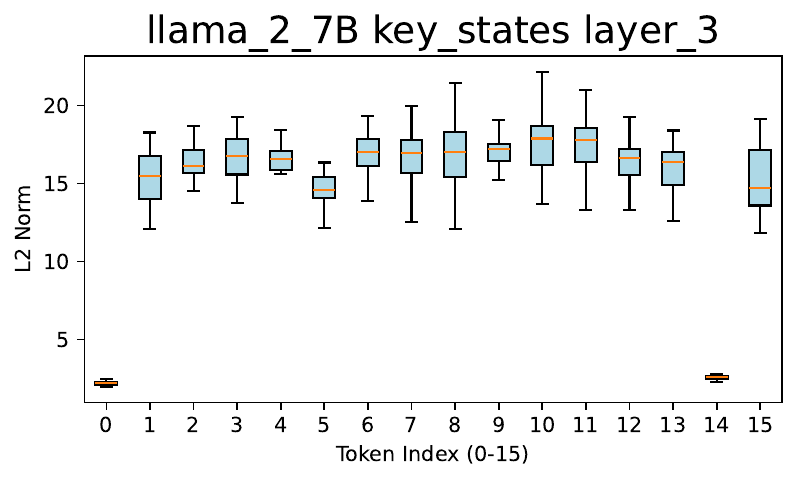}
    \end{subfigure}
    \begin{subfigure}{0.32\textwidth}
        \centering
    \includegraphics[width=\linewidth]{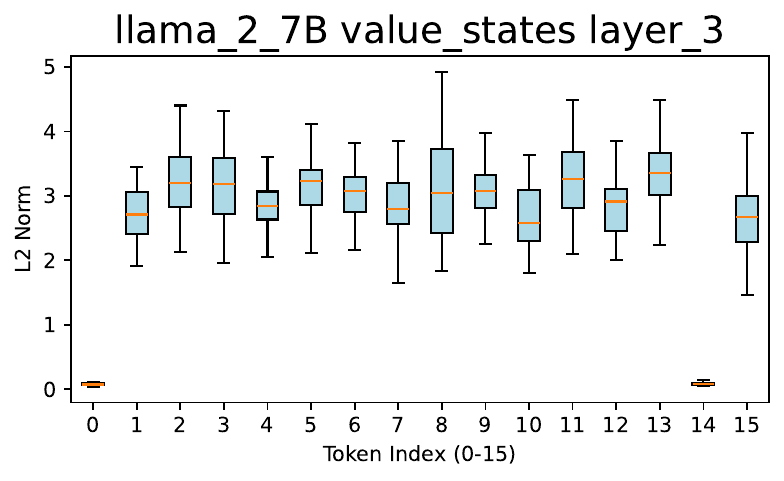}
    \end{subfigure}
    \vspace{-3mm}
    \caption{L2 norm distributions of Queries, Keys, and Values for all heads in the attention layer 3 of LLaMA2-7B across first 15 tokens.
    The sink tokens (0 and 14) display significantly smaller norms compared to non-sink tokens.}
    \vspace{-5mm}
\label{QKV}
\end{figure}

\cite{xiao2023efficient} characterizes the widely observed attention sink phenomenon, where LLMs tend to assign disproportionately high attention to the initial token, as shown in Figure \ref{attn_sink}. 
Further research \citep{an2025systematic, guo2024active} suggests that attention sinks have a profound impact on the compression of LLMs.
Specifically, the corresponding KVs require higher precision during quantization for adequate protection \citep{hooper2025kvquant,su2025rotatekv, duanmu2024skvq,su2025akvq} and must also be preserved during pruning \citep{xiao2023efficient, zhang2023h2o}. 
Additionally, as shown in Figure \ref{hidden_atates_1}, attention sink tokens have been found to exhibit extreme activation outliers \citep{bondarenko2021understanding,sun2024massive, an2025systematic}, which significantly impact activation quantization \citep{liu2024intactkv,li2024flash}. 
Although the practice of retaining the original precision of KVs for sinks tokens has proven effective in mitigating performance degradation, existing practices and understanding still have several limitations:
\textit{\textbf{(1)}} inadequate systematic analysis of attention sinks and their related phenomena during LLM inference;
\textit{\textbf{(2)}} lack of in-depth understanding of the mutual impact between attention sinks and KV cache quantization;
\textit{\textbf{(3)}} current implementations focus on statically persevering the KVs of first few tokens, which has proven insufficient in light of recent findings \citep{sun2024massive, yu2024unveiling} indicating that attention sinks can occur at other positions, as shown in Figure \ref{attn_sink}.

In this work, we aim to deepen the understanding of the role of attention sinks during LLM inference and their interplay with KV quantization, addressing the aforementioned limitations and advancing research in LLM compression and interpretability.
We begin by elucidating the role of attention sinks through an examination of the cross-layer evolution of various types of extreme activation outliers in Section \ref{section3}.
Previous studies \citep{sun2024massive, guo2024active} have identified that extreme activation outliers manifest in the hidden states between decoders.
As shown in Figure \ref{hidden_atates_2}, these outliers emerge and stabilize in the intermediate decoder layers, maintaining a persistent presence at sink tokens and LLM-specific channels, exhibiting large and consistent magnitudes.
Given their distinctive properties and their central role in our analysis, we refer to them as \textit{\textbf{stable outliers}} for clarity.
Interestingly, we further observe that stable outliers follow a structured progression across decoder layers, evolving through the stages of emergence, stabilization, and dissipation, driven by extreme activation outliers originating from the down-projection layer in the feed-forward network.
Additionally, attention sinks emerge and persist throughout the stabilization stage, serving as the core mechanism for maintaining stability.
As shown in Figure \ref{QKV}, this mechanism simultaneously imposes distinct numerical characteristics on the Queries, Keys, and Values of sink tokens, inevitably affecting KV quantization.

Then, in Section \ref{section4}, we provide a comprehensive analysis of the interplay between attention sinks and KV cache quantization.
We quantitatively assess the impact of attention sinks on various quantization schemes through quantization error analysis.
Our study also confirms that quantization significantly disrupts the implicit attention biases introduced by attention sinks.
Finally, based on our enhanced understanding, we introduce \textit{\textbf{KVSink}}, a plug-and-play method that effectively predict sink tokens by leveraging the intrinsic relationship between attention sinks and stable outliers.
Our contributions are summarized as follows:

\textbullet We advance the understanding of extreme activation outliers and attention sinks, elucidating the role of attention sinks during the stabilization phase of stable outliers, and clarifying the mechanism by which attention sinks influence KV cache quantization.

\textbullet To the best of our knowledge, this is the first work to thoroughly analyze and reveal the mutual influence between attention sinks and KV cache quantization. 
Our work not only deepens the understanding of attention sink preservation in KV cache quantization but also provides valuable insights for the development of more refined approaches in the future.

\textbullet Extensive experiments demonstrate that KVSink addresses the limitations of existing Preserve-First-N (PFN) strategy, providing more effective preservation with negligible overhead.
Additionally, KVSink further refines the well-established KVQuant method, leading to improved perplexity (PPL) and reduced dependence on 16-bit numerical outliers.
\section{Preliminary}
\label{preliminary}
\begin{wrapfigure}{r}{0.2\textwidth} 
\vspace{-6mm}
    \centering
    \includegraphics[width=\linewidth]{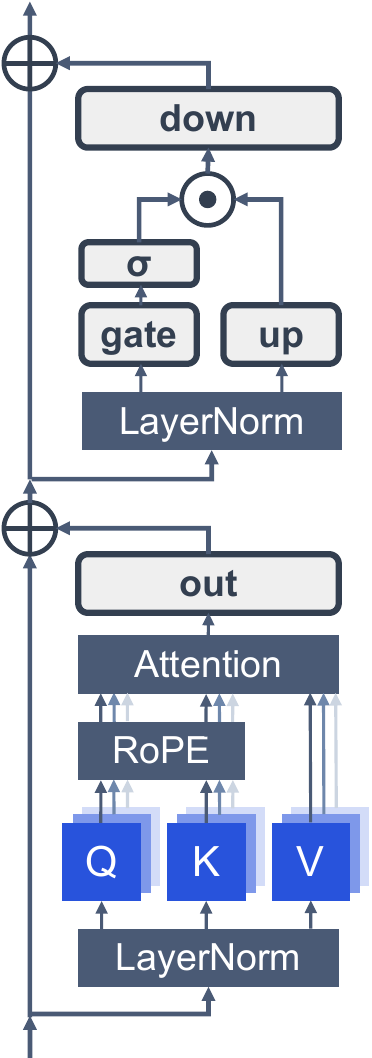}  
    \vspace{-7mm}
    \caption{Transformer decoder.}
    \label{fig:transformer}
    \vspace{-3mm}
\end{wrapfigure}

\textbf{Transformer decoder.}
LLMs are typically structured as a stack of Transformer decoder blocks \citep{vaswani2017attention}, each consisting of a multi-head self-attention (MHSA) layer and a feed-forward network (FFN) layer.
We use the widely adopted LLaMA \citep{dubey2024llama} architecture as an illustrative example, with a concise depiction provided in Figure \ref{fig:transformer}.
After tokenization and embedding, the input to the first decoder can be represented as \( H^{0} = \{ h^0_1, h^0_2, \dots, h^0_n \} \in \mathbb{R}^{n \times d} \), where \( h_i \) denotes the \( i \)-th input token, \( d \) is the embedding dimension, and \( n \) is the length of the tokenized input sequence. 
Then, the output of the \( l \)-th decoder block, \( H^l \in \mathbb{R}^{n \times d} \), is given by:
\begin{equation}
H^{l} = \text{FFN}\left( \text{LN}_{ffn} \left( H^{l'} \right) \right) + H^{l'}, 
\end{equation}
\begin{equation}
H^{l'} = O^l + H^{l-1}, O^l = \text{MHSA}\left( \text{LN}_{mhsa}\left( H^{l-1} \right) \right),
\end{equation}
where \( 1 \leq l \leq L \), with \( L \) denoting the total number of blocks.
LN refers to layer normalization (with LLaMA employing pre-norm), \( O^l \) representing the output of the MHSA, and \( H^{l'} \) denoting the output of residual summations after the MHSA.
Previous studies have found that Transformer-based models tend to learn structured activation outliers in \( H^{l} \) and \( H^{l'} \), which give rise to specific attention patterns that focus on special tokens, including BERT \citep{kovaleva2019revealing}, Vision Transformer (ViT) \citep{bondarenko2023quantizable}, and LLM \citep{sun2024massive}.
A more detailed discussion of related works on this topic is provided in Appendix \ref{related_work_1}.

\textbf{Multi-head self-attention.}
After LN, the input \( H^{l-1} \) is projected through the weight matrices \( W^{l}_Q, W^{l}_K, W^{l}_V \in \mathbb{R}^{d \times d} \) to generate the Queries, Keys, and Values, which are then divided into \( K \) heads, denoted as \( Q^{l,k}, K^{l,k}, V^{l,k} \), for \( 1 \leq k \leq K \).
The MHSA is computed as:
\begin{equation}
A^{l,k} = \text{Softmax}\left(\frac{Q^{l,k} K^{l,k^T}}{\sqrt{d_k}} + M \right), \quad O^l = \text{Concat}_{k=1}^{K} \left( A^{l,k} V^{l,k} \right)W^{l}_O,
\end{equation}
where \( M\) represents the attention mask, and \( d_k = d / K \).
For simplicity, the rotation position encoding (RoPE) \citep{su2024roformer} applied to the Queries and Keys is omitted here.

\textbf{Feed-forward network.}
Next, \( H^{l'} \) is passed through a new LN and subsequently enters the feed-forward network layer:
\begin{equation}
H^l = \text{FFN}\left( \text{LN}_{ffn}(H^{l'}) \right) + H^{l'} = \left( \sigma \left( \text{LN}_{ffn}(H^{l'}) W_g \right) \odot \text{LN}_{ffn}(H^{l'}) W_u \right) W_d + H^{l'} ,
\end{equation}
where \( W_g \), \( W_u \), and \( W_d \) are the weight matrices for the gating, up-projection, and down-projection. 
\( \sigma \) denotes the activation function, and \( \odot \) represents the Hadamard product.

\textbf{KV cache.}  
The inference process comprises two stages: the prefill phase and the decoding phase. 
In the prefill phase, the LLM processes the token sequence generated from the input and produces the initial output token. 
Each attention layer \( l \) computes and caches the KV tensors \( K^{l}_{\text{cache}} \) and \( V^{l}_{\text{cache}} \).
In the decoding phase, the model takes the newly generated token as input.
Let \( t^{l} \in \mathbb{R}^{1 \times d} \) denote the input embedding of the \( l \)-th attention layer.
Each attention layer computes the Queries, Keys, and Values  \( t^{l}_Q \), \( t^{l}_K \), and \( t^{l}_V \) as:
\begin{equation}
    t^{l}_Q = t^{l} \cdot W^{l}_Q, \quad t^{l}_K = t^{l} \cdot W^{l}_K, \quad t^{l}_V = t^{l} \cdot W^{l}_V.
\end{equation}
Then, \( t^{l}_K \) and \( t^{l}_V \) are used to update the KV cache, with the complete KV supporting subsequent MHSA computations:
\begin{equation}
    K^{l}_{\text{cache}} \leftarrow \mathrm{concat}(K^{l}_{\text{cache}}, t^{l}_K),
    V^{l}_{\text{cache}} \leftarrow \mathrm{concat}(V^{l}_{\text{cache}}, t^{l}_V).
\end{equation}
The growing size of the KV cache presents significant challenges in terms of memory usage and access latency, underscoring the need for efficient compression. 
Low-bit quantization has emerged as an effective approach, with related works discussed in Appendix \ref{related_2}.
\section{Attention Sinks and Extreme Activation Outliers }
\label{section3}
To enhance the understanding and preservation of attention sinks, we first present our findings on the intrinsic relationship between the cross-layer evolution of different types of extreme activation outliers in Section \ref{Extreme Activation Outliers}. 
Subsequently, we elucidate the core mechanism through which attention sinks maintain the stability of stable outliers in Section \ref{Stabilization}.
\subsection{Cross-Layer Evolution of Extreme Activation Outliers }
\label{Extreme Activation Outliers}
We begin by specifying the activations in which each type of outlier occurs, using the same notation as in Section \ref{preliminary}:

(1) The input to the down-projection layer, denoted as \( X^{l}_{\text{d\_in}} \), is given by
\begin{equation}
X^{l}_{\text{d\_in}} =  \sigma \left( \text{LN}_{ffn}(H^{l'}) W_g \right) \odot \text{LN}_{ffn}(H^{l'}) W_u ,
\end{equation}
(2) The output of the down-projection layer, denoted as \( X^{l}_{\text{d\_out}} \), is given by
\begin{equation}
X^{l}_{\text{d\_out}} = X^{l}_{\text{d\_in}} W_d,
\end{equation}
(3) The output of the residual summation after the MHSA, denoted as \( H^{l'} \),

(4) The output of the residual summation after the FFN, denoted as \( H^{l} \), where \textit{\textbf{stable outliers}} emerge.
Notably, it satisfies  
\(
H^l = X^{l}_{\text{d\_out}} + H^{l'}.
\)

\begin{figure*}[t!]
    \vspace{-7mm}
        \centering
\includegraphics[width=1\linewidth]{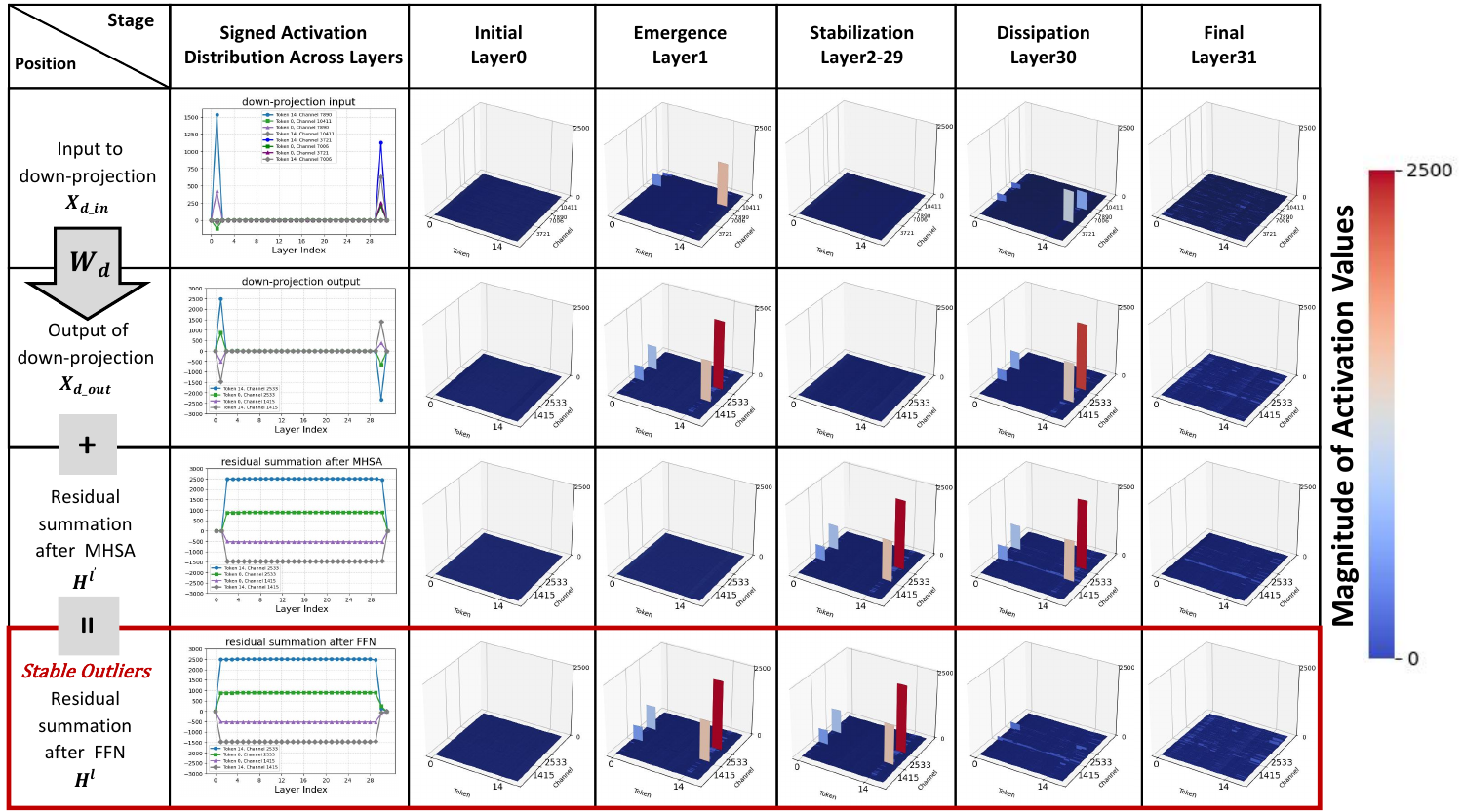}
    \label{down_in_outlier}
\vspace{-6mm}
    \caption{Visualizations of various types of extreme activation outliers in LLaMA2-7B. 
    The left side illustrates the cross-layer distribution of various types of extreme activation outliers, while the right side demonstrates their behavior in different stages.
    }
    \vspace{-4mm}
\label{stages}
\end{figure*}

While all these outliers appear at attention sink tokens and exhibit significantly larger magnitudes, they differ in their locations within the model and follow distinct cross-decoder-layer distribution patterns.
As shown on the left side of Figure \ref{stages}, outliers in \( X^{l}_{\text{d\_in}} \) and \( X^{l}_{\text{d\_out}} \) appear in the early and late layers, while outliers in \( H^{l'} \) and \( H^{l} \) remain present across the intermediate layers.
Interestingly, we found that the cross-layer evolution of these outliers unveils intrinsic relationships.
Stable outliers, driven by outliers in \( X^{l}_{\text{d\_in}} \) and \( X^{l}_{\text{d\_out}} \), undergo a structured progression, which we categorize into five stages: initial, emergence, stabilization, dissipation, and final (as shown on the right side of Figure \ref{stages}).

In the initial stage, no noticeable outliers are present.
Then, in the emergence stage, extreme outliers first appear in \( X^{l}_{\text{d\_in}} \), subsequently propagating to \( X^{l}_{\text{d\_out}} \), and ultimately inducing the emergence of stable outliers in \( H^{l} \) through the residual connection.
The stabilization stage extends across the intermediate layers, encompassing the majority of the decoder layers.  
In this stage, outliers persist in \( H^{l'} \) and \( H^{l} \), while \( X^{l}_{\text{d\_in}} \) and \( X^{l}_{\text{d\_out}} \) no longer exhibit extreme outliers. 
Instead, their magnitudes decrease significantly, resulting in minimal contribution to the variation in extreme outliers.
During the dissipation stage, extreme outliers re-emerge in \( X^{l}_{\text{d\_in}} \), and their propagation to \( X^{l}_{\text{d\_out}} \) generates outliers at the same positions as those observed during the emergence stage, with similar magnitudes but opposite signs.
After the residual summation, this results in a significant reduction or disappearance of the stable outliers.
Finally, in the last stage, no noticeable extreme outliers remain, and the model is ready to generate the output.
Overall, these various extreme activation outliers demonstrate systematic relationships and interactions. 
Experimental results on additional models and inputs can be found in Appendix \ref{cross_layer}.
\subsection{Attention Sinks and the Stabilization of Stable Outliers}
\label{Stabilization}
\begin{figure}[t!]
    \vspace{-9mm}
    \centering    
    \begin{subfigure}{0.21\textwidth}
        \centering
    \includegraphics[width=\linewidth]{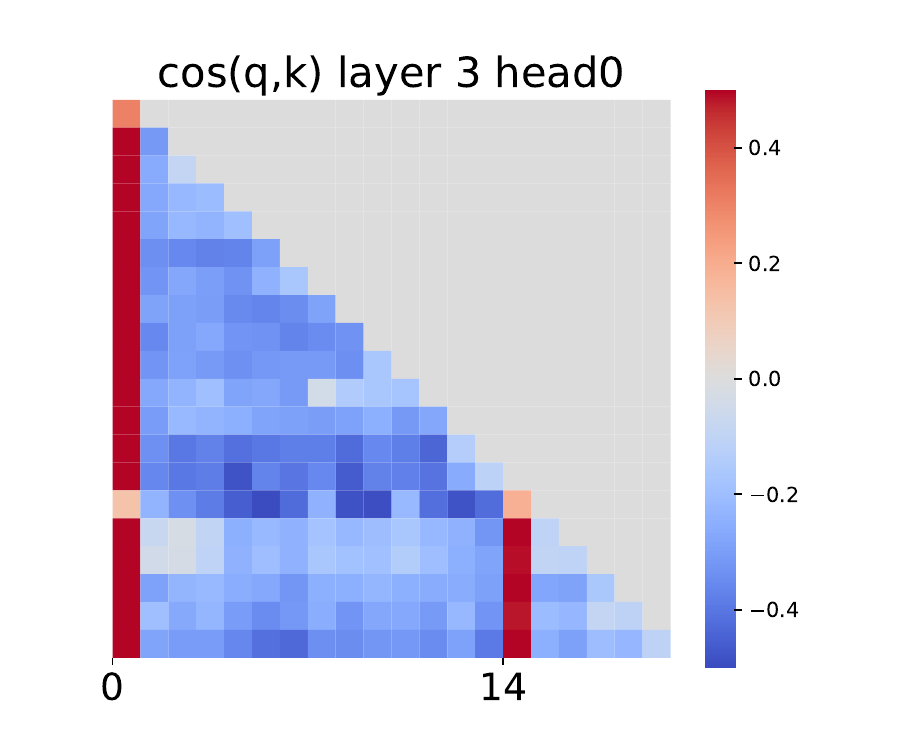}
    \vspace{-5mm}
    \caption{Cosine similarity}
    \label{3.2Cosine similarity}
    \end{subfigure}
    \begin{subfigure}{0.35\textwidth}
        \centering
    \includegraphics[width=\linewidth]{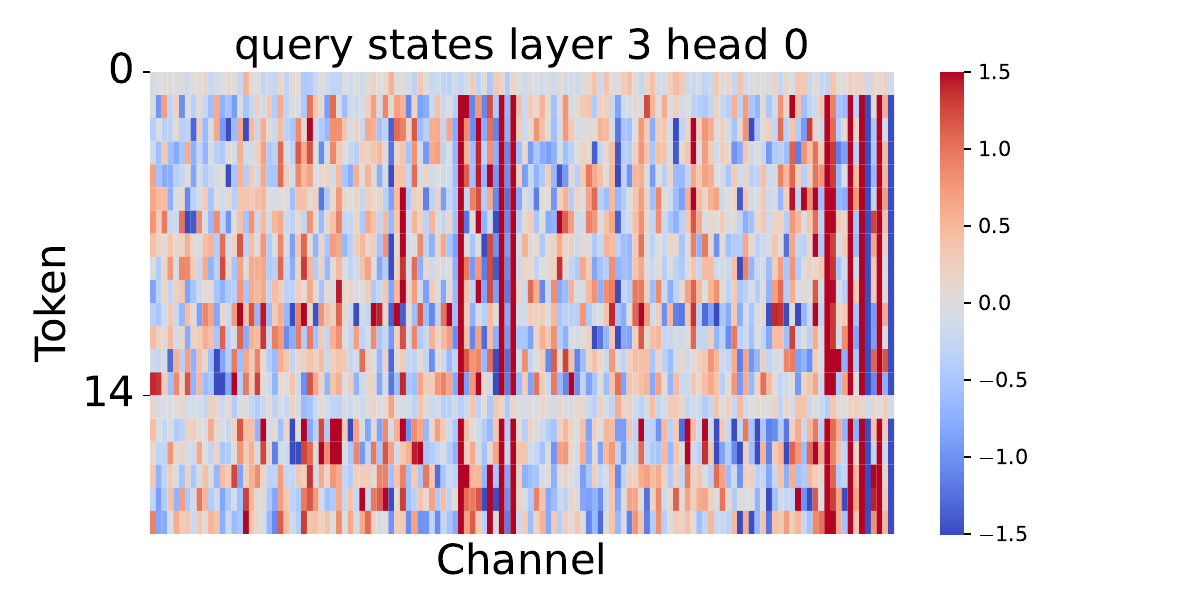}
        \vspace{-5mm}
    \caption{Query states}

    \label{3.2Query states}
    \end{subfigure}
    \begin{subfigure}{0.35\textwidth}
        \centering
    \includegraphics[width=\linewidth]{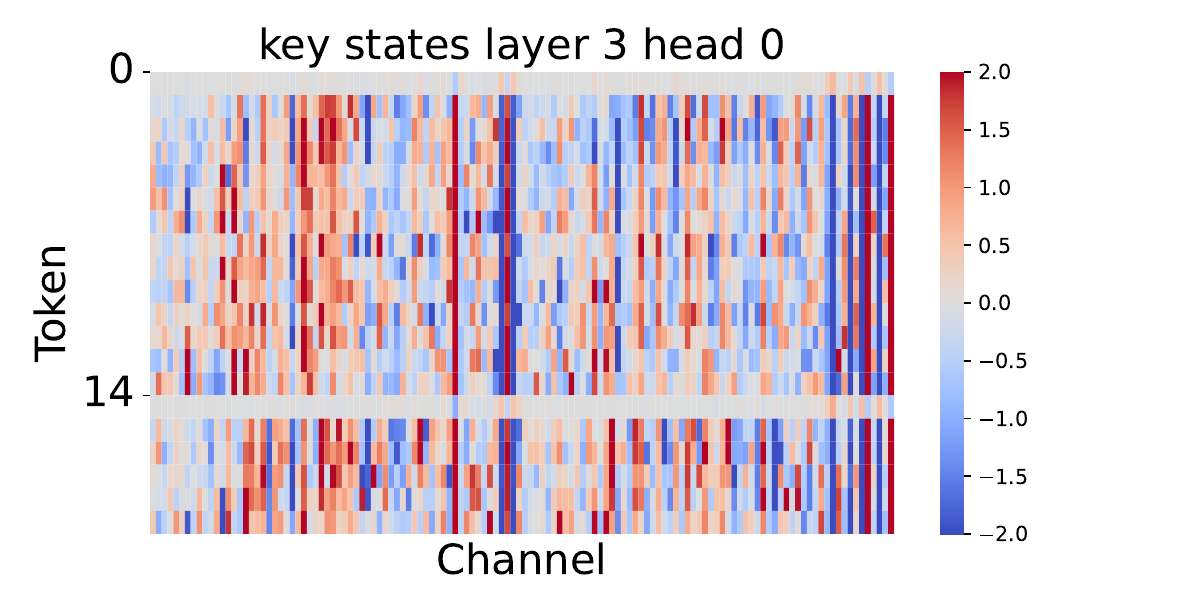}
    \vspace{-5mm}
    \caption{Key states}
    \label{3.2Key states}
    \end{subfigure}
    \begin{subfigure}{0.21\textwidth}
        \centering
    \includegraphics[width=\linewidth]{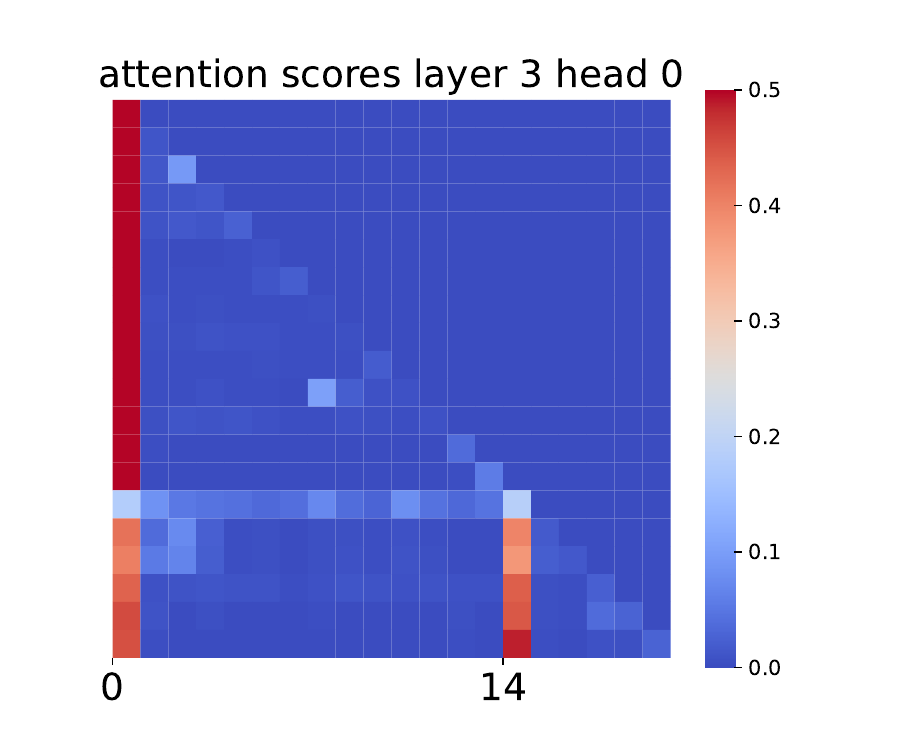}
    \vspace{-5mm}
    \caption{Attention scores}
    \label{3.2Attention scores}
    \end{subfigure}
    \begin{subfigure}{0.35\textwidth}
        \centering
    \includegraphics[width=\linewidth]{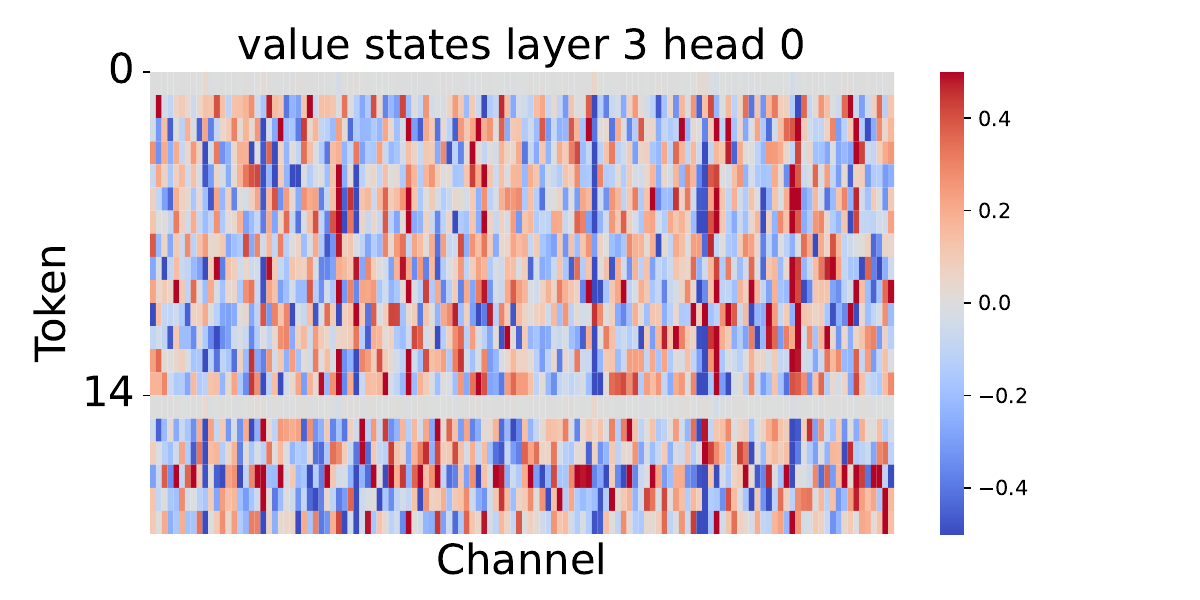}
    \vspace{-5mm}
    \caption{Value states}
    \label{3.2Value states}
    \end{subfigure}
    \begin{subfigure}{0.35\textwidth}
        \centering
    \includegraphics[width=\linewidth]{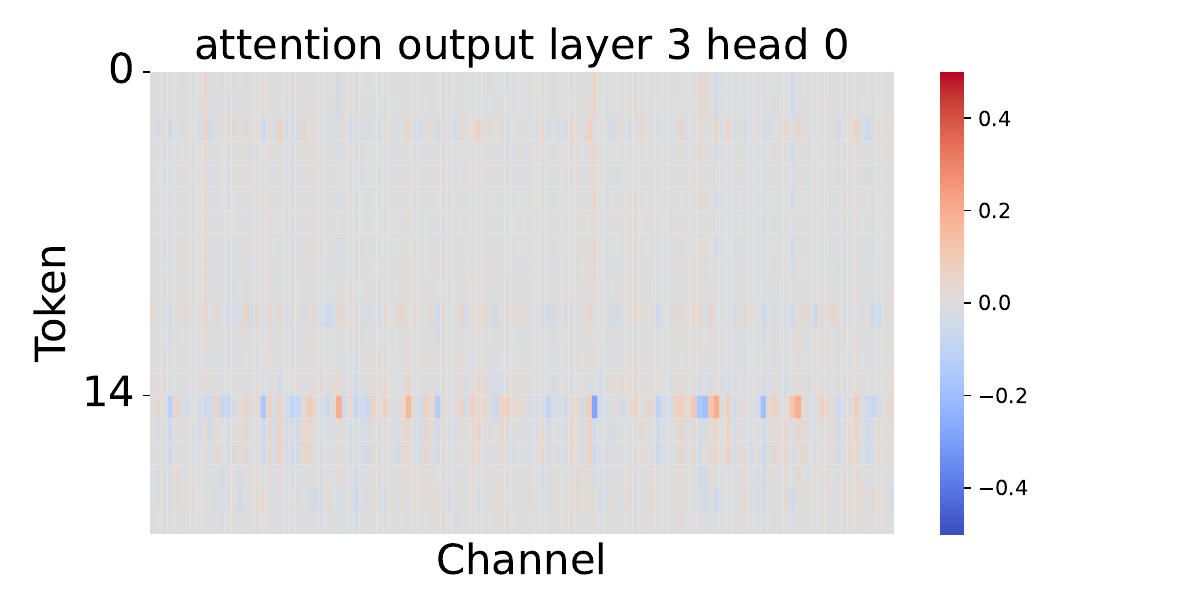}
    \vspace{-5mm}
    \caption{Attention outputs}
    \label{3.2Attention outputs}
    \end{subfigure}
    \vspace{-2mm}
    \caption{
    (\ref{3.2Query states}), (\ref{3.2Key states}), and (\ref{3.2Value states}) illustrate QKV suppression.
    (\ref{3.2Cosine similarity}) highlights the high cosine similarity of QK.  
    (\ref{3.2Attention outputs}) visualizes the attention output.
    }
    \vspace{-2mm}
\label{3.2}
\end{figure}

A notable characteristic of stable outliers is that during the stabilization phase, they remain consistently present, with their values varying only slightly.
This implies that during the stabilization phase, the FFN and MHSA layers make minimal updates to the hidden states. 
Motivated by this and drawing inspiration from previous research \citep{bondarenko2023quantizable}, we conclude that LLMs leverage attention sinks to achieve this behavior in the MHSA layer.
Specifically, this process is governed by the following two key mechanisms:

\textbf{QKV suppression.}  
As shown in Figures \ref{QKV} and \ref{3.2}, the Queries, Keys, and Values of sink tokens exhibit significantly smaller norms compared to non-sink tokens.  

\textbf{High cosine similarity of QK.}  
Although the norms of Queries and Keys are small, the cosine similarity between the Queries of non-sink tokens and the Keys of the sink tokens remains high \citep{gu2024attention}, resulting in large attention scores, as shown in Figure \ref{3.2Cosine similarity}.

Due to these two mechanisms, a small number of sink tokens exhibit extremely high attention scores but small Values, while the remaining tokens receive lower attention scores, resulting in attention outputs with small values, as shown in Figure \ref{3.2Attention outputs}.
This perspective aligns with previous research \citep{bondarenko2023quantizable} on extreme outliers in pre-LLM Transformers, but we arrive at this conclusion from a novel angle by analyzing the stabilization of stable outliers during LLM inference.
Notably, the QKV suppression mechanism imposes distinct numerical characteristics on the Queries, Keys, and Values of sink tokens, which is the fundamental reason for their sensitivity to quantization. 
Additional experimental results on QKV suppression and high cosine similarity of QK are provided in Appendix \ref{QKV Suppression}.
\begin{table}[t]
\centering
\resizebox{0.96\columnwidth}{!}{%
\begin{tabular}{@{}cc|ccc|ccc|ccc@{}}
\toprule
\multicolumn{2}{c|}{\begin{tabular}[c]{@{}c@{}}Quantization\\ Schemes\end{tabular}} & \multicolumn{3}{c|}{\begin{tabular}[c]{@{}c@{}}Per-Token Key\\ Quantization Error\end{tabular}} & \multicolumn{3}{c|}{\begin{tabular}[c]{@{}c@{}}Per-Token Value\\ Quantization Error\end{tabular}} & \multicolumn{3}{c}{\begin{tabular}[c]{@{}c@{}}Per-Channel Key\\ Quantization Error\end{tabular}} \\ \midrule
\multicolumn{1}{c|}{\multirow{4}{*}{Dynamic}} & Bits & overall & \begin{tabular}[c]{@{}c@{}}w/o\\ Sink tokens\end{tabular} & \begin{tabular}[c]{@{}c@{}}w/ \\ Sink tokens\end{tabular} & Overall & \begin{tabular}[c]{@{}c@{}}w/o\\ Sink tokens\end{tabular} & \begin{tabular}[c]{@{}c@{}}w/ \\ Sink tokens\end{tabular} & Overall & \begin{tabular}[c]{@{}c@{}}w/o\\ Sink tokens\end{tabular} & \begin{tabular}[c]{@{}c@{}}w/ \\ Sink tokens\end{tabular} \\ \cmidrule(l){2-11} 
\multicolumn{1}{c|}{} & 4 & 1.30 & 1.38 & 0.07 & 0.10 & 0.11 & 0.01 & 0.28 & 0.26 & 0.31 \\
\multicolumn{1}{c|}{} & 3 & 9.00 & 9.57 & 0.50 & 0.68 & 0.72 & 0.08 & 2.03 & 1.78 & 2.30 \\
\multicolumn{1}{c|}{} & 2 & 27.58 & 29.32 & 1.37 & 2.53 & 2.68 & 0.23 & 7.05 & 6.52 & 7.58 \\ \midrule
\multicolumn{1}{c|}{\multirow{4}{*}{Static}} & Bits & overall & \begin{tabular}[c]{@{}c@{}}w/o\\ Sink tokens\end{tabular} & - & Overall & \begin{tabular}[c]{@{}c@{}}w/o\\ Sink tokens\end{tabular} & - & Overall & \begin{tabular}[c]{@{}c@{}}w/o\\ Sink tokens\end{tabular} & - \\ \cmidrule(l){2-11} 
\multicolumn{1}{c|}{} & 4 & 86.73 & 16.39 & - & 9.75 & 3.10 & - & 9.02 & 5.17 & - \\
\multicolumn{1}{c|}{} & 3 & 89.60 & 20.41 & - & 10.09 & 3.45 & - & 10.24 & 8.26 & - \\
\multicolumn{1}{c|}{} & 2 & 107.36 & 46.14 & - & 11.30 & 6.35 & - & 15.83 & 11.44 & - \\ \bottomrule
\end{tabular}%
}
\vspace{-2mm}
\caption{The experiments are conducted using round-to-nearest (RTN) integer quantization and the mean squared error (MSE) metric on the LLaMA2-7B model, with the values in the table scaled by a factor of 100.
The quantization group size is uniformly set to 16, and global min-max is used for static quantization. 
Under dynamic quantization, \textit{w/o Sink tokens} refers to MSE for quantization groups that exclude sink tokens, while \textit{w/ Sink tokens} refers to MSE solely for groups containing sink tokens.
Under static quantization, \textit{w/o Sink tokens} indicates that sink tokens are excluded during both the calibration and quantization.
}
\vspace{-3mm}
\label{tab:error}
\end{table}
\section{KV Cache Quantization and Attention Sinks}
\label{section4}
\subsection{Impact of Attention Sinks on KV Cache Quantization }
Due to QKV suppression, the inclusion of sink tokens within a quantization group can expand the quantization range and exacerbate quantization errors, ultimately leading to further performance degradation. 
This arises from the trade-off between range and precision.
Recent KV cache quantization methods adopt diverse quantization schemes \citep{liu2024kivi,hooper2025kvquant,duanmu2024skvq}, yet the impact of attention sinks across these schemes remains largely unexplored.
In this section, we conduct an comprehensive study of the impact of attention sinks on widely adopted per-token and per-channel approaches under both dynamic and static quantization through quantization error analysis (presented in Table \ref{tab:error}).
An overview of low-bit quantization is provided in Appendix \ref{Preliminary on Low-Bit Quantization}.

\textbf{Per-token quantization.}
For per-token dynamic quantization, since quantization parameters are computed within each group corresponding to a single token, the QKV suppression of sink tokens does not influence other tokens.
However, in static quantization, where parameters are calibrated and fixed throughout quantization, the influence of attention sinks propagates across all quantization groups, exacerbating performance degradation. 
As shown in Table \ref{tab:error}, excluding sink tokens during static per-token quantization significantly reduces quantization error, decreasing by up to 81.1\% for Keys and 68.2\% for Values.

\textbf{Per-channel Key quantization.} 
Based on the observation that the Keys exhibit outliers in specific channels, while Values do not, some research \citep{hooper2025kvquant,liu2024kivi} employs per-channel quantization for Keys.
In per-channel dynamic Key quantization, groups containing sink tokens experience larger quantization errors, while those without remain unaffected. 
As shown in Table \ref{tab:error}, quantization errors increase by 16.3\% to 29.2\% compared to those without sink tokens.
In per-channel static Key quantization, similar to per-token static quantization, excluding the impact of sink tokens leads to a reduction in quantization error of up to 42.7\%.
\begin{figure}[t]
\vspace{-9mm}
    \centering
    \begin{subfigure}{0.37\textwidth}  
        \centering
        \includegraphics[width=\linewidth]{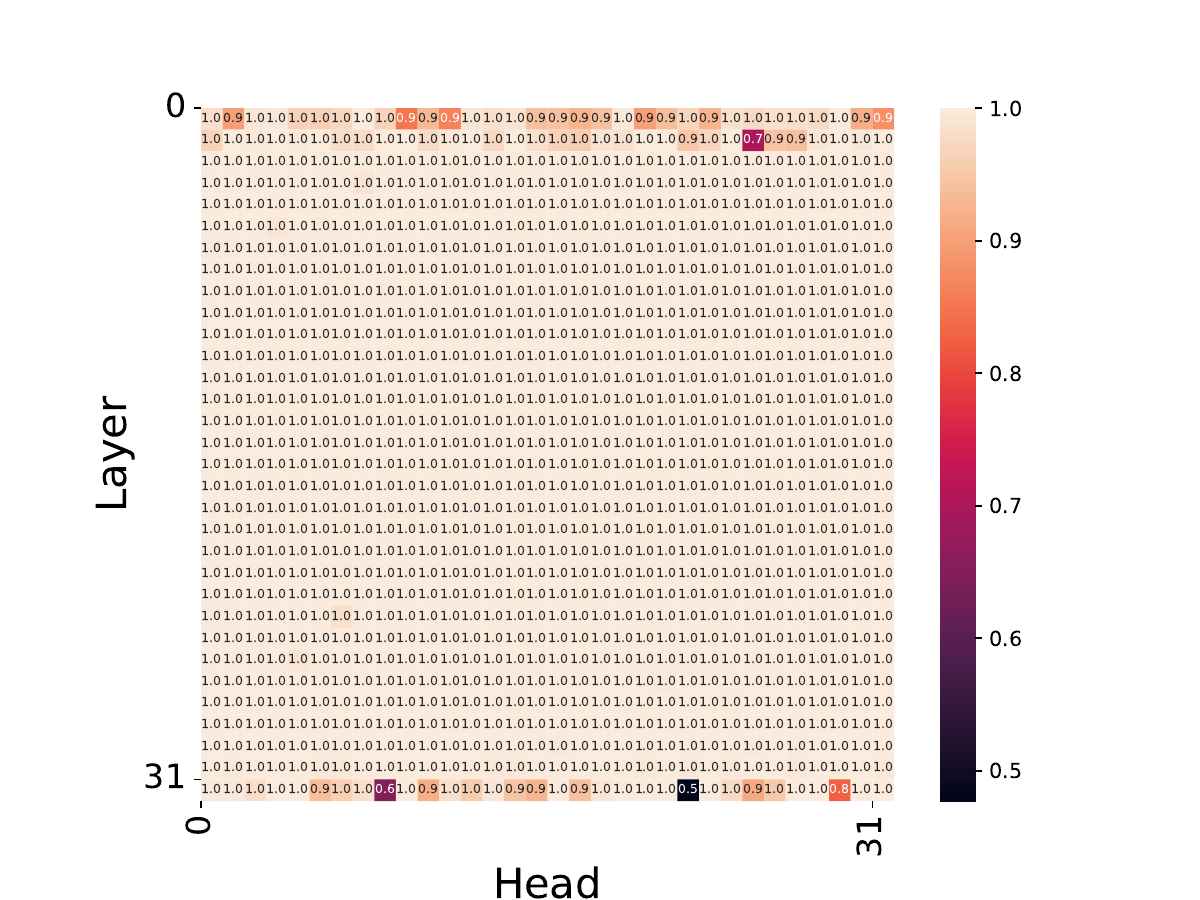}
        \caption{Average cosine similarity}
        \label{7left}
    \end{subfigure}
    \hfill
    \begin{subfigure}{0.62\textwidth}  
        \centering
        \begin{minipage}{\linewidth}
            \centering
            \begin{subfigure}{0.32\textwidth}
                \centering
                \includegraphics[width=\linewidth]{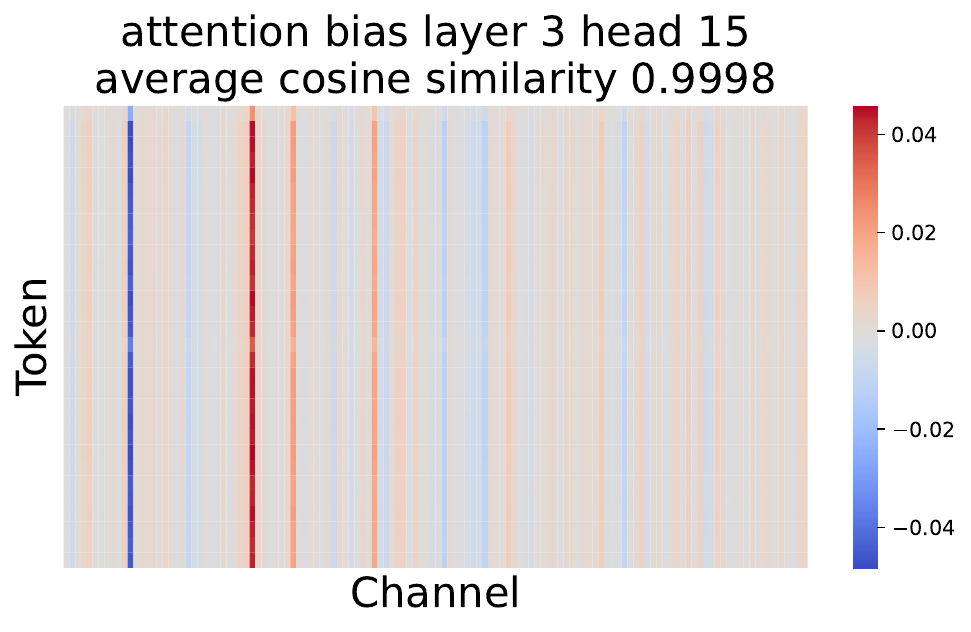}
            \end{subfigure}
            \begin{subfigure}{0.33\textwidth}
                \centering
                \includegraphics[width=\linewidth]{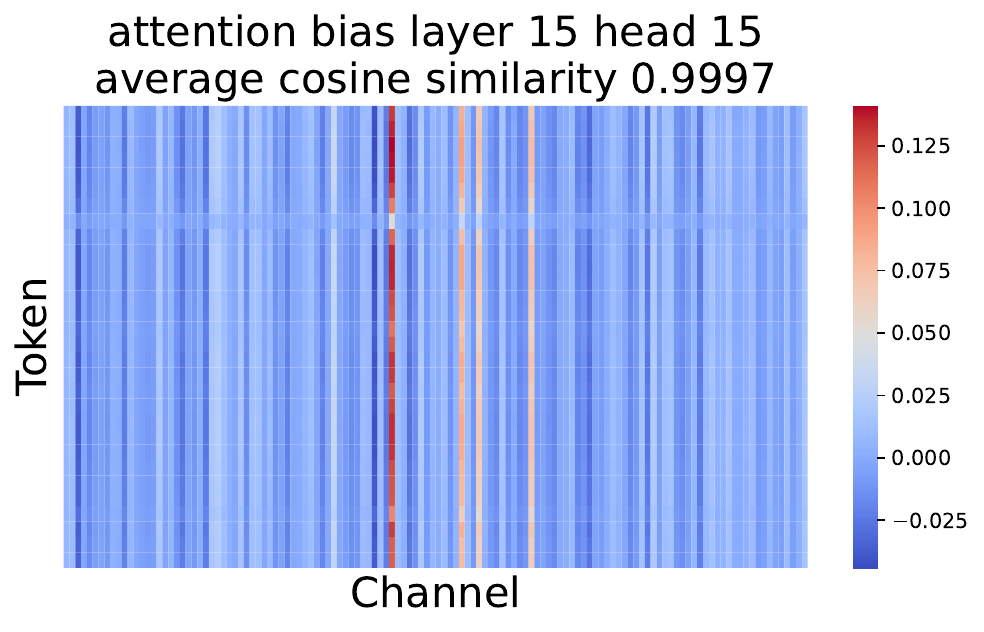}
            \end{subfigure}
            \begin{subfigure}{0.31\textwidth}
                \centering
                \includegraphics[width=\linewidth]{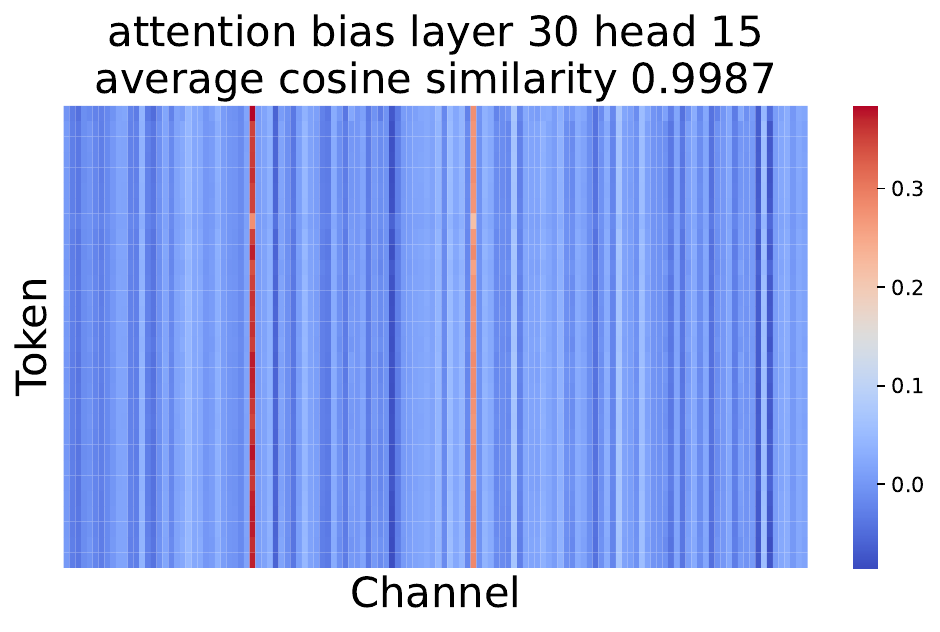}
            \end{subfigure}
        \end{minipage}
        \begin{minipage}{\linewidth}
            \centering
            \begin{subfigure}{0.32\textwidth}
                \centering
                \includegraphics[width=\linewidth]{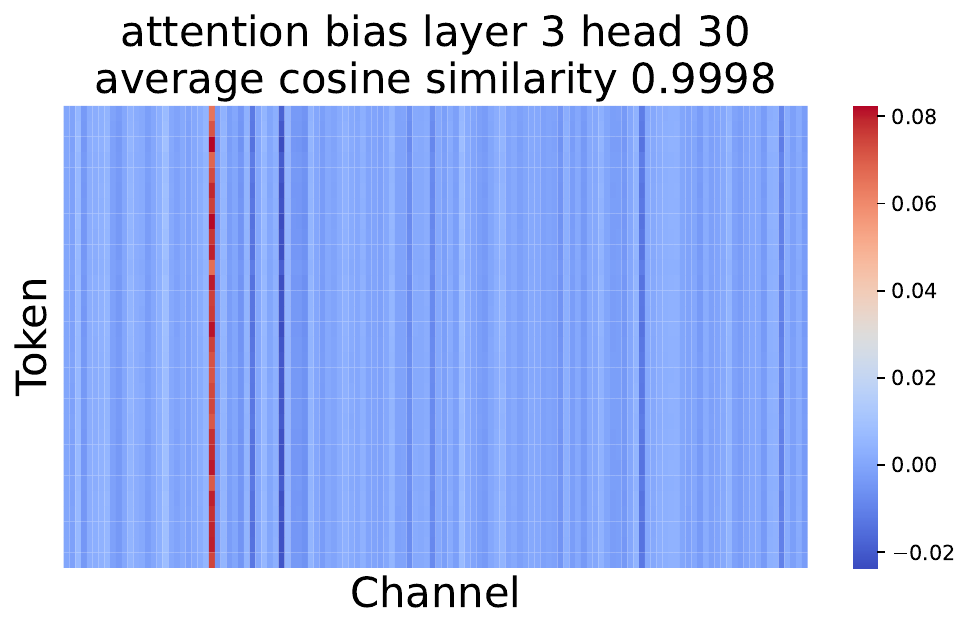}
            \end{subfigure}
            \begin{subfigure}{0.32\textwidth}
                \centering
                \includegraphics[width=\linewidth]{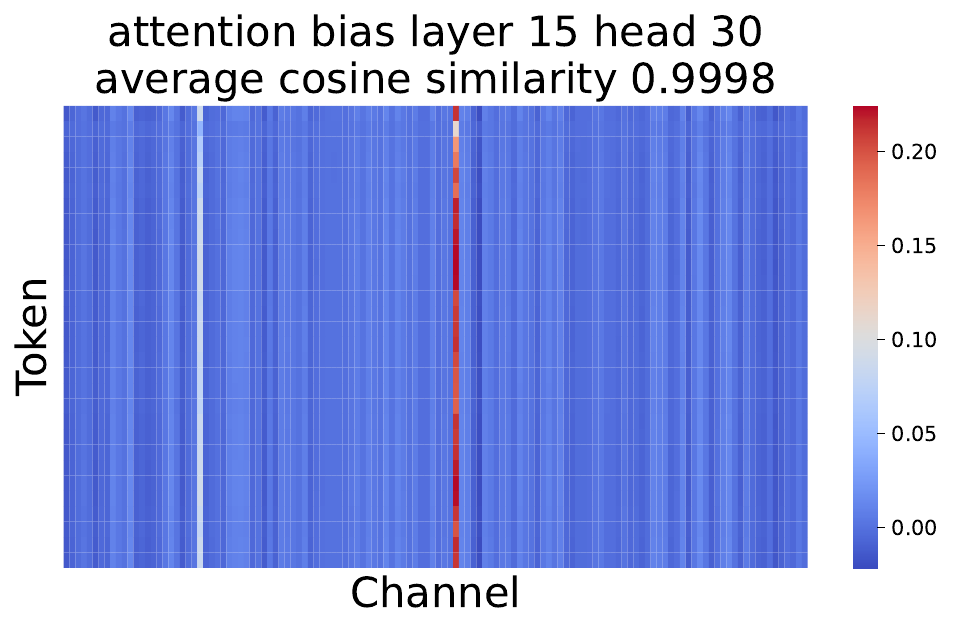}
            \end{subfigure}
            \begin{subfigure}{0.32\textwidth}
                \centering
                \includegraphics[width=\linewidth]{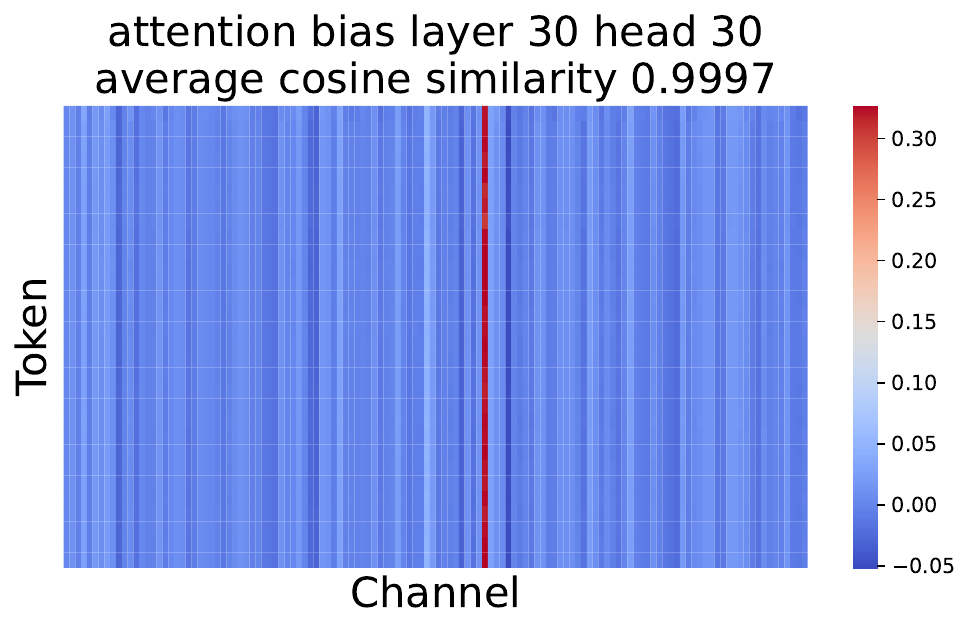}
            \end{subfigure}
        \end{minipage}
        \caption{Attention biases visualization}
    \label{7right}        
    \end{subfigure}
    \vspace{-6mm}
    \caption{
    (\ref{7left}) depicts the average cosine similarity of \( \sum_{i \in S} p_i^t v_i \) across all tokens for each head on LLaMA2-7B. 
    (\ref{7right}) visualizes the attention biases for several example heads.}
    \label{attention_bias}
\end{figure}
\begin{figure}[t]
\vspace{-2mm}
    \centering    
    \begin{subfigure}{0.19\textwidth}
        \centering
    \includegraphics[width=\linewidth]{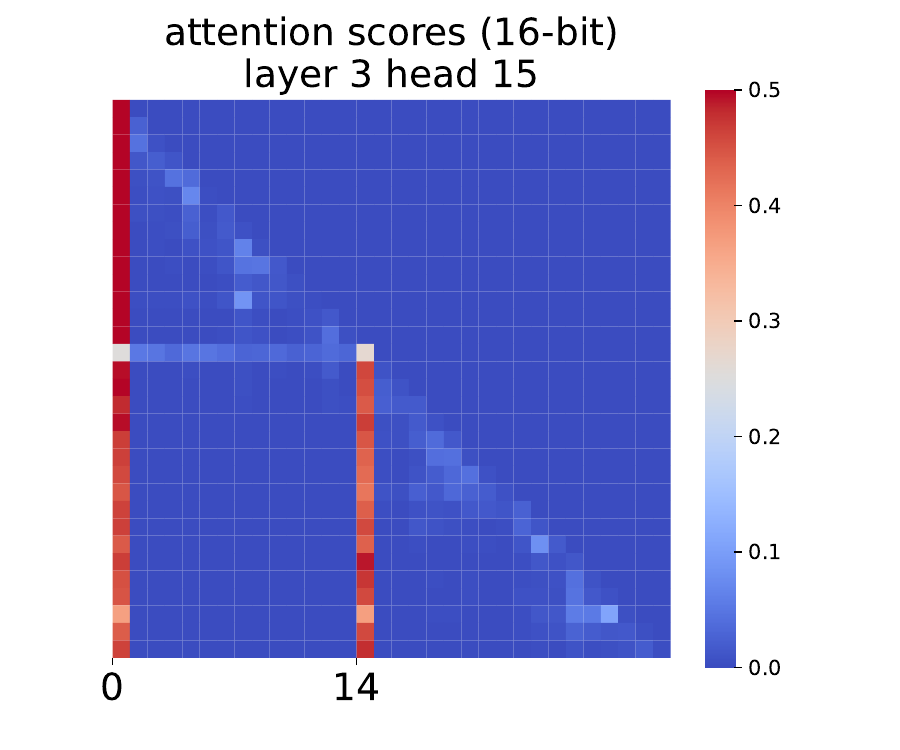}
    \end{subfigure}
    \begin{subfigure}{0.19\textwidth}
        \centering
    \includegraphics[width=\linewidth]{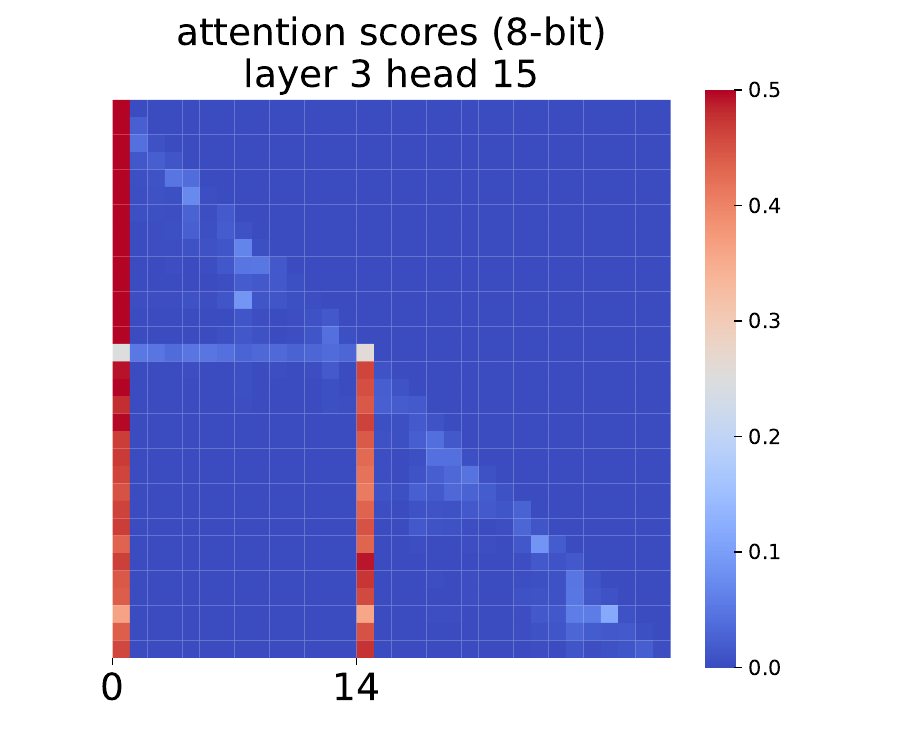}
    \end{subfigure}
    \begin{subfigure}{0.19\textwidth}
        \centering
    \includegraphics[width=\linewidth]{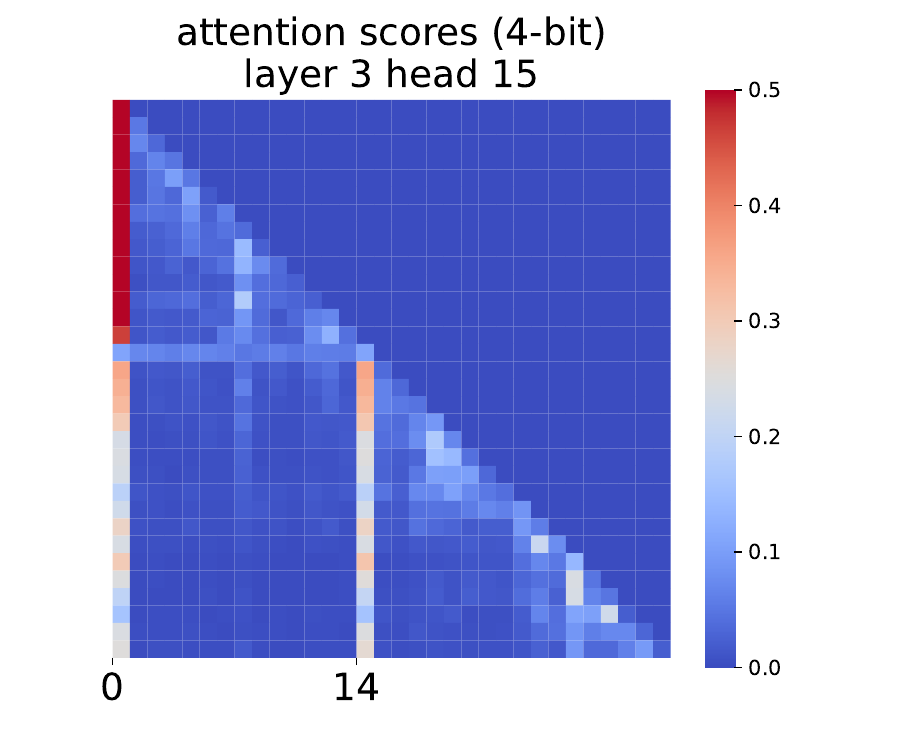}
    \end{subfigure}
    \begin{subfigure}{0.19\textwidth}
        \centering
    \includegraphics[width=\linewidth]{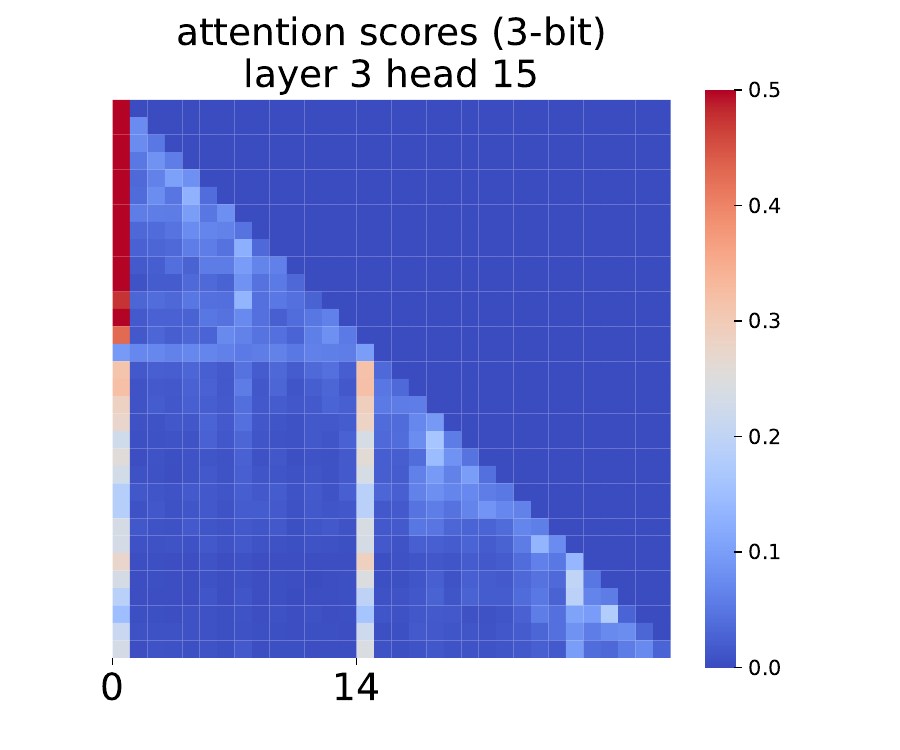}
    \end{subfigure}
    \begin{subfigure}{0.19\textwidth}
        \centering
    \includegraphics[width=\linewidth]{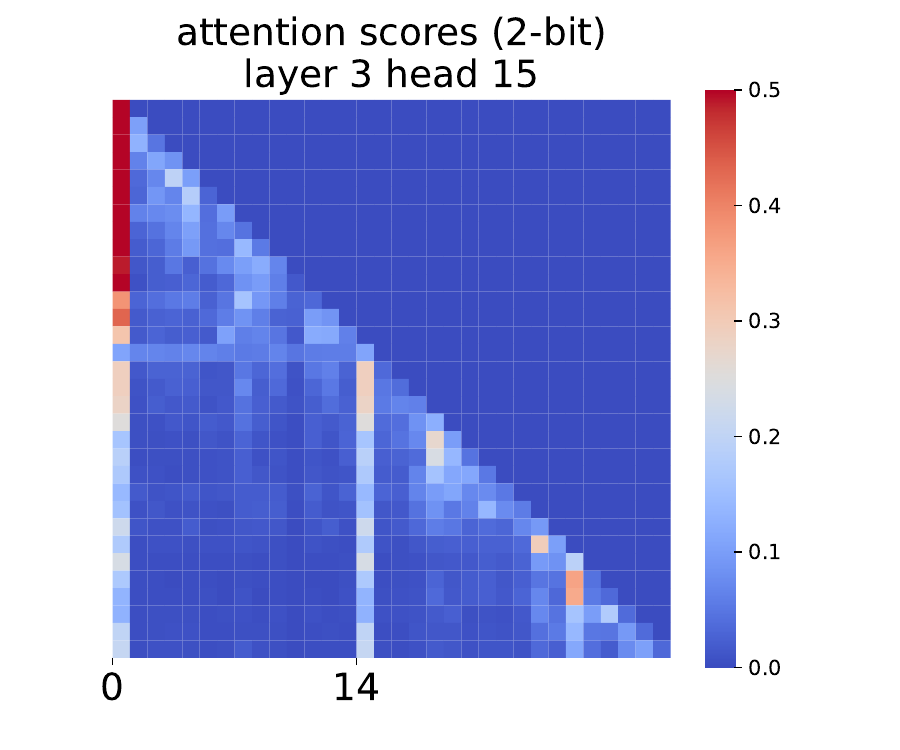}
    \end{subfigure}
    \begin{subfigure}{0.19\textwidth}
        \centering
    \includegraphics[width=\linewidth]{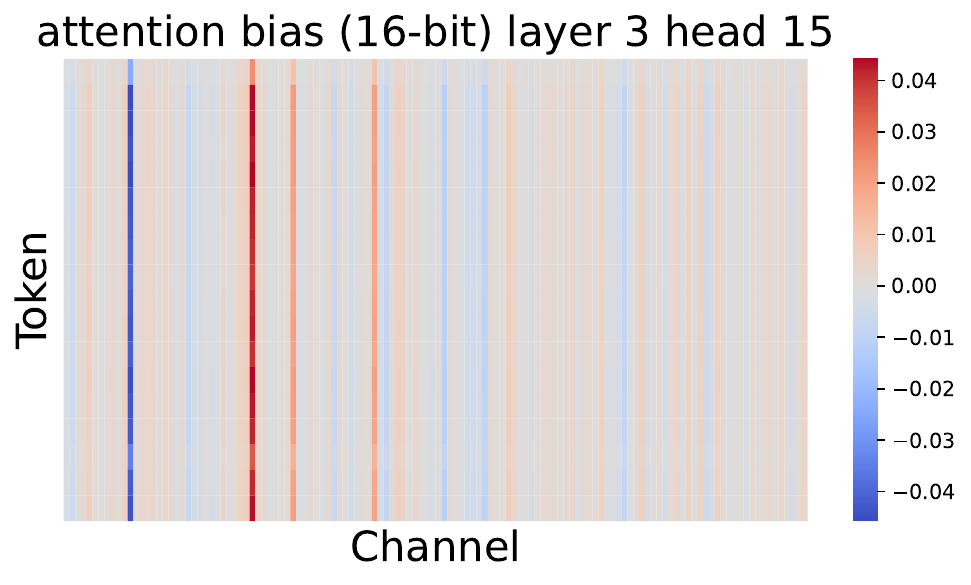}
    \caption{16-bit}
    \end{subfigure}
    \begin{subfigure}{0.19\textwidth}
        \centering
    \includegraphics[width=\linewidth]{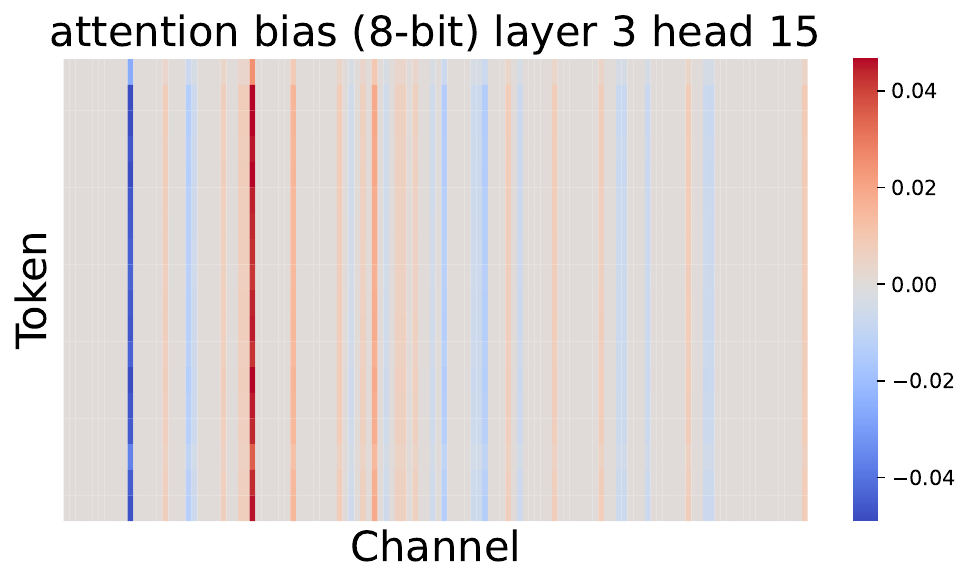}
    \caption{8-bit}
    \end{subfigure}
    \begin{subfigure}{0.19\textwidth}
        \centering
    \includegraphics[width=\linewidth]{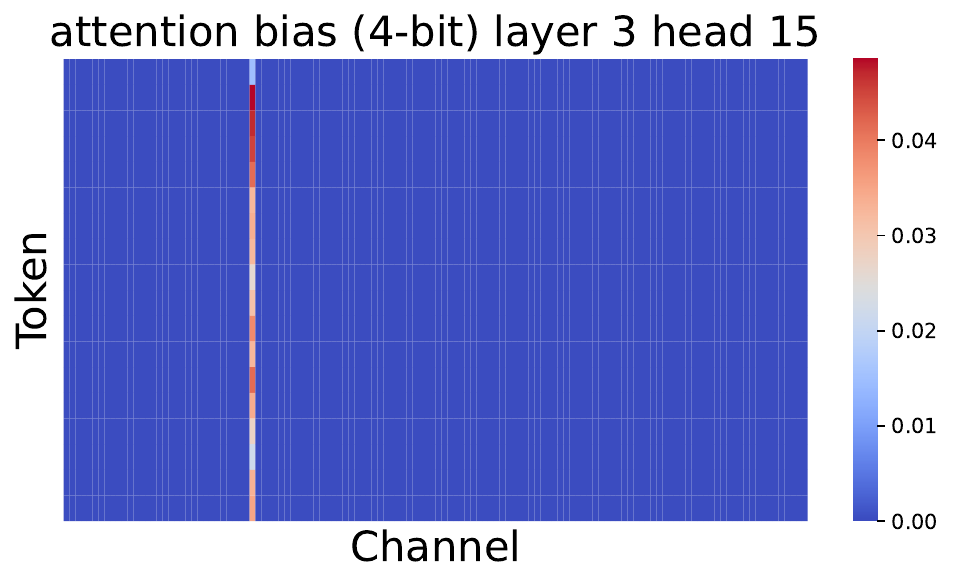}
    \caption{4-bit}
    \end{subfigure}
    \begin{subfigure}{0.19\textwidth}
        \centering
    \includegraphics[width=\linewidth]{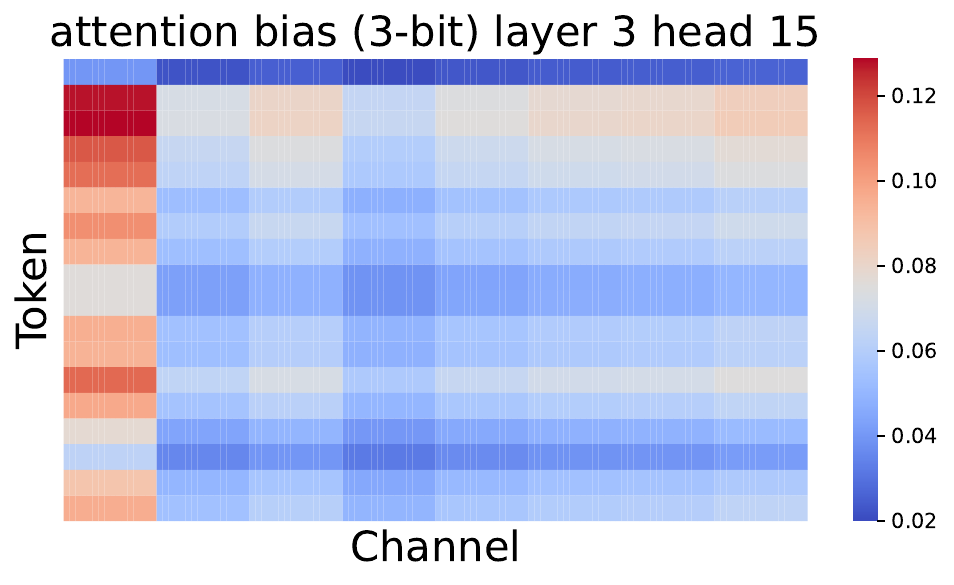}
    \caption{3-bit}
    \end{subfigure}
    \begin{subfigure}{0.19\textwidth}
        \centering
    \includegraphics[width=\linewidth]{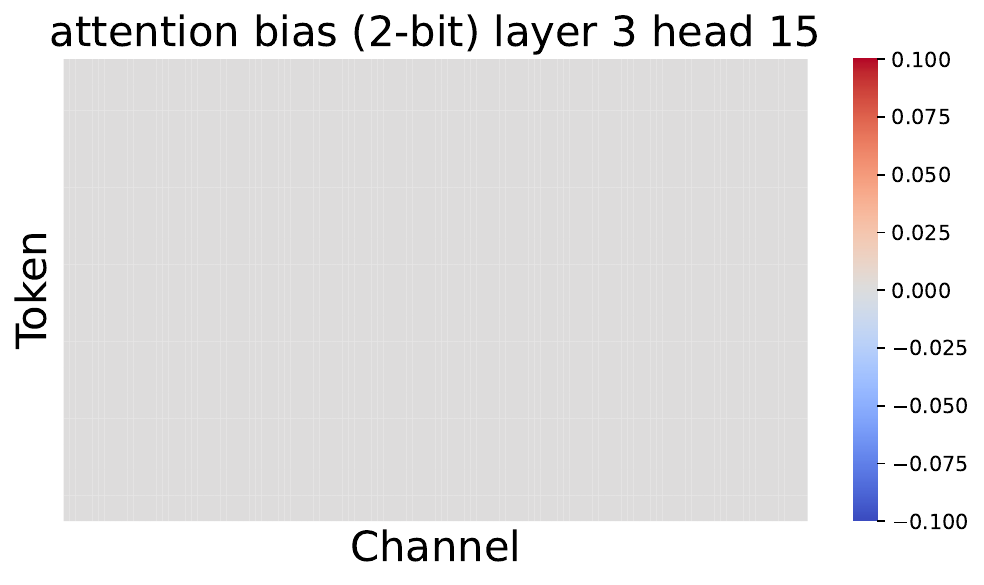}
    \caption{2-bit}
    \end{subfigure}
    \vspace{-3mm}
    \caption{The impact of KV cache quantization on attention scores and attention biases across various bit-widths, employing static per-token quantization with a group size of 16.
    }
    \vspace{-3mm}
\label{quant_attn_bias}
\end{figure}
\subsection{Effect of KV Cache Quantization on Attention Sinks }
\label{nfluence of KV Quantization on Attention Sinks}
Several previous research \citep{sun2024massive,gu2024attention,an2025systematic} suggest that attention sinks and extreme activation outliers introduce implicit attention biases and demonstrate that incorporating explicit learnable biases during training can effectively eliminate outliers. 
Building on this insight, we first validate the presence of attention biases induced by attention sinks during inference and then analyze the impact of KV cache quantization on these biases.
The attention output of token \( t \) can be expressed as:
\begin{equation}
\text{Attention}(Q, K, V)^t = \sum_{i \leq t} p_i^t v_i =
\sum_{\substack{i \notin S}} p_i^t v_i + \sum_{i \in S} p_i^t v_i,
\label{equaiton:attention sink}
\end{equation}
where \( p_i^t \) represents the attention score of the Query token \( t \) and the Key token \( i \), \( S \) denotes the set of sink tokens and \( v_i \) denotes the Value of token \( i \). 
As shown in Equation \ref{equaiton:attention sink}, the presence of attention sinks influences the attention output of each token \( t \) through \( \sum_{i \in S} p_i^t v_i \).
To verify whether this represents attention biases, we calculate the average cosine similarity of \( \sum_{i \in S} p_i^t v_i \) across all tokens for each attention head.
As shown in Figure \ref{attention_bias}, we find that for each head, \( \sum_{i \in S} p_i^t v_i \) remains highly consistent across all tokens when attention sinks emerge, confirming that it represents the attention biases.

Next, we analyze the impact of KV cache quantization on attention sinks and attention biases. 
As shown in Figure \ref{quant_attn_bias}, quantization significantly affects both, with the impact becoming more pronounced as the bit-width decreases. 
Notably, since the biases introduced by attention sinks persist across all subsequent tokens and may contain global or other crucial information \citep{darcet2023vision}, their influence on attention computation remains continuous and significant.
\subsection{KVSink}
\begin{wrapfigure}{r}{0.30\textwidth} 
\vspace{-6mm}
    \centering
    \includegraphics[width=\linewidth]{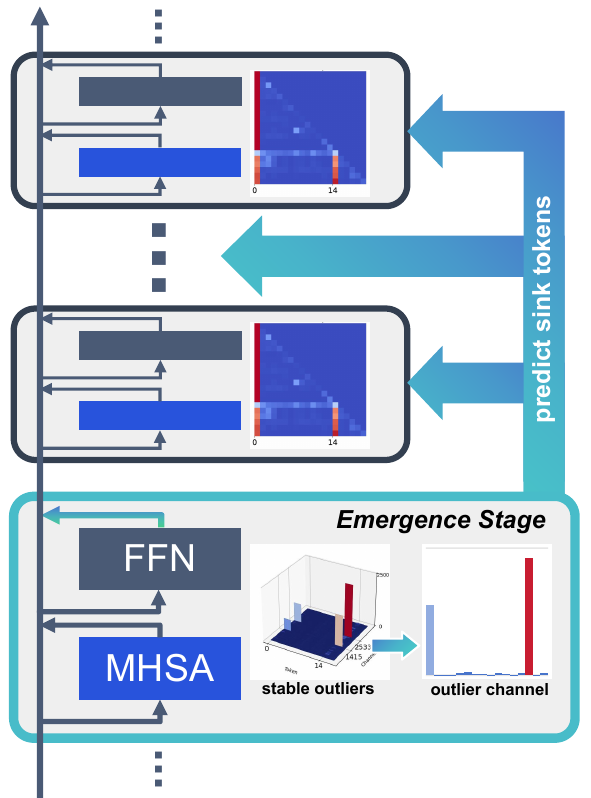}  
    \vspace{-7mm}
    \caption{Overview of KVSink.}
    \label{fig:kvsink}
    \vspace{-3mm}
\end{wrapfigure}
\label{KVSink}
Given the profound interplay between attention sinks and KV cache quantization, it is crucial to implement effective preservation mechanisms during quantization.
Existing approaches \citep{hooper2025kvquant,duanmu2024skvq} statically preserve the first few tokens (PFN), overlooking the potential presence of attention sinks at other positions.
Moreover, relying on attention scores to identify attention sinks is not a practical solution. 
First, dynamically identifying sink tokens incurs significant overhead.
Second, attention computations rely on optimized CUDA kernels, such as FlashAttention \citep{dao2022flashattention,dao2023flashattention}, which do not expose intermediate results.
To address this challenge, we propose \textit{\textbf{KVSink}}, with an overview illustrated in Figure \ref{fig:kvsink}.
As discussed in Section \ref{section3}, both stable outliers and attention sinks manifest on sink tokens. 
Building on this, KVSink initially identifies the emergence of stable outliers and subsequently uses them as indicators to predict the positions of sink tokens.
Notably, identification of outliers is both highly efficient and accurate.
\textit{\textbf{(1)}} Outliers exhibit extremely large magnitudes and are sparse in occurrence, which enables their efficient detection through a straightforward top-k sorting approach.
\textit{\textbf{(2)}} The identification needs to be performed only once during inference at the emergence stage. 
Additionally, the emergence stage layer can be pre-identified and treated as a static pattern, as it is input-independent (see Appendix \ref{cross_layer}).
\textit{\textbf{(3)}} For a given LLM, outliers consistently emerge in specific fixed channels \citep{sun2024massive,an2025systematic}. 
This allows identification to be restricted to a single pre-identified channel.
\textit{\textbf{(4)}} Since the initial input sequence is typically sufficiently long to capture all attention sinks, performing outlier identification only during the prefill phase offers a more computationally efficient approach.
The complete algorithm for KVSink is presented in Appendix \ref{Algorithm}.
\vspace{-2mm}
\section{Experiments}
\vspace{-2mm}
\subsection{Experiment Settings}
\vspace{-3mm}
To assess the benefits of KVSink, we first compare it with the existing Preserve-First-N (PFN) solution across various LLMs and KV cache quantization schemes.
Additionally, to further highlight the improvements KVSink provides over well-established methods, we conduct experiments using the KVQuant method \citep{hooper2025kvquant}. 
The efficiency analysis of KVSink is presented in Appendix \ref{Efficiency Analysis}.

\vspace{-1mm}
\textbf{Models, tasks, and datasets.}
We evaluate KVSink across seven models: LLaMA2-7B/13B/70B, LLaMA2-7B-chat, Mistral-7B, LLaMA3-8B and LLaMA3.1-8B-instruct \citep{touvron2023llama,chaplot2023albert,dubey2024llama}. 
The evaluation includes PPL tests conducted on the WikiText-2 and C4 datasets \citep{merity2016wikitext, 2020t5}.

\vspace{-1mm}
\textbf{Comparison of KVSink with the PFN strategy.} 
We compare the PPL reduction achieved by KVSink and PFN using the same number of tokens for preservation, with values set to 0, 5, 10, 15, and 20.
The evaluation is conducted based on three basic quantization schemes using RTN INT2/INT4 quantization: per-token Key and Value static, per-token Key and Value dynamic, and per-channel Key with per-token Value static.
The quantization group size is uniformly set to 128, and the Wikitext-2 dataset is used for the PPL test. 

\vspace{-1mm}
\textbf{Application of KVSink to the KVQuant method.}
KVQuant \citep{hooper2025kvquant} employs a range of advanced techniques for ultra-low-bit KV cache quantization, delivering state-of-the-art performance, including non-uniform quantization, per-vector dense-and-sparse quantization for isolating numerical outliers, and per-channel quantization for Keys along with per-token quantization for Values.
However, to preserve attention sinks, KVQuant only excludes the first token during calibration and quantization, a limitation that KVSink aims to enhance.
We apply KVSink to the 2-bit KVQuant method with various settings for numerical outliers, including 1\%, 0.5\%, 0.1\%, and no isolation of numerical outliers. 
The evaluation is performed using PPL tests on the Wikitext-2 and C4 datasets.

\begin{figure}[t]
\vspace{-4mm}
    \centering    
    \begin{subfigure}{0.16\textwidth}
        \centering
    \includegraphics[width=\linewidth]{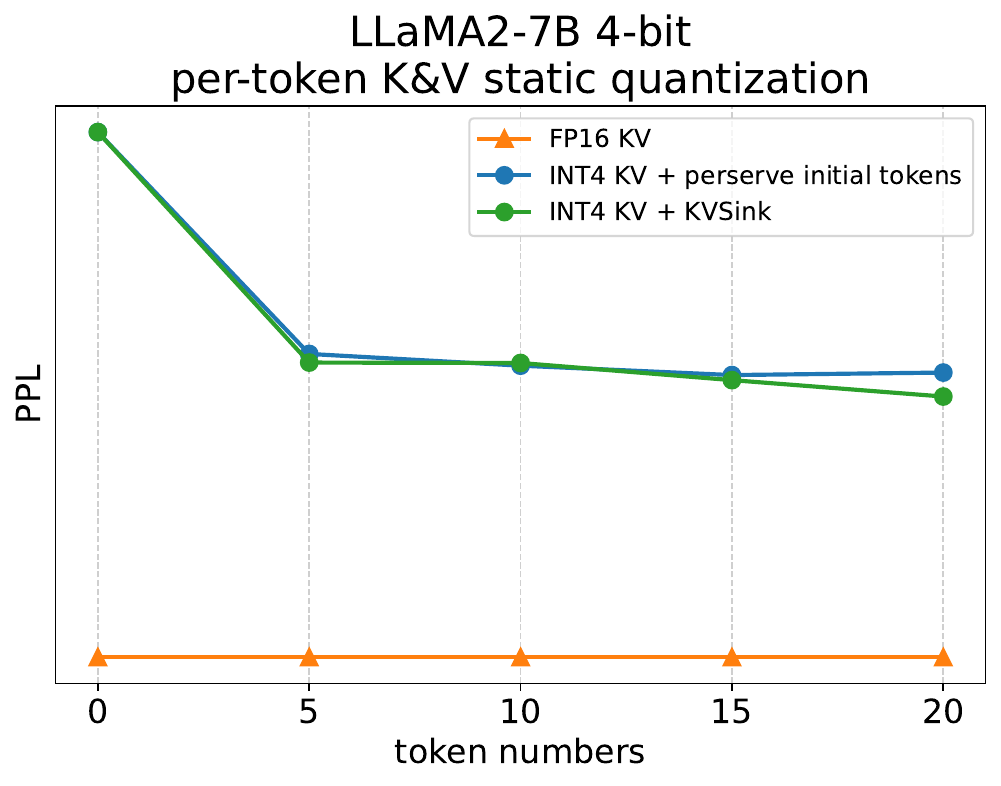}
    \end{subfigure}
    \begin{subfigure}{0.16\textwidth}
        \centering
    \includegraphics[width=\linewidth]{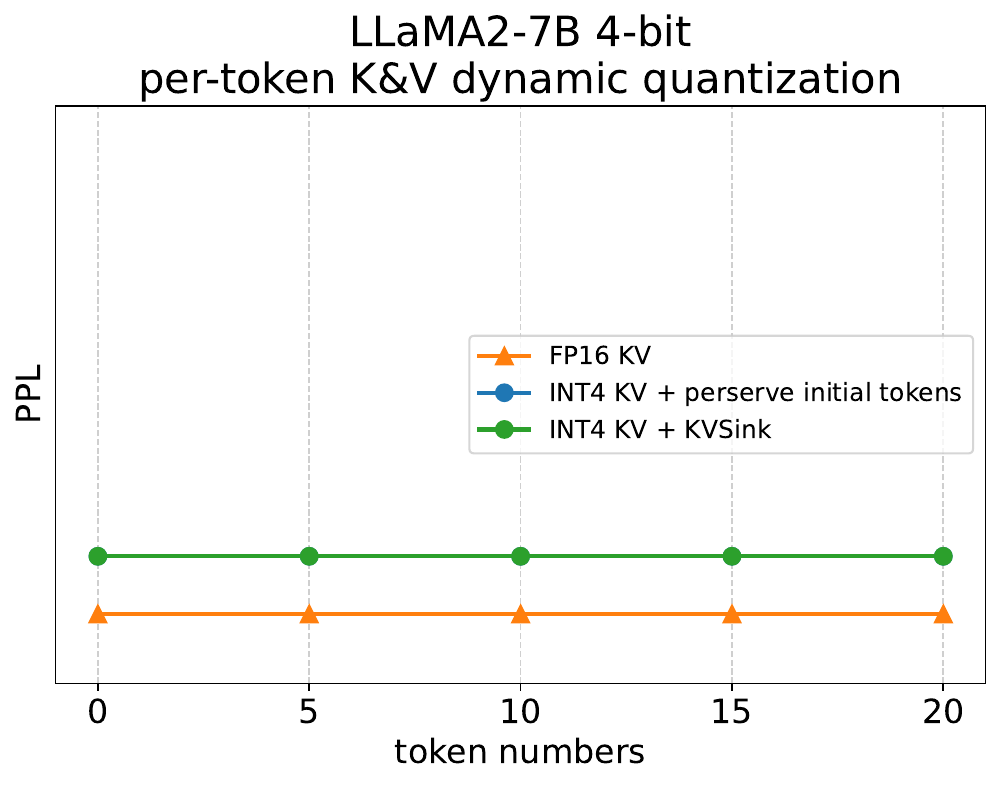}
    \end{subfigure}
    \begin{subfigure}{0.16\textwidth}
        \centering
    \includegraphics[width=\linewidth]{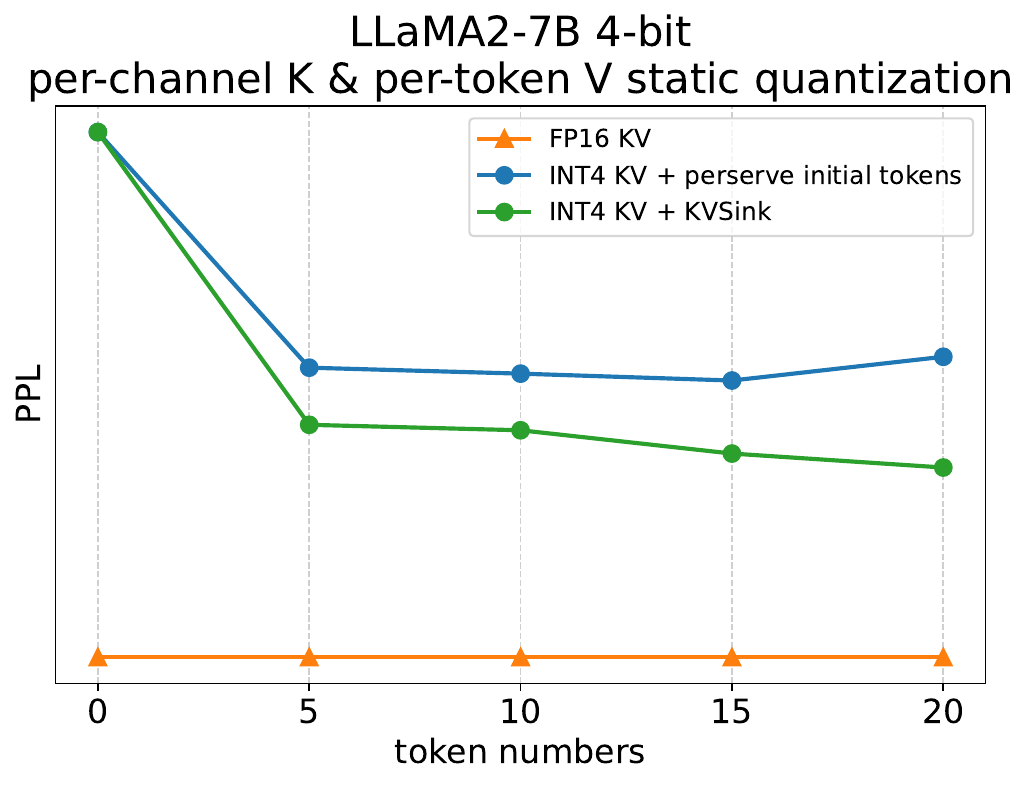}
    \end{subfigure}
    \begin{subfigure}{0.16\textwidth}
        \centering
    \includegraphics[width=\linewidth]{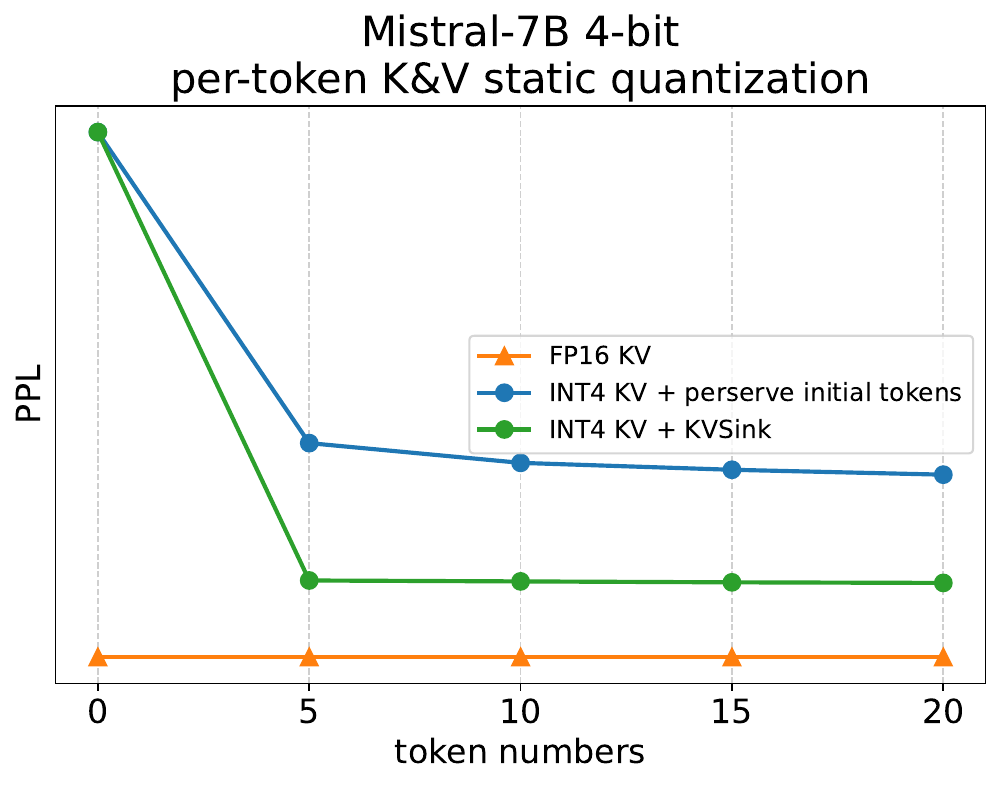}
    \end{subfigure}
    \begin{subfigure}{0.16\textwidth}
        \centering
    \includegraphics[width=\linewidth]{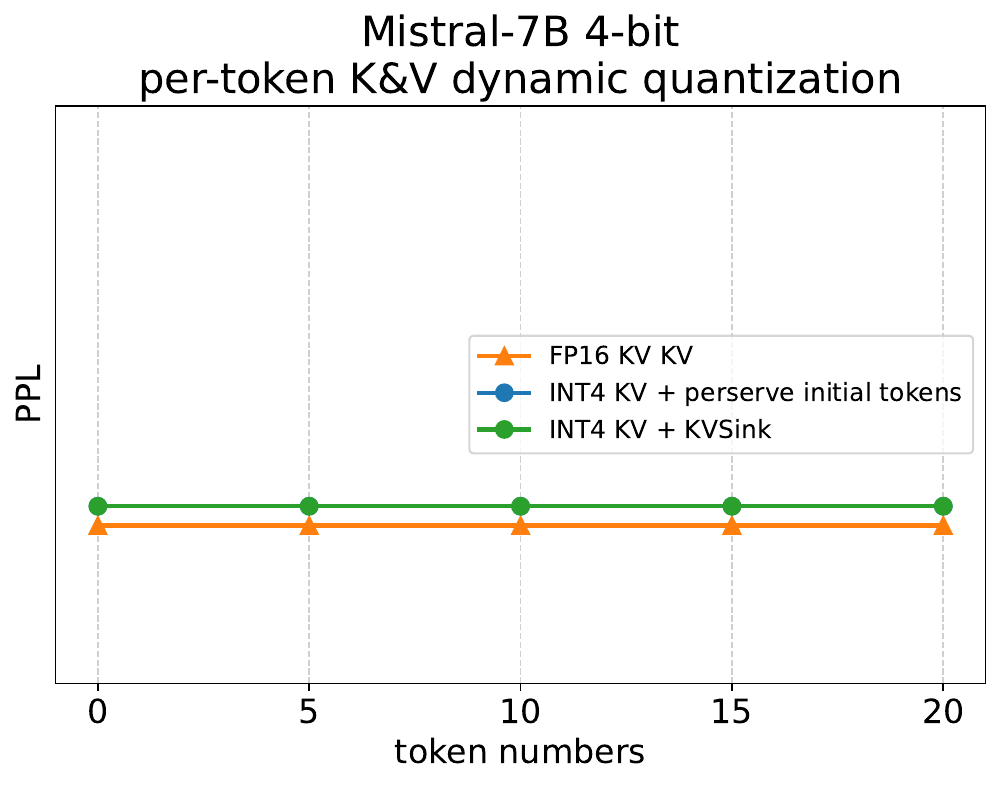}
    \end{subfigure}
    \begin{subfigure}{0.16\textwidth}
        \centering
    \includegraphics[width=\linewidth]{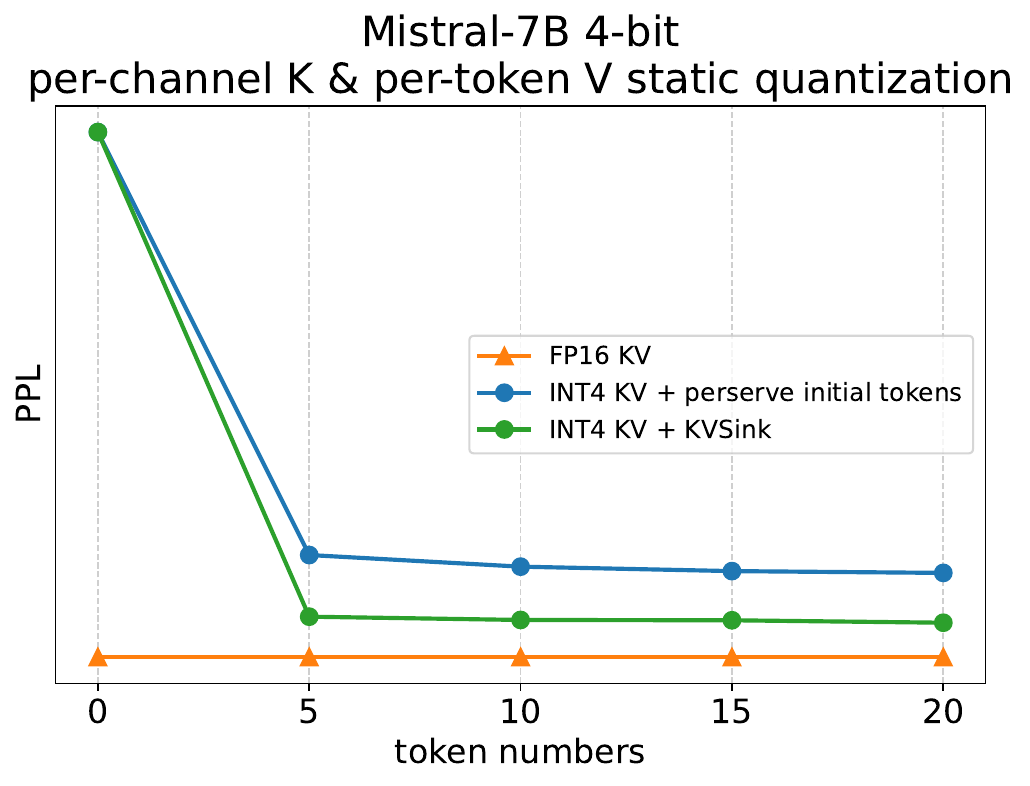}
    \end{subfigure}
    \begin{subfigure}{0.16\textwidth}
        \centering
    \includegraphics[width=\linewidth]{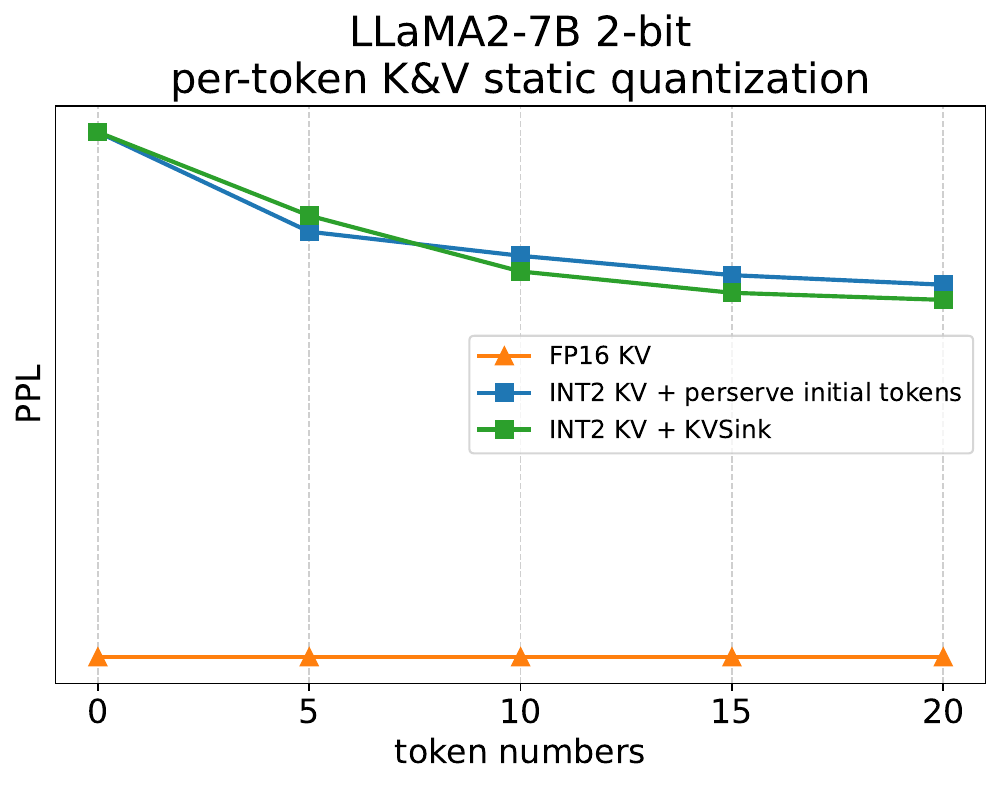}
    \end{subfigure}
    \begin{subfigure}{0.16\textwidth}
        \centering
    \includegraphics[width=\linewidth]{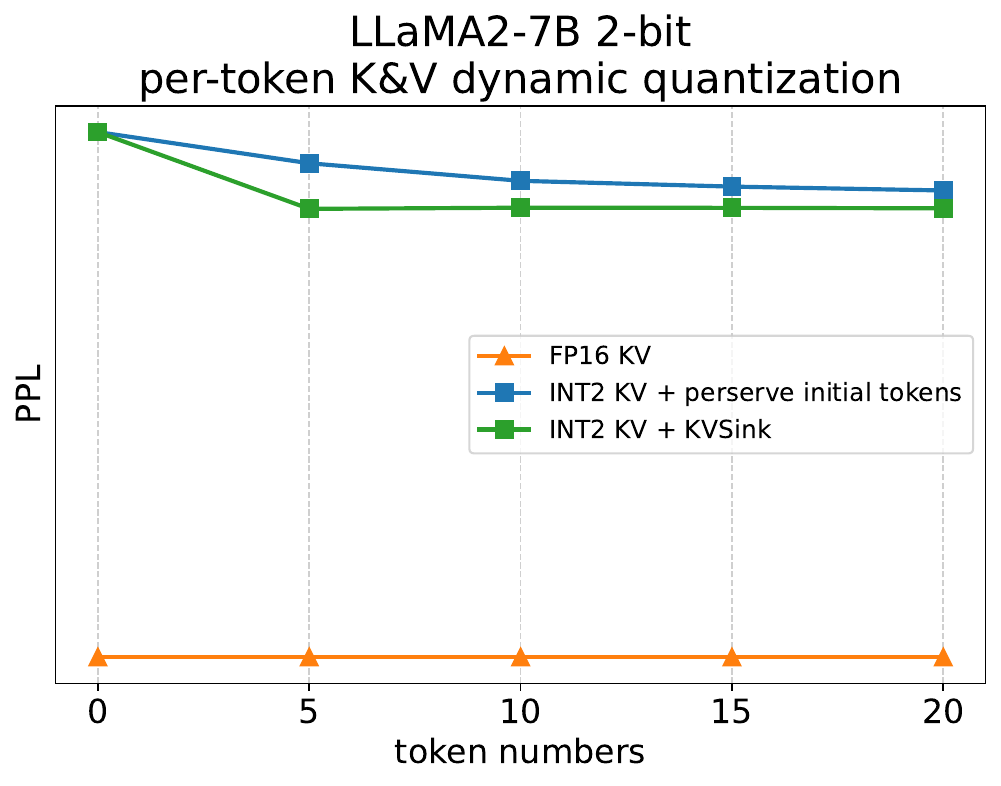}
    \end{subfigure}
    \begin{subfigure}{0.16\textwidth}
        \centering
    \includegraphics[width=\linewidth]{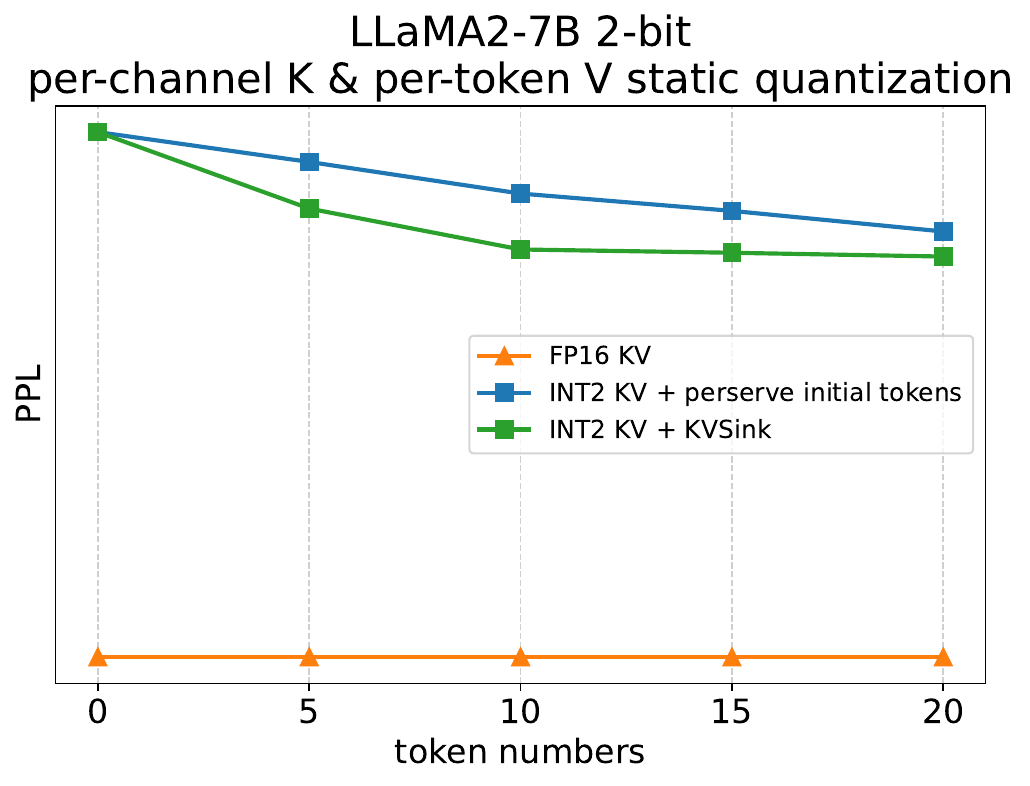}
    \end{subfigure}
    \begin{subfigure}{0.16\textwidth}
        \centering
    \includegraphics[width=\linewidth]{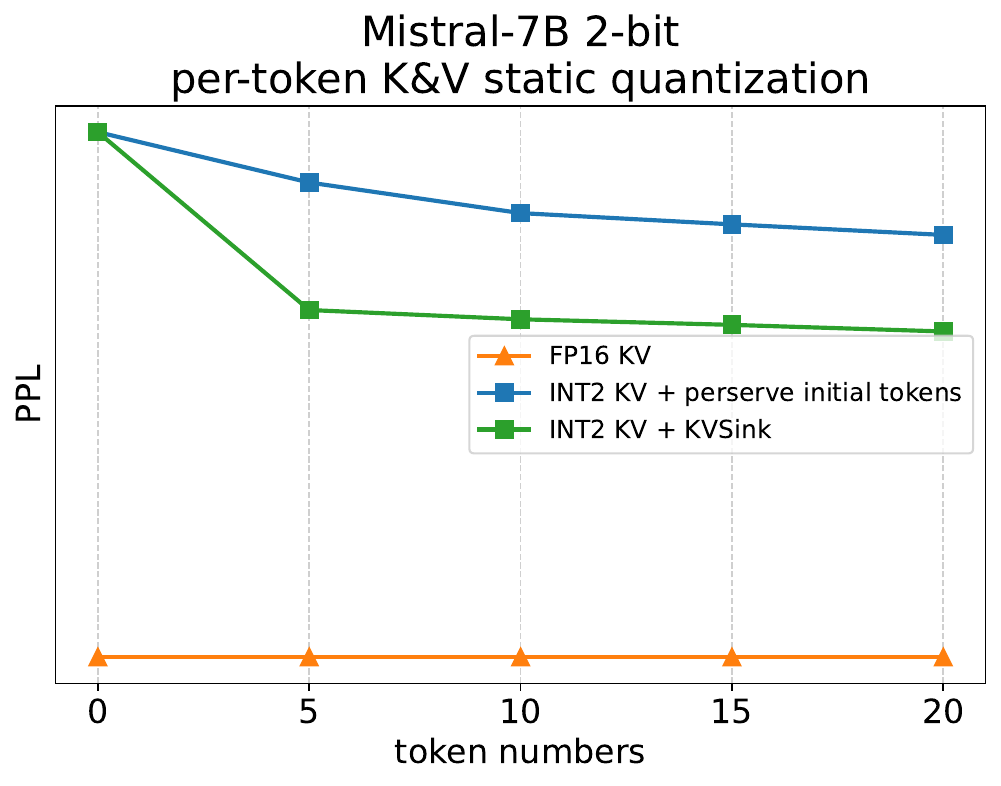}
    \end{subfigure}
    \begin{subfigure}{0.16\textwidth}
        \centering
    \includegraphics[width=\linewidth]{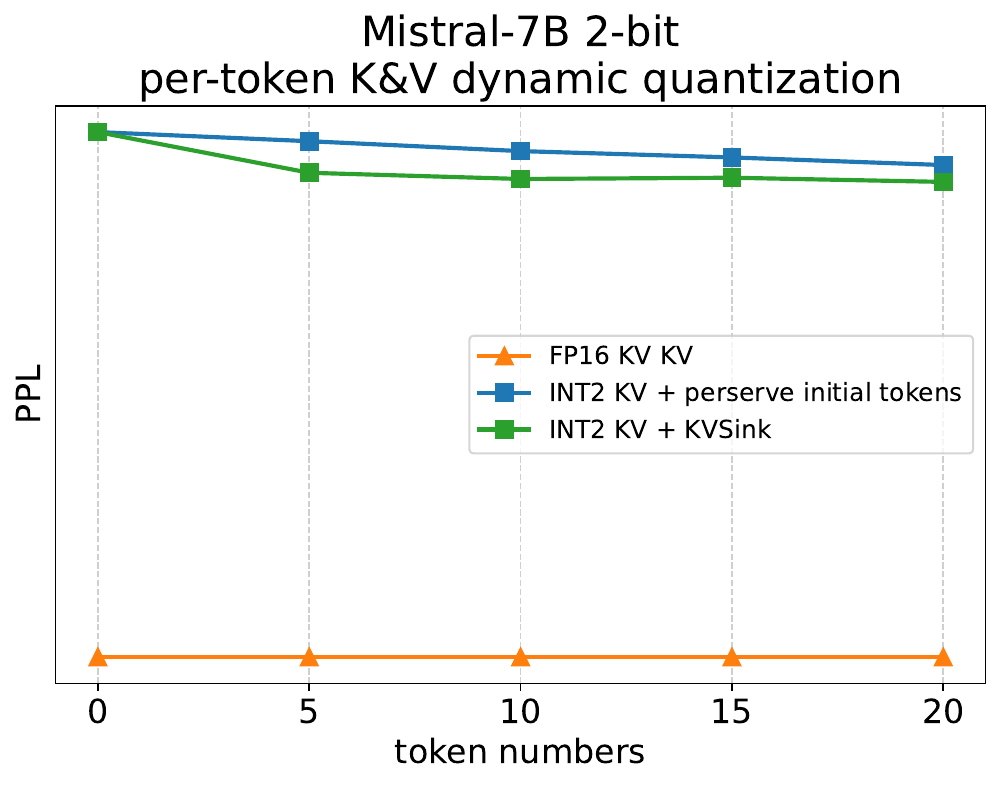}
    \end{subfigure}
    \begin{subfigure}{0.16\textwidth}
        \centering
    \includegraphics[width=\linewidth]{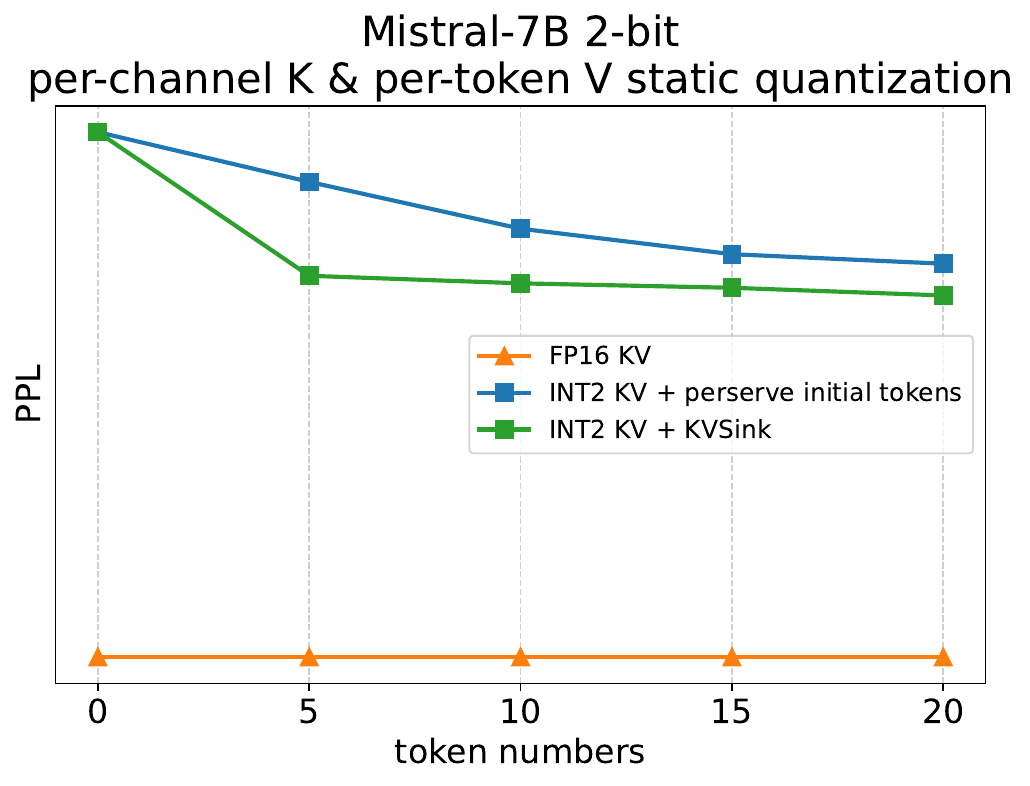}
    \end{subfigure}
    \begin{subfigure}{0.16\textwidth}
        \centering
    \includegraphics[width=\linewidth]{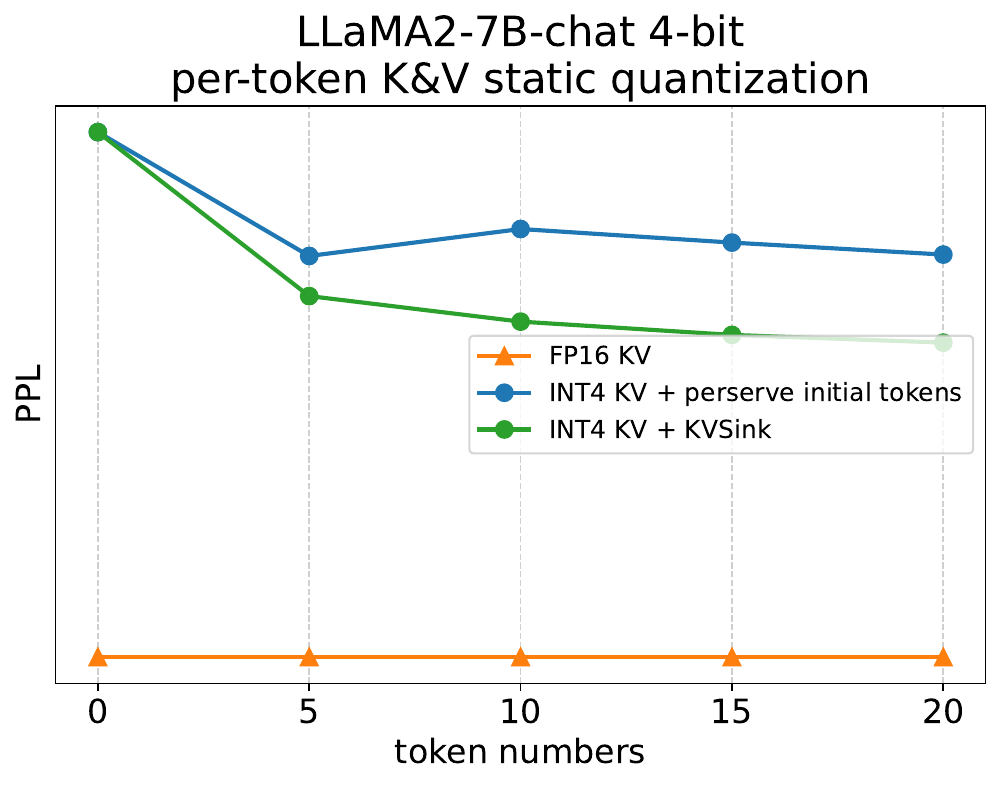}
    \end{subfigure}
    \begin{subfigure}{0.16\textwidth}
        \centering
    \includegraphics[width=\linewidth]{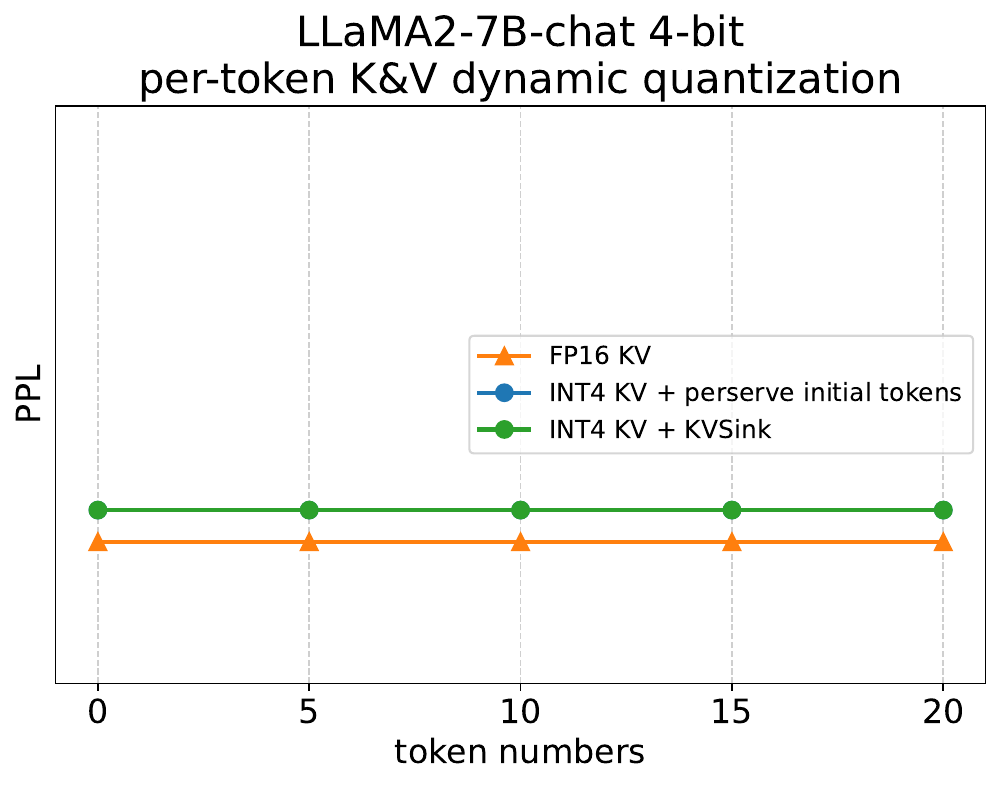}
    \end{subfigure}
    \begin{subfigure}{0.16\textwidth}
        \centering
    \includegraphics[width=\linewidth]{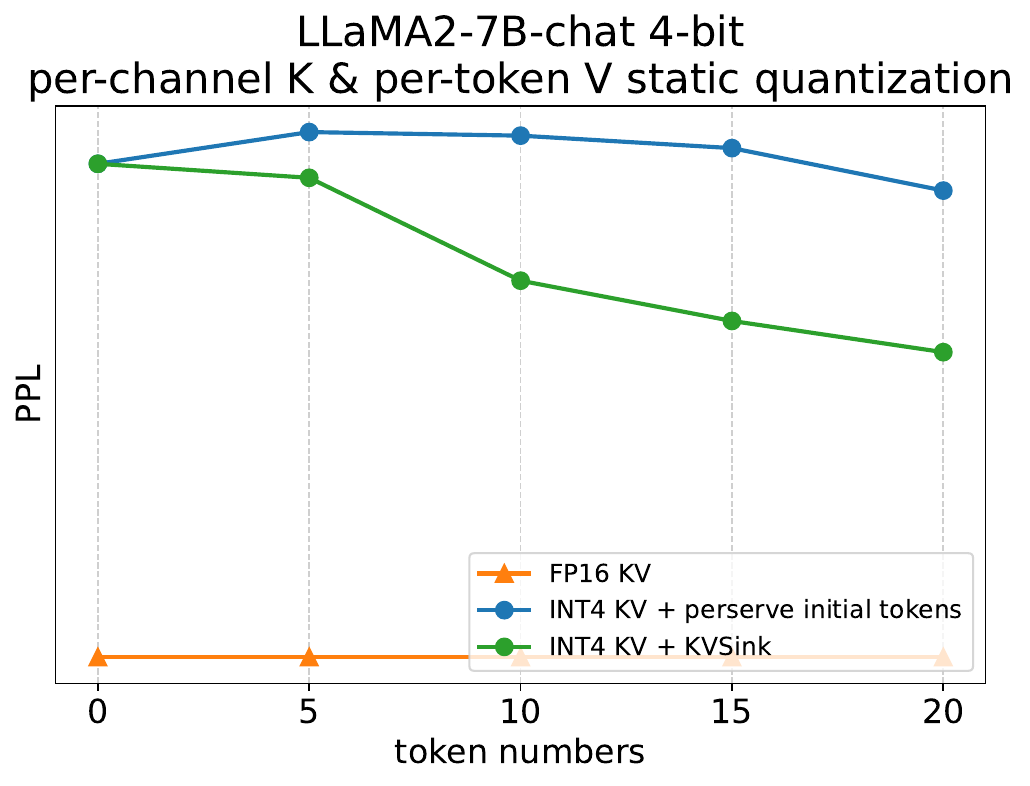}
    \end{subfigure}
    \begin{subfigure}{0.16\textwidth}
        \centering
    \includegraphics[width=\linewidth]{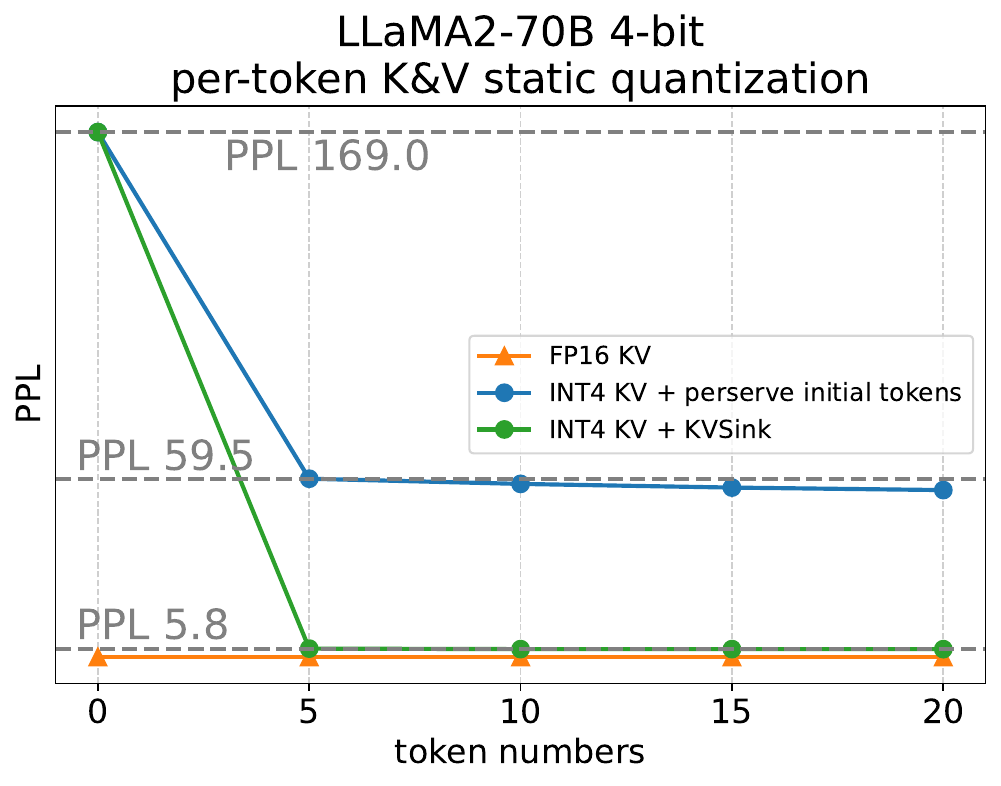}
    \end{subfigure}
    \begin{subfigure}{0.16\textwidth}
        \centering
    \includegraphics[width=\linewidth]{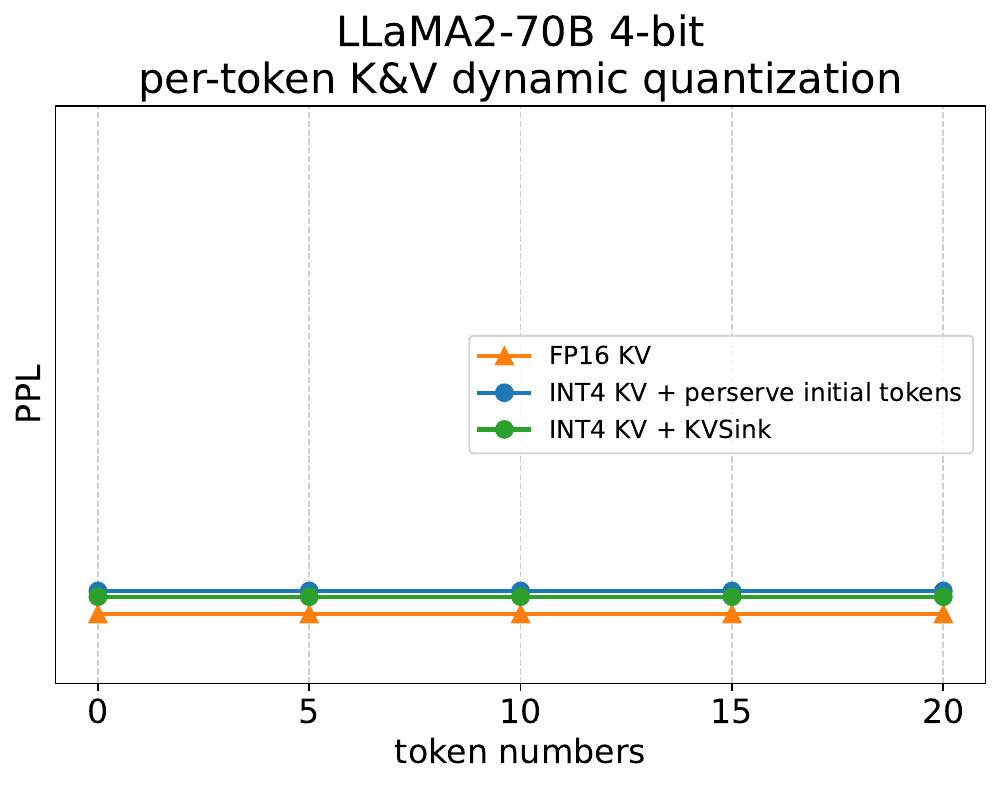}
    \end{subfigure}
    \begin{subfigure}{0.16\textwidth}
        \centering
    \includegraphics[width=\linewidth]{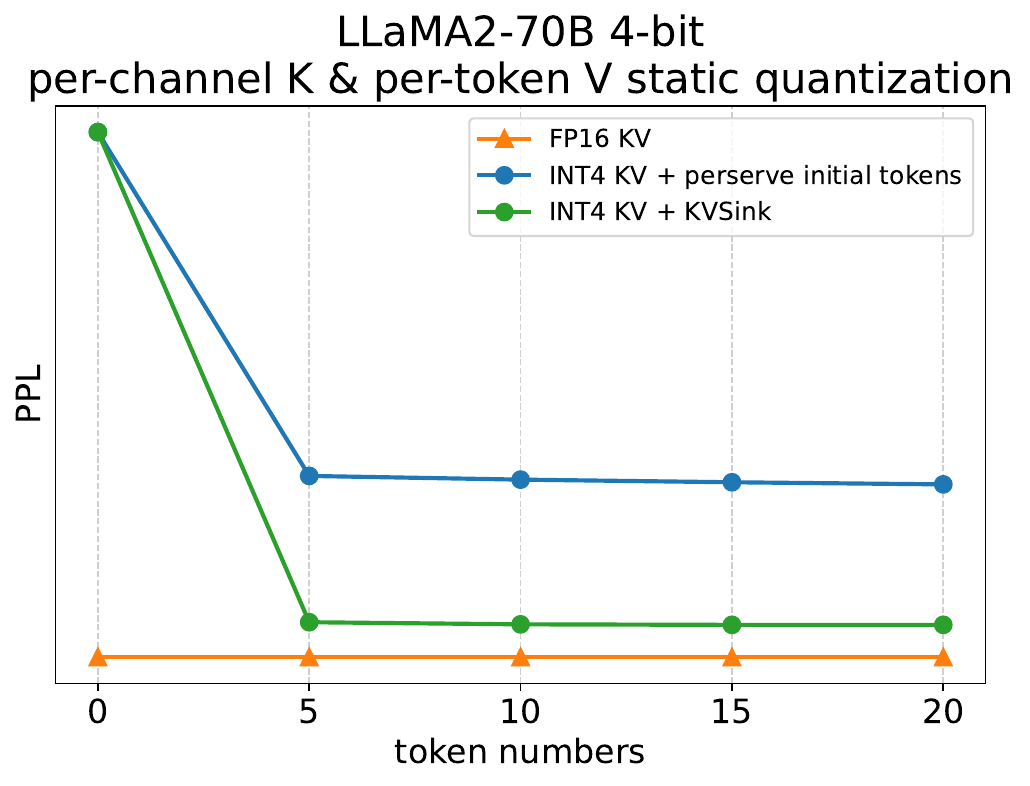}
    \end{subfigure}    
    \begin{subfigure}{0.16\textwidth}
        \centering
    \includegraphics[width=\linewidth]{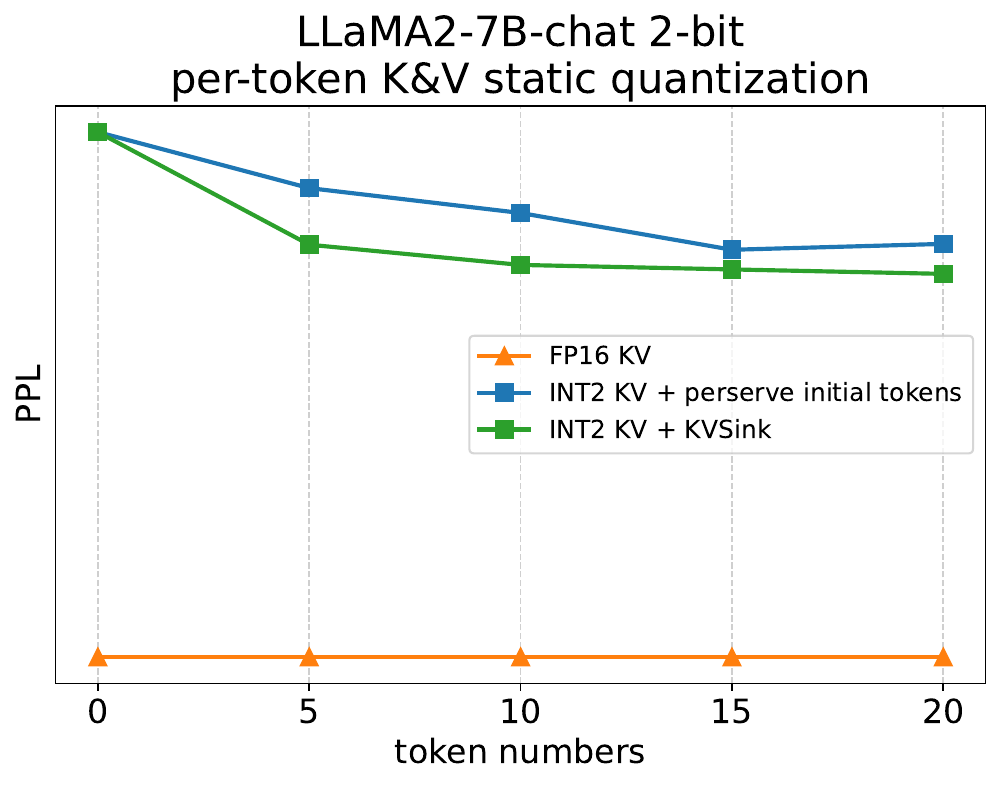}
    \end{subfigure}
    \begin{subfigure}{0.16\textwidth}
        \centering
    \includegraphics[width=\linewidth]{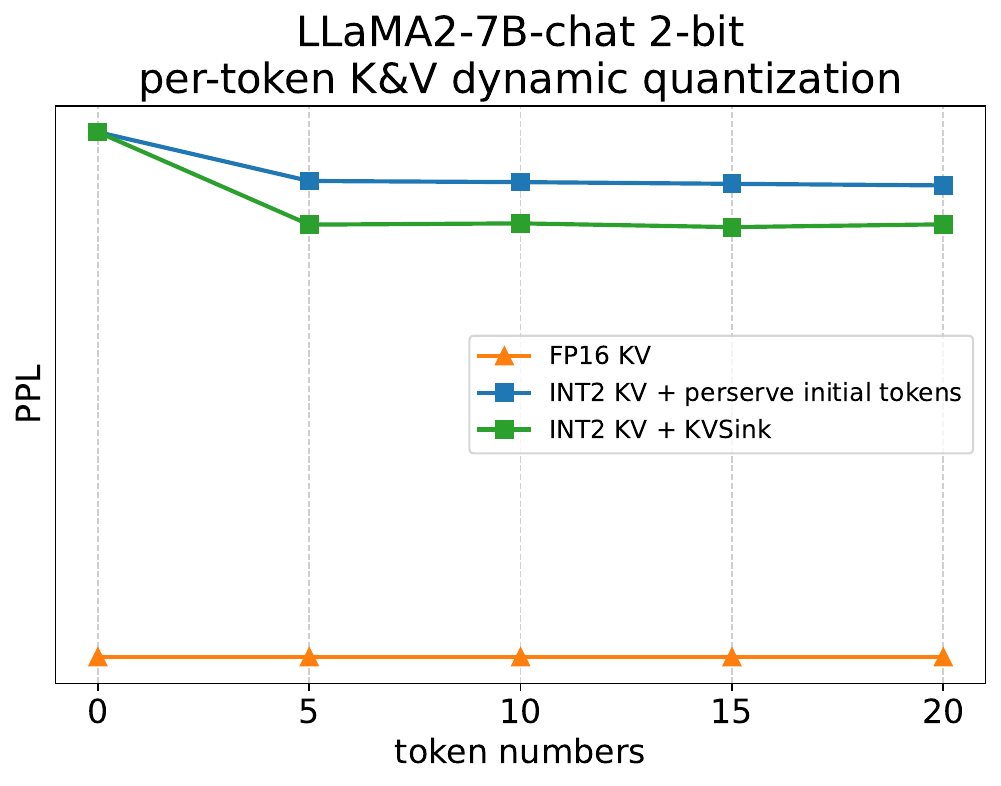}
    \end{subfigure}
    \begin{subfigure}{0.16\textwidth}
        \centering
    \includegraphics[width=\linewidth]{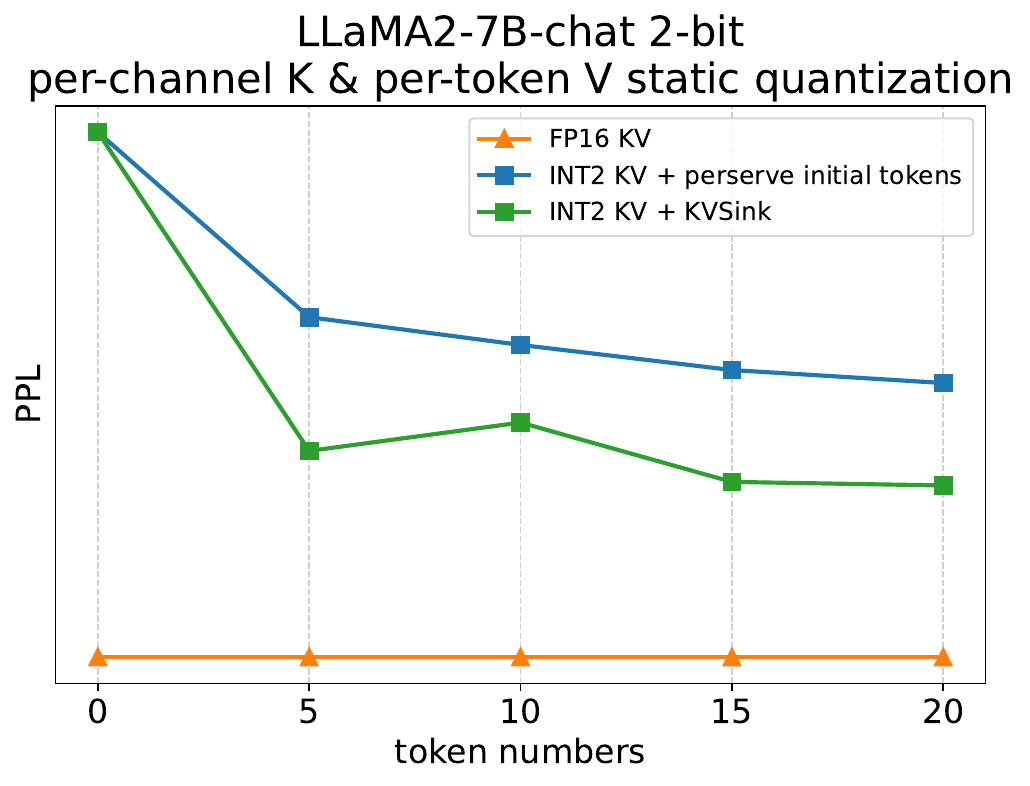}
    \end{subfigure}
    \begin{subfigure}{0.16\textwidth}
        \centering
    \includegraphics[width=\linewidth]{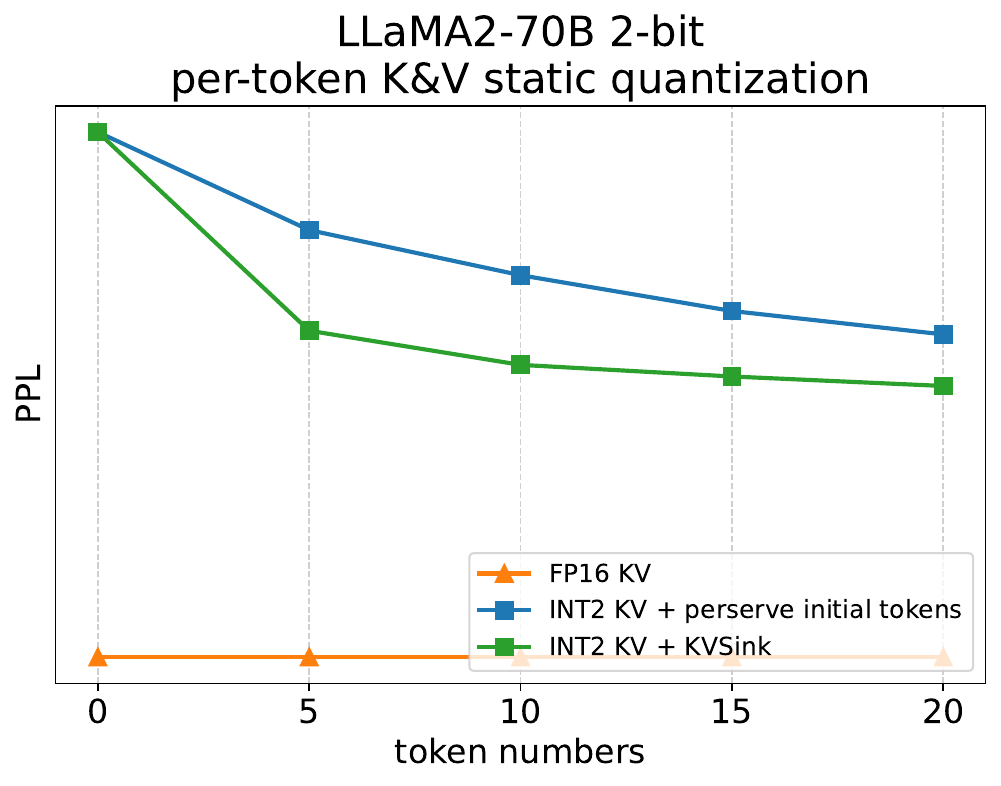}
    \end{subfigure}
    \begin{subfigure}{0.16\textwidth}
        \centering
    \includegraphics[width=\linewidth]{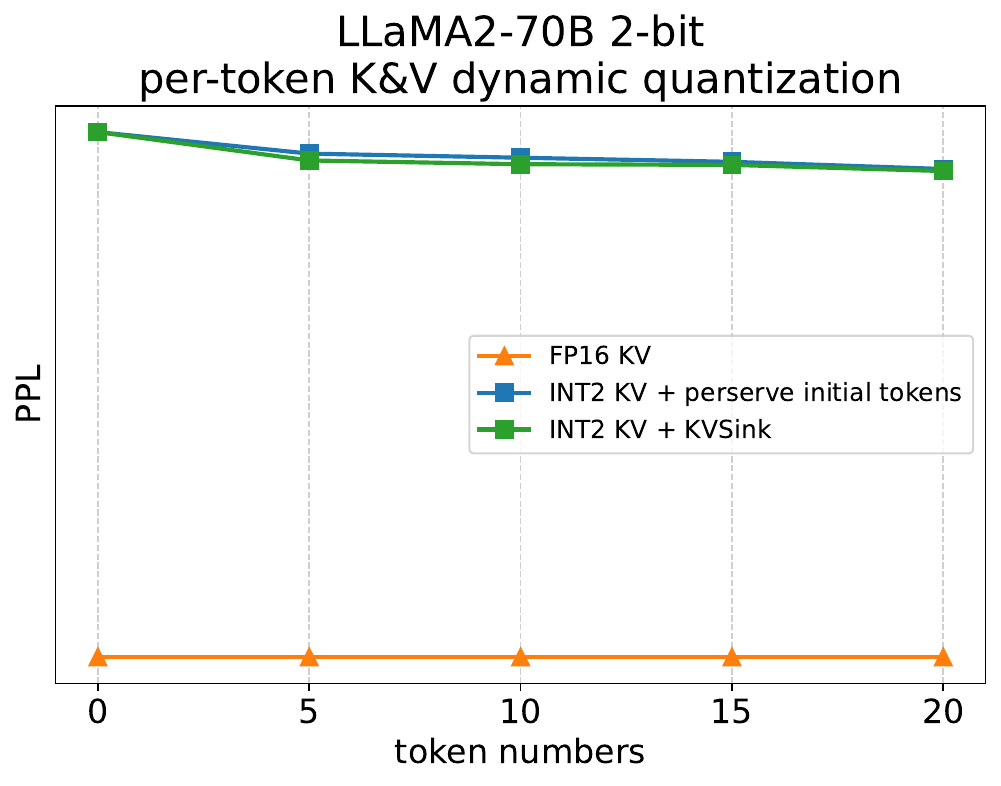}
    \end{subfigure}
    \begin{subfigure}{0.16\textwidth}
        \centering
    \includegraphics[width=\linewidth]{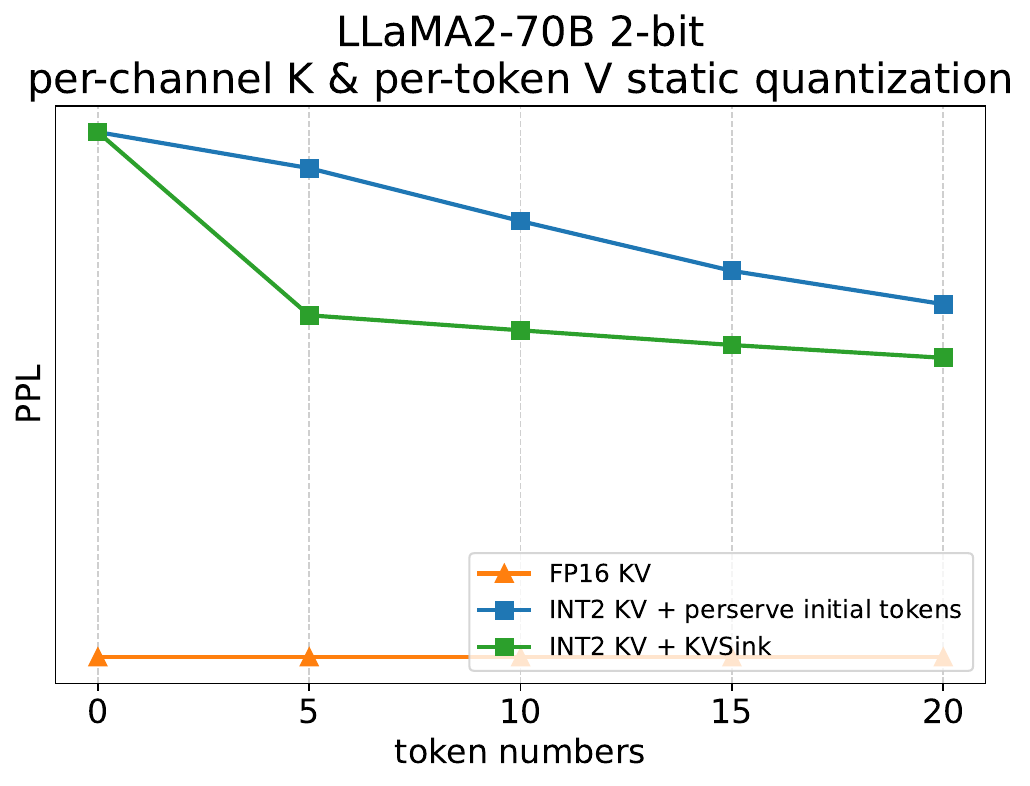}
    \end{subfigure}  
    \vspace{-3mm}
    \caption{Comparison of KVSink with the PFN strategy.
    The orange line denotes the PPL of the FP16 KV cache, green indicates the use of KVSink, and the blue line represents the PFN method. 
    A PPL closer to that of FP16 reflects better performance.}
\label{RTNPPL}
\vspace{-2mm}
\end{figure}
\begin{table}[t]
\centering
\resizebox{0.96\columnwidth}{!}{%
\begin{tabular}{@{}l|cccc|cccc|c@{}}
\toprule
\multicolumn{1}{c|}{\multirow{2}{*}{Methods}} & \multicolumn{4}{c|}{WikiText-2 PPL $\downarrow$} & \multicolumn{4}{c|}{C4 PPL $\downarrow$} & \multirow{2}{*}{\begin{tabular}[c]{@{}c@{}}KV\\ Avg. Bits\end{tabular}} \\ \cmidrule(lr){2-9}
\multicolumn{1}{c|}{} & \begin{tabular}[c]{@{}c@{}}LLaMA2\\ 7B\end{tabular} & \begin{tabular}[c]{@{}c@{}}LLaMA2\\ 13B\end{tabular} & \begin{tabular}[c]{@{}c@{}}LLaMA3\\ 8B\end{tabular} & \begin{tabular}[c]{@{}c@{}}LLaMA3.1\\ 8B-Instruct\end{tabular} & \begin{tabular}[c]{@{}c@{}}LLaMA2\\ 7B\end{tabular} & \begin{tabular}[c]{@{}c@{}}LLaMA2\\ 13B\end{tabular} & \begin{tabular}[c]{@{}c@{}}LLaMA3\\ 8B\end{tabular} & \begin{tabular}[c]{@{}c@{}}LLaMA3.1\\ 8B-Instruct\end{tabular} &  \\ \midrule
\multicolumn{1}{c|}{16-bit} & 5.12 & 4.57 & 5.75 & 6.75 & 6.63 & 6.05 & 7.37 & 8.03 & 16 \\ \midrule
KVQuant & 7.46 & 16.26 & 7.31 & 9.75 & 9.46 & 26.29 & 8.53 & 10.43 & \multirow{2}{*}{2-2.02} \\
\multicolumn{1}{r|}{\textit{\textbf{+ KVSink-5}}} & 6.52 & 5.10 & 6.77 & 8.83 & 7.93 & 6.47 & 7.74 & 9.57 &  \\ \midrule
KVQuant-0.1\% & 5.73 & 5.00 & 6.84 & 8.80 & 7.11 & 6.39 & 8.06 & 9.25 & \multirow{2}{*}{2.04-2.06} \\
\multicolumn{1}{r|}{\textit{\textbf{+ KVSink-5}}} & 5.60 & 4.85 & 6.56 & 8.25 & 7.10 & 6.28 & 7.87 & 8.80 &  \\ \midrule
KVQuant-0.5\% & 5.62 & 4.86 & 6.51 & 7.91 & 7.11 & 6.24 & 7.87 & 8.71 & \multirow{2}{*}{2.16-2.19} \\
\multicolumn{1}{r|}{\textit{\textbf{+ KVSink-5}}} & 5.54 & 4.76 & 6.34 & 7.69 & 6.95 & 6.22 & 7.81 & 8.65 &  \\ \midrule
KVQuant-1\% & 5.53 & 4.80 & 6.37 & 7.58 & 6.94 & 6.23 & 7.82 & 8.53 & \multirow{2}{*}{2.32-2.35} \\
\multicolumn{1}{r|}{\textit{\textbf{+ KVSink-5}}} & 5.44 & 4.71 & 6.22 & 7.47 & 6.81 & 6.19 & 7.78 & 8.50 &  \\ \bottomrule
\end{tabular}%
}
\vspace{-2mm}
\caption{Application to the KVQuant method.
KVSink-5 denotes the use of 5 sink tokens.}
\label{kvquant}
\vspace{-5mm}
\end{table}
\vspace{-2mm}
\subsection{Main Results}
\vspace{-2mm}
\textbf{Comparison with the PFN strategy.} 
As shown in Figure \ref{RTNPPL}, KVSink outperforms PFN in almost all cases. 
For instance, when evaluating LLaMA2-70B with per-token KV static 4-bit quantization, preserving only 5 sink tokens with KVSink results in a PPL reduction of 163.2, with only a marginal increase of 2.5 in PPL compared to the FP16 baseline. 
In contrast, the PFN strategy exhibits the PPL of 59.5, as it fails to account for sink tokens at other positions.
Additionally, the experiments also show that preserving only 5 sink tokens with KVSink is sufficient in most cases.

\textbf{Application to the KVQuant method.} 
As shown in Table \ref{kvquant}, integrating KVSink with the well-established KVQuant method provides two key benefits.
First, while KVSink introduces only minimal modifications, it consistently improves PPL, with the benefits becoming more pronounced as fewer numerical outliers are preserved.
Second, KVSink reduces the dependence on preserving FP16 numerical outliers. 
For example, with the 0.1\% setting, the performance with KVSink is maintained or even exceeds that of the 0.5\%. 
Notably, reducing the preservation of FP16 outliers leads to more efficient compression.
\vspace{-2mm}
\section{Conclusion}
\vspace{-2mm}
In this study, we enhance the understanding of attention sink preservation in KV cache quantization and propose KVSink to improve the existing PFN solution. 
We elucidate the intrinsic relationships in the cross-layer evolution of different types of extreme activation outliers and highlight the pivotal role of attention sinks during the stabilization stage of stable outliers. 
We also thoroughly analyze the mutual influence between attention sinks and KV cache quantization. 
Experimental results demonstrate that KVSink outperforms the existing PFN strategy and improves established KV cache quantization techniques.

\newpage
\appendix
\section{Related Work on Attention Sinks and Extreme Activation Outliers in Transformer-Based Models}
\label{related_work_1}
Previous studies have revealed the widespread presence of extreme activation outliers in Transformer-based models \citep{bondarenko2021understanding}, including BERT \citep{devlin2019bert,kovaleva2019revealing,clark2019does}, Vision Transformer (ViT) \citep{dosovitskiy2020image,bondarenko2023quantizable,sun2024massive,darcet2023vision}, and LLM \citep{sun2024massive,guo2024active,yang2024mitigating}, with substantial attention concentrating on these outliers, forming attention sinks.

Numerous studies have sought to elucidate this behavior in Transformer models.
In pre-LLM research, \cite{bondarenko2021understanding}, as a pioneering study, identified the bottleneck in activation quantization of Transformers caused by extreme outliers and uncovered the intrinsic relationship between attention focus pattern and these outliers.
\cite{clark2019does} demonstrates that BERT-like Transformers tend to focus on the special [SEP] token. 
This behavior effectively acts as a "no-op" for attention heads that are unable to extract the patterns they were trained to detect from the specific passage of text.
\cite{bondarenko2023quantizable} found that outliers and attention focus arise as attention heads attempt to learn a “no-op” or a partial update of the residual. 
In this process, strong outliers emerge due to the softmax function.

StreamLLM \citep{xiao2023efficient} has conducted an initial investigation revealing the presence of attention sinks in LLMs. 
It suggests that LLMs tend to treat initial token as attention sink because the model tends to dump unnecessary attention values to specific tokens.
\cite{sun2024massive} conducted an in-depth study on massive activations in LLMs and ViTs, demonstrating that these activations give rise to attention sinks and lead to implicit attention biases.
\cite{gu2024attention} found that attention sinks function more like Key biases, storing extra attention scores that may be non-informative and not contribute to the Value computation. 
\cite{an2025systematic} categorize three types of outliers—activation outliers, weight outliers, and attention outliers—revealing their intrinsic connections and collective impact on the attention mechanism.

Building on these works, we elucidate the role of attention sinks by analyzing the cross-layer evolution of different types of extreme activation outliers, offering a novel perspective that has not been explored in prior research. 
Furthermore, our study delves into their interaction with KV cache quantization, offering valuable insights for future investigations.
\section{Related Work on KV Cache Quantization}
Low-bit quantization reduces the bit-width of the KV cache representation, effectively decreasing its size and thereby mitigating memory usage and access bottlenecks.
However, this process inevitably introduces quantization errors, leading to performance degradation.
Existing methods explore various approaches to mitigate the impact of quantization errors in KV cache quantization.
These methods can generally be classified into two main types based on the selected quantization dimension of the Keys.

KVQuant \citep{hooper2025kvquant} and KIVI \citep{liu2024kivi} both observe that Keys exhibit outliers in specific channels, while Values do not. 
Based on this observation, both methods utilize per-channel quantization for Keys to reduce the quantization difficulty.
KVQuant employs non-uniform static quantization and incorporates several optimization techniques to address the challenges associated with static quantization. 
These techniques include pre-RoPE quantization, per-vector dense-and-sparse quantization, and attention-sink-aware quantization, among others.
KIVI employs dynamic integer quantization, where KV cache quantization is applied after accumulating a specified number of local tokens.

On the other hand, per-token KV cache quantization often uses token-level mixed-precision quantization to preserve the precision of the KVs of critical tokens, thereby minimizing the loss of essential information.
SKVQ \citep{duanmu2024skvq}, RotateKV \citep{su2025rotatekv}, MiKV \citep{yang2024no} and ZipCache \citep{he2024zipcache} are all approaches that focus on per-token mixed-precision quantization. 
SKVQ introduces clipped dynamic quantization with channel reordering, preserving high precision for both the initial and most recent tokens.
RotateKV utilizes outlier-aware Hadamard-transform-based rotation to reduce the quantization difficulty of the Keys, while preserving high precision for KV pairs associated with attention sink tokens.
MiKV assesses token importance using accumulated attention scores, similar to the approach used in H2O \citep{zhang2023h2o}, and subsequently employs relatively higher bit-widths to preserve the KVs of important tokens.
ZipCache utilizes normalized attention scores to more accurately identify salient tokens and incorporates an efficient approximation of the saliency metric.

Although most of these methods preserve higher precision for the KVs of sink tokens or tokens with high attention scores, they lack a clear explanation of the underlying principles.
In contrast, our work offers a deeper understanding of the interaction between attention sinks and KV quantization, while also enhancing the existing Preserve-First-N solution.
\label{related_2}
\section{Overview on Low-Bit Quantization }
\begin{figure}[t!]
    \vspace{-7mm}
    \centering    
    \begin{subfigure}{0.14\textwidth}
        \centering
    \includegraphics[width=1\linewidth]{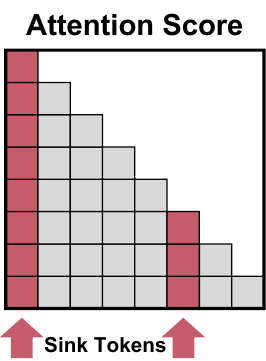}
    \caption{Sinks}
    \label{attn_sink_quant}
    \end{subfigure}
    \begin{subfigure}{0.27\textwidth}
        \centering
    \includegraphics[width=1\linewidth]{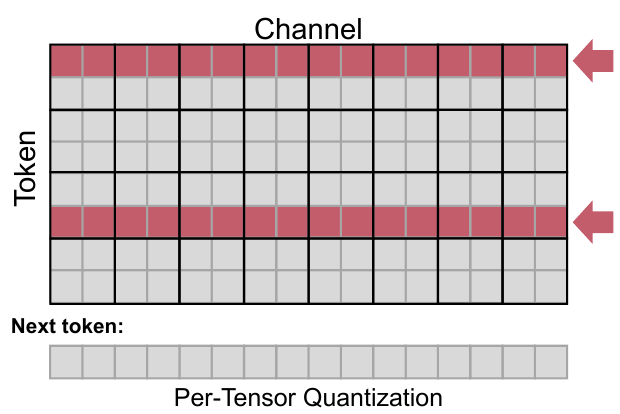}
    \caption{Per-tensor (block)}
    \label{per-tensor}
    \end{subfigure}
    \begin{subfigure}{0.27\textwidth}
        \centering
    \includegraphics[width=1\linewidth]{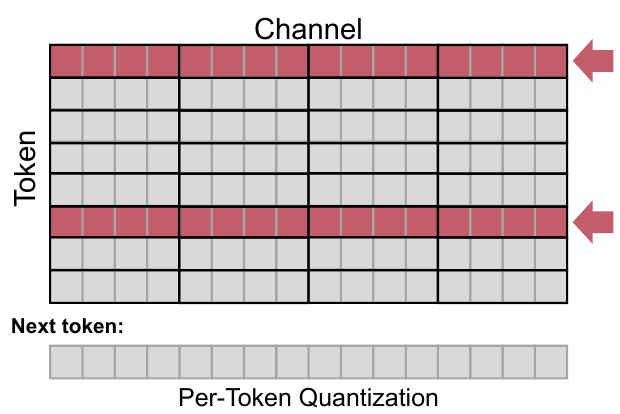}
    \caption{Per-token}
    \label{per-token}
    \end{subfigure}
    \begin{subfigure}{0.27\textwidth}
        \centering
    \includegraphics[width=1\linewidth]{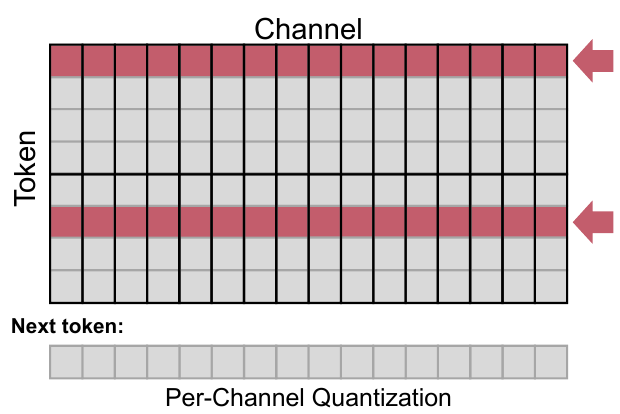}
    \caption{Per-channel}
    \label{per-channel}
    \end{subfigure}
    \vspace{-3mm}
    \caption{KV quantization based on different dimensionalities. 
    The tokens highlighted in red represent the Keys or Values of sink tokens.
    }
\label{quant_dim}
\end{figure}
\label{Preliminary on Low-Bit Quantization}
In this section, we introduce the quantization process using the commonly employed asymmetric integer quantization technique.
The \( n \)-bit asymmetric integer quantization and dequantization processes, where \( n \in \mathbb{N} \), can be expressed as:
\begin{equation}
Q(X) = clamp\left(\left\lfloor \frac{X}{{scale}} \right\rceil + zero, 0, 2^n - 1 \right),
\end{equation}
\begin{equation}
X' = scale \cdot (Q(X) - zero),
\end{equation}
\begin{equation}
    scale = \frac{clipped\_max(X) - clipped\_min(X)}{2^n - 1} ,
\end{equation}
\begin{equation}
    zero = - \left\lfloor \frac{clipped\_min(X)}{scale} \right\rceil,
\end{equation}
where $\left\lfloor\cdot\right\rceil$ indicates round operation. 
$Q(X)$ and $X'$ denote the quantized and dequantized values of $X$, respectively.
The $clamp$ operation ensures that the values are constrained within the specified range. 
The operations $clipped\_max(X)$ and $clipped\_min(X)$ denote the operations that truncate the maximum and minimum values of $X$. 

\textbf{Quantization dimension.} 
Quantization can be applied along various dimensions, with common approaches in KV cache quantization including per-tensor, per-token, and per-channel methods, as briefly shown in Figure \ref{quant_dim}.
Per-tensor quantization utilizes a single set of quantization parameters for the entire tensor (or block).
Since LLMs perform autoregressive inference, per-token KV quantization has become a widely adopted approach \citep{duanmu2024skvq, su2025akvq, he2024zipcache}.
Recent studies \citep{liu2024kivi, hooper2025kvquant} have identified outliers along the channel dimension in the Keys, with per-channel quantization shown to effectively mitigate quantization errors.
Furthermore, quantization granularity can be improved by defining smaller groupings along the quantization dimension, which reduces quantization errors but introduces additional overhead due to the increased number of quantization parameters.

\textbf{Dynamic and static quantization.}
Dynamic quantization adjusts the quantization parameters during inference, offering greater flexibility but potentially leading to less efficient performance compared to static quantization due to the additional computational demands of online adjustment.
In contrast, static quantization involves estimating the range by passing a few batches of calibration data through the model prior to inference.
This approach enhances inference efficiency, as the quantization parameters are pre-calculated and remain fixed during inference. 
However, it may result in higher quantization errors due to the inability to adapt to varying input distributions during runtime.

\textbf{Impact of attention sinks on KV cache quantization.}
As shown in Figure \ref{quant_dim}, the abnormal value characteristics of sink tokens resulting from QKV suppression can significantly impact quantization when sink tokens are included in quantization groups or when quantization parameters calibrated using sink tokens are applied. 
The impact of sink tokens varies significantly across different quantization schemes, as discussed in Section \ref{section4}.
\section{Additional Experimental Results on Cross-Layer Evolution of Extreme Activation Outliers}
\begin{table}[t]
\vspace{-5mm}
\centering
\resizebox{0.8\columnwidth}{!}{%
\begin{tabular}{@{}c|cccc@{}}
\toprule
Model & Total Layers & \begin{tabular}[c]{@{}c@{}}Emergence Stage Layer\\ of Stable Outliers\end{tabular} & Hidden Size & \begin{tabular}[c]{@{}c@{}}Outlier Channels\\ of Stable Outliers\end{tabular} \\ \midrule
LLaMA2-7B & 32 & 1 & 4096 & 2533, 1415 \\
LLaMA2-13B & 40 & 3 & 5120 & 4743, 2100 \\
Mistral-7B & 32 & 1 & 4096 & 2070, 3398 \\
LLaMA3-8B & 32 & 1 & 4096 & 788, 1384, 4062 \\
LLaMA3.1-8B-instruct & 32 & 1 & 4096 & 788, 1384, 4062 \\
LLaMA3.2-1B & 16 & 1 & 2048 & 400, 698, 2029, 1159 \\
LLaMA3.2-3B & 28 & 1 & 3072 & 588, 1016, 3046, 1731 \\ \bottomrule
\end{tabular}%
}
\vspace{-2mm}
\caption{Emergence stage and outlier channels of stable outliers for several models.}
\vspace{-5mm}
\label{tab:stage}
\end{table}
In this section, we present additional experiments on the cross-layer evolution of different types of extreme activation outliers across various inputs and models.
We conduct experiments using two distinct prompts. These prompts are:

\textbullet  prompt 1:
"The following are multiple choice questions (with answers) about machine learning.\textbackslash n \textbackslash n A 6-sided die is rolled 15 times and the results are: side 1 comes up 0 times;"

\textbullet  prompt 2:
"Summer is warm.\textbackslash n Winter is cold.\textbackslash n Spring is mild.\textbackslash n Autumn is crisp.\textbackslash n The sun rises early in the summer.\textbackslash n The days are short in the winter.\textbackslash n "

As shown in Figures \ref{cross_llama_2_7b_p1}, \ref{cross_llama_2_7b_p2}, and \ref{outlier_appendix}, different inputs do not affect the layers at each stage or the channels where stable outliers emerge.
Therefore, they can be used as pre-identified static features during inference.
Table \ref{tab:stage} shows the emergence stage layers and outlier channels for several models used in KVSink.

We then validate the behavior of each type of outlier at each stage on additional models, including models of different sizes, fine-tuned models, and models using Grouped-Query-Attention (GQA) \citep{ainslie2023gqa}.
As shown in Figures \ref{outlier_appendix}, \ref{cross_llama_2_13b_p1}, and \ref{cross_llama_3.2_3b_p1}, validation across additional models confirms our findings that stable outliers, driven by outliers in \( X^{l}_{\text{d\_in}} \) and \( X^{l}_{\text{d\_out}} \), undergo a key process of emergence, stabilization, and dissipation.
Building on this, the behavior varies slightly across models. 
For instance, in the LLaMA3 series models, the dissipation stage occurs in the final layer, with no distinct final stage.
In the Mistral models, the dissipation stage spans multiple layers, whereas for the other models in our experiments, this stage is confined to a single layer.
These minor differences do not impact KVSink's ability to predict sink tokens during the emergence stage.
\label{cross_layer}
\section{Additional Experimental Results on QKV Suppression and High Cosine Similarity of QK }  
In this section, we present additional experimental results on QKV suppression and the high cosine similarity of QK.
First, as shown in Figures \ref{QKV_appendix_0}, \ref{QKV_appendix_state_1}, \ref{QKV_appendix_2}, and \ref{QKV_appendix_state_2}, QKV suppression persists across layers when attention sinks occur for different inputs (as detailed in Section \ref{stages}).
Second, as shown in Figures \ref{QKV_appendix_0}, \ref{QKV_appendix_2}, \ref{QKV_appendix_3}, and \ref{QKV_appendix_4}, QKV suppression persists across different models.
A notable observation is the occurrence of unusually large norms in the Queries and Values of sink tokens in the final layer, a phenomenon that has not been fully explored in previous research.
We hypothesize that this may suggest inherent differences among various sink tokens, and it could be explored as part of future work.

\begin{figure}[t]
    \vspace{-5mm}
    \centering    
    \begin{subfigure}{0.16\textwidth}
        \centering
    \includegraphics[width=\linewidth]{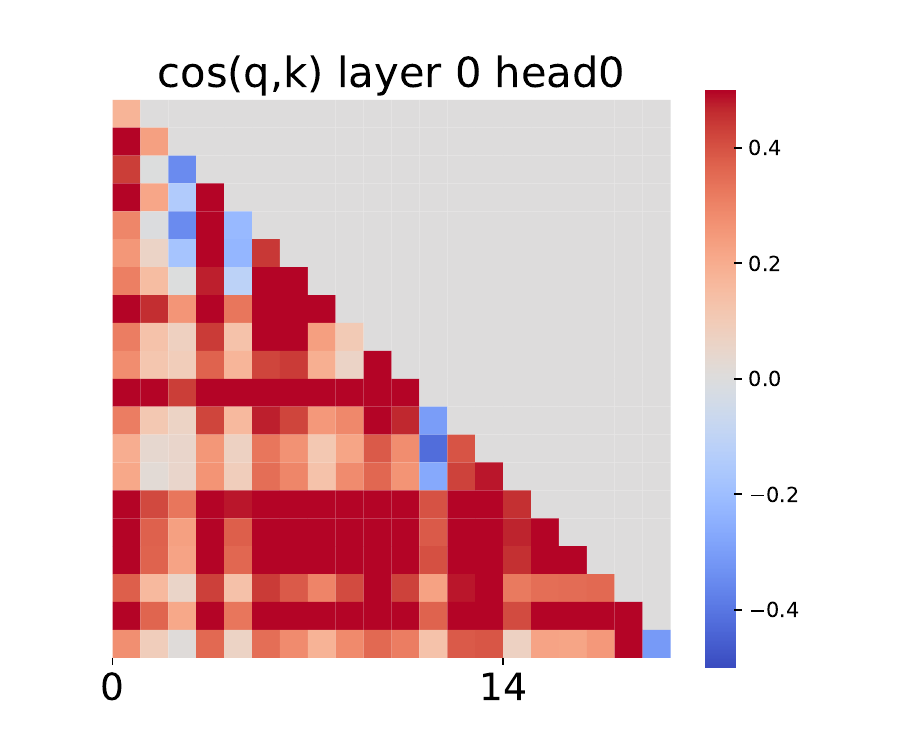}
    \end{subfigure}
    \begin{subfigure}{0.16\textwidth}
        \centering
    \includegraphics[width=\linewidth]{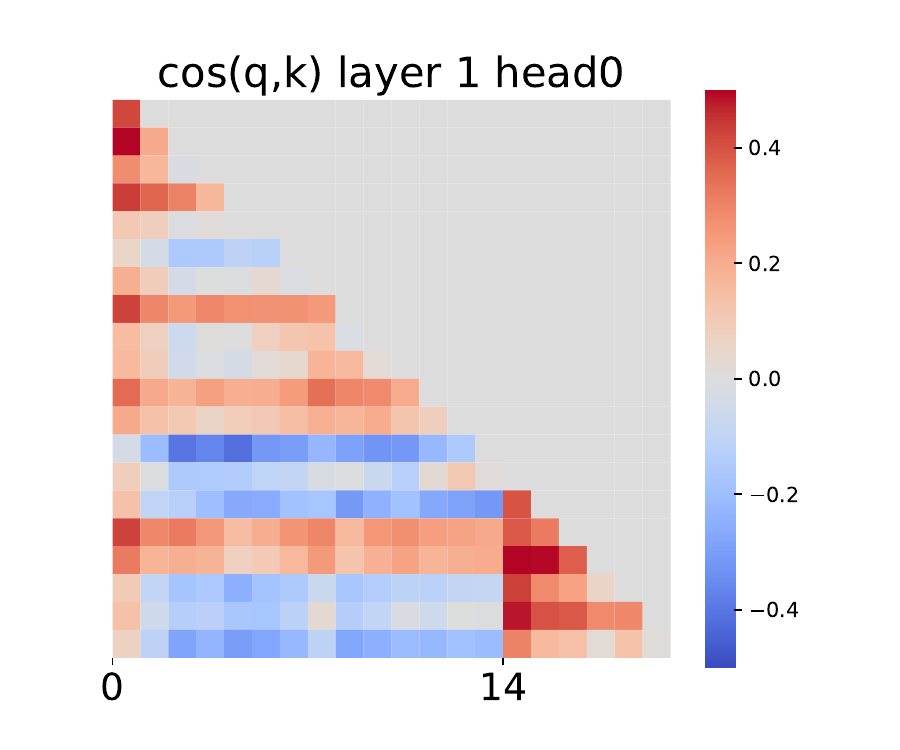}
    \end{subfigure}
    \begin{subfigure}{0.16\textwidth}
        \centering
    \includegraphics[width=\linewidth]{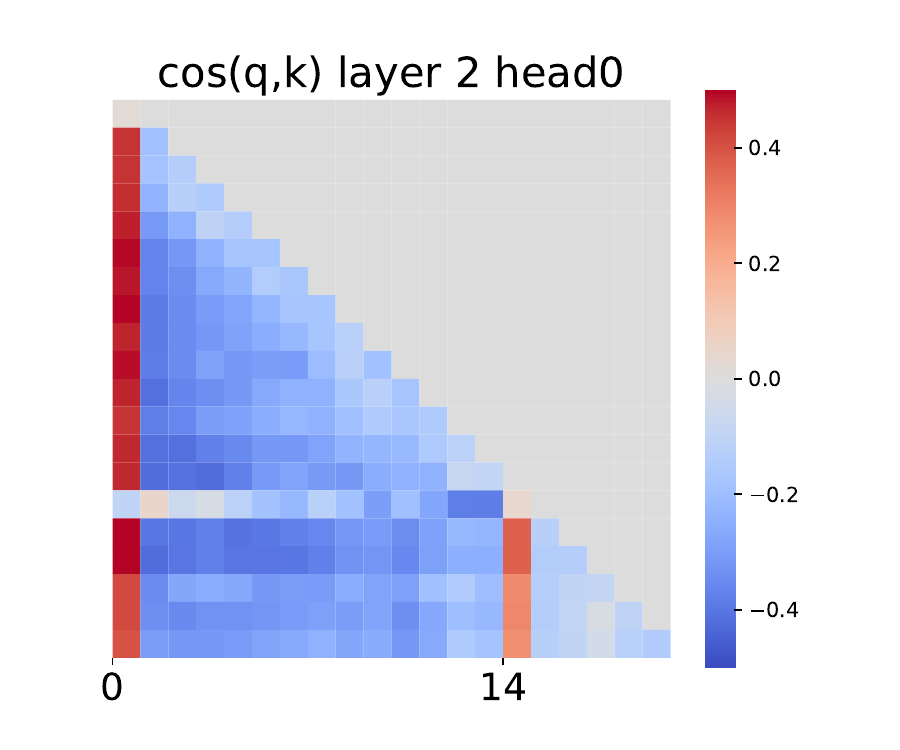}
    \end{subfigure}    
    \begin{subfigure}{0.16\textwidth}
        \centering
    \includegraphics[width=\linewidth]{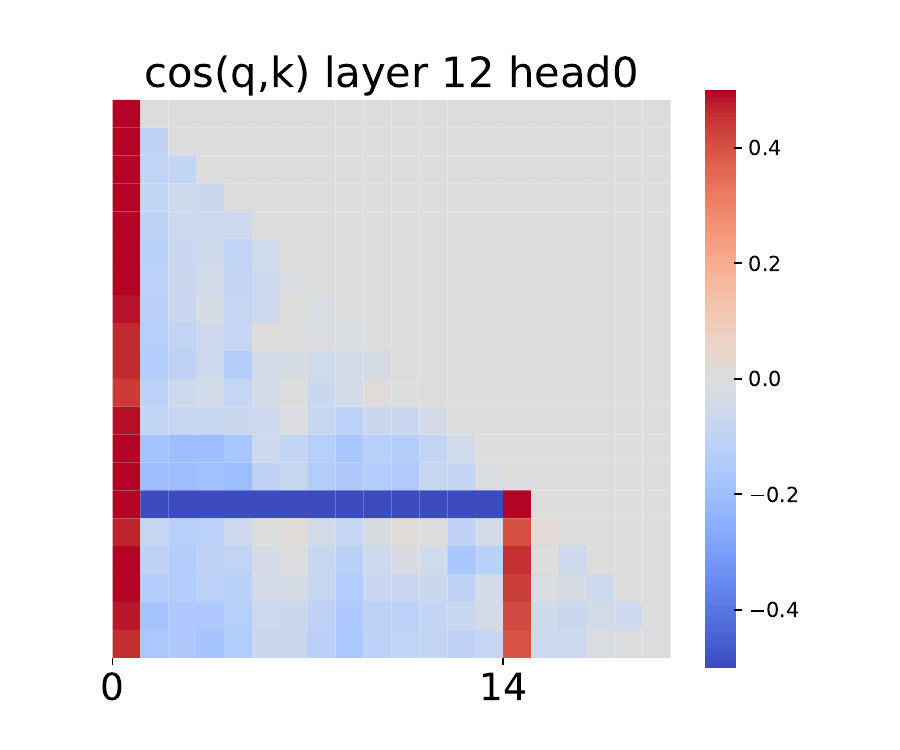}
    \end{subfigure}
    \begin{subfigure}{0.16\textwidth}
        \centering
    \includegraphics[width=\linewidth]{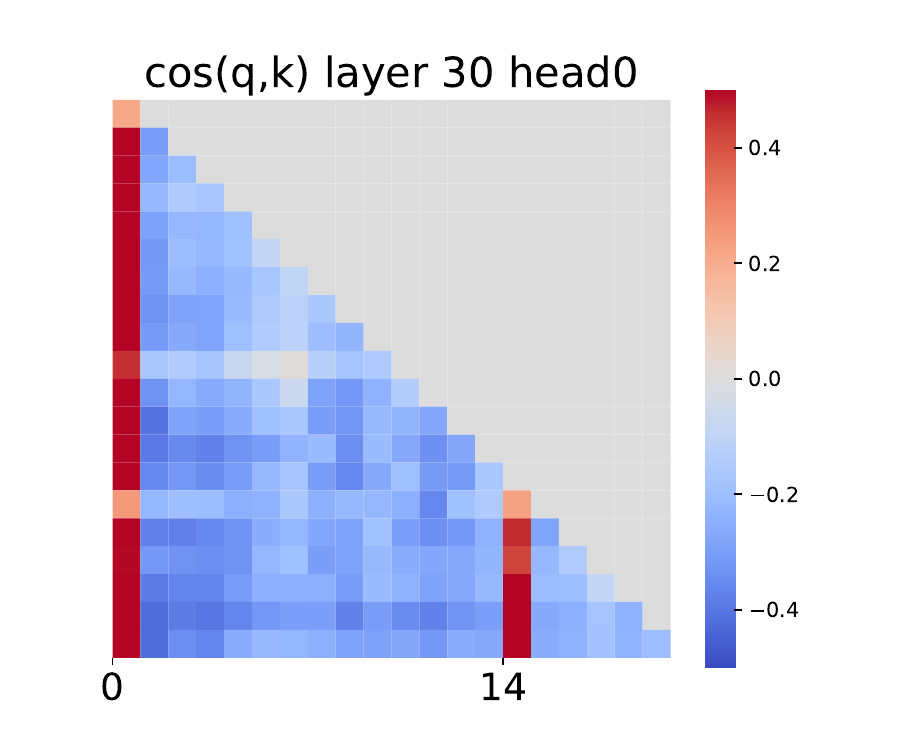}
    \end{subfigure}
    \begin{subfigure}{0.16\textwidth}
        \centering
    \includegraphics[width=\linewidth]{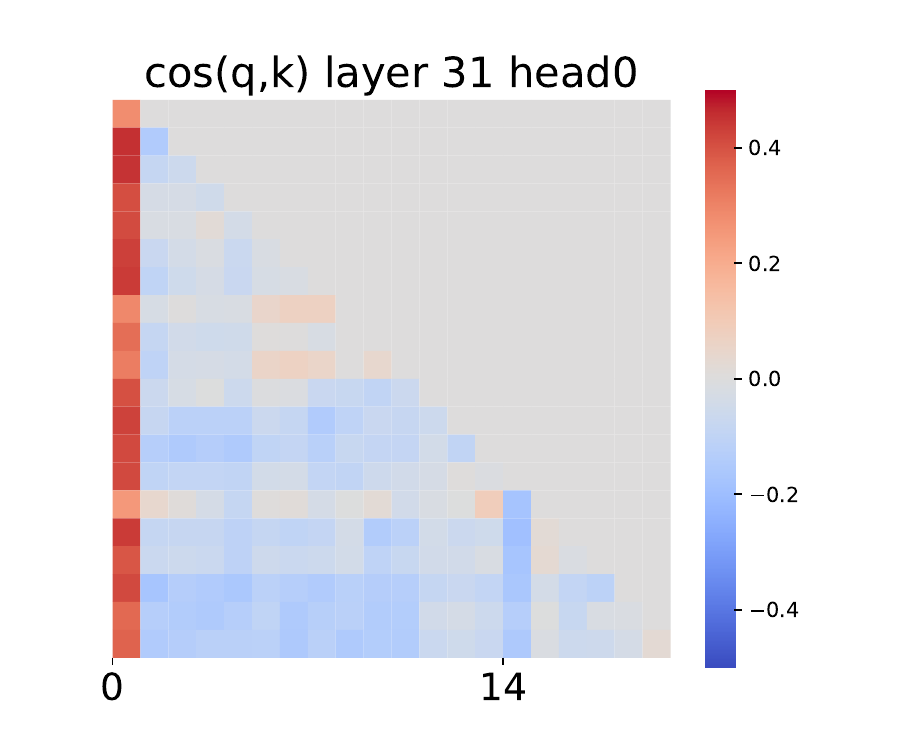}
    \end{subfigure}    
    \begin{subfigure}{0.16\textwidth}
        \centering
    \includegraphics[width=\linewidth]{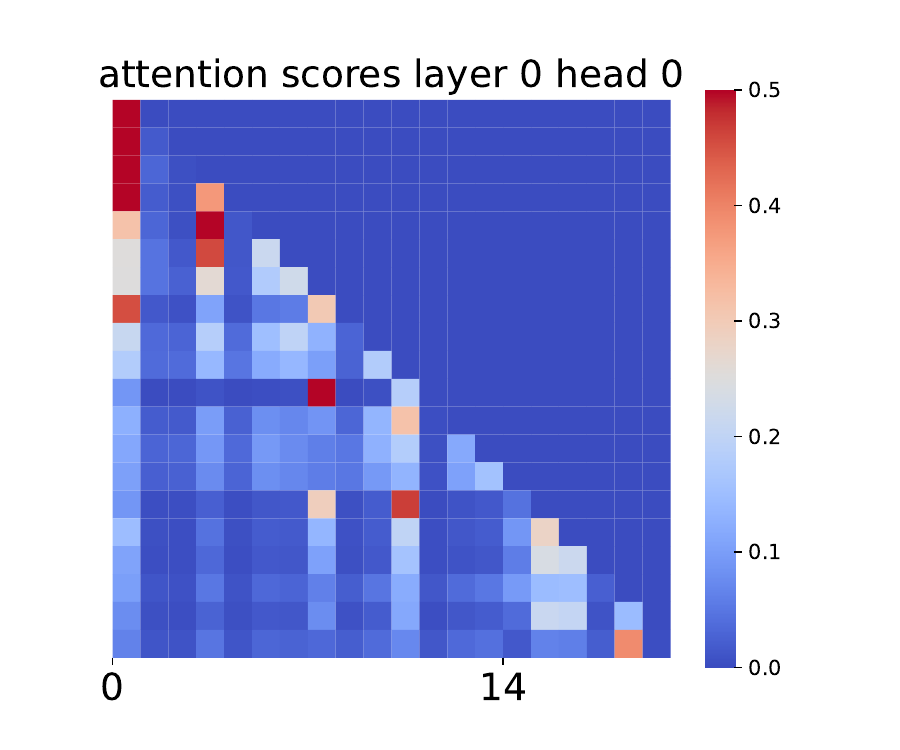}
    \end{subfigure}
    \begin{subfigure}{0.16\textwidth}
        \centering
    \includegraphics[width=\linewidth]{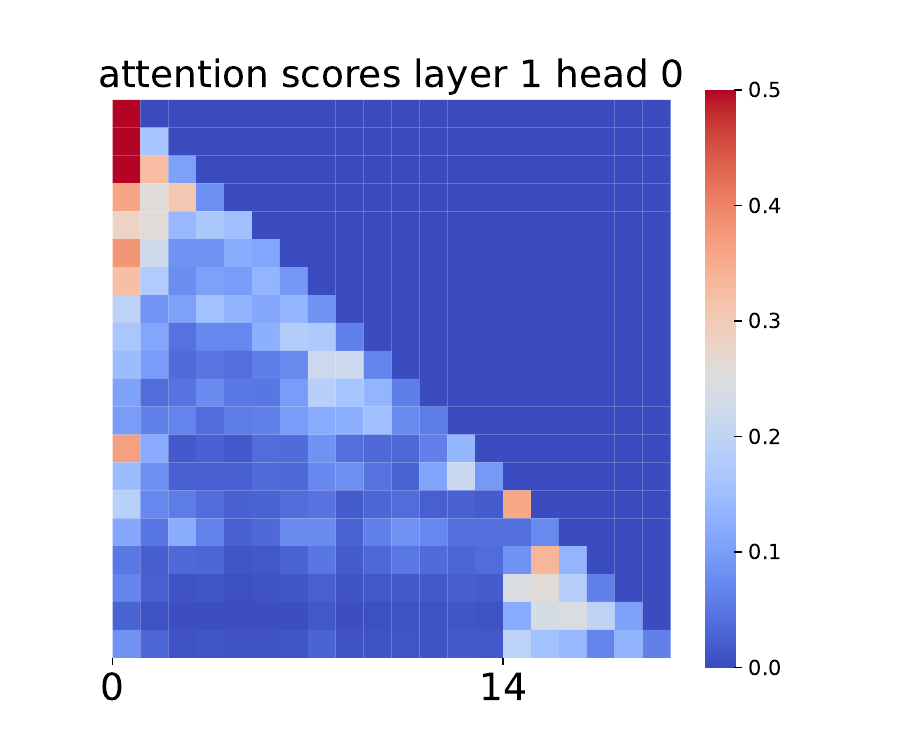}
    \end{subfigure}
    \begin{subfigure}{0.16\textwidth}
        \centering
    \includegraphics[width=\linewidth]{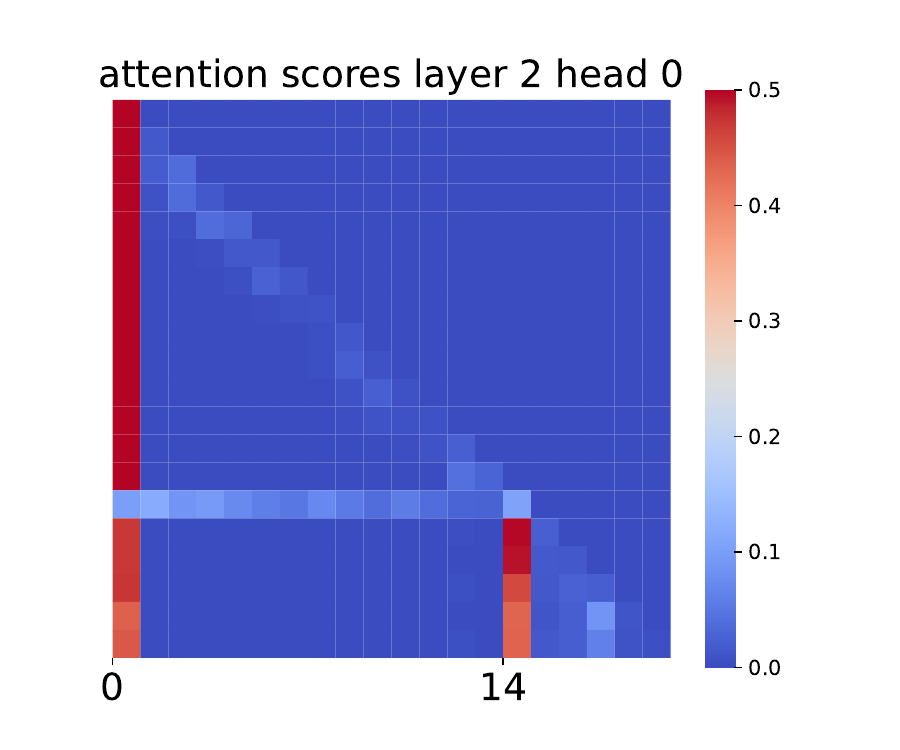}
    \end{subfigure}   
    \begin{subfigure}{0.16\textwidth}
        \centering
    \includegraphics[width=\linewidth]{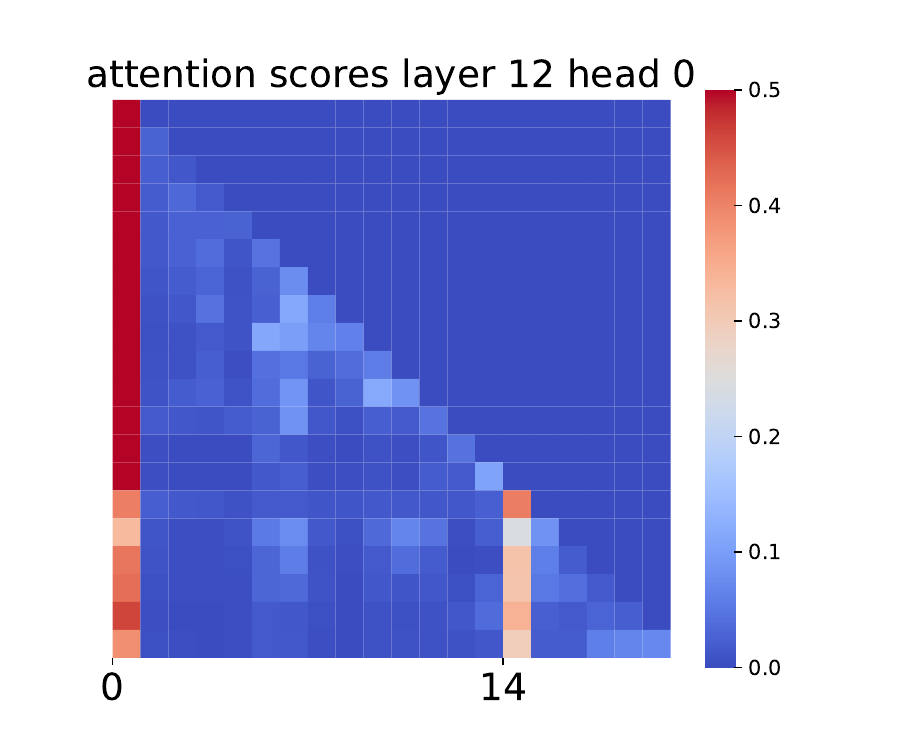}
    \end{subfigure}
    \begin{subfigure}{0.16\textwidth}
        \centering
    \includegraphics[width=\linewidth]{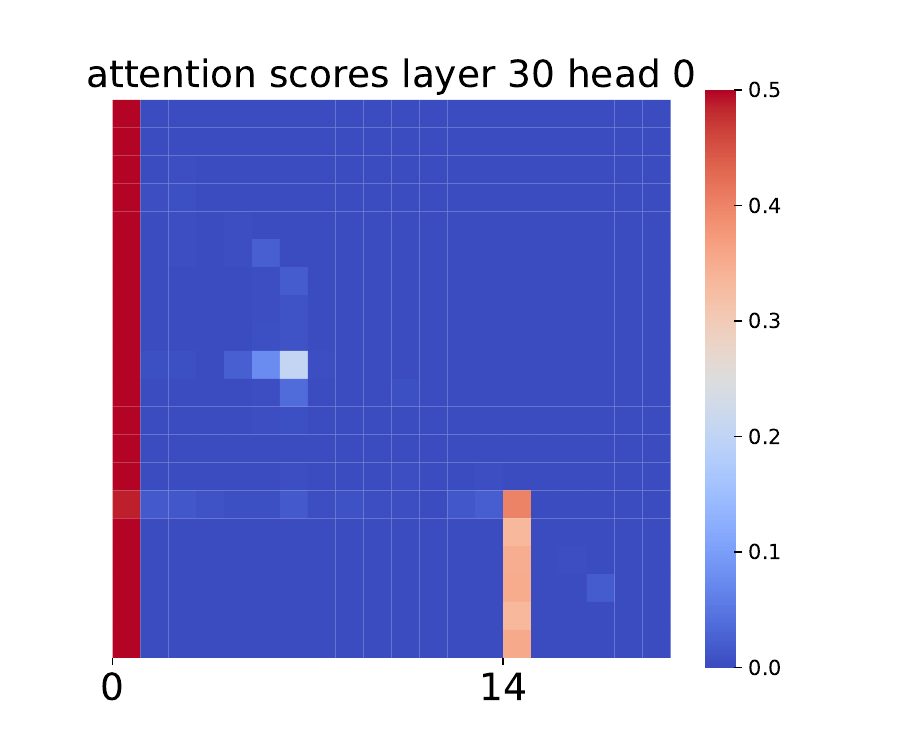}
    \end{subfigure}
    \begin{subfigure}{0.16\textwidth}
        \centering
    \includegraphics[width=\linewidth]{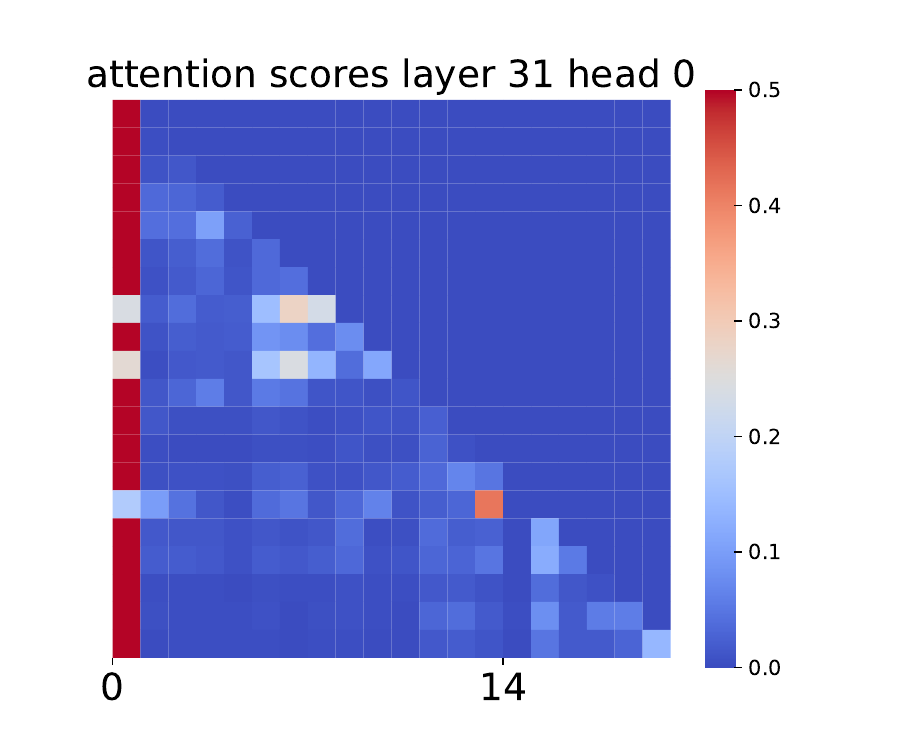}
    \end{subfigure}
    \vspace{-3mm}
    \caption{Additional visualizations of the attention scores and the high cosine similarity of QK in LLaMA2-7B.
    }
\label{cos_qk_appendix}
\end{figure}

In Figure \ref{cos_qk_appendix}, we present additional visualizations of the attention scores and the high cosine similarity of QK in LLaMA2-7B, confirming the correlation between these two factors.
\label{QKV Suppression}
\section{Efficiency Analysis of KVSink}
This section presents experimental evaluation and analysis of KVSink’s efficiency. 
We conduct experiments on multiple models using a 4 × A100 (80GB) setup and evaluate performance on the Wikitext-2 dataset, with input sequences segmented into 4K tokens.
For time efficiency, we measure the average prefill latency and the time required for KVSink's outlier identification operation under various configurations of sink token quantities.
For memory efficiency, we theoretically estimate the original KV cache memory consumption under 2-bit quantization and quantify the additional memory overhead introduced by KVSink’s preservation mechanism.

As shown in Table \ref{tab:efficiency}, leveraging our enhanced understanding of the attention sinks mechanism, KVSink involves only minimal computations, can be efficiently implemented using PyTorch, and exhibits a negligible impact on time efficiency.
The impact on memory efficiency is also minimal. 
As the context length increases, this impact could diminish further, as the number of preserved tokens remains fixed.
\label{Efficiency Analysis}
\begin{table}[t]
\resizebox{\columnwidth}{!}{%
\begin{tabular}{@{}c|cccc|cccc|cccc@{}}
\toprule
\multirow{3}{*}{Model} & \multicolumn{4}{c|}{LLaMA2-7B} & \multicolumn{4}{c|}{LLaMA2-13B} & \multicolumn{4}{c}{LLaMA2-70B} \\ \cmidrule(l){2-13} 
 & \multicolumn{2}{c|}{prefill time} & \multicolumn{2}{c|}{KV Cache memory} & \multicolumn{2}{c|}{prefill time} & \multicolumn{2}{c|}{KV Cache memory} & \multicolumn{2}{c|}{prefill time} & \multicolumn{2}{c}{kv cache memory} \\ \cmidrule(l){2-13} 
 & \begin{tabular}[c]{@{}c@{}}w/o\\ KVSink\end{tabular} & \multicolumn{1}{c|}{\begin{tabular}[c]{@{}c@{}}+\\ KVSink\end{tabular}} & \begin{tabular}[c]{@{}c@{}}w/o\\ KVSink\end{tabular} & \begin{tabular}[c]{@{}c@{}}+\\ KVSink\end{tabular} & \begin{tabular}[c]{@{}c@{}}w/o\\ KVSink\end{tabular} & \multicolumn{1}{c|}{\begin{tabular}[c]{@{}c@{}}+\\ KVSink\end{tabular}} & \begin{tabular}[c]{@{}c@{}}w/o\\ KVSink\end{tabular} & \begin{tabular}[c]{@{}c@{}}+\\ KVSink\end{tabular} & \begin{tabular}[c]{@{}c@{}}w/o\\ KVSink\end{tabular} & \multicolumn{1}{c|}{\begin{tabular}[c]{@{}c@{}}+\\ KVSink\end{tabular}} & \begin{tabular}[c]{@{}c@{}}w/o\\ KVSink\end{tabular} & \begin{tabular}[c]{@{}c@{}}+\\ KVSink\end{tabular} \\ \midrule
KVSink-1 & \multirow{3}{*}{676.78} & \multicolumn{1}{c|}{+ 0.04} & \multirow{3}{*}{256} & + 0.5 & \multirow{3}{*}{1090.70} & \multicolumn{1}{c|}{+ 0.04} & \multirow{3}{*}{400} & + 0.78 & \multirow{3}{*}{4481.05} & \multicolumn{1}{c|}{+ 0.04} & \multirow{3}{*}{160} & + 0.31 \\ \cmidrule(lr){3-3} \cmidrule(lr){5-5} \cmidrule(lr){7-7} \cmidrule(lr){9-9} \cmidrule(lr){11-11} \cmidrule(l){13-13} 
KVSink-5 &  & \multicolumn{1}{c|}{+ 0.04} &  & + 2.5 &  & \multicolumn{1}{c|}{+ 0.05} &  & + 3.91 &  & \multicolumn{1}{c|}{+ 0.05} &  & + 1.56 \\ \cmidrule(lr){3-3} \cmidrule(lr){5-5} \cmidrule(lr){7-7} \cmidrule(lr){9-9} \cmidrule(lr){11-11} \cmidrule(l){13-13} 
KVSink-20 &  & \multicolumn{1}{c|}{+ 0.05} &  & + 10 &  & \multicolumn{1}{c|}{+ 0.05} &  & + 15.63 &  & \multicolumn{1}{c|}{+ 0.05} &  & + 6.25 \\ \midrule
\multirow{3}{*}{Model} & \multicolumn{4}{c|}{Mistral-7B} & \multicolumn{4}{c|}{LLaMA3-8B} & \multicolumn{4}{c}{LLaMA3.2-3B} \\ \cmidrule(l){2-13} 
 & \multicolumn{2}{c|}{prefill time} & \multicolumn{2}{c|}{KV cache memory} & \multicolumn{2}{c|}{prefill time} & \multicolumn{2}{c|}{KV cache memory} & \multicolumn{2}{c|}{prefill time} & \multicolumn{2}{c}{KV cache memory} \\ \cmidrule(l){2-13} 
 & \begin{tabular}[c]{@{}c@{}}w/o\\ KVSink\end{tabular} & \multicolumn{1}{c|}{\begin{tabular}[c]{@{}c@{}}+\\ KVSink\end{tabular}} & \begin{tabular}[c]{@{}c@{}}w/o\\ KVSink\end{tabular} & \begin{tabular}[c]{@{}c@{}}+\\ KVSink\end{tabular} & \multicolumn{1}{c|}{\begin{tabular}[c]{@{}c@{}}w/o\\ KVSink\end{tabular}} & \multicolumn{1}{c|}{\begin{tabular}[c]{@{}c@{}}+\\ KVSink\end{tabular}} & \begin{tabular}[c]{@{}c@{}}w/o\\ KVSink\end{tabular} & \begin{tabular}[c]{@{}c@{}}+\\ KVSink\end{tabular} & \begin{tabular}[c]{@{}c@{}}w/o\\ KVSink\end{tabular} & \multicolumn{1}{c|}{\begin{tabular}[c]{@{}c@{}}+\\ KVSink\end{tabular}} & \begin{tabular}[c]{@{}c@{}}w/o\\ KVSink\end{tabular} & \begin{tabular}[c]{@{}c@{}}+\\ KVSink\end{tabular} \\ \midrule
KVSink-1 & \multirow{3}{*}{686.32} & \multicolumn{1}{c|}{+ 0.04} & \multirow{3}{*}{64} & + 0.13 & \multirow{3}{*}{707.95} & \multicolumn{1}{c|}{+ 0.04} & \multirow{3}{*}{64} & + 0.13 & \multirow{3}{*}{430.74} & \multicolumn{1}{c|}{+ 0.04} & \multirow{3}{*}{56} & + 0.11 \\ \cmidrule(lr){3-3} \cmidrule(lr){5-5} \cmidrule(lr){7-7} \cmidrule(lr){9-9} \cmidrule(lr){11-11} \cmidrule(l){13-13} 
KVSink-5 &  & \multicolumn{1}{c|}{+ 0.05} &  & + 0.63 &  & \multicolumn{1}{c|}{+ 0.05} &  & + 0.63 &  & \multicolumn{1}{c|}{+ 0.05} &  & + 0.55 \\ \cmidrule(lr){3-3} \cmidrule(lr){5-5} \cmidrule(lr){7-7} \cmidrule(lr){9-9} \cmidrule(lr){11-11} \cmidrule(l){13-13} 
KVSink-20 &  & \multicolumn{1}{c|}{+ 0.05} &  & + 2.5 &  & \multicolumn{1}{c|}{+0.05} &  & + 2.5 &  & \multicolumn{1}{c|}{+ 0.05} &  & + 2.19 \\ \bottomrule
\end{tabular}%
}
\vspace{-3mm}
\caption{Efficiency analysis of KVSink.
KVSink-N indicates the preservation of N tokens. 
Time is reported in milliseconds (ms) and memory in megabytes (MB).}
\label{tab:efficiency}
\vspace{-3mm}
\end{table}
\section{KVSink Algorithm}
The prefill phase with KVsink is illustrated in Algorithm \ref{Prefill Phase with KVSink}. 
For clarity, the multi-head mechanism is omitted in the algorithm.
Note that if static quantization is applied, sink tokens should also be excluded during quantization parameter calibration.
\label{Algorithm}
\begin{algorithm}
\caption{Prefill Phase with KVSink}
\begin{algorithmic}[1]
\STATE \textbf{Parameters:} Number of decoder layers: \( L \),
Emergence stage layer: \( l_E \), 
Hidden size: \( d \), 
Outlier channel: \( c \), 
Number of token length: \( n \),
Number of tokens for preservation: \( k \),
Weights: \( W_Q, W_K, W_V, W_O \).
\STATE \textbf{Input:} 
Input to decoder 0: \( H^{0} \in \mathbb{R}^{n \times d} \).
\STATE \textbf{Output:}
Output of decoder \( L \): \( H^L \in \mathbb{R}^{n \times d} \).
\STATE \textbf{Initialize:} \( S_{\text{outliers}} = \emptyset \), \( S_{\text{sink}} = \emptyset \)
\FOR{\( l = 1 \) \textbf{to} \( L \)}
    \STATE \( H^{l} \gets \text{LayerNorm}^{l}_{mhsa}(H^{l-1}) \)
    \STATE \( Q^{l} \gets H^{l} \cdot W_Q\),
    \( K^{l} \gets  H^{l} \cdot W_K\),
    \( V^{l} \gets H^{l} \cdot W_V\)
    \STATE \( K^{l}[ \text{token} \notin S_{\text{sink}} ] \gets \text{quantize}((K^{l}[ \text{token} \notin S_{\text{sink}} ]) \)
    \STATE \( V^{l}[ \text{token} \notin S_{\text{sink}} ] \gets \text{quantize}(V^{l}[ \text{token} \notin S_{\text{sink}} ]) \)
    \STATE \( K^{l} \gets K^{l}_{\text{quant}}\), \( V^{l} \gets V^{l}_{\text{quant}}\)
    \STATE \( H^{l} \gets \text{Attention}(Q^{l}, K^{l}_{\text{quant}}, V^{l}_{\text{quant}}) \)
    \STATE \( H^{l} \gets H^{l} \cdot W_O + H^{l-1}\)
    \STATE \( H^{l} \gets \text{FFN}(\text{LayerNorm}^{l}_{ffn}(H^{l}) ) + H^{l}\)    
    \IF{l = \( l_E \)}
        \STATE \( S_{\text{outliers}} \gets \left\{ (i, c) \mid |H^{l}_{i,c}| \in \text{Top-k}(|H^{l}_{i,c}|) \right\} \)
        \STATE \( S_{\text{sink}} \gets \{ i \mid (i, c) \in S_{\text{outliers}} \} \)
    \ENDIF
\ENDFOR
\RETURN \( H^L \)
\end{algorithmic}
\label{Prefill Phase with KVSink}
\end{algorithm}
{
\begin{figure*}[t]
        \centering
\includegraphics[width=1\linewidth]{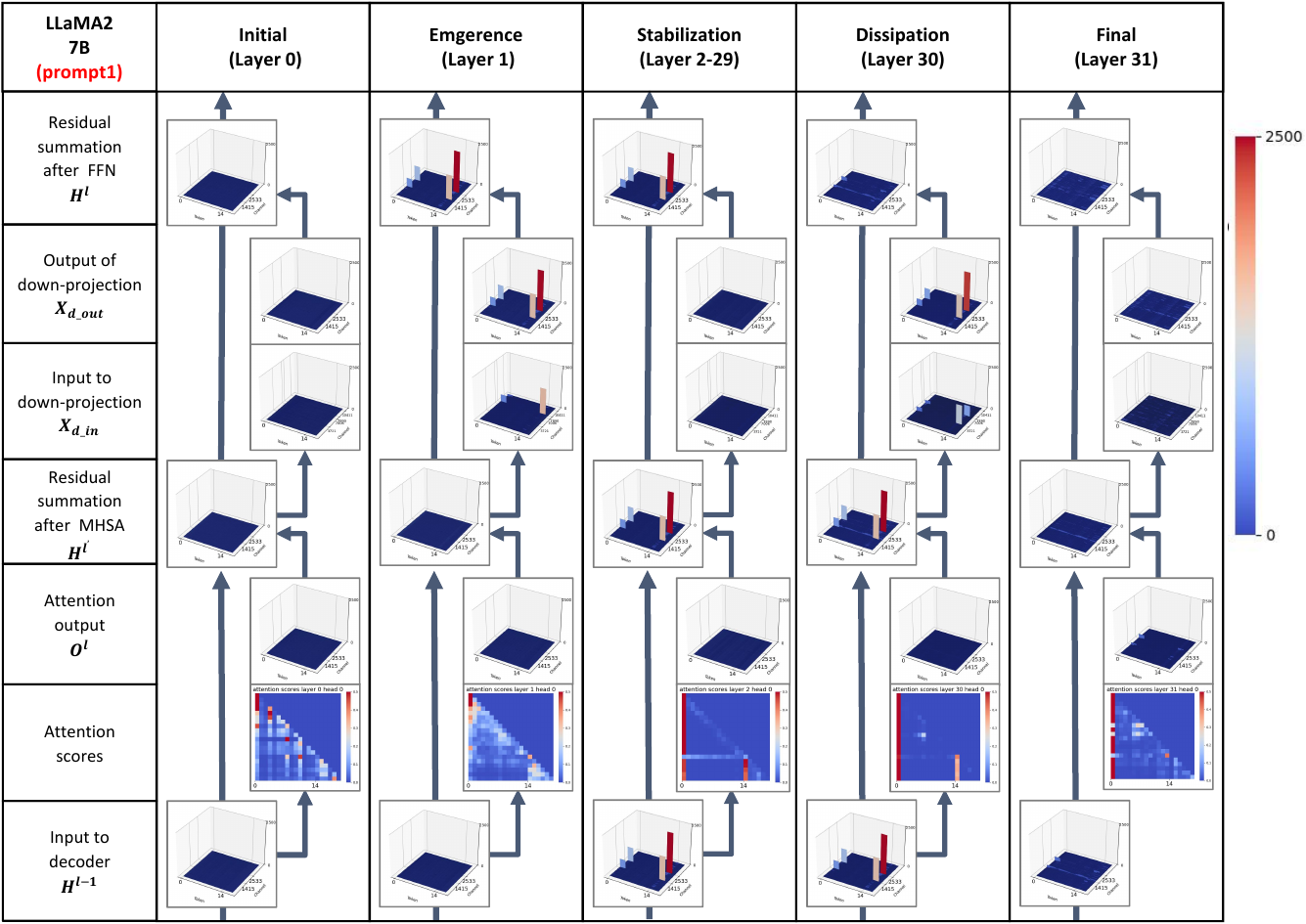}
\caption{Visualizations of the cross-layer evolution of extreme activation outliers in LLaMA2-7B with Prompt 1.}
\label{cross_llama_2_7b_p1}    
\end{figure*}
\begin{figure*}[t]
        \centering
\includegraphics[width=1\linewidth]{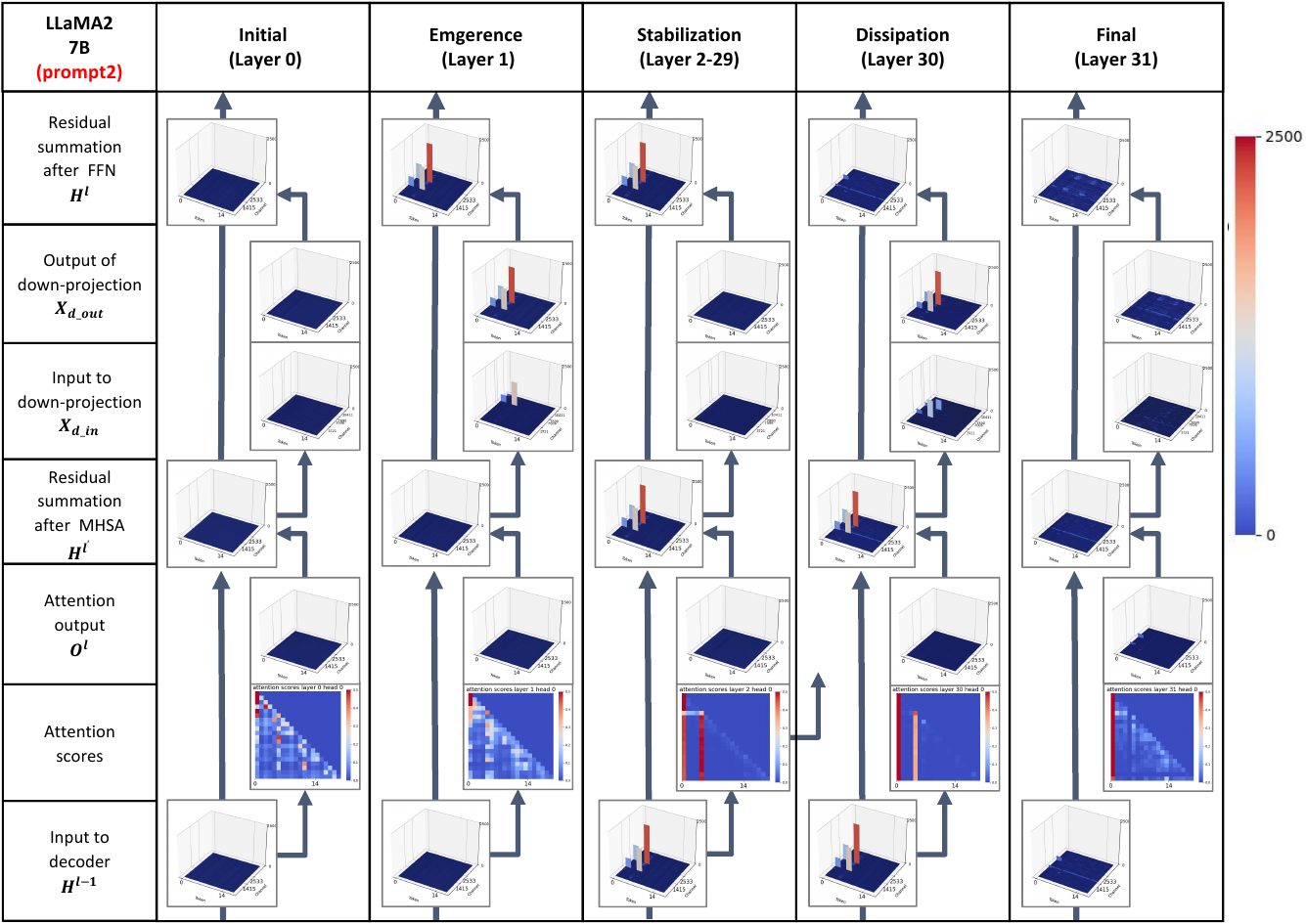}
\caption{Visualizations of the cross-layer evolution of extreme activation outliers in LLaMA2-7B with Prompt 2.}
\label{cross_llama_2_7b_p2}
\end{figure*}
\begin{figure*}[t]
        \centering
\includegraphics[width=1\linewidth]{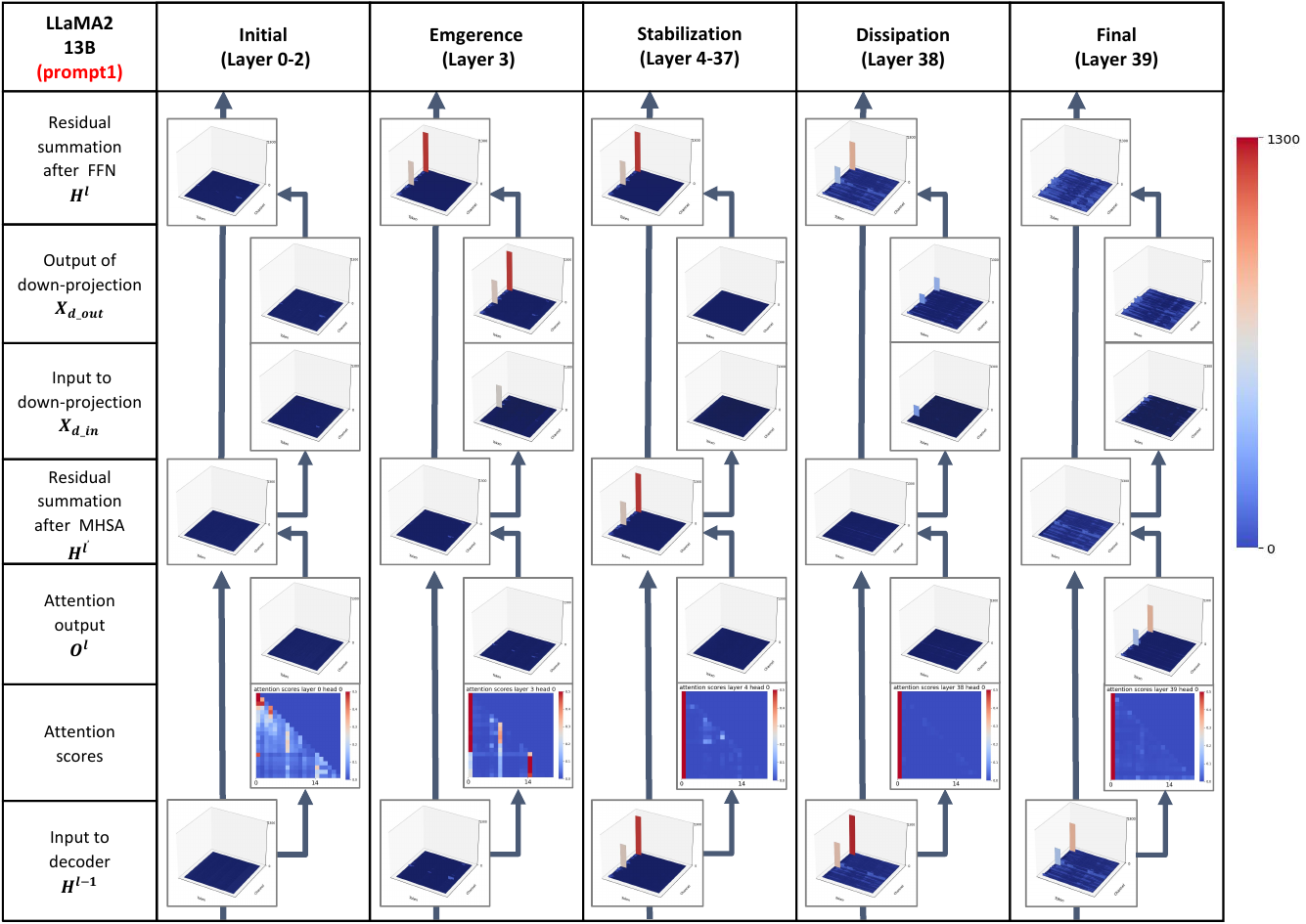}
\caption{Visualizations of the cross-layer evolution of extreme activation outliers in LLaMA2-13B with Prompt 1.}
\label{cross_llama_2_13b_p1}    
\end{figure*}
\begin{figure*}[t]
        \centering
\includegraphics[width=1\linewidth]{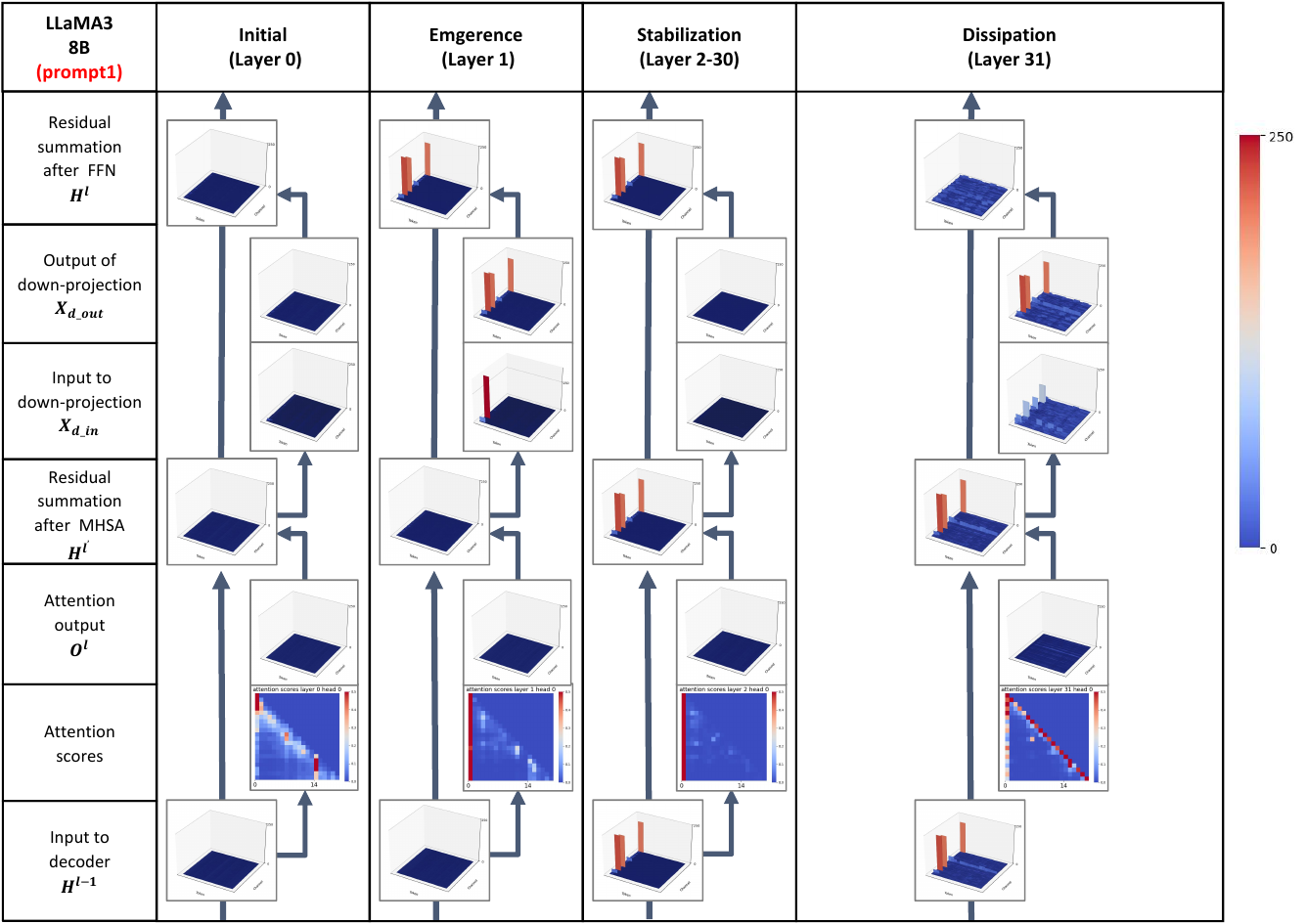}
\caption{Visualizations of the cross-layer evolution of extreme activation outliers in LLaMA3-8B with Prompt 1.}
    \label{cross_llama_3_8b_p1}
\end{figure*}
\begin{figure*}[t]
        \centering
\includegraphics[width=1\linewidth]{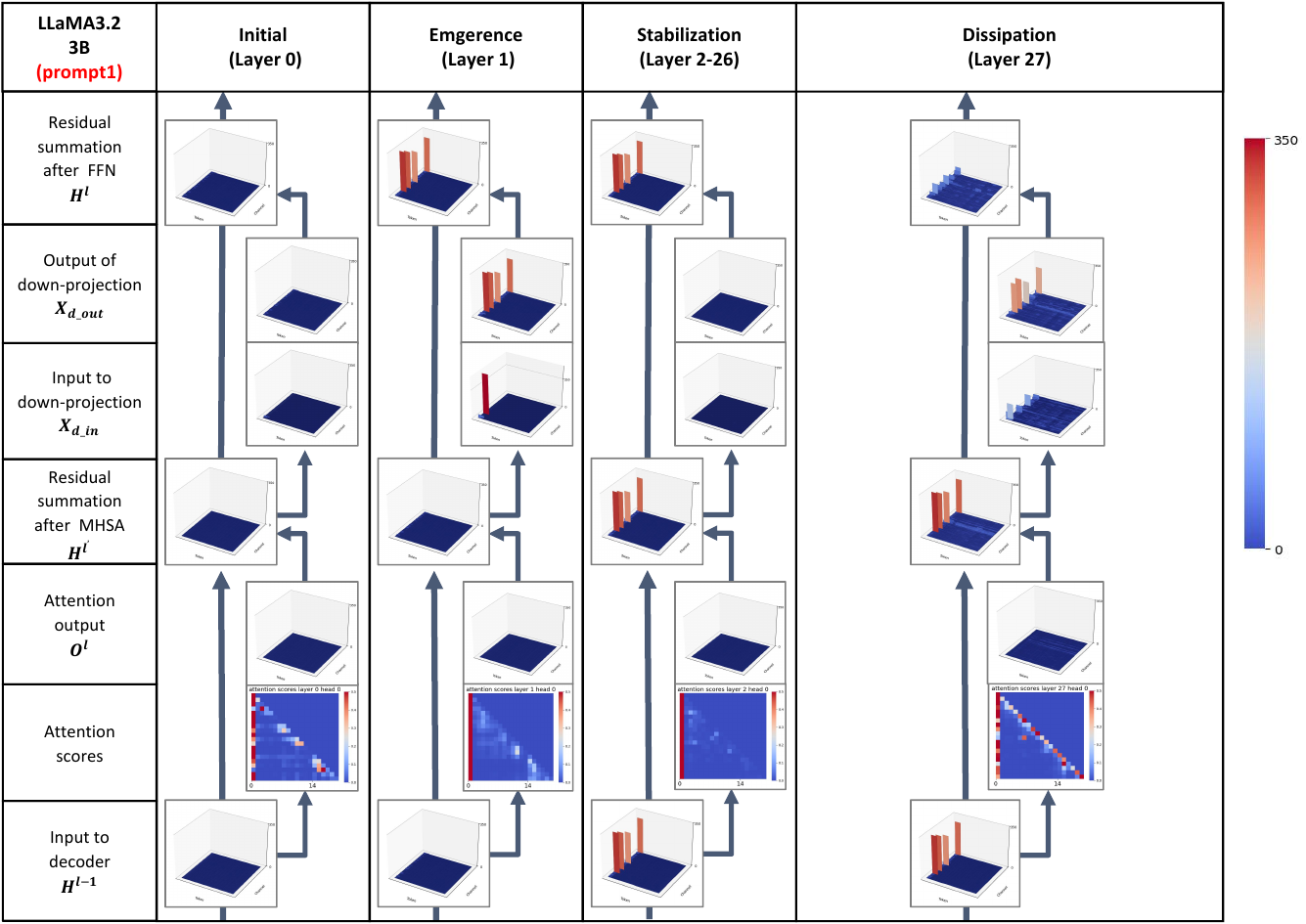}
\caption{Visualizations of the cross-layer evolution of extreme activation outliers in LLaMA3.2-3B with Prompt 1.}
    \label{cross_llama_3.2_3b_p1}
\end{figure*}
\begin{figure}[t]
    \centering    
    \begin{subfigure}{0.24\textwidth}
        \centering
    \includegraphics[width=\linewidth]{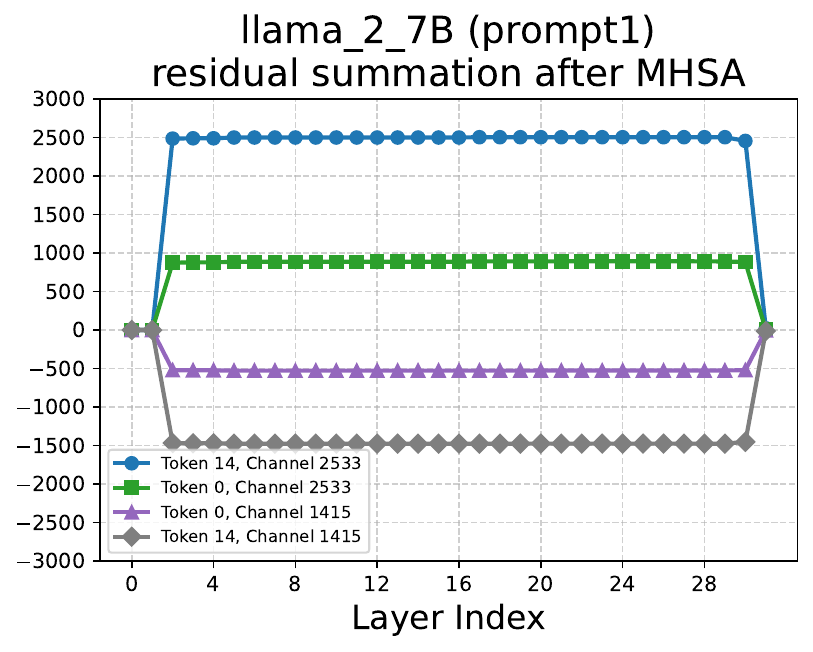}
    \end{subfigure}
    \begin{subfigure}{0.24\textwidth}
        \centering
    \includegraphics[width=\linewidth]{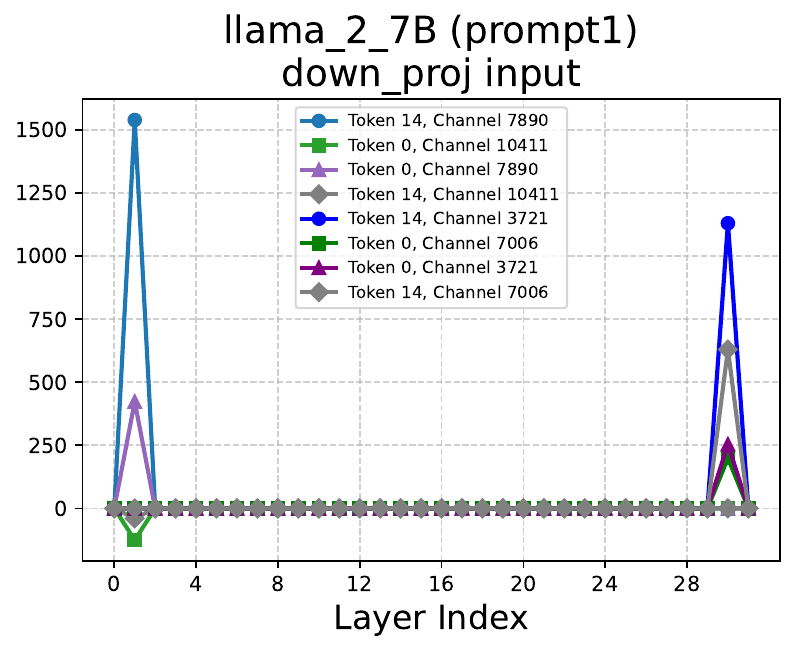}
    \end{subfigure}
    \begin{subfigure}{0.24\textwidth}
        \centering
    \includegraphics[width=\linewidth]{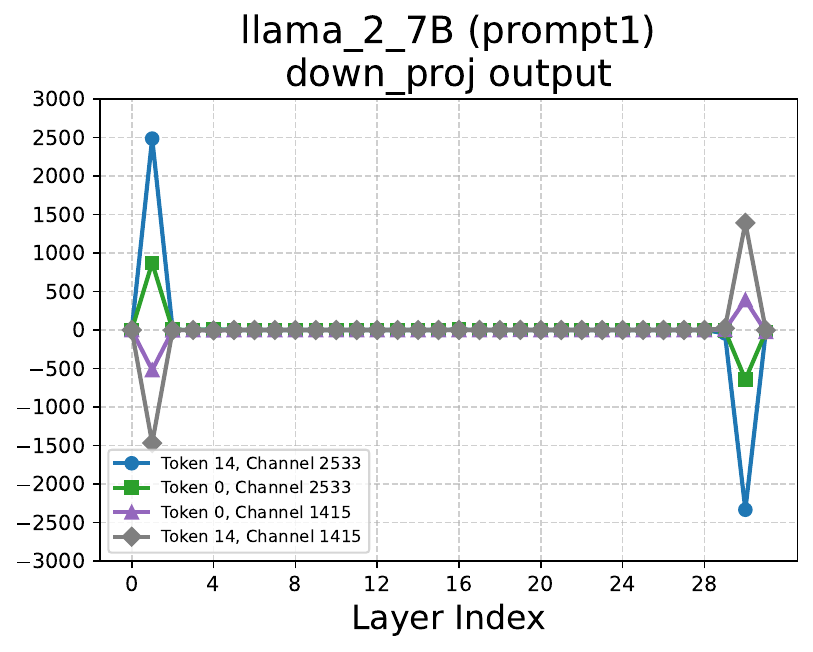}
    \end{subfigure}
    \begin{subfigure}{0.24\textwidth}
        \centering
    \includegraphics[width=\linewidth]{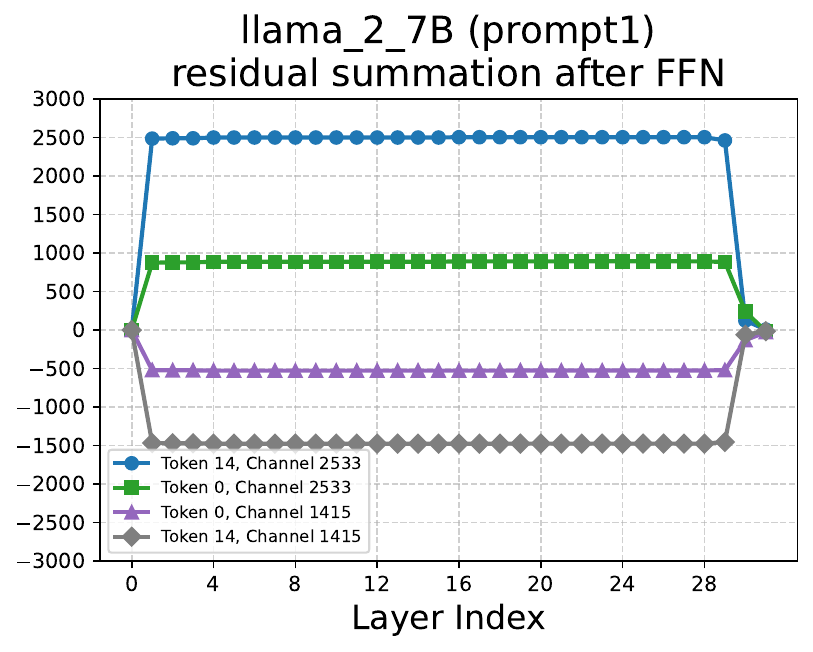}
    \end{subfigure}    
    \begin{subfigure}{0.24\textwidth}
        \centering
    \includegraphics[width=\linewidth]{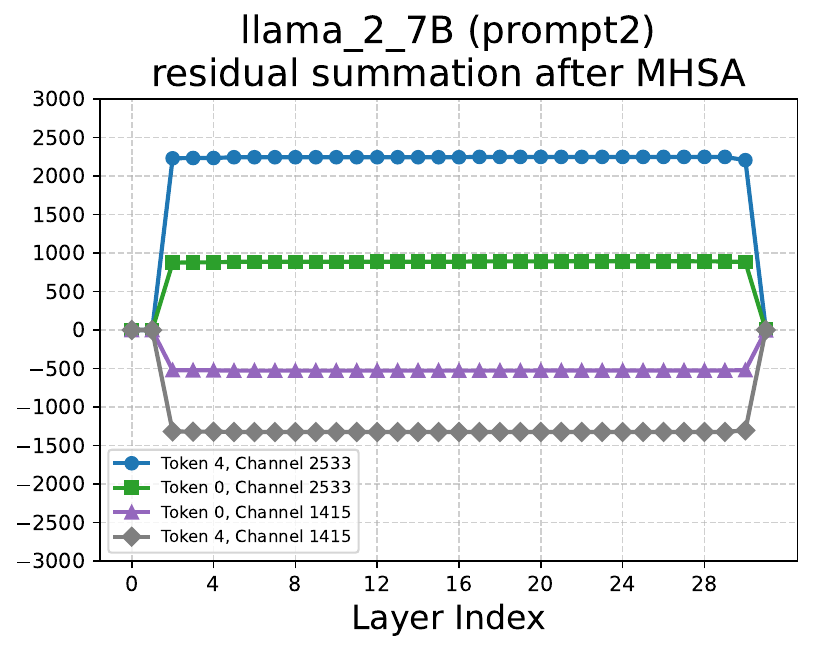}
    \end{subfigure}
    \begin{subfigure}{0.24\textwidth}
        \centering
    \includegraphics[width=\linewidth]{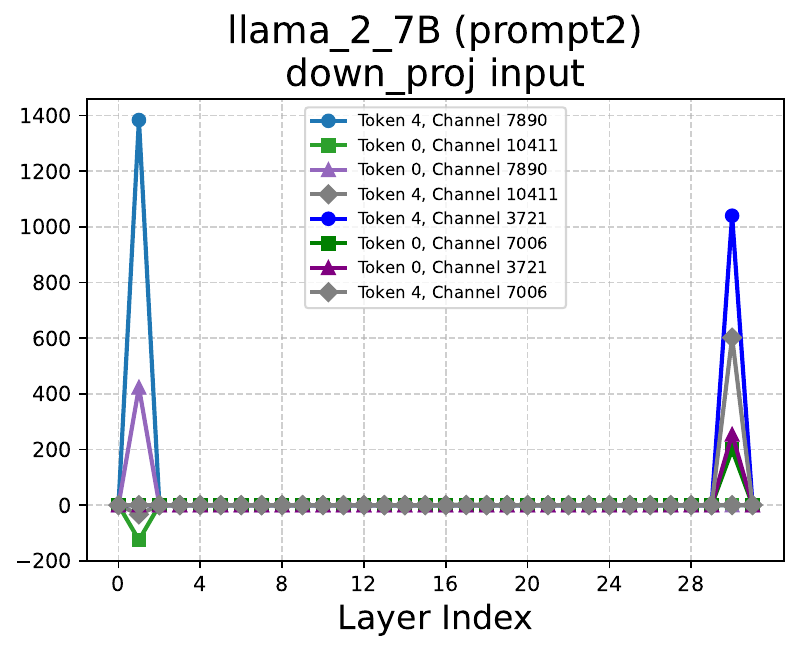}
    \end{subfigure}
    \begin{subfigure}{0.24\textwidth}
        \centering
    \includegraphics[width=\linewidth]{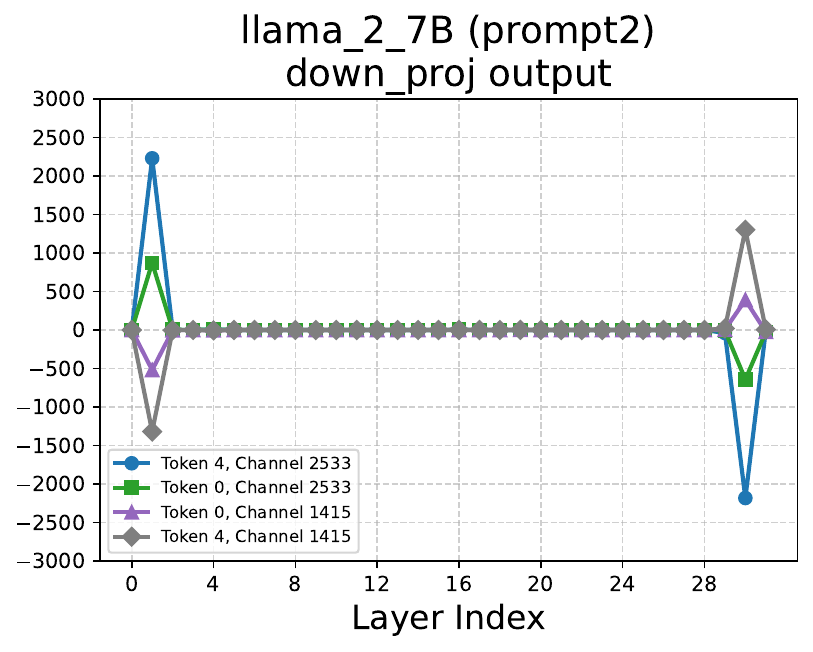}
    \end{subfigure}
    \begin{subfigure}{0.24\textwidth}
        \centering
    \includegraphics[width=\linewidth]{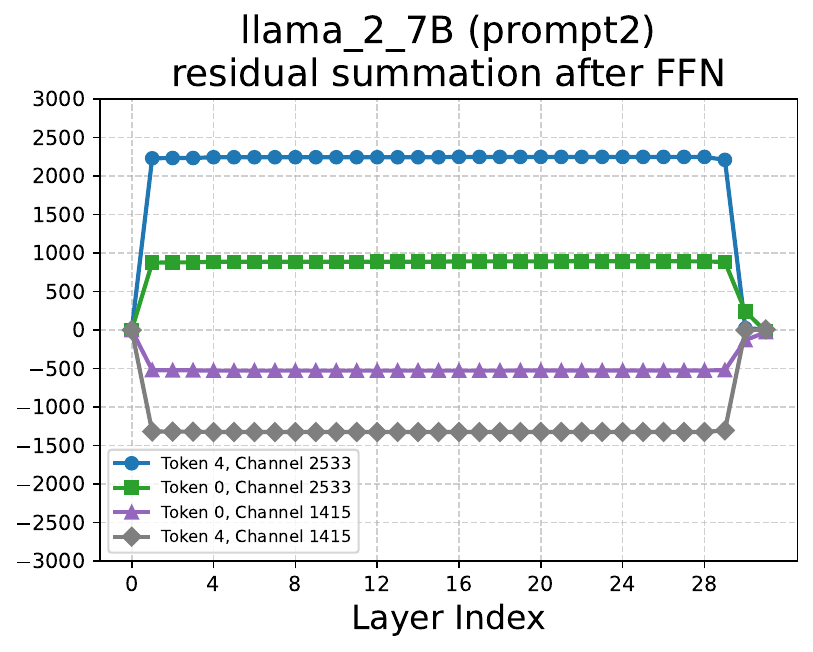}
    \end{subfigure}    
    \begin{subfigure}{0.24\textwidth}
        \centering
    \includegraphics[width=\linewidth]{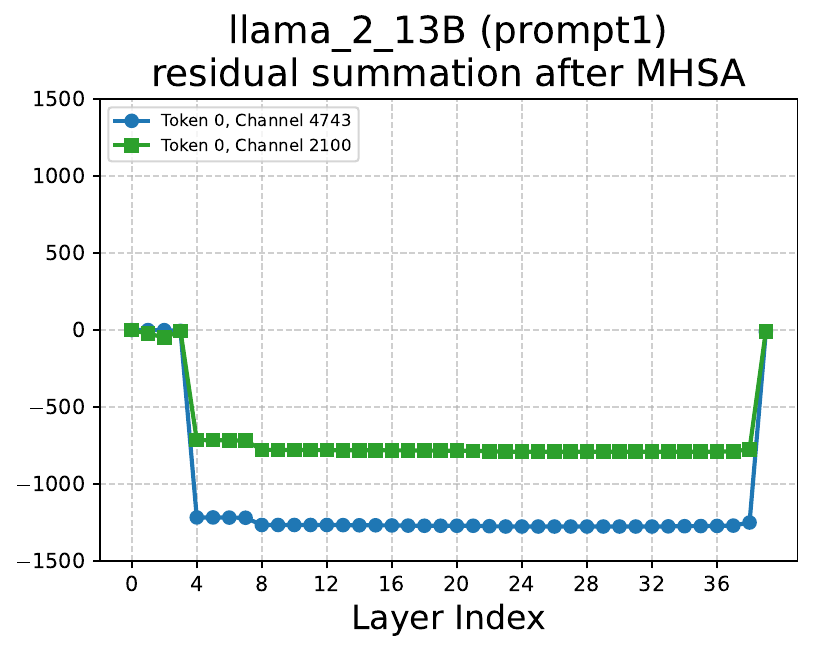}
    \end{subfigure}
    \begin{subfigure}{0.24\textwidth}
        \centering
    \includegraphics[width=\linewidth]{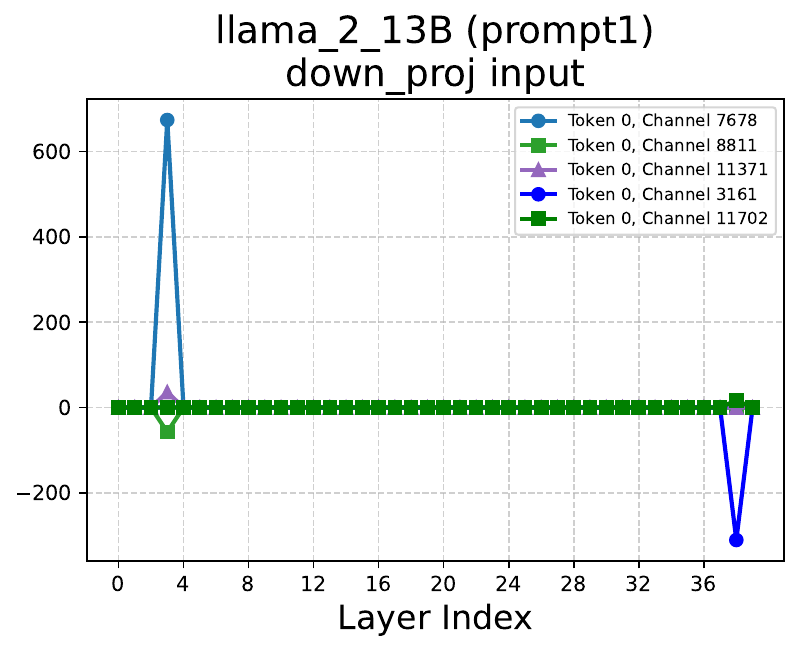}
    \end{subfigure}
    \begin{subfigure}{0.24\textwidth}
        \centering
    \includegraphics[width=\linewidth]{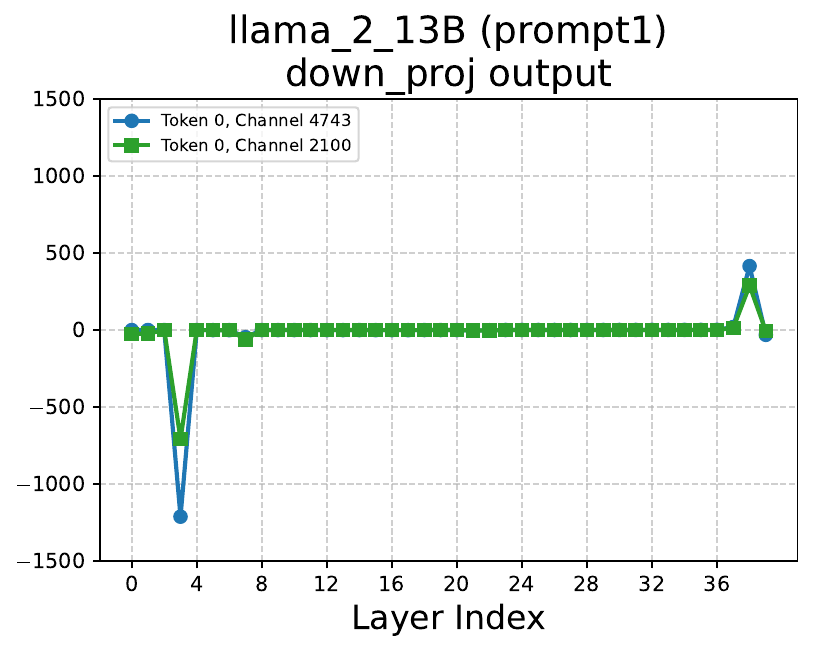}
    \end{subfigure}
    \begin{subfigure}{0.24\textwidth}
        \centering
    \includegraphics[width=\linewidth]{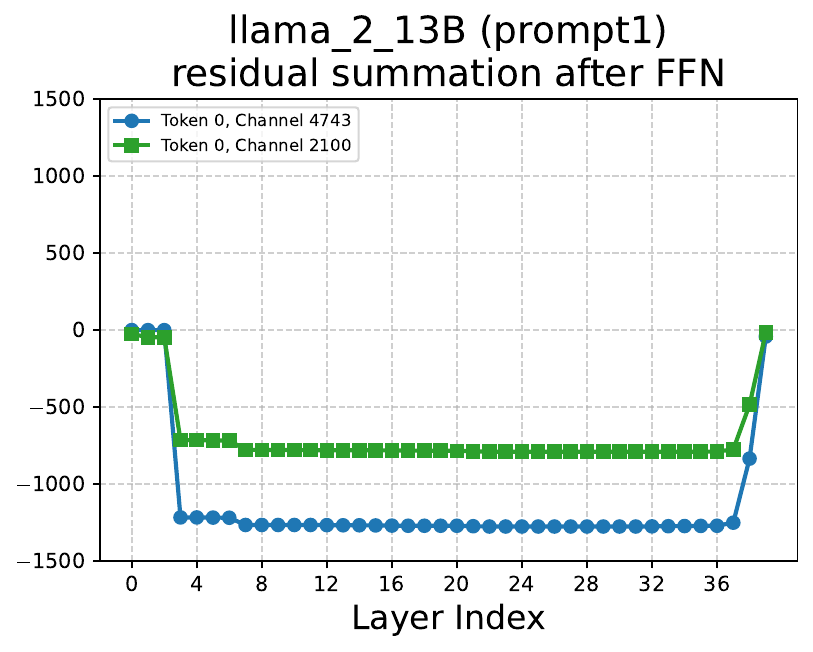}
    \end{subfigure}    
    \begin{subfigure}{0.24\textwidth}
        \centering
    \includegraphics[width=\linewidth]{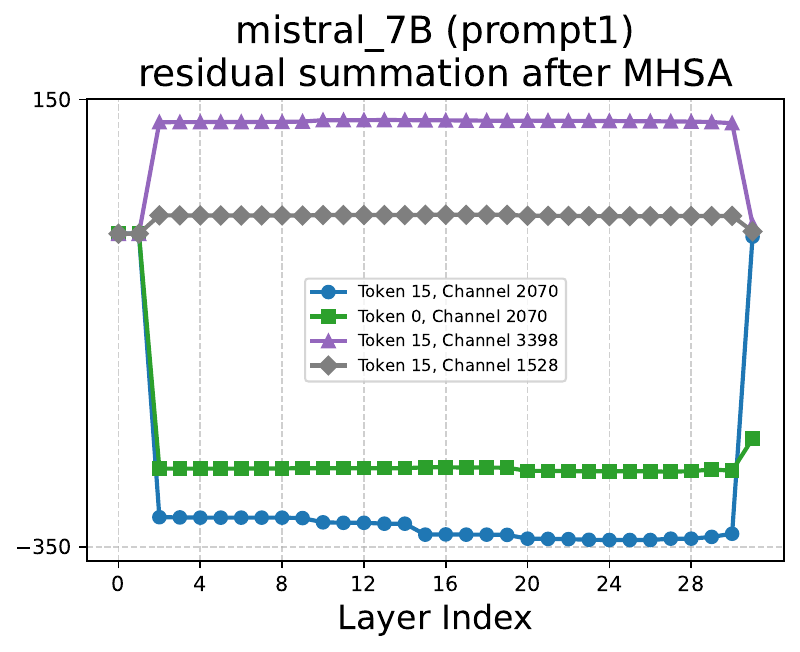}
    \end{subfigure}
    \begin{subfigure}{0.24\textwidth}
        \centering
    \includegraphics[width=\linewidth]{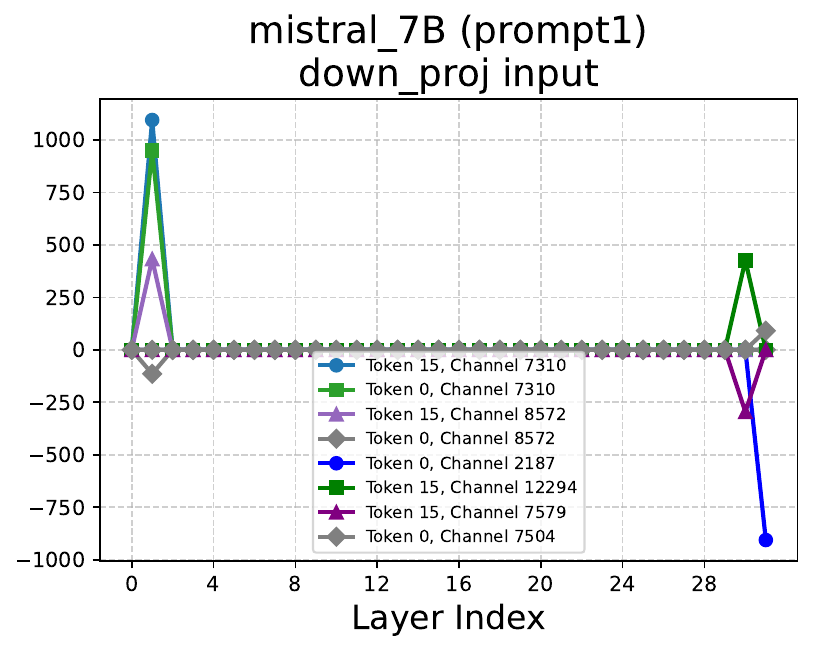}
    \end{subfigure}
    \begin{subfigure}{0.24\textwidth}
        \centering
    \includegraphics[width=\linewidth]{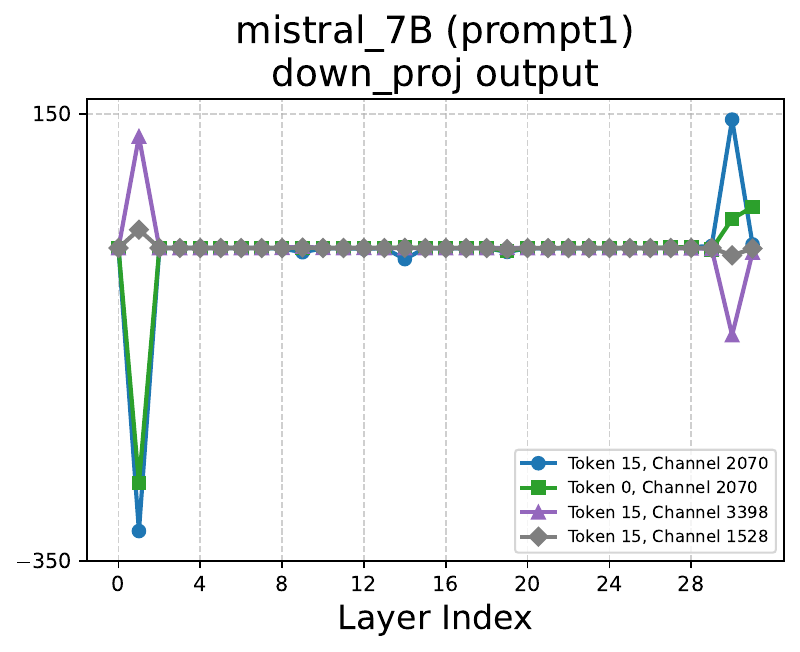}
    \end{subfigure}
    \begin{subfigure}{0.24\textwidth}
        \centering
    \includegraphics[width=\linewidth]{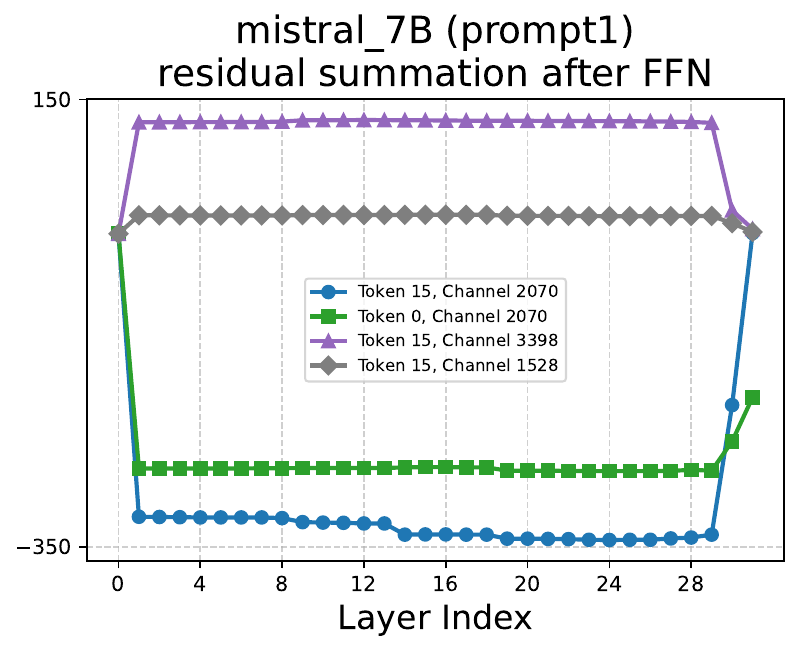}
    \end{subfigure}    
    \begin{subfigure}{0.24\textwidth}
        \centering
    \includegraphics[width=\linewidth]{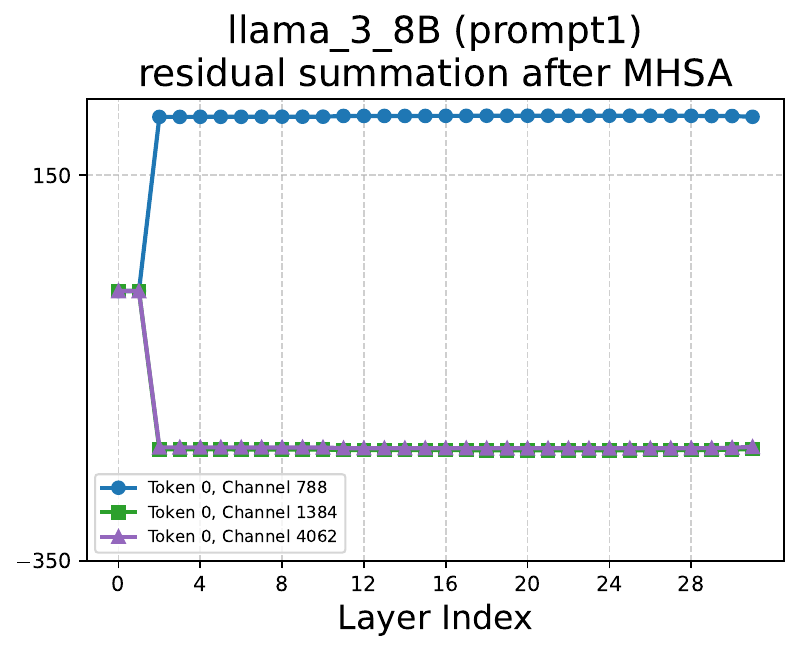}
    \end{subfigure}
    \begin{subfigure}{0.24\textwidth}
        \centering
    \includegraphics[width=\linewidth]{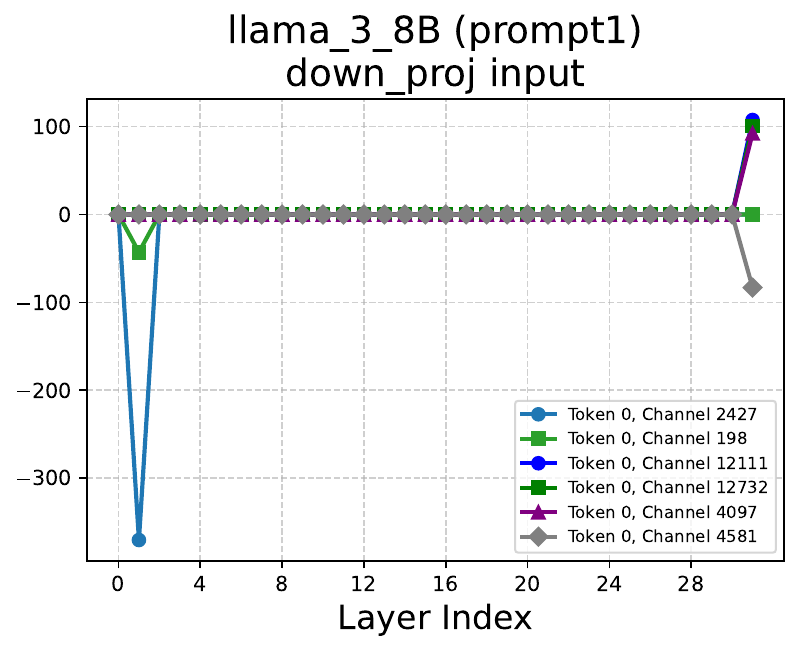}
    \end{subfigure}
    \begin{subfigure}{0.24\textwidth}
        \centering
    \includegraphics[width=\linewidth]{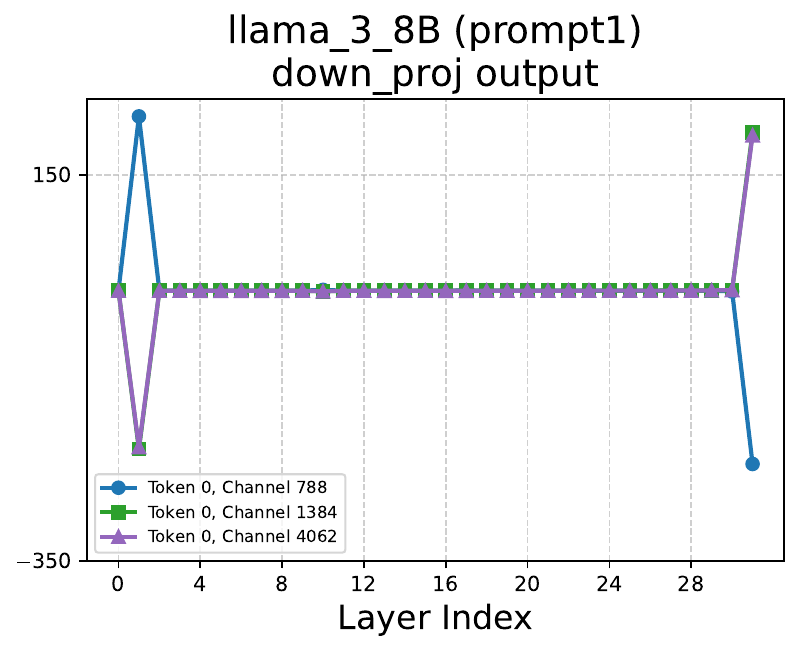}
    \end{subfigure}  
    \begin{subfigure}{0.24\textwidth}
        \centering
    \includegraphics[width=\linewidth]{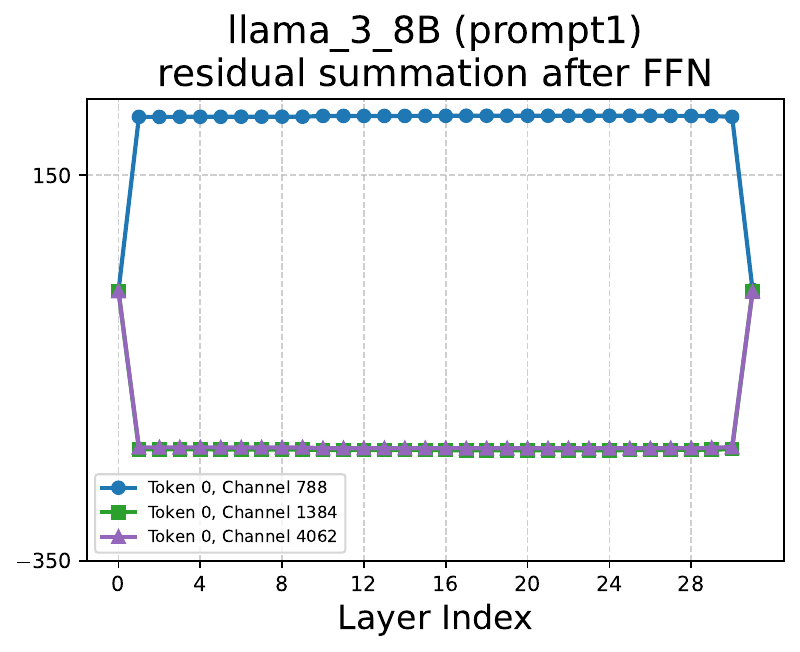}
    \end{subfigure}    
    \begin{subfigure}{0.24\textwidth}
        \centering
    \includegraphics[width=\linewidth]{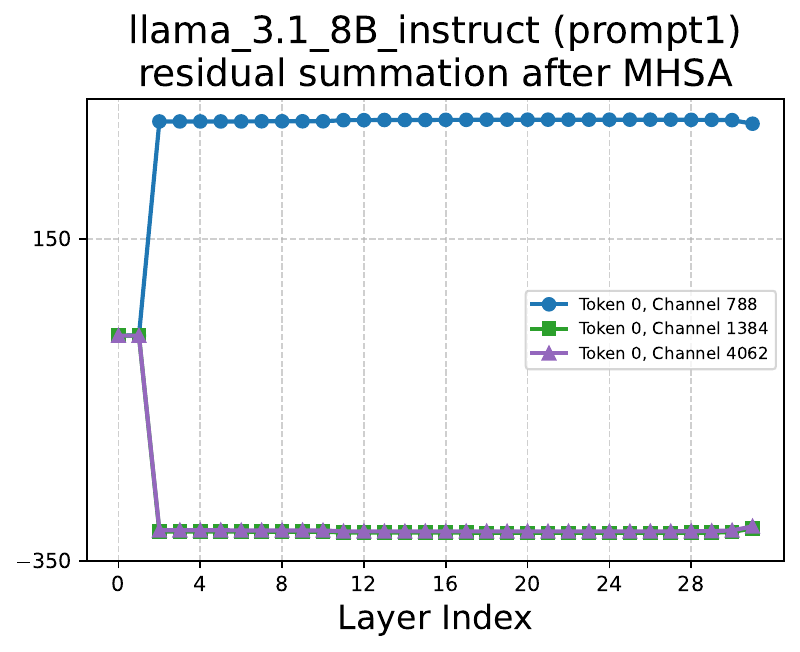}
    \end{subfigure}
    \begin{subfigure}{0.24\textwidth}
        \centering
    \includegraphics[width=\linewidth]{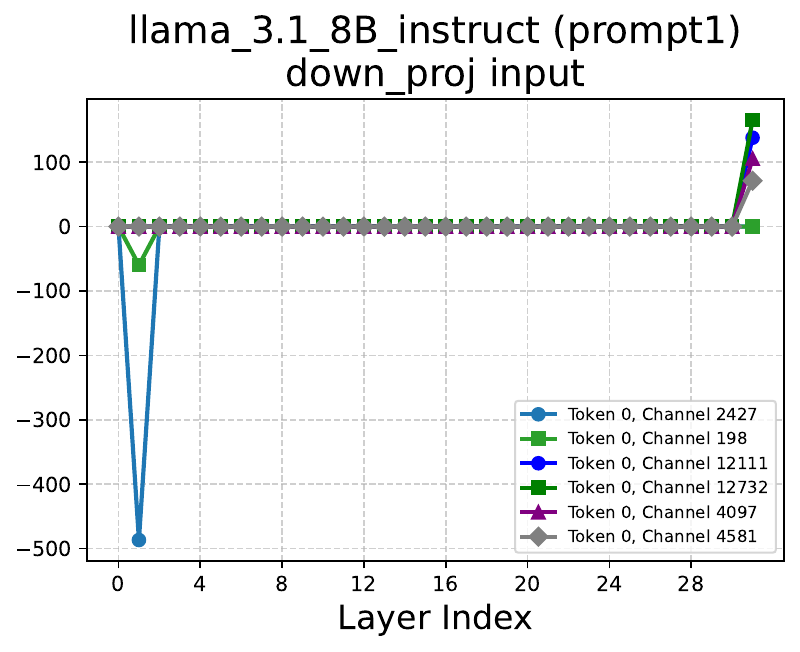}
    \end{subfigure}
    \begin{subfigure}{0.24\textwidth}
        \centering
    \includegraphics[width=\linewidth]{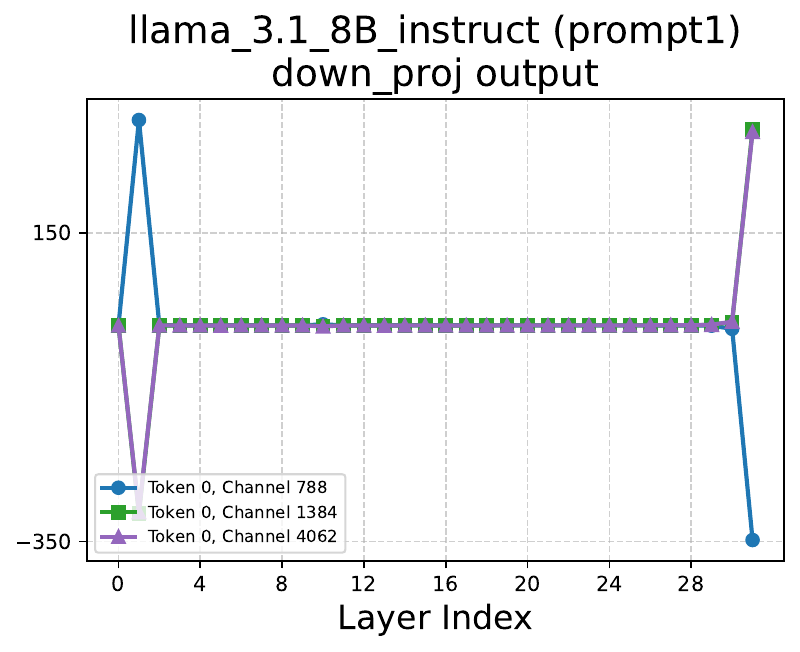}
    \end{subfigure}
    \begin{subfigure}{0.24\textwidth}
        \centering
    \includegraphics[width=\linewidth]{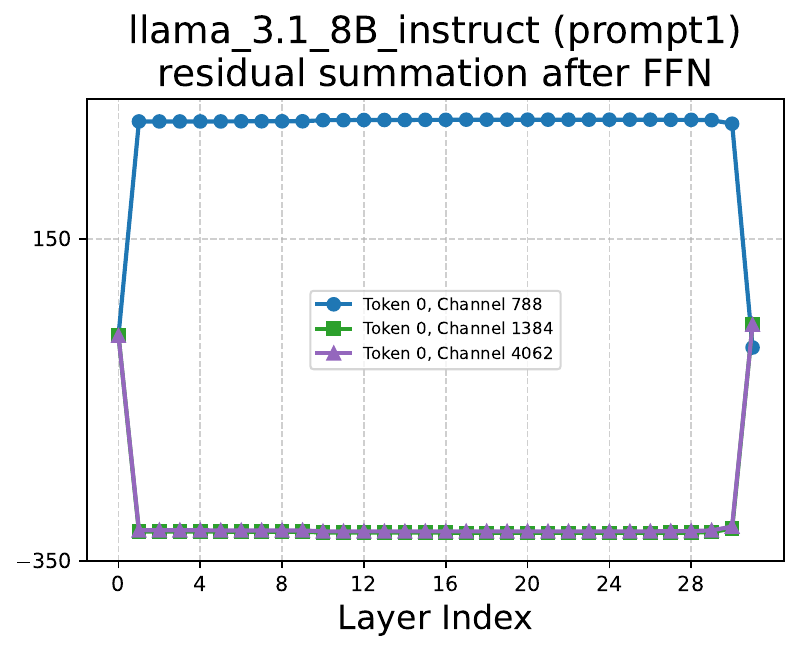}
    \end{subfigure}    
    \begin{subfigure}{0.24\textwidth}
        \centering
    \includegraphics[width=\linewidth]{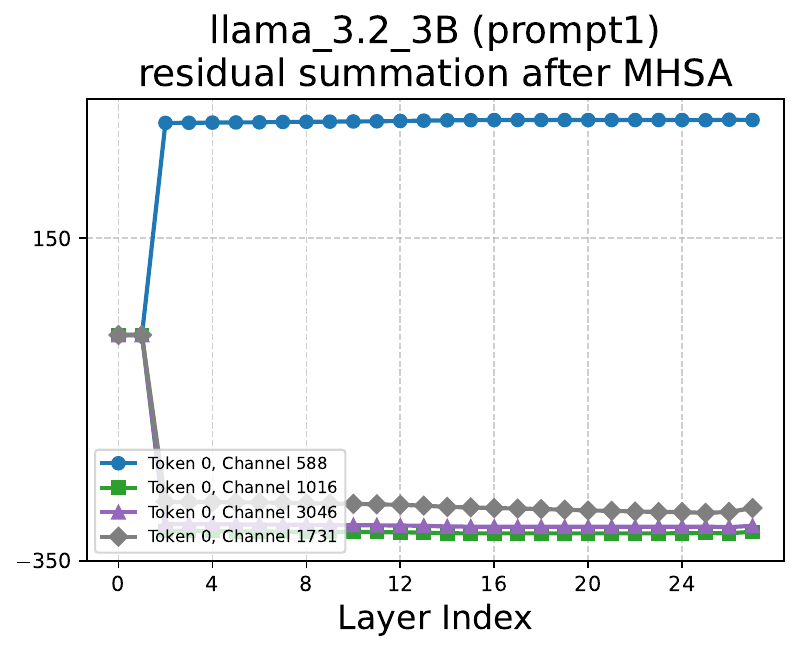}
    \end{subfigure}
    \begin{subfigure}{0.24\textwidth}
        \centering
    \includegraphics[width=\linewidth]{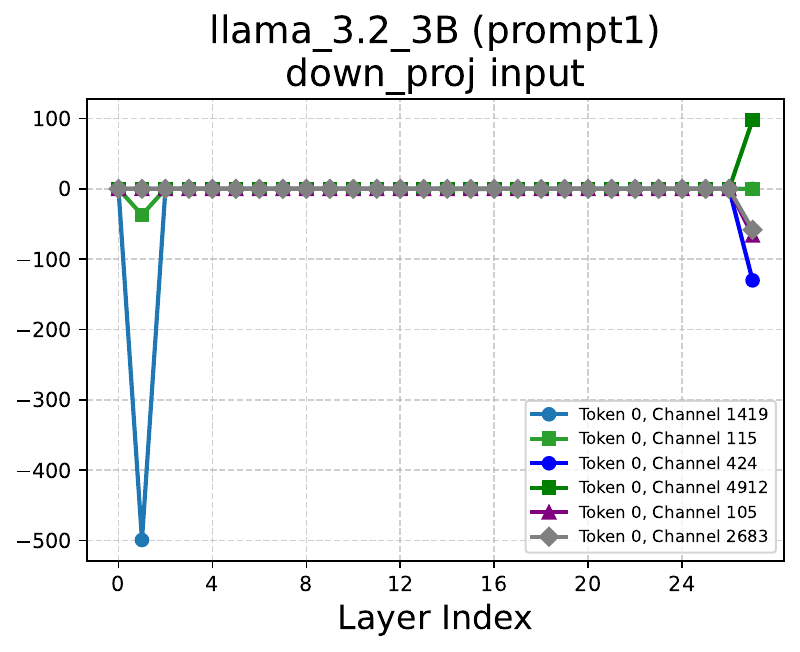}
    \end{subfigure}
    \begin{subfigure}{0.24\textwidth}
        \centering
    \includegraphics[width=\linewidth]{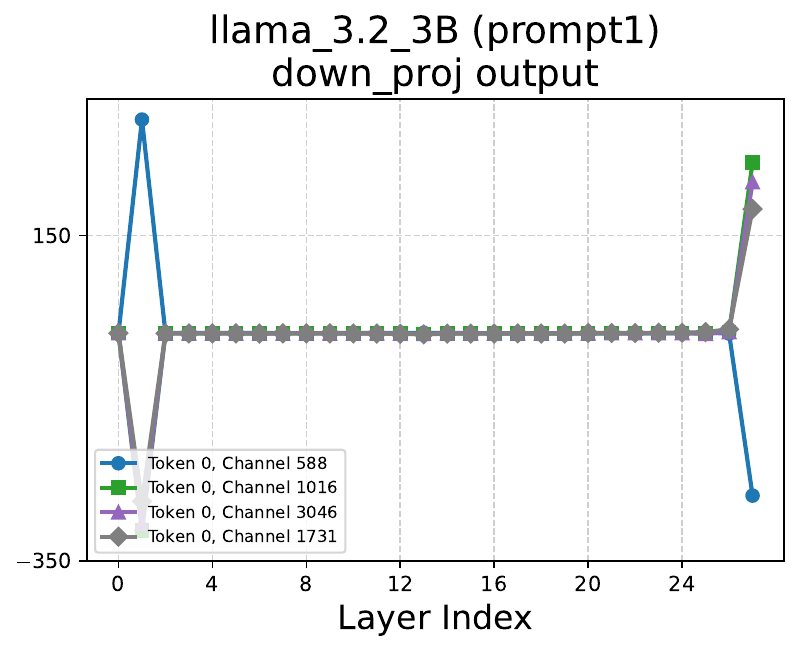}
    \end{subfigure}
    \begin{subfigure}{0.24\textwidth}
        \centering
    \includegraphics[width=\linewidth]{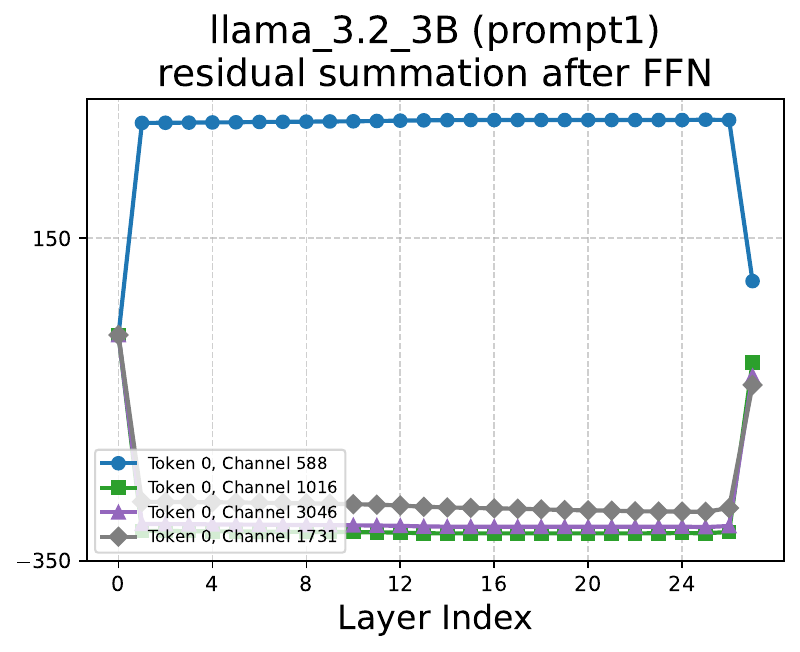}
    \end{subfigure}    
    \caption{Distribution of extreme activation outliers across decoder layers in multiple models.}
\label{outlier_appendix}
\end{figure}
\begin{figure}[t]
    \centering    
    \begin{subfigure}{0.32\textwidth}
        \centering
    \includegraphics[width=\linewidth]{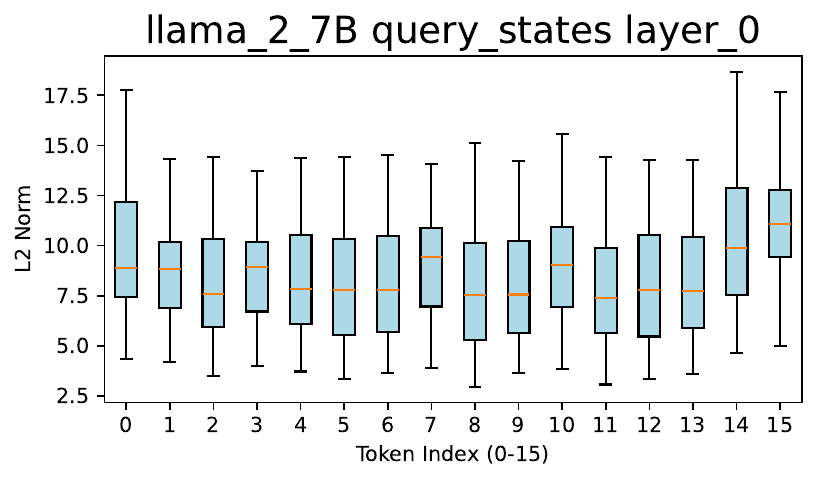}
    \end{subfigure}
    \begin{subfigure}{0.32\textwidth}
        \centering
    \includegraphics[width=\linewidth]{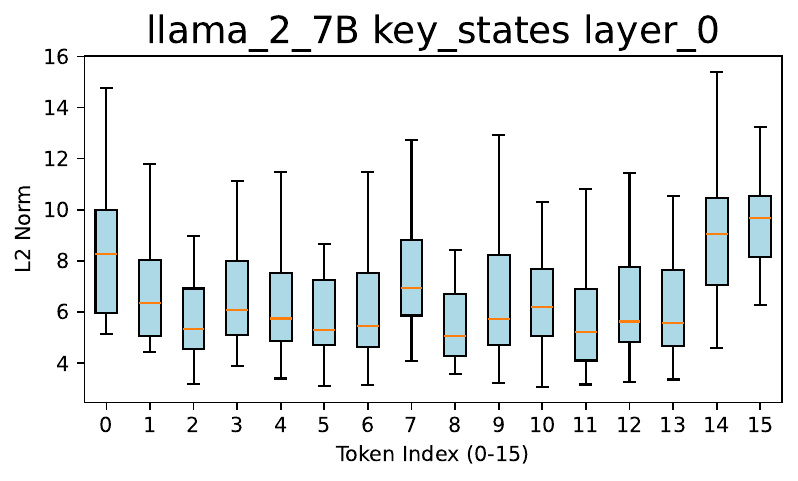}
    \end{subfigure}
    \begin{subfigure}{0.32\textwidth}
        \centering
    \includegraphics[width=\linewidth]{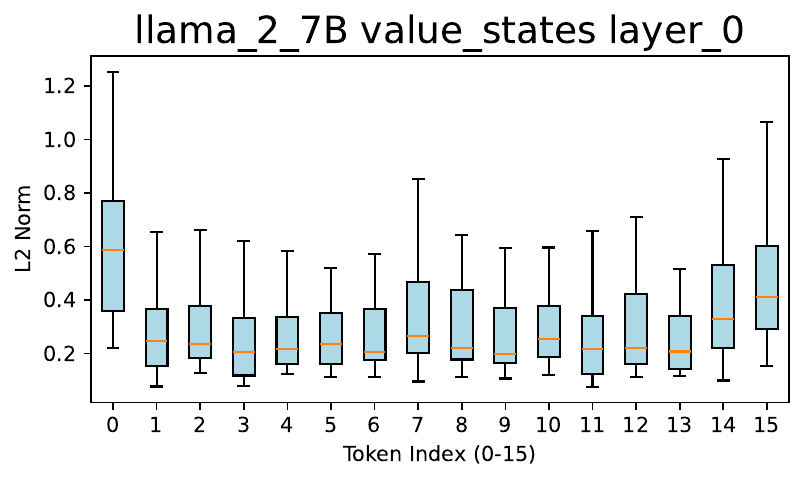}
    \end{subfigure}
    \begin{subfigure}{0.32\textwidth}
        \centering
    \includegraphics[width=\linewidth]{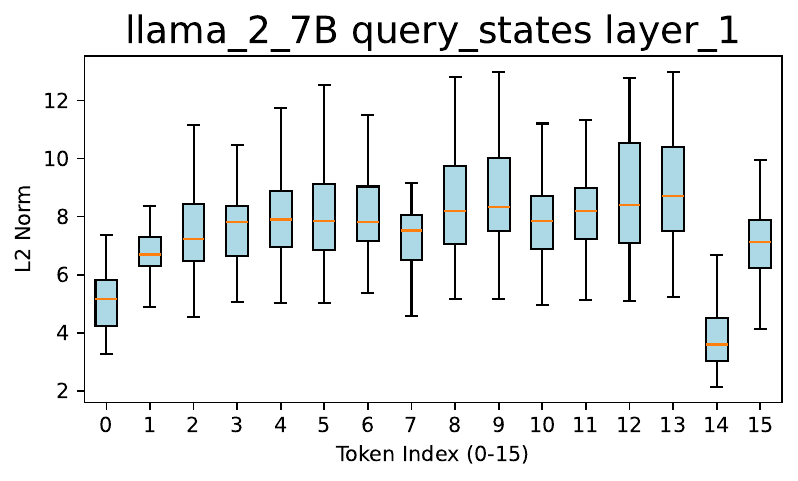}
    \end{subfigure}
    \begin{subfigure}{0.32\textwidth}
        \centering
    \includegraphics[width=\linewidth]{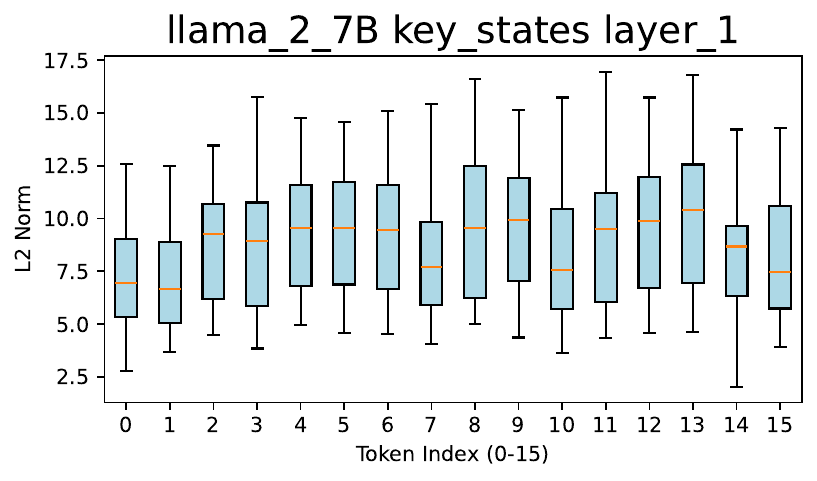}
    \end{subfigure}
    \begin{subfigure}{0.32\textwidth}
        \centering
    \includegraphics[width=\linewidth]{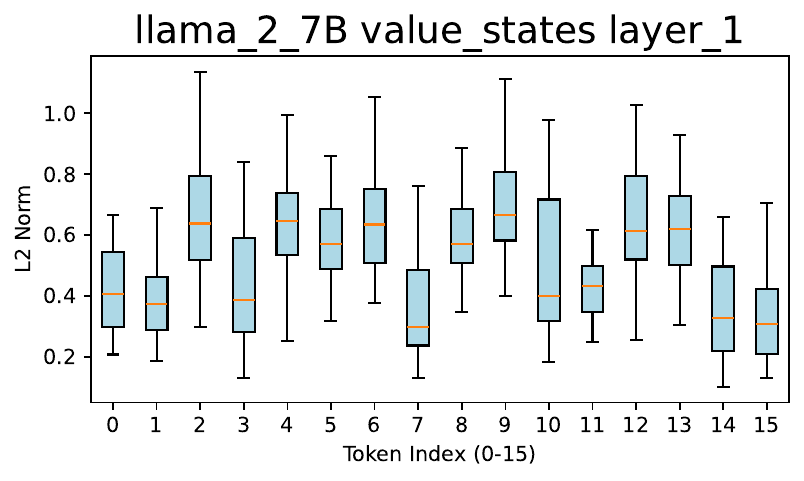}
    \end{subfigure}
    \begin{subfigure}{0.32\textwidth}
        \centering
    \includegraphics[width=\linewidth]{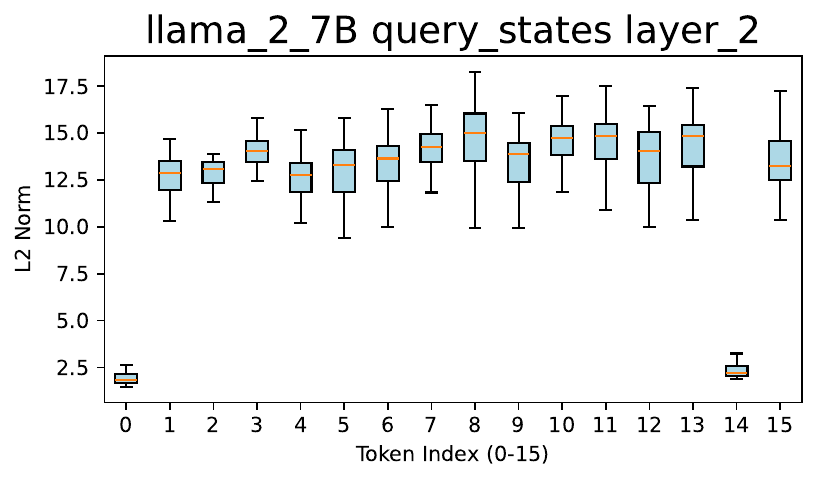}
    \end{subfigure}
    \begin{subfigure}{0.32\textwidth}
        \centering
    \includegraphics[width=\linewidth]{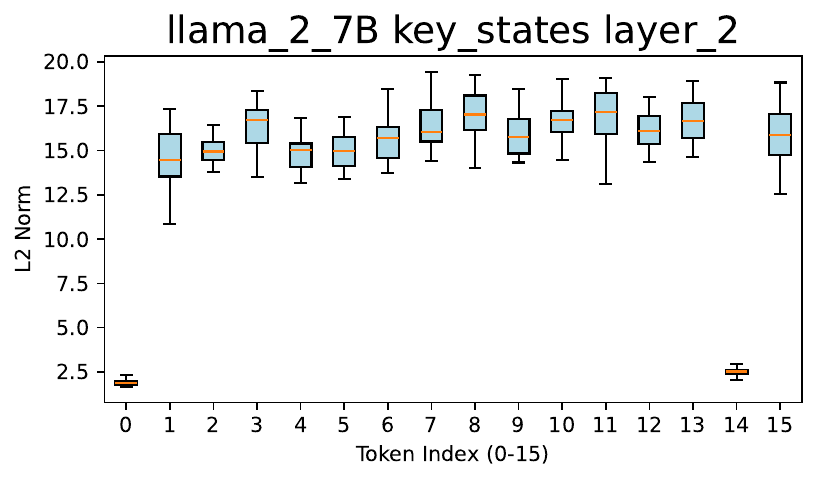}
    \end{subfigure}
    \begin{subfigure}{0.32\textwidth}
        \centering
    \includegraphics[width=\linewidth]{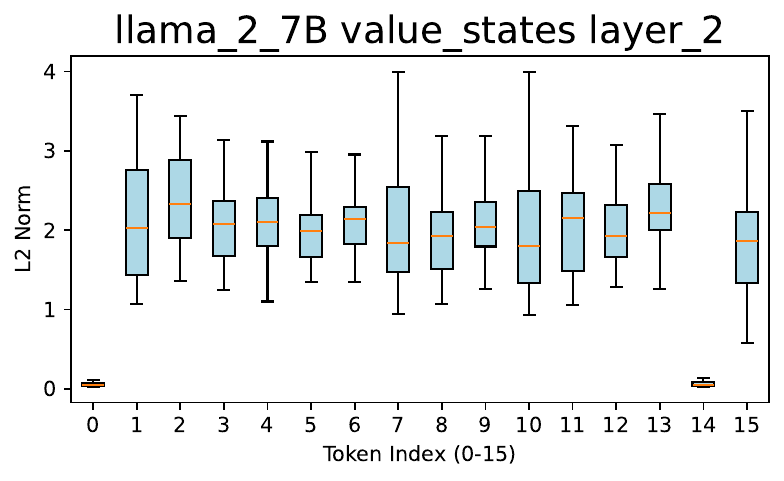}
    \end{subfigure}
    \begin{subfigure}{0.32\textwidth}
        \centering
    \includegraphics[width=\linewidth]{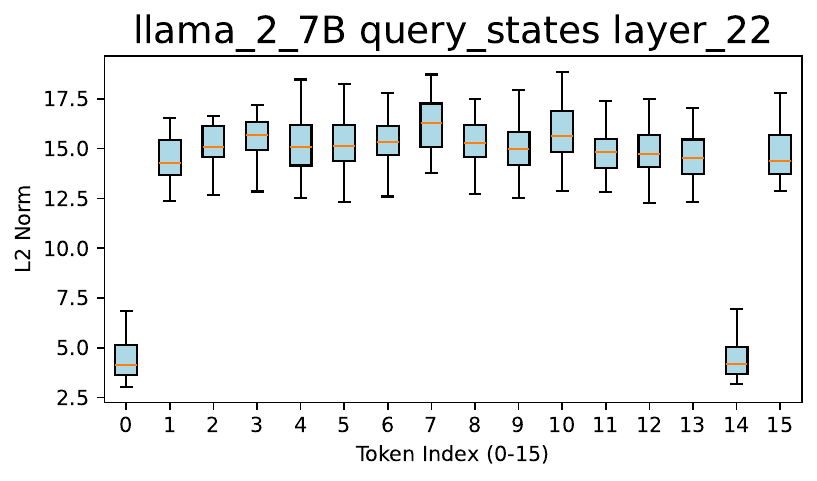}
    \end{subfigure}
    \begin{subfigure}{0.32\textwidth}
        \centering
    \includegraphics[width=\linewidth]{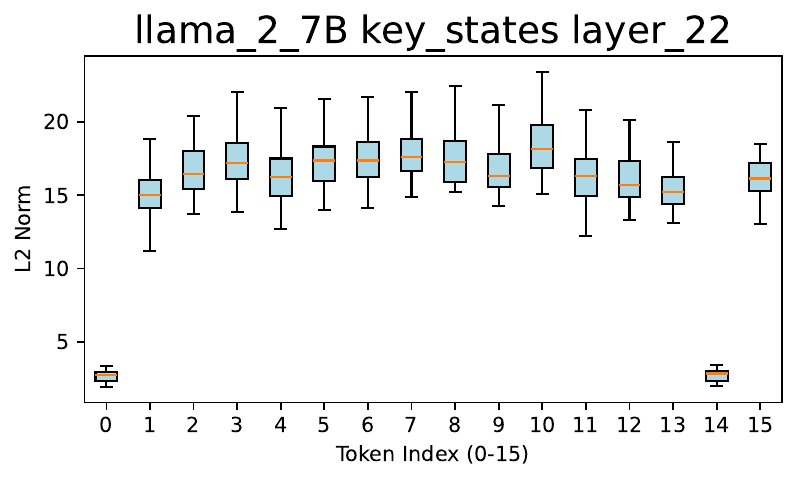}
    \end{subfigure}
    \begin{subfigure}{0.32\textwidth}
        \centering
    \includegraphics[width=\linewidth]{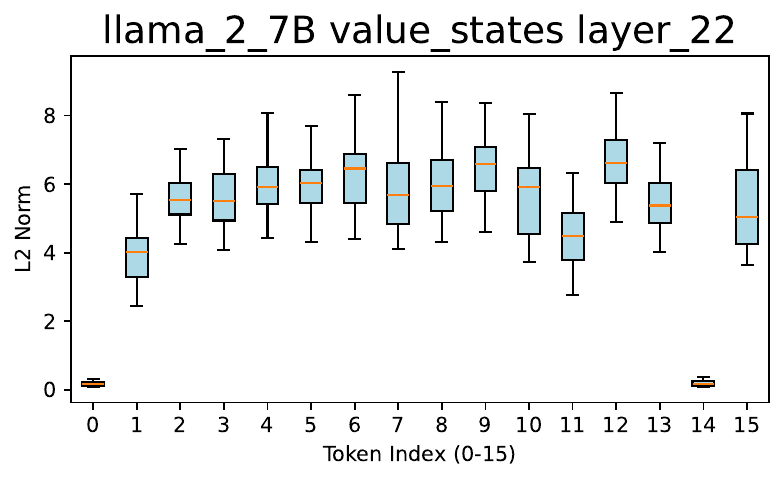}
    \end{subfigure}
    \begin{subfigure}{0.32\textwidth}
        \centering
    \includegraphics[width=\linewidth]{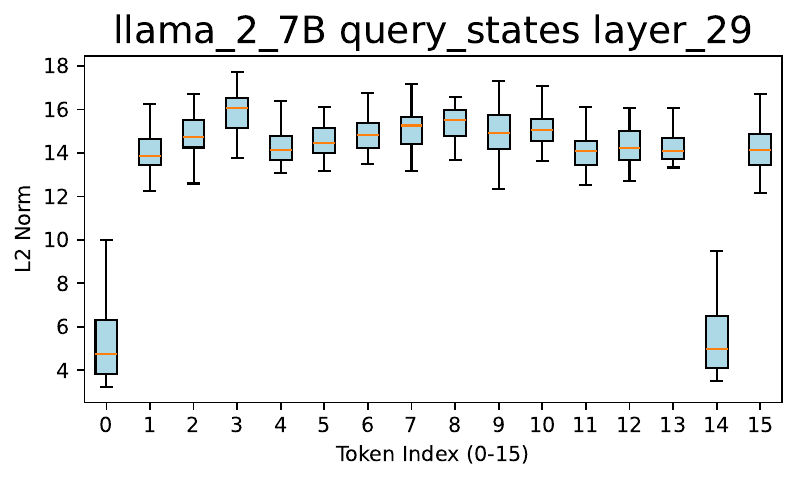}
    \end{subfigure}
    \begin{subfigure}{0.32\textwidth}
        \centering
    \includegraphics[width=\linewidth]{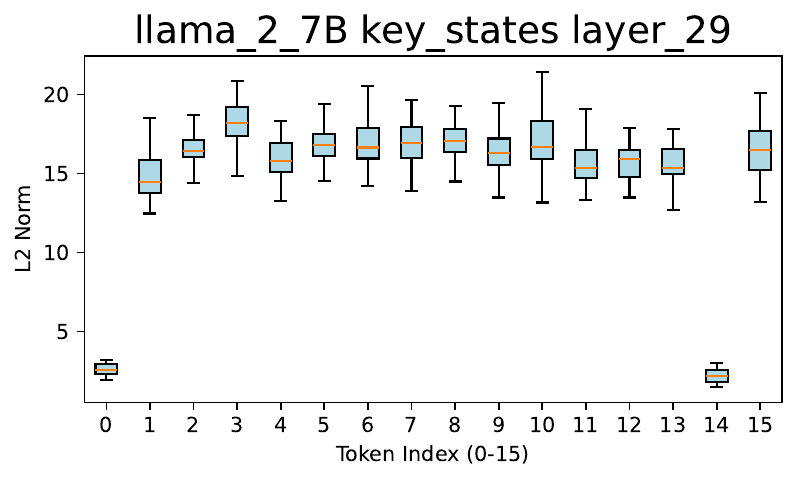}
    \end{subfigure}
    \begin{subfigure}{0.32\textwidth}
        \centering
    \includegraphics[width=\linewidth]{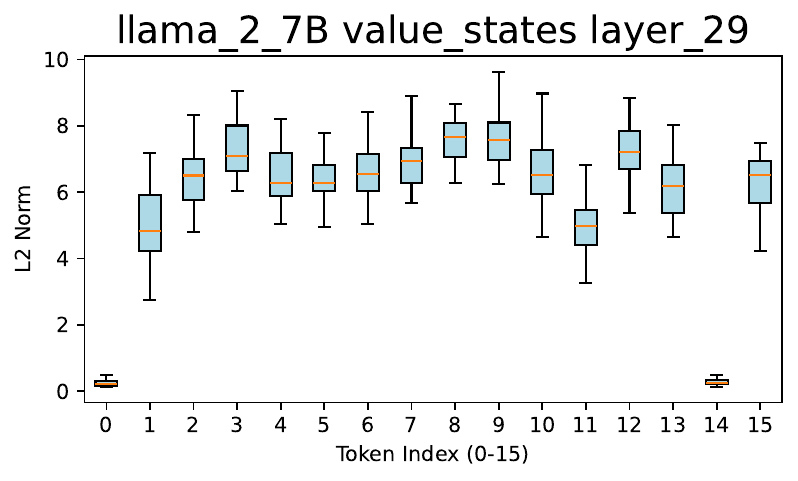}
    \end{subfigure}
    \begin{subfigure}{0.32\textwidth}
        \centering
    \includegraphics[width=\linewidth]{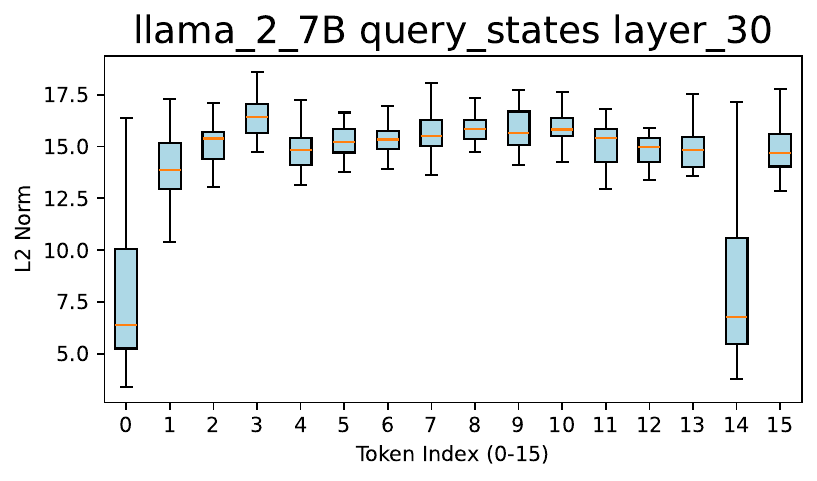}
    \end{subfigure}
    \begin{subfigure}{0.32\textwidth}
        \centering
    \includegraphics[width=\linewidth]{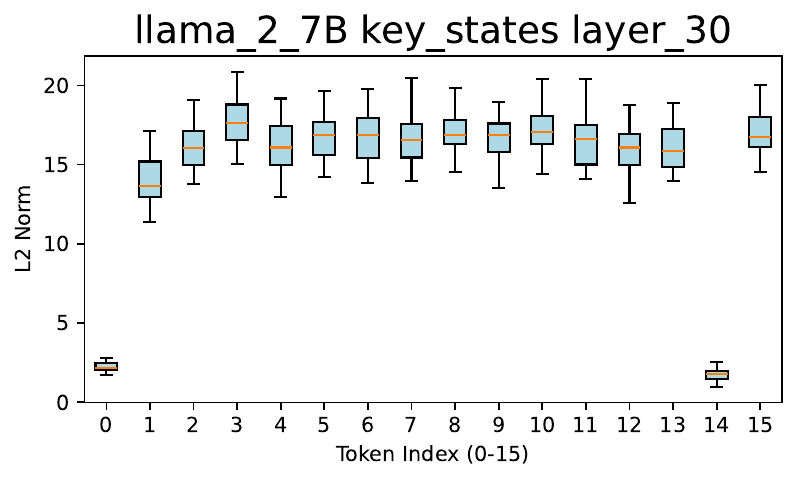}
    \end{subfigure}
    \begin{subfigure}{0.32\textwidth}
        \centering
    \includegraphics[width=\linewidth]{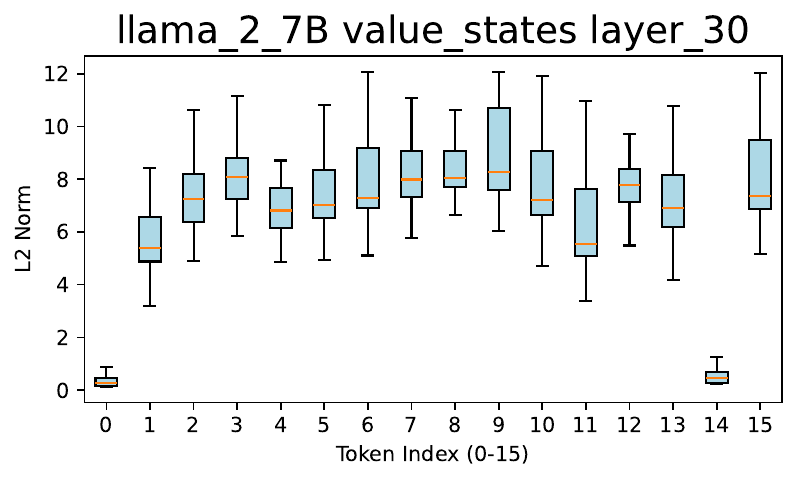}
    \end{subfigure}
    \begin{subfigure}{0.32\textwidth}
        \centering
    \includegraphics[width=\linewidth]{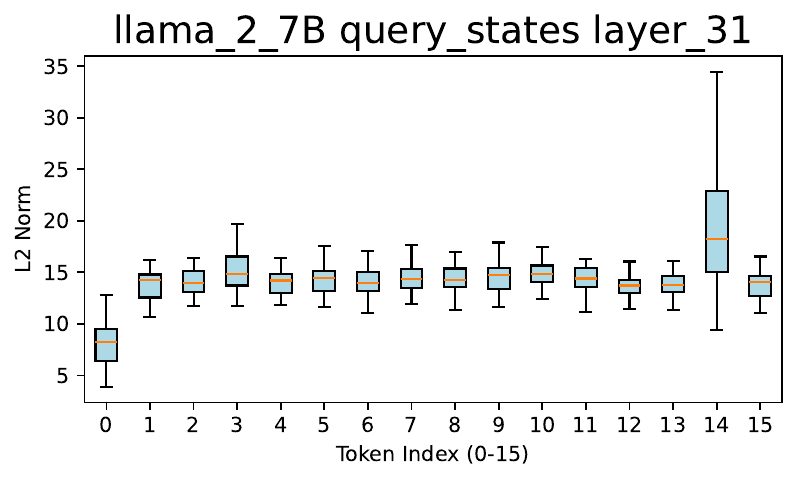}
    \end{subfigure}
    \begin{subfigure}{0.32\textwidth}
        \centering
    \includegraphics[width=\linewidth]{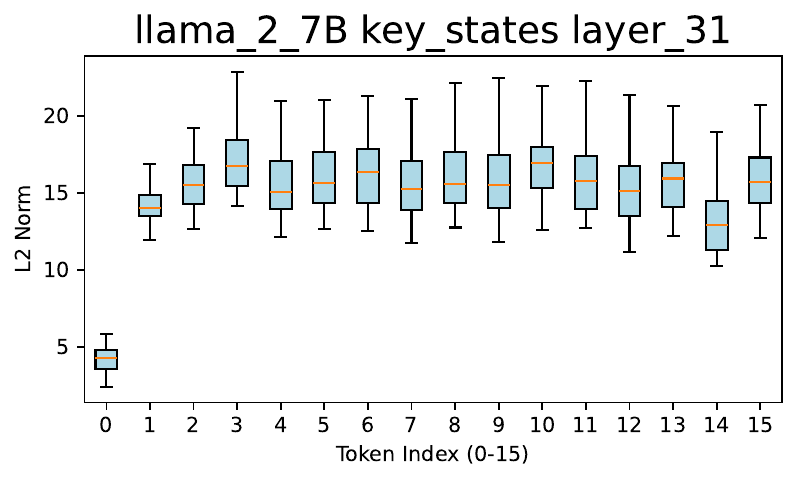}
    \end{subfigure}
    \begin{subfigure}{0.32\textwidth}
        \centering
    \includegraphics[width=\linewidth]{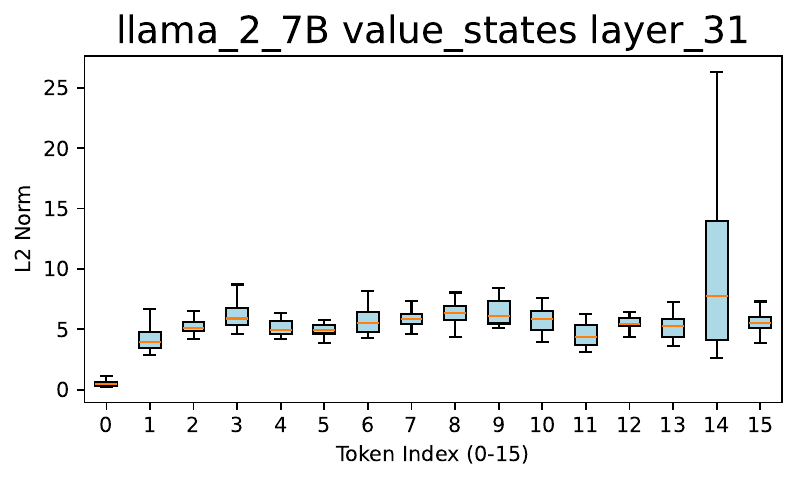}
    \end{subfigure}
    \caption{L2 norm distributions of Queries, Keys, and Values for LLaMA2-7B using Prompt 1, with attention sinks occurring in layers beyond layer 0 and 1, at tokens 0 and 14.}
\label{QKV_appendix_0}
\end{figure}
\begin{figure}[t]
    \centering    
    \begin{subfigure}{0.37\textwidth}
        \centering
    \includegraphics[width=\linewidth]{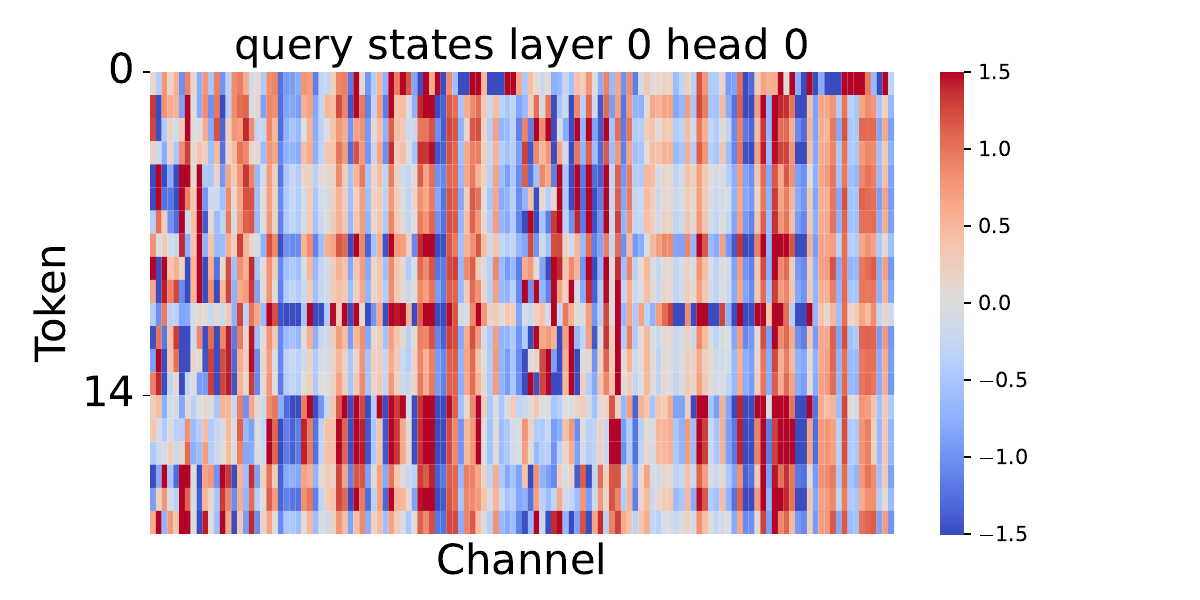}
    \end{subfigure}
    \hspace{-11mm} 
    \begin{subfigure}{0.37\textwidth}
        \centering
    \includegraphics[width=\linewidth]{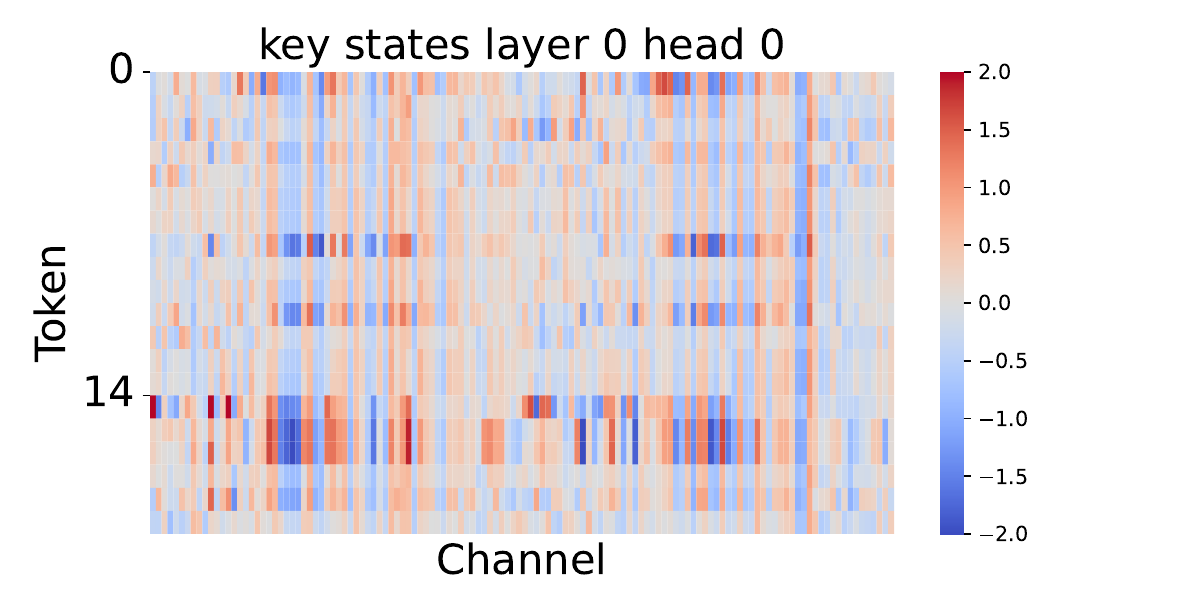}
    \end{subfigure}
    \hspace{-11mm} 
    \begin{subfigure}{0.37\textwidth}
        \centering
    \includegraphics[width=\linewidth]{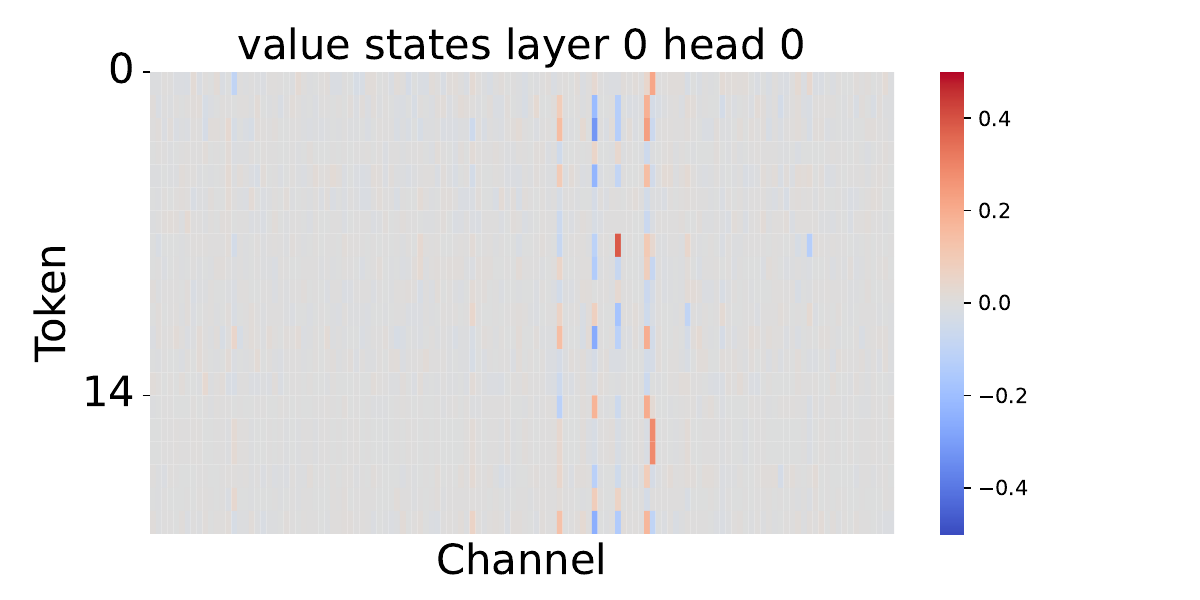}
    \end{subfigure}
    \begin{subfigure}{0.37\textwidth}
        \centering
    \includegraphics[width=\linewidth]{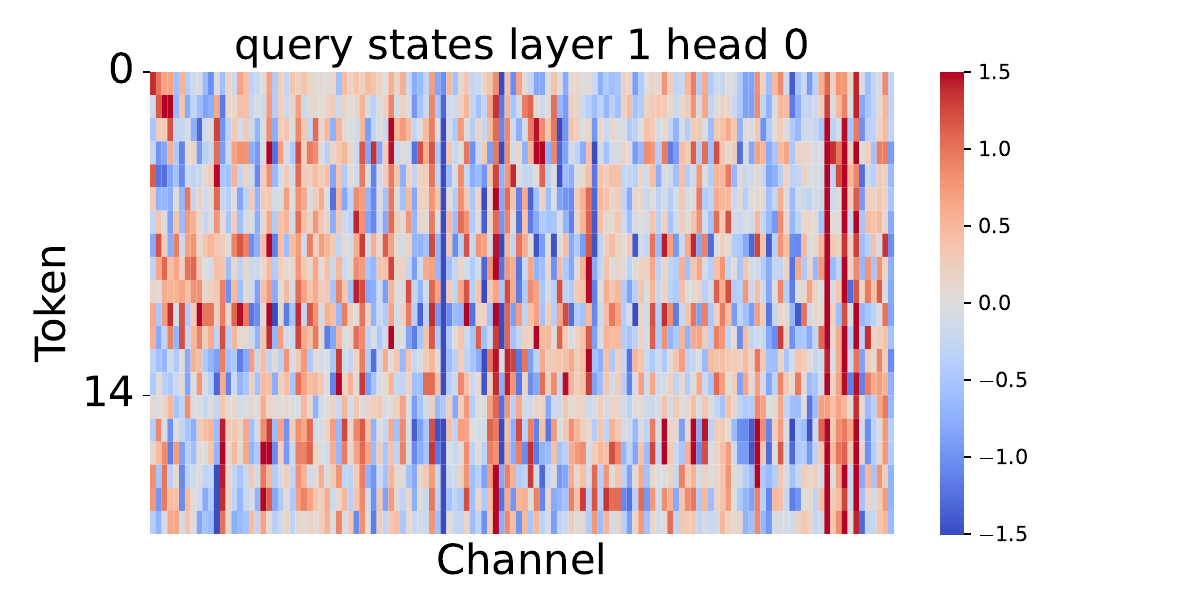}
    \end{subfigure}
    \hspace{-11mm} 
    \begin{subfigure}{0.37\textwidth}
        \centering
    \includegraphics[width=\linewidth]{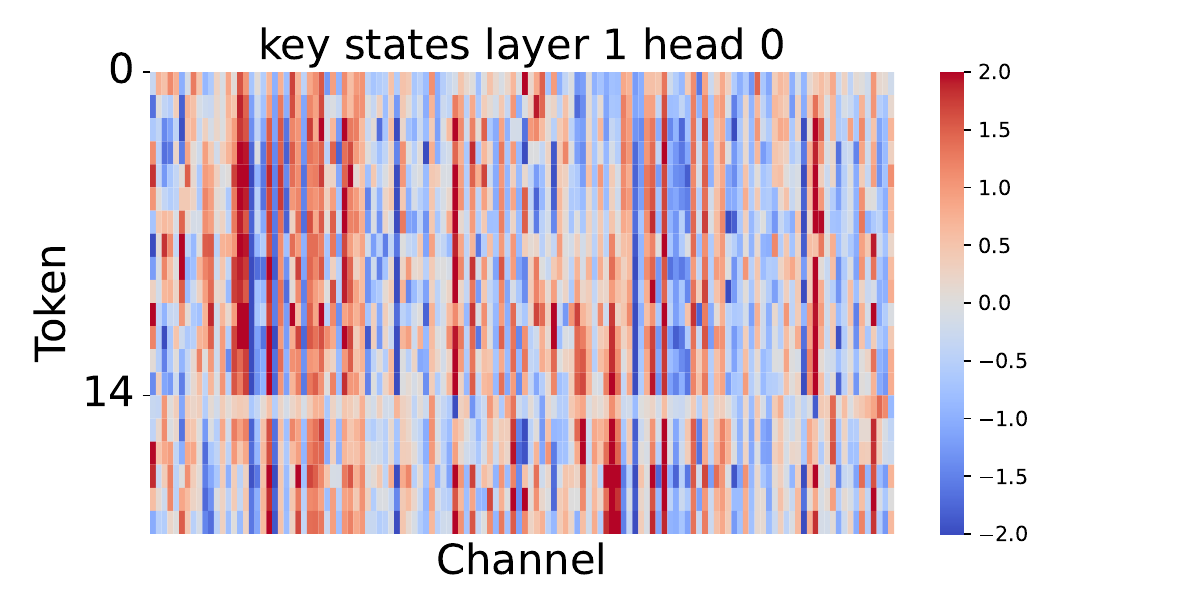}
    \end{subfigure}
    \hspace{-11mm} 
    \begin{subfigure}{0.37\textwidth}
        \centering
    \includegraphics[width=\linewidth]{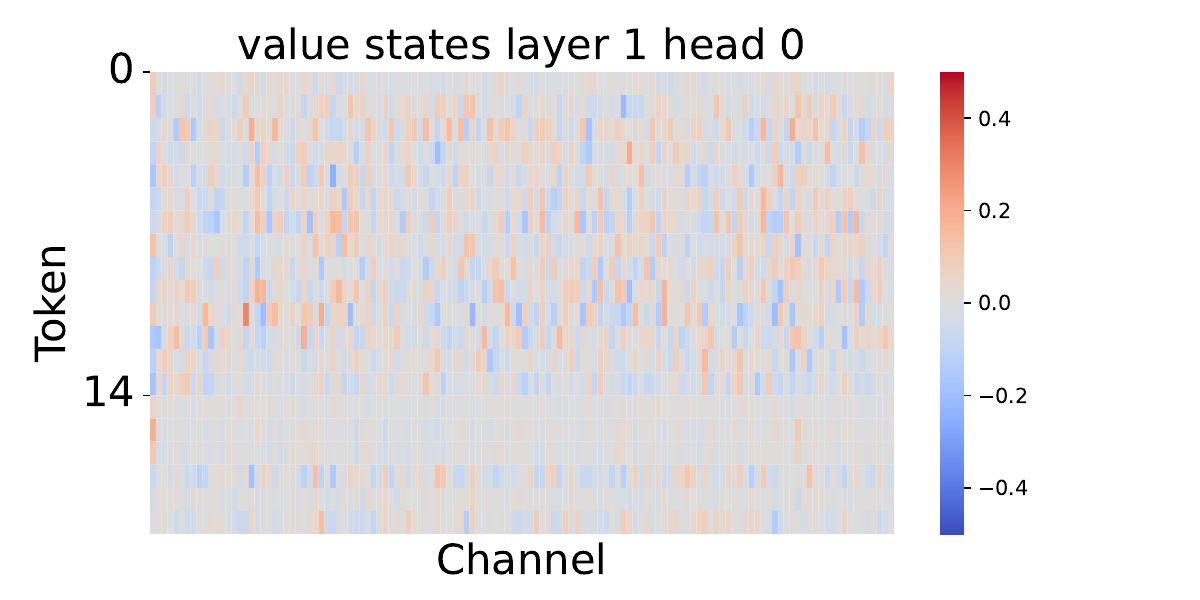}
    \end{subfigure}
    \begin{subfigure}{0.37\textwidth}
        \centering
    \includegraphics[width=\linewidth]{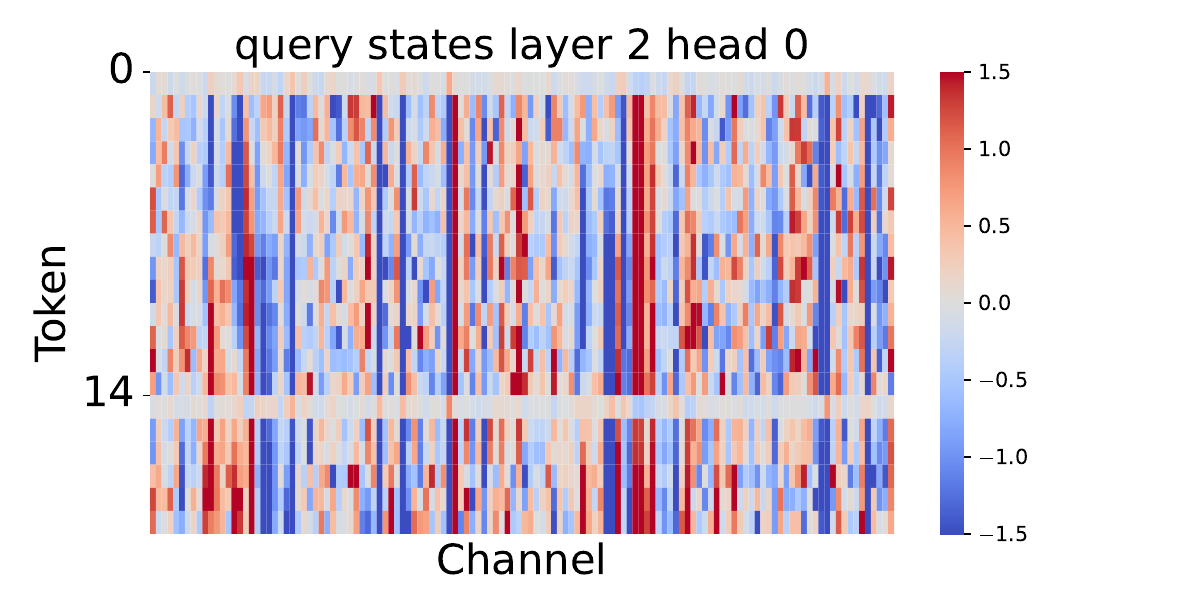}
    \end{subfigure}
    \hspace{-11mm} 
    \begin{subfigure}{0.37\textwidth}
        \centering
    \includegraphics[width=\linewidth]{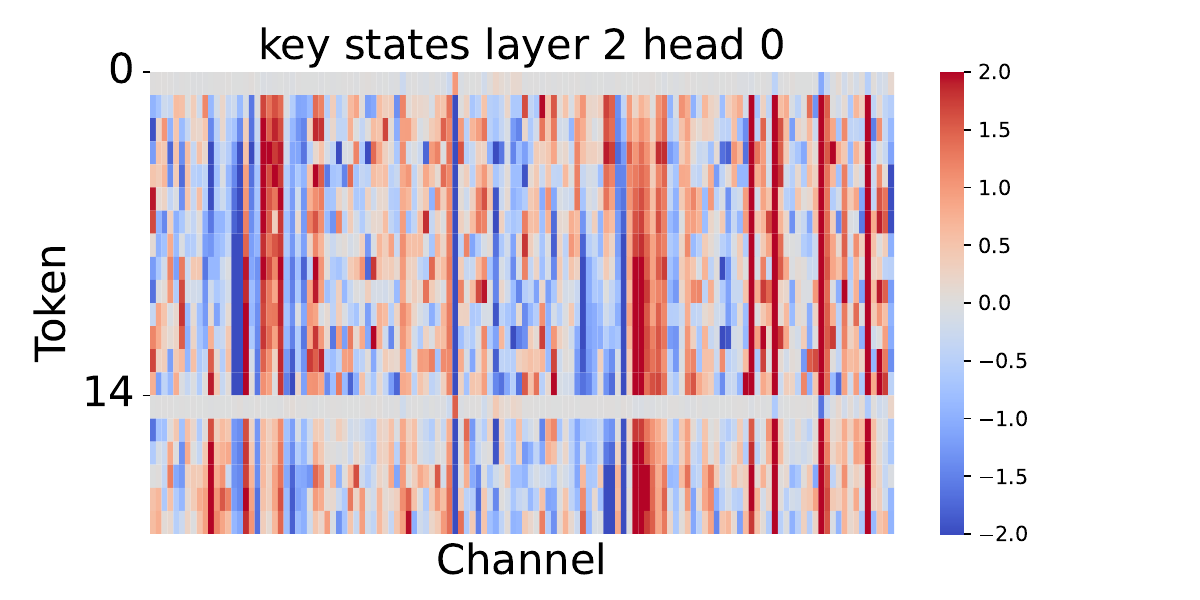}
    \end{subfigure}
    \hspace{-11mm} 
    \begin{subfigure}{0.37\textwidth}
        \centering
    \includegraphics[width=\linewidth]{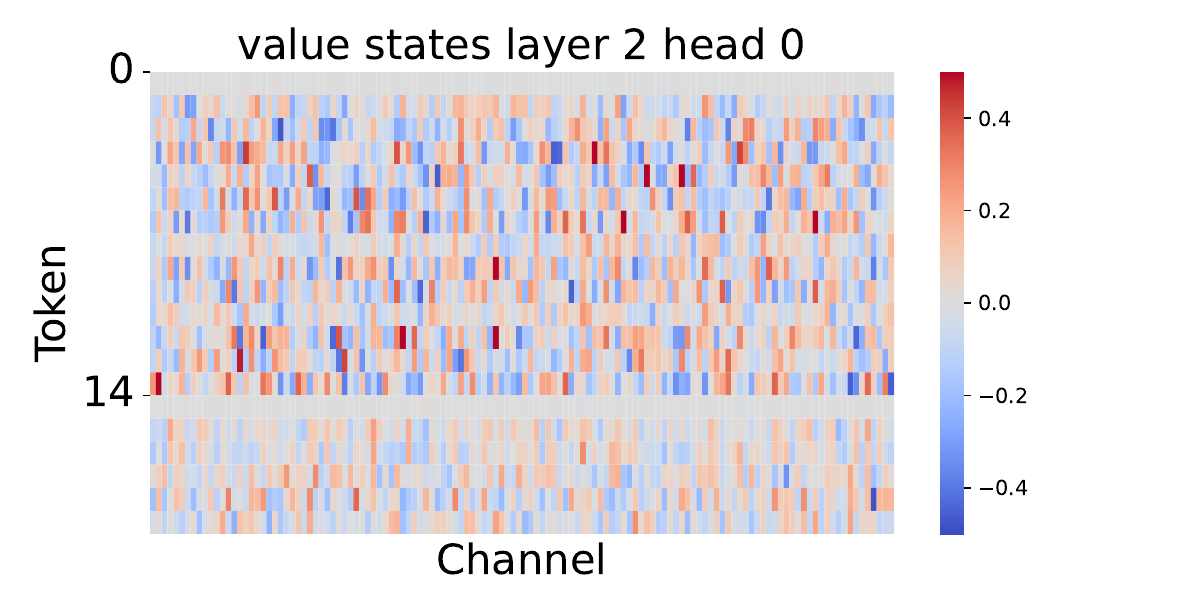}
    \end{subfigure}
    \begin{subfigure}{0.37\textwidth}
        \centering
    \includegraphics[width=\linewidth]{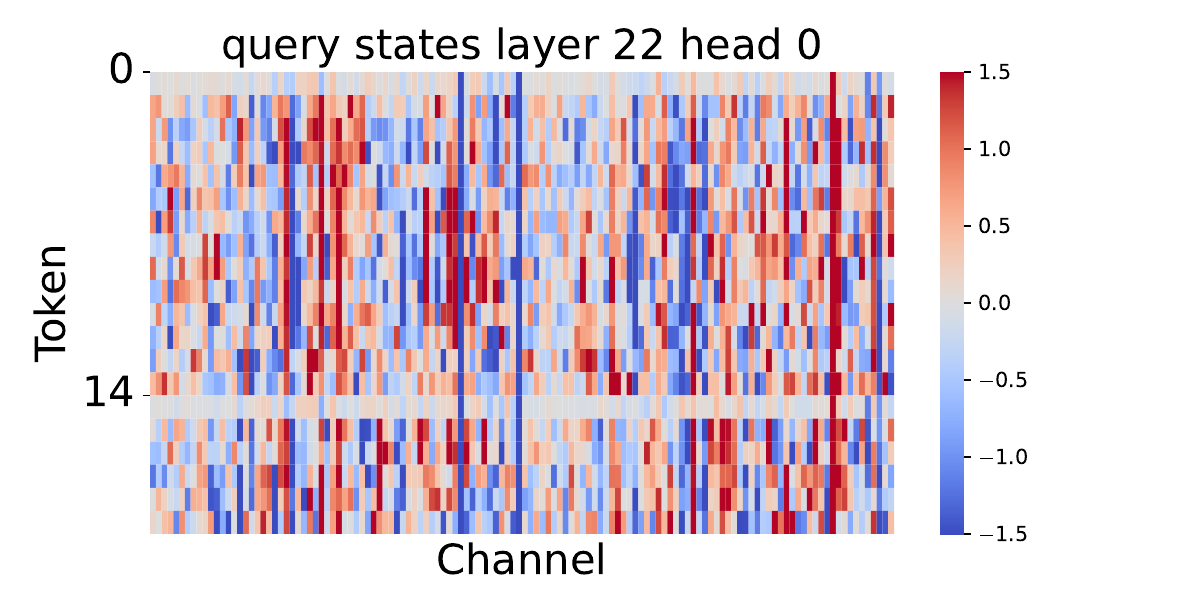}
    \end{subfigure}
    \hspace{-11mm} 
    \begin{subfigure}{0.37\textwidth}
        \centering
    \includegraphics[width=\linewidth]{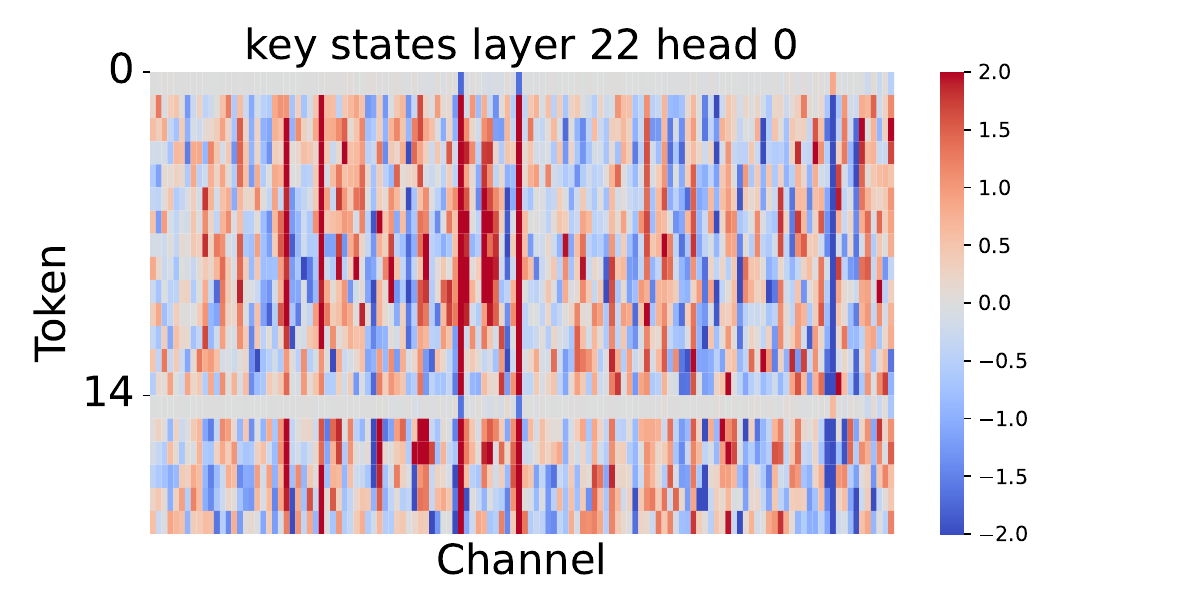}
    \end{subfigure}
    \hspace{-11mm} 
    \begin{subfigure}{0.37\textwidth}
        \centering
    \includegraphics[width=\linewidth]{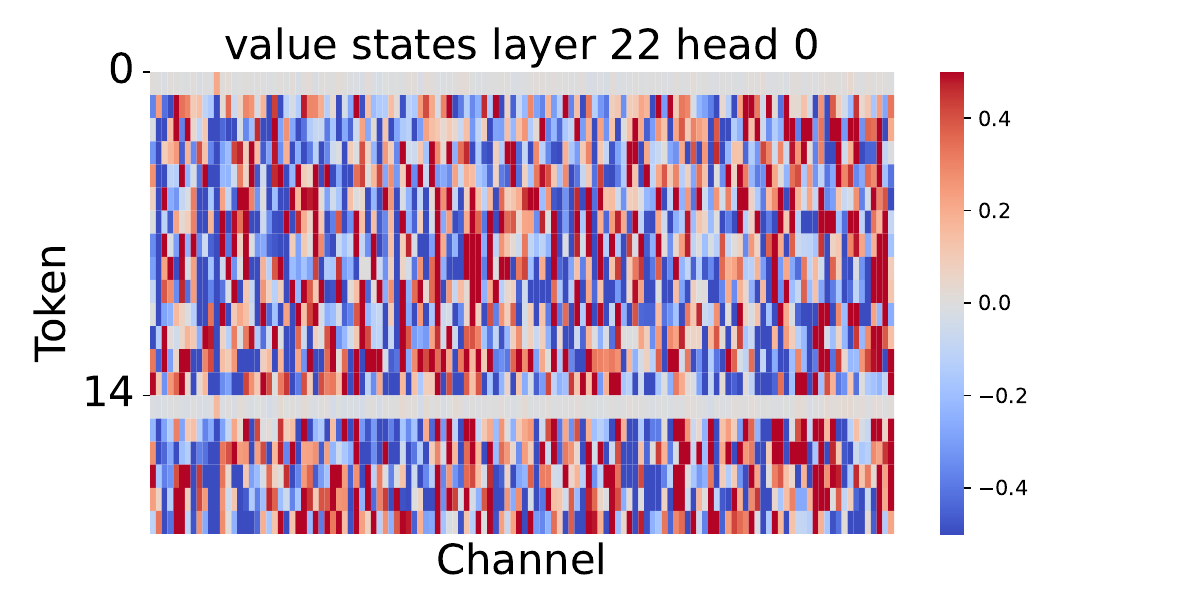}
    \end{subfigure}
    \begin{subfigure}{0.37\textwidth}
        \centering
    \includegraphics[width=\linewidth]{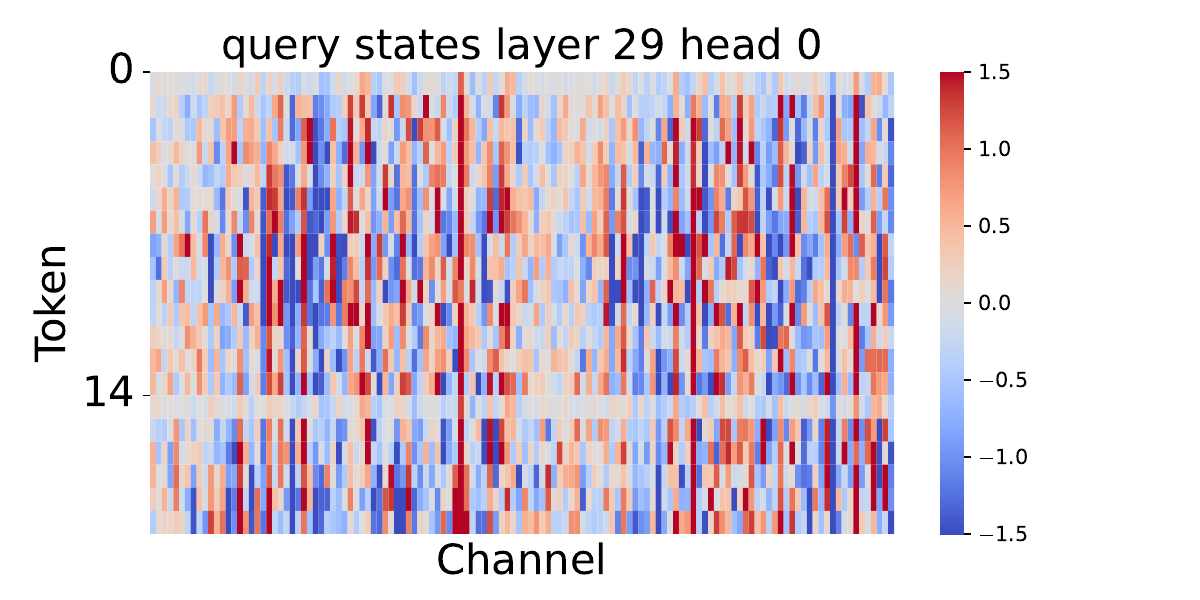}
    \end{subfigure} 
    \hspace{-11mm} 
    \begin{subfigure}{0.37\textwidth}
        \centering
    \includegraphics[width=\linewidth]{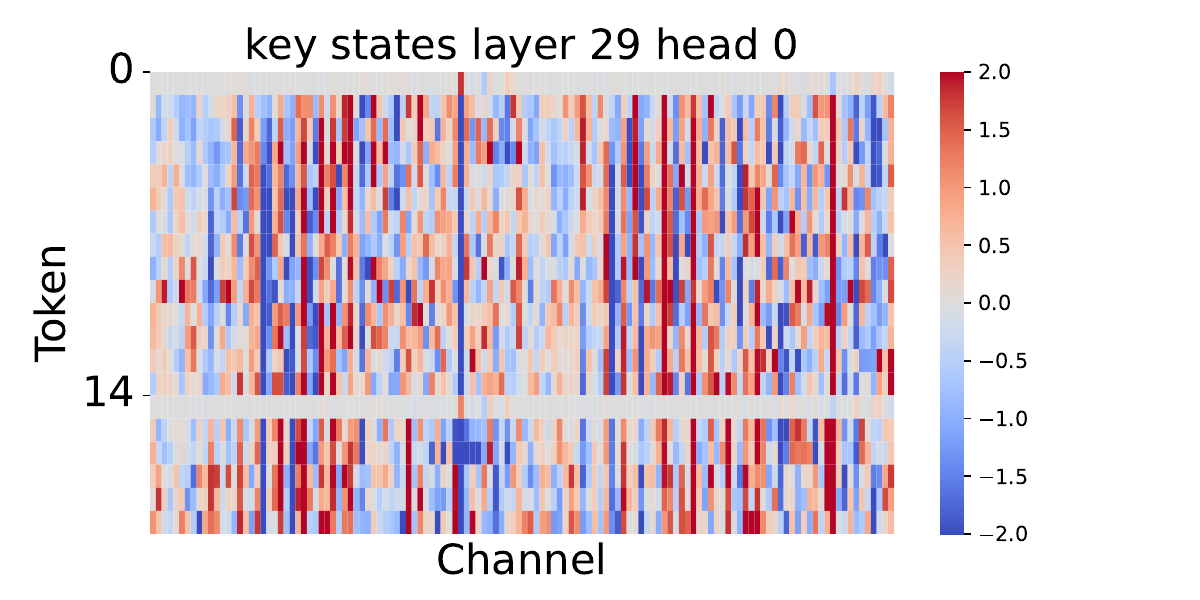}
    \end{subfigure}
    \hspace{-11mm} 
    \begin{subfigure}{0.37\textwidth}
        \centering
    \includegraphics[width=\linewidth]{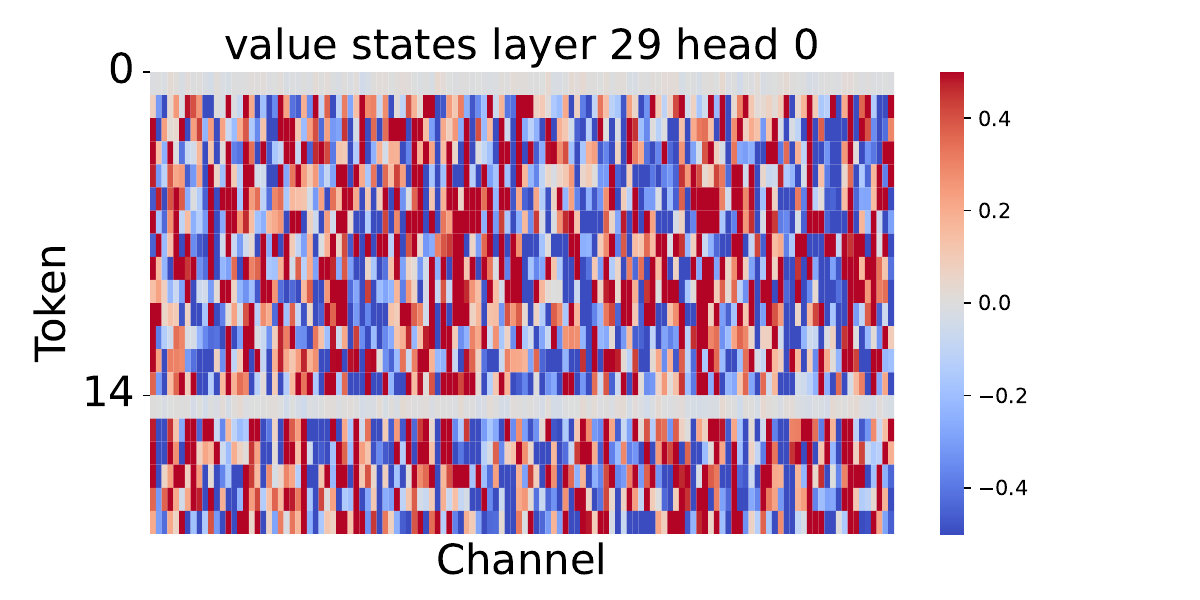}
    \end{subfigure}
    \begin{subfigure}{0.37\textwidth}
        \centering
    \includegraphics[width=\linewidth]{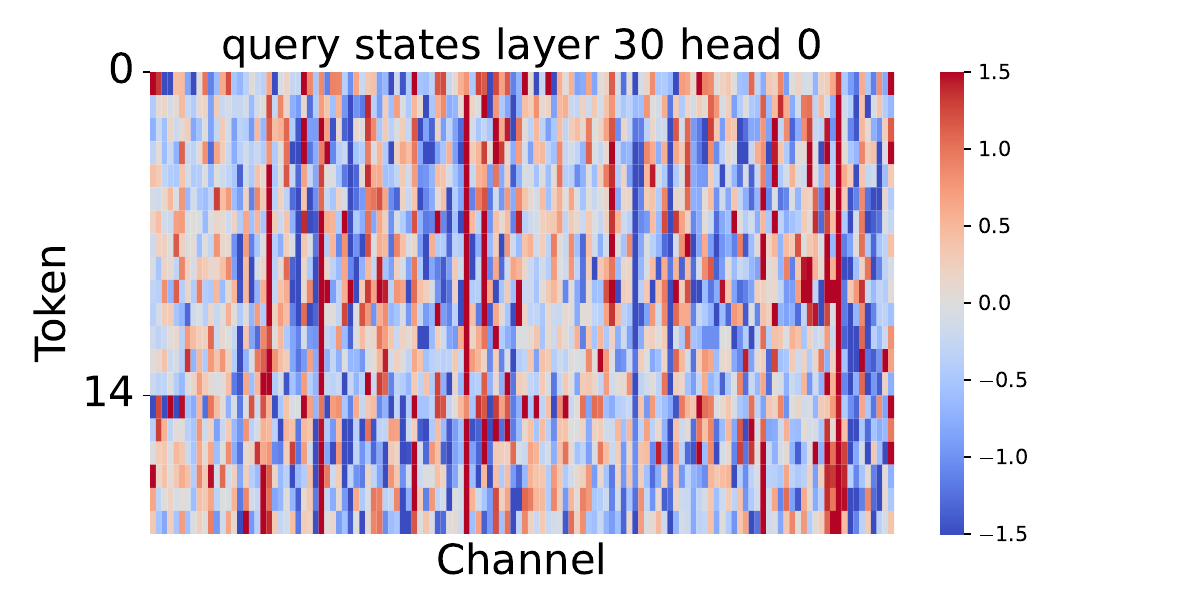}
    \end{subfigure}
    \hspace{-11mm} 
    \begin{subfigure}{0.37\textwidth}
        \centering
    \includegraphics[width=\linewidth]{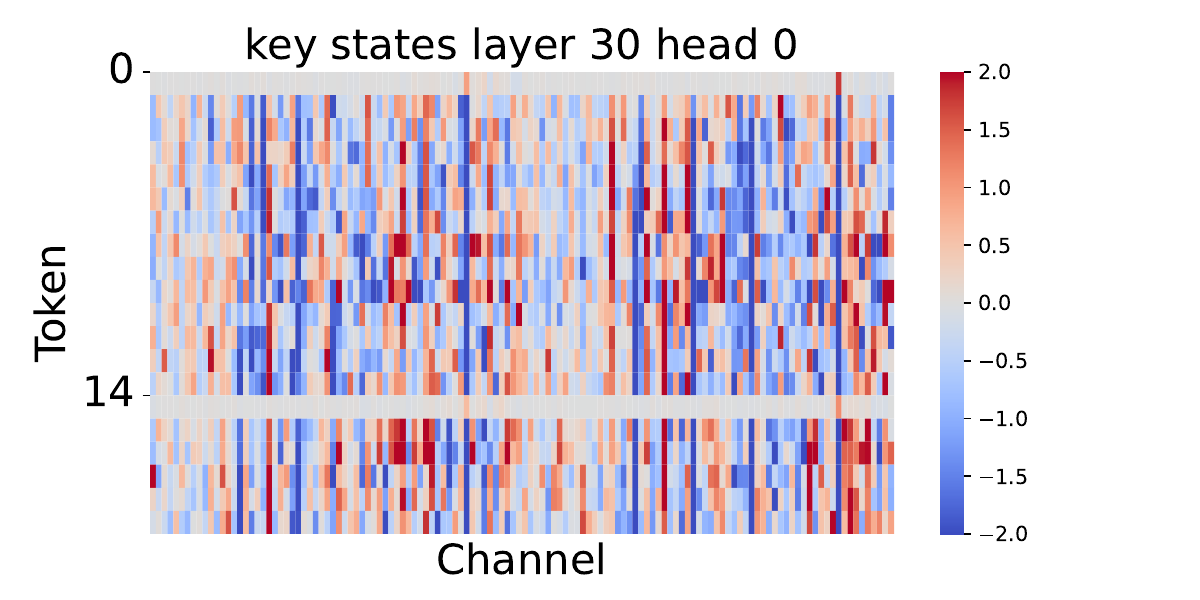}
    \end{subfigure}
    \hspace{-11mm} 
    \begin{subfigure}{0.37\textwidth}
        \centering
    \includegraphics[width=\linewidth]{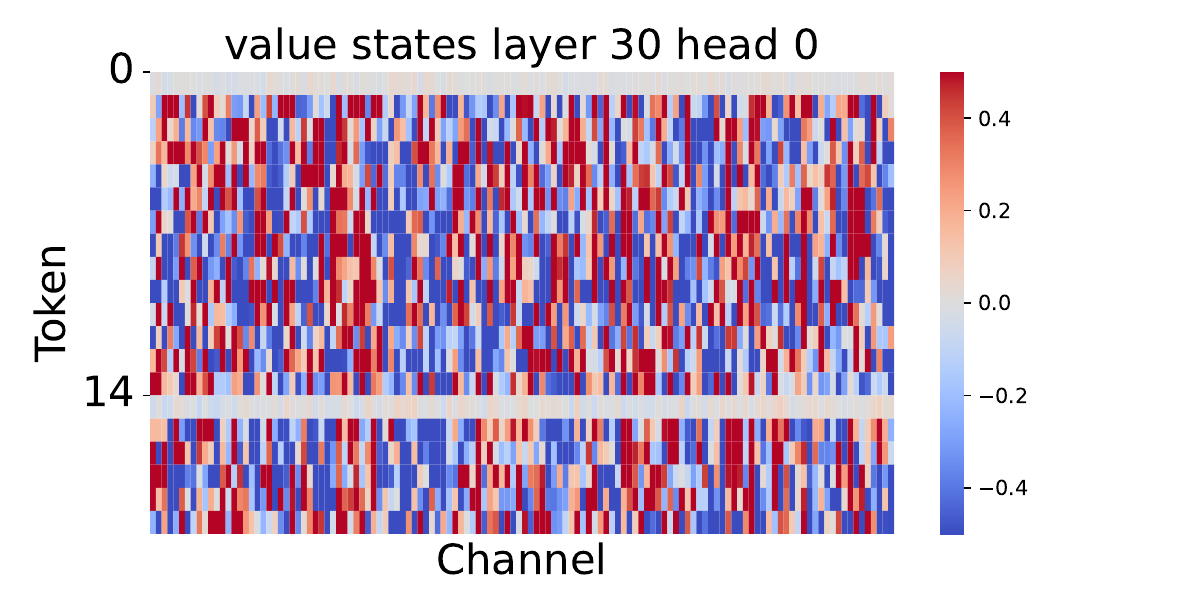}
    \end{subfigure}
    \begin{subfigure}{0.37\textwidth}
        \centering
    \includegraphics[width=\linewidth]{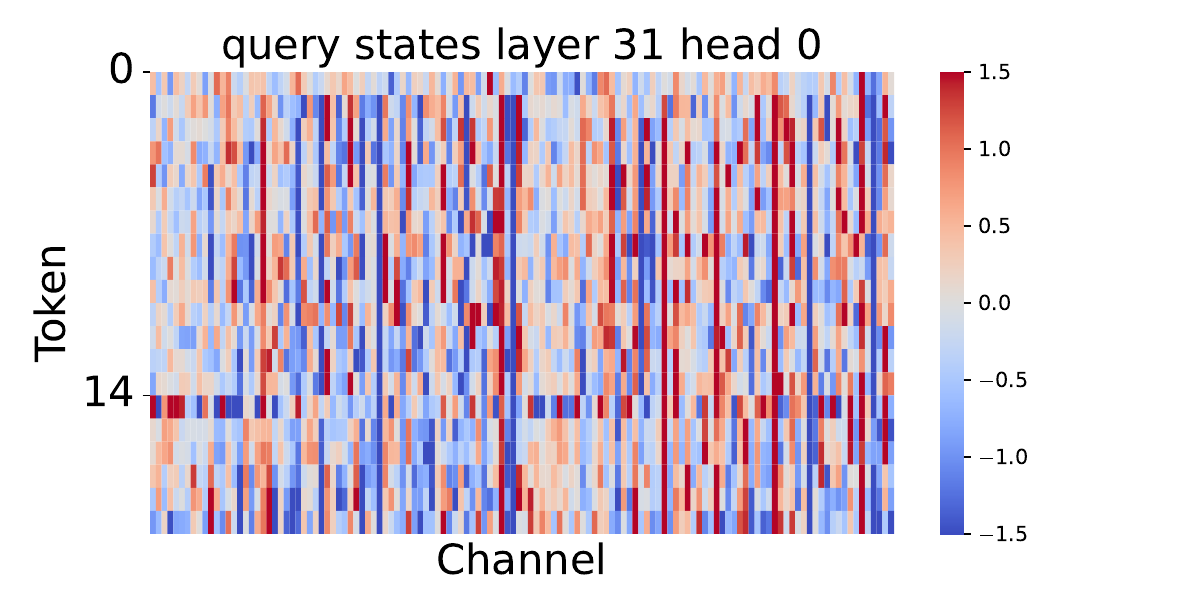}
    \end{subfigure}
    \hspace{-11mm} 
    \begin{subfigure}{0.37\textwidth}
        \centering
    \includegraphics[width=\linewidth]{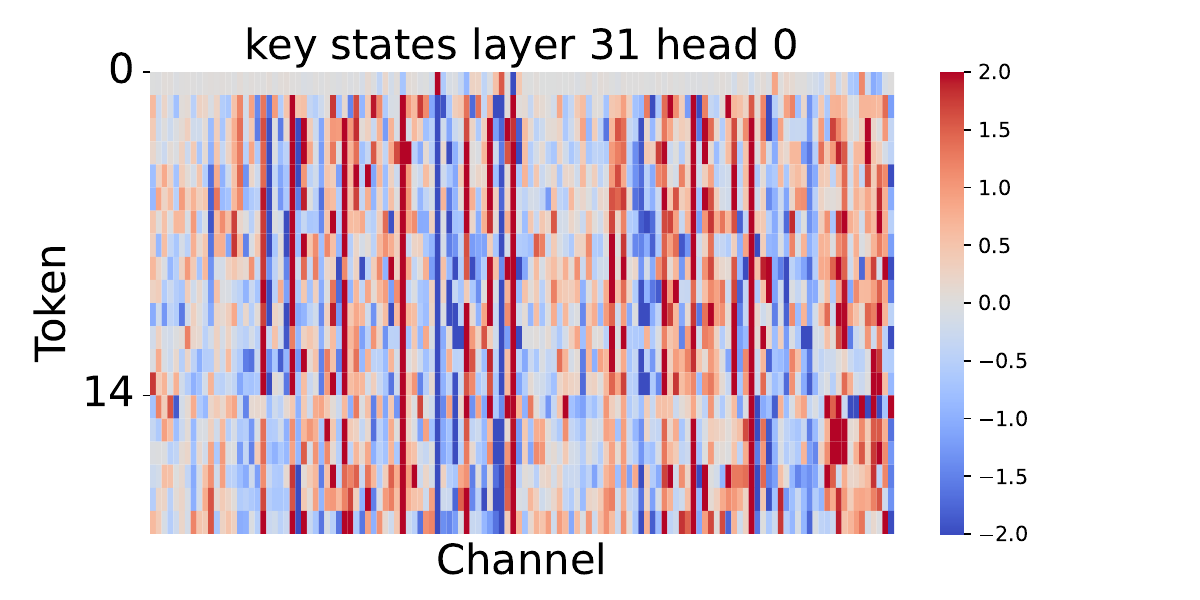}
    \end{subfigure}
    \hspace{-11mm} 
    \begin{subfigure}{0.37\textwidth}
        \centering
    \includegraphics[width=\linewidth]{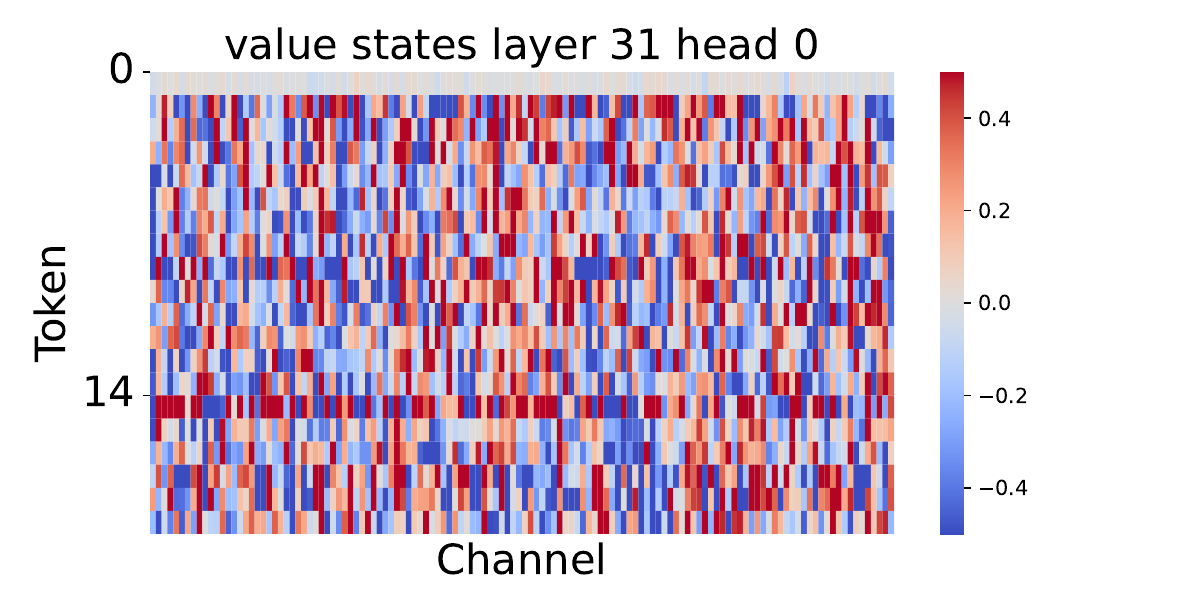}
    \end{subfigure}
    \caption{Queries, Keys, and Values for LLaMA2-7B using Prompt 1,with attention sinks occurring in layers beyond layer 0 and 1, at tokens 0 and 14.}
\label{QKV_appendix_state_1}
\end{figure}
\begin{figure}[t]
    \centering    
    \begin{subfigure}{0.32\textwidth}
        \centering
    \includegraphics[width=\linewidth]{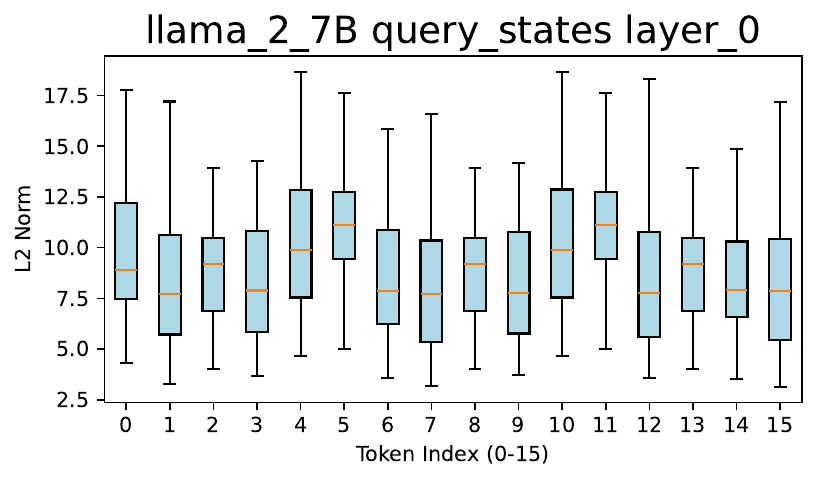}
    \end{subfigure}
    \begin{subfigure}{0.32\textwidth}
        \centering
    \includegraphics[width=\linewidth]{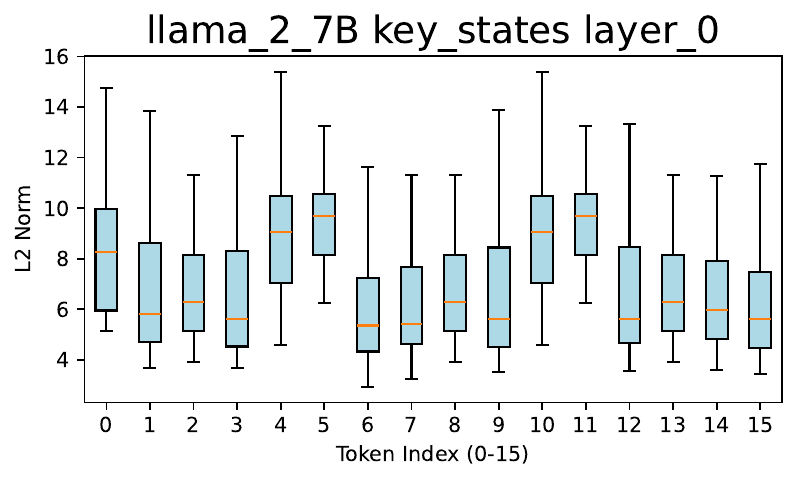}
    \end{subfigure}
    \begin{subfigure}{0.32\textwidth}
        \centering
    \includegraphics[width=\linewidth]{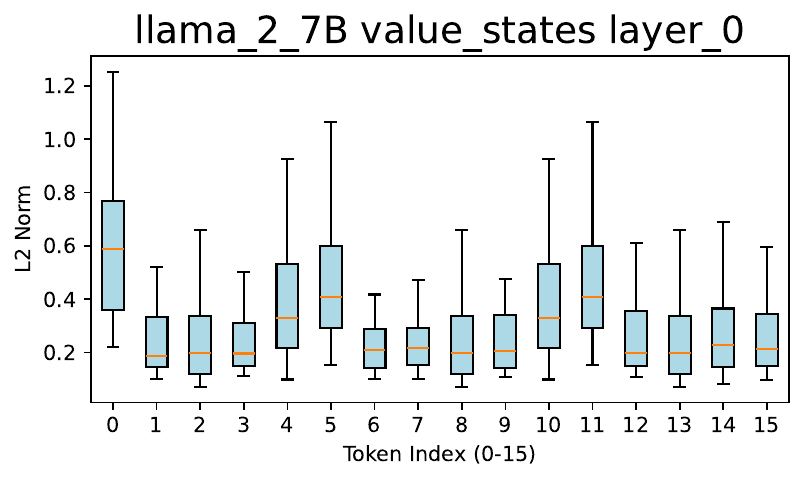}
    \end{subfigure}
    \begin{subfigure}{0.32\textwidth}
        \centering
    \includegraphics[width=\linewidth]{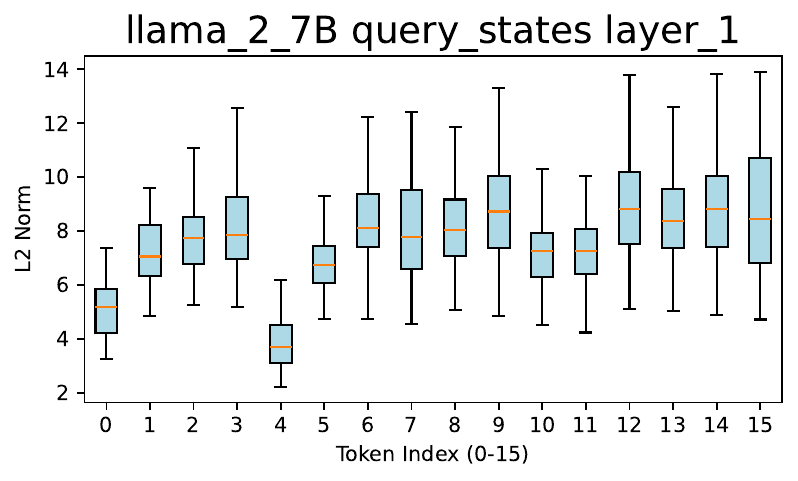}
    \end{subfigure}
    \begin{subfigure}{0.32\textwidth}
        \centering
    \includegraphics[width=\linewidth]{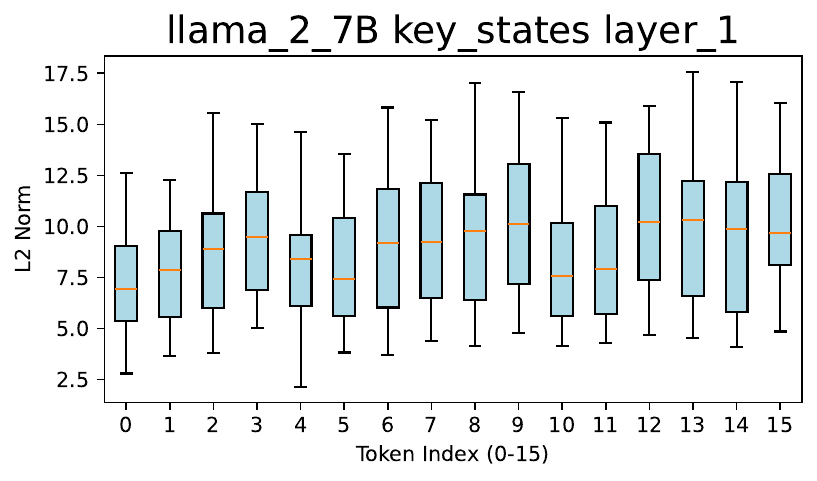}
    \end{subfigure}
    \begin{subfigure}{0.32\textwidth}
        \centering
    \includegraphics[width=\linewidth]{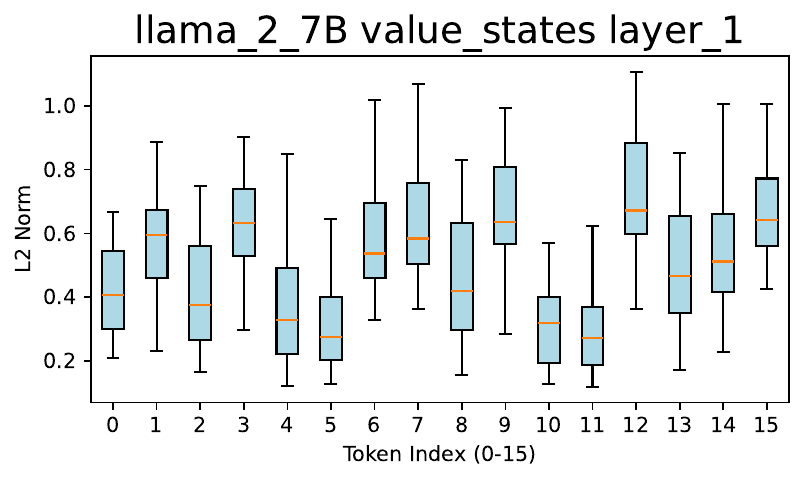}
    \end{subfigure}
    \begin{subfigure}{0.32\textwidth}
        \centering
    \includegraphics[width=\linewidth]{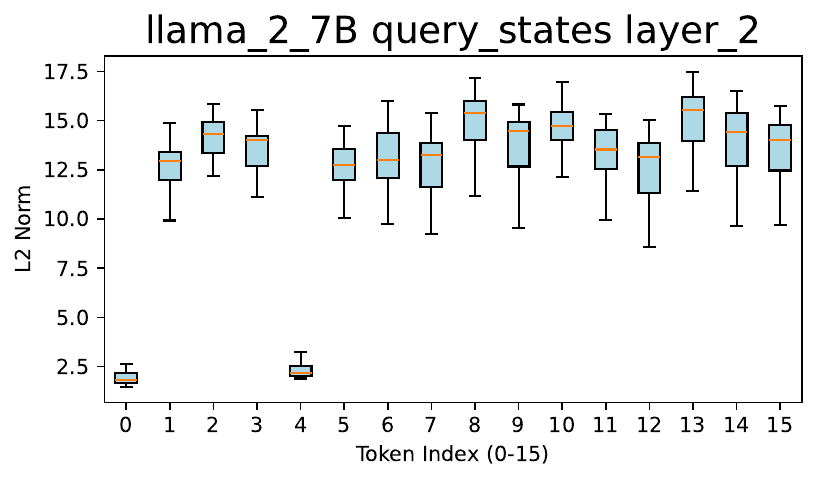}
    \end{subfigure}
    \begin{subfigure}{0.32\textwidth}
        \centering
    \includegraphics[width=\linewidth]{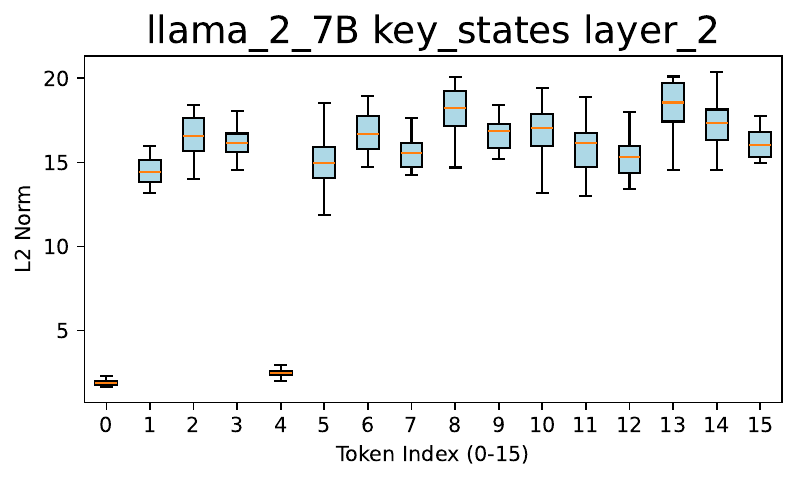}
    \end{subfigure}
    \begin{subfigure}{0.32\textwidth}
        \centering
    \includegraphics[width=\linewidth]{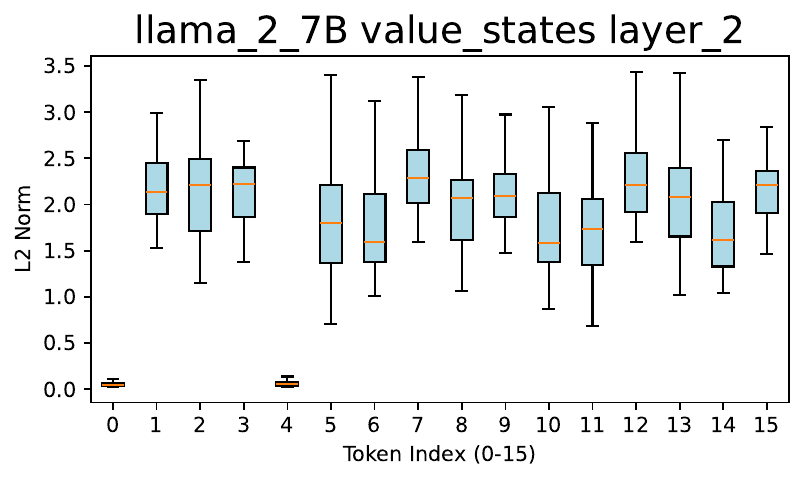}
    \end{subfigure}
    \begin{subfigure}{0.32\textwidth}
        \centering
    \includegraphics[width=\linewidth]{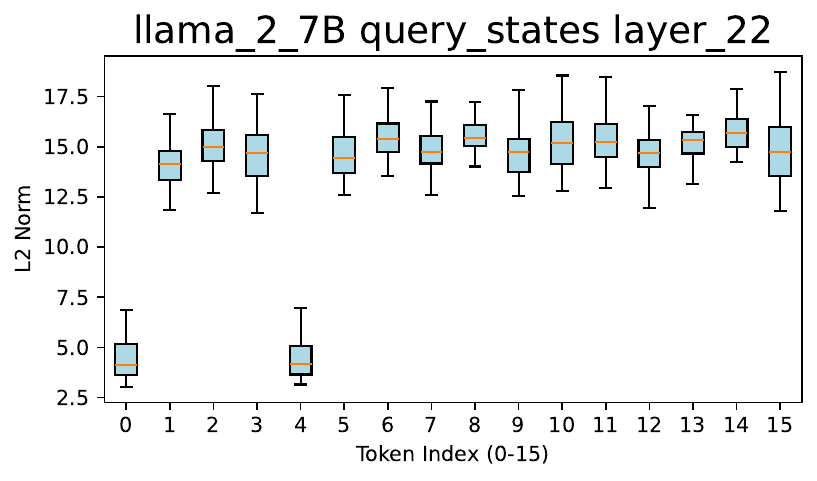}
    \end{subfigure}
    \begin{subfigure}{0.32\textwidth}
        \centering
    \includegraphics[width=\linewidth]{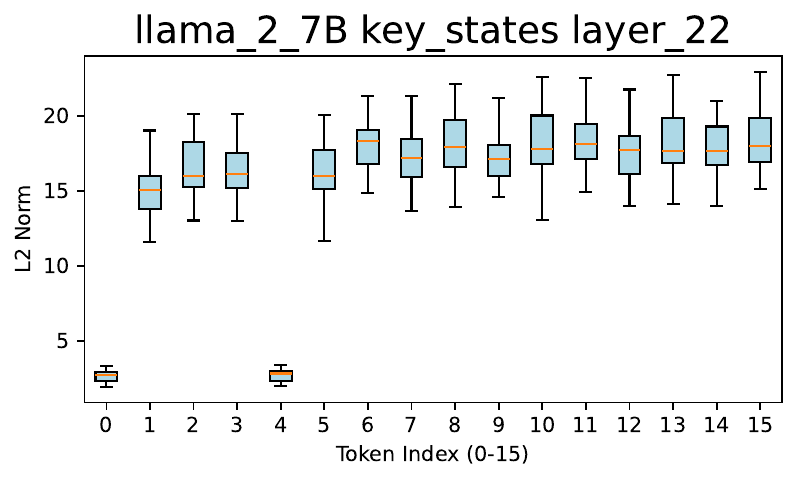}
    \end{subfigure}
    \begin{subfigure}{0.32\textwidth}
        \centering
    \includegraphics[width=\linewidth]{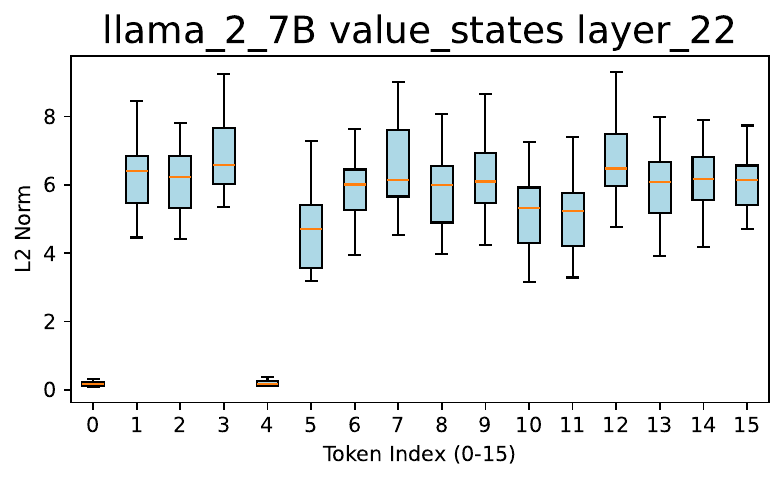}
    \end{subfigure}
    \begin{subfigure}{0.32\textwidth}
        \centering
    \includegraphics[width=\linewidth]{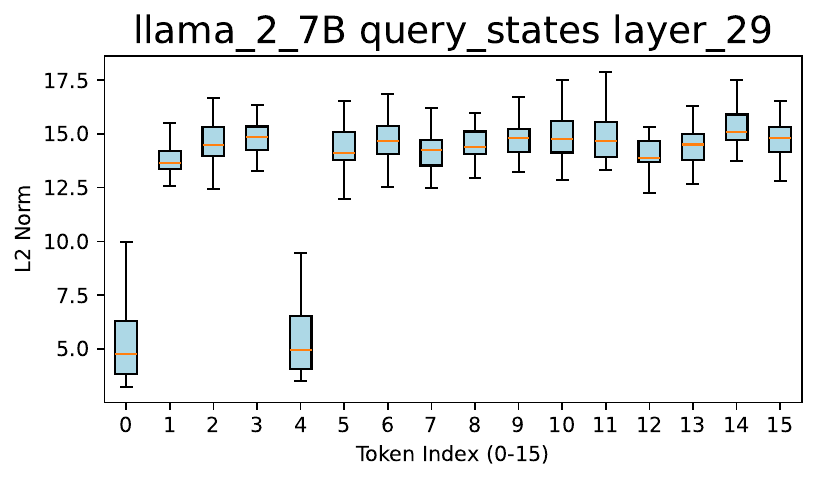}
    \end{subfigure}
    \begin{subfigure}{0.32\textwidth}
        \centering
    \includegraphics[width=\linewidth]{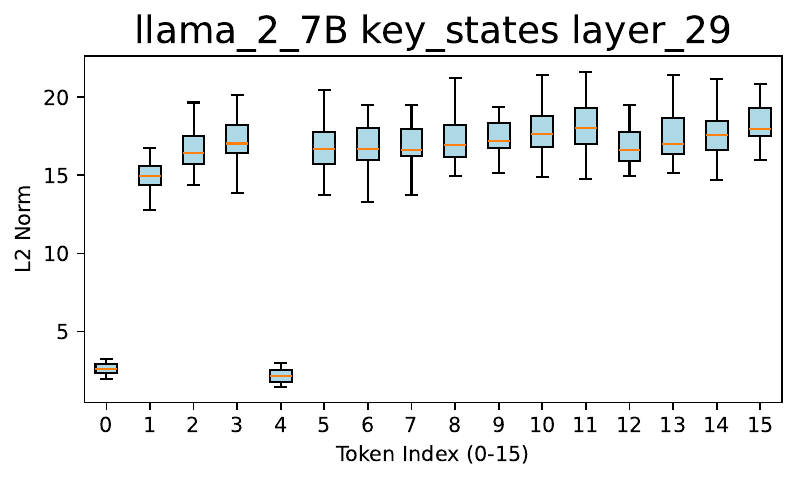}
    \end{subfigure}
    \begin{subfigure}{0.32\textwidth}
        \centering
    \includegraphics[width=\linewidth]{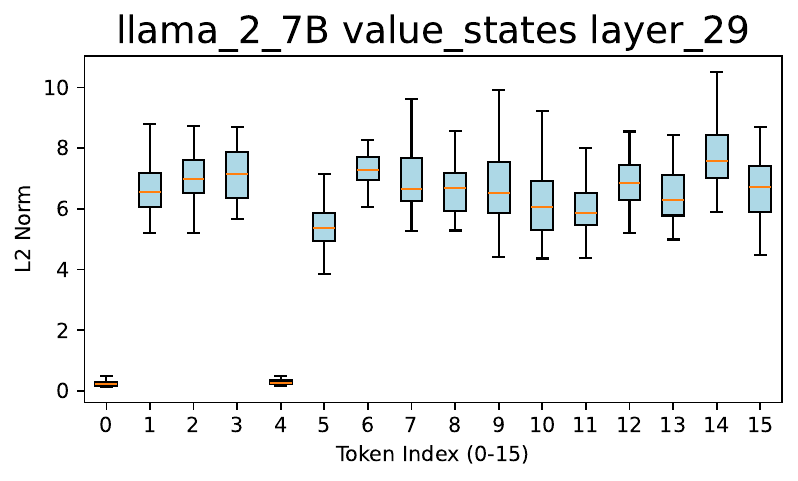}
    \end{subfigure}
    \begin{subfigure}{0.32\textwidth}
        \centering
    \includegraphics[width=\linewidth]{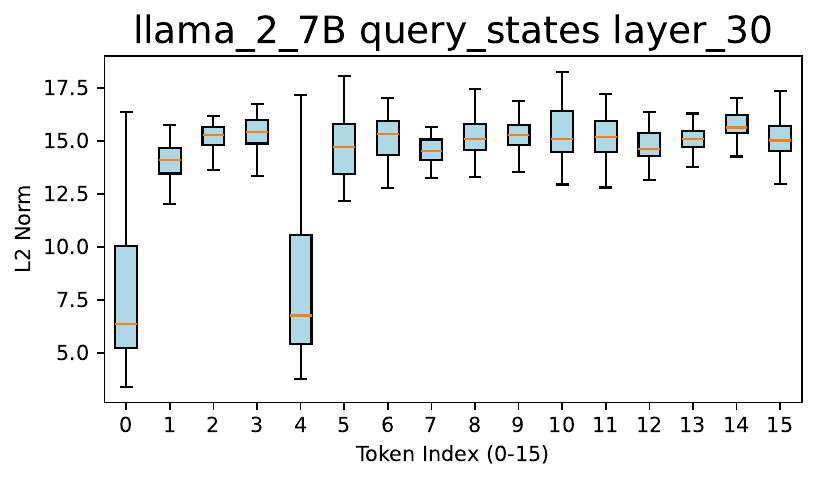}
    \end{subfigure}
    \begin{subfigure}{0.32\textwidth}
        \centering
    \includegraphics[width=\linewidth]{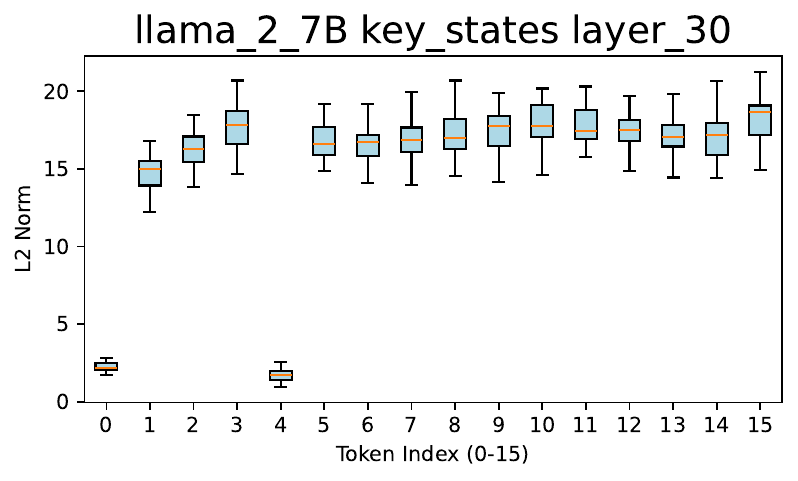}
    \end{subfigure}
    \begin{subfigure}{0.32\textwidth}
        \centering
    \includegraphics[width=\linewidth]{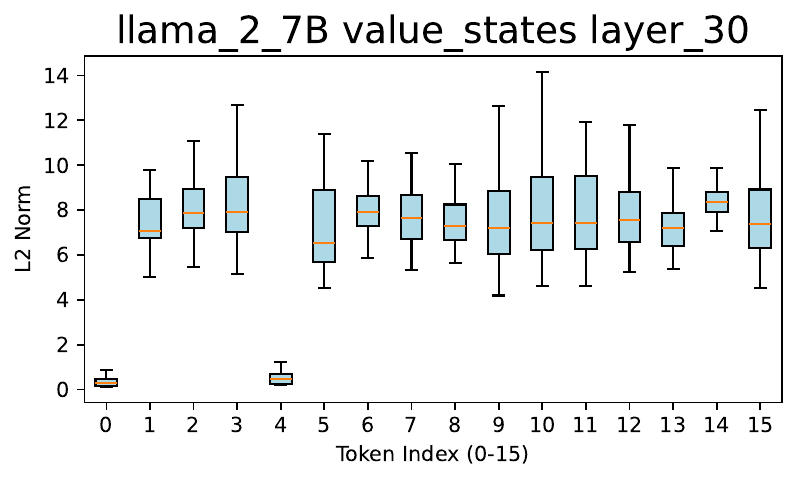}
    \end{subfigure}
    \begin{subfigure}{0.32\textwidth}
        \centering
    \includegraphics[width=\linewidth]{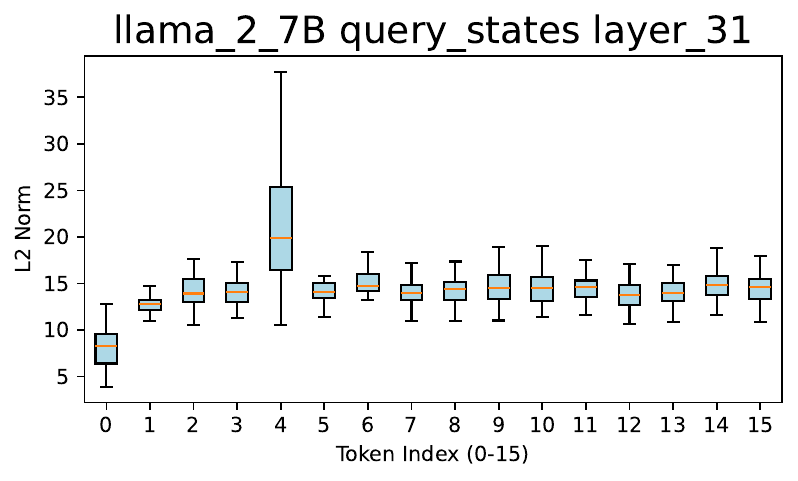}
    \end{subfigure}
    \begin{subfigure}{0.32\textwidth}
        \centering
    \includegraphics[width=\linewidth]{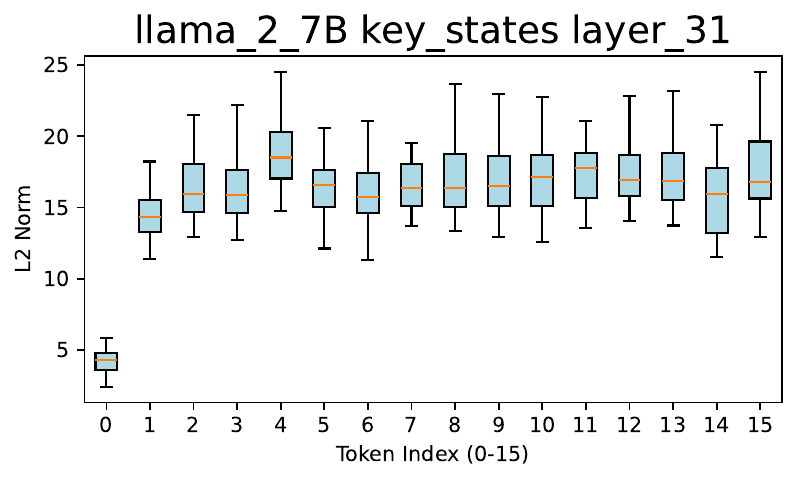}
    \end{subfigure}
    \begin{subfigure}{0.32\textwidth}
        \centering
    \includegraphics[width=\linewidth]{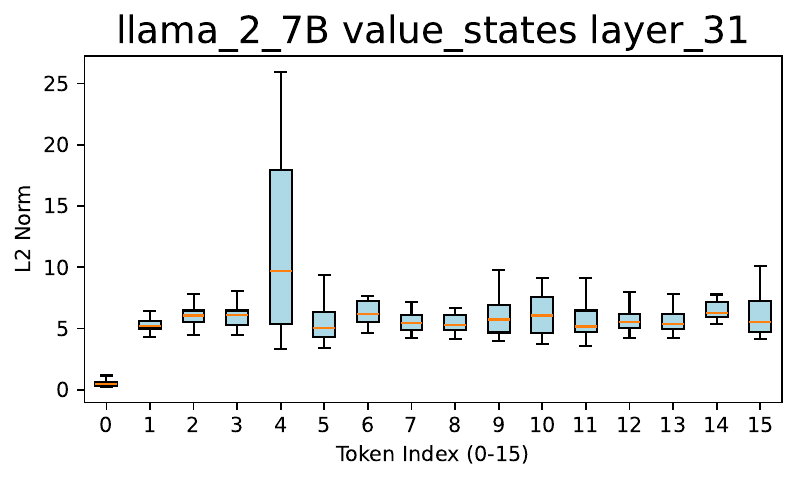}
    \end{subfigure}
    \caption{L2 norm distributions of Queries, Keys, and Values for LLaMA2-7B using Prompt 2, with attention sinks occurring in layers beyond layer 0 and 1, at tokens 0 and 4.}
\label{QKV_appendix_2}
\end{figure}
\begin{figure}[t]
    \centering    
    \begin{subfigure}{0.37\textwidth}
        \centering
    \includegraphics[width=\linewidth]{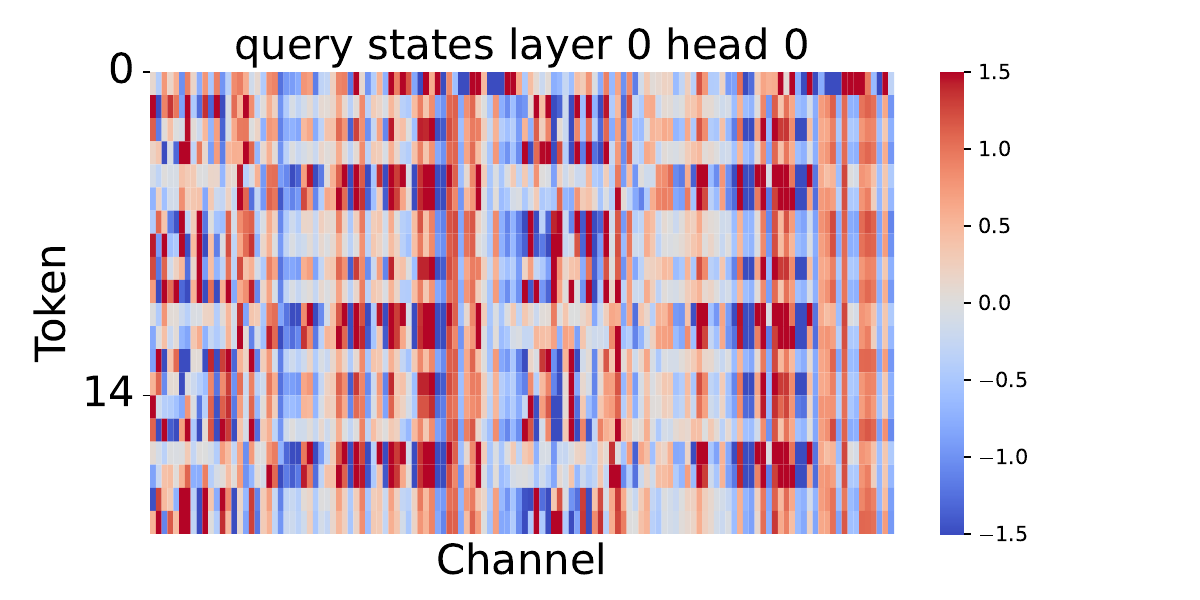}
    \end{subfigure}
    \hspace{-11mm} 
    \begin{subfigure}{0.37\textwidth}
        \centering
    \includegraphics[width=\linewidth]{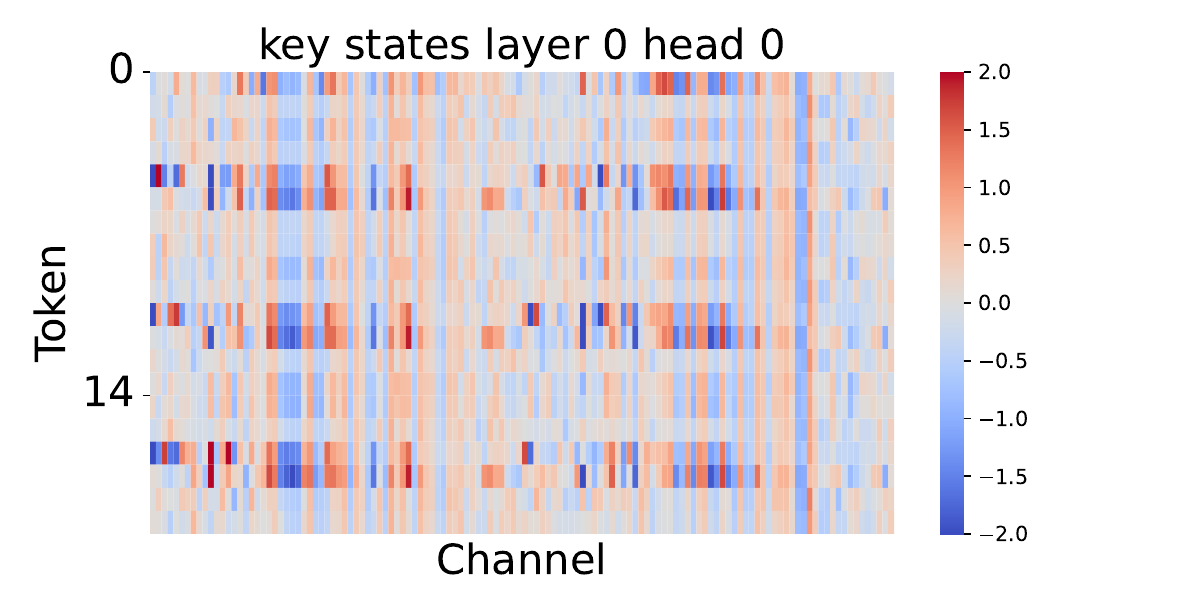}
    \end{subfigure}
    \hspace{-11mm} 
    \begin{subfigure}{0.37\textwidth}
        \centering
    \includegraphics[width=\linewidth]{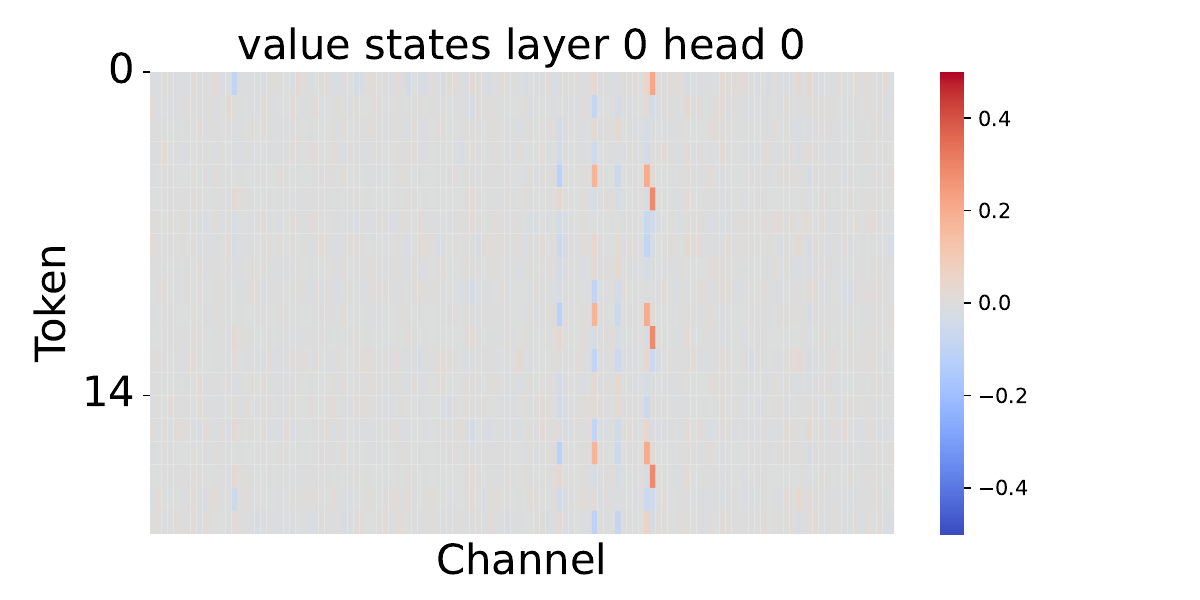}
    \end{subfigure}
    \begin{subfigure}{0.37\textwidth}
        \centering
    \includegraphics[width=\linewidth]{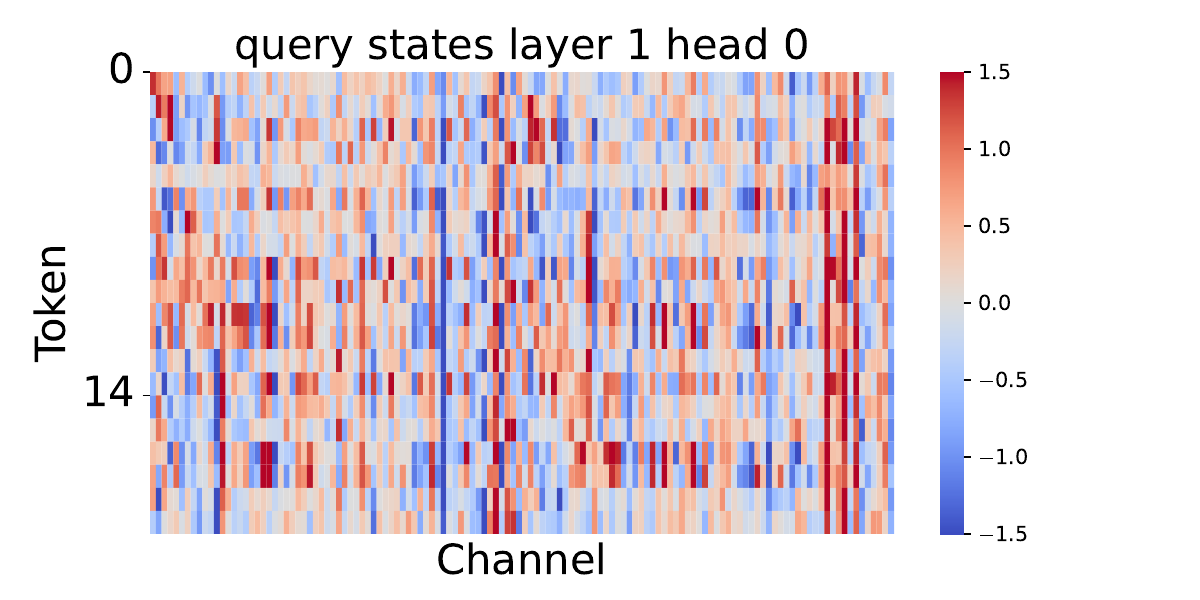}
    \end{subfigure}
    \hspace{-11mm} 
    \begin{subfigure}{0.37\textwidth}
        \centering
    \includegraphics[width=\linewidth]{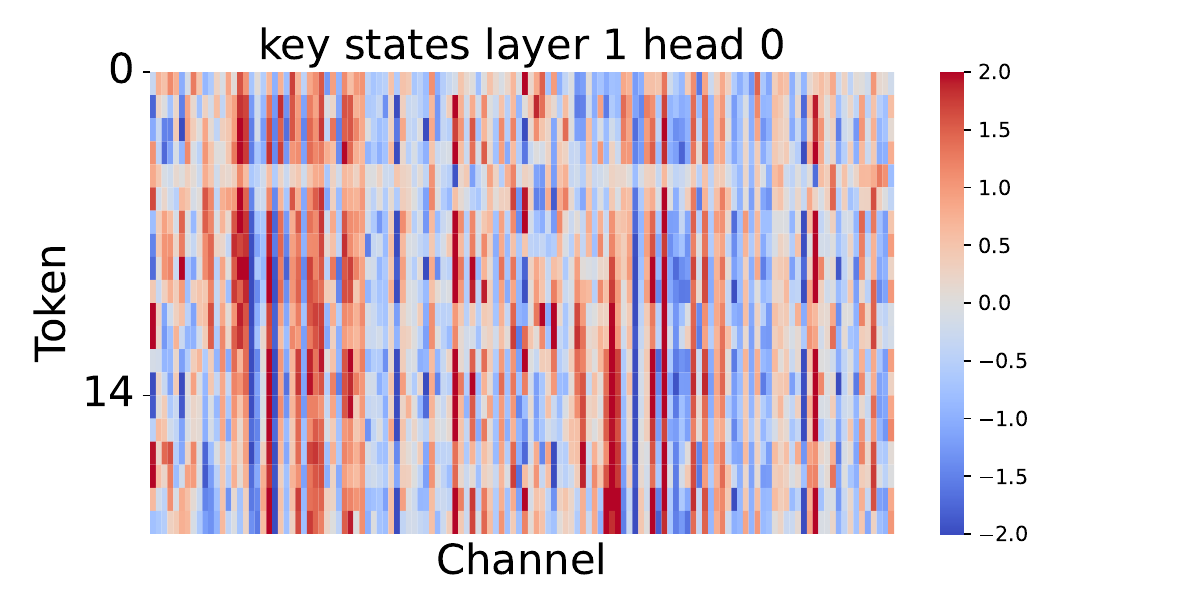}
    \end{subfigure}
    \hspace{-11mm} 
    \begin{subfigure}{0.37\textwidth}
        \centering
    \includegraphics[width=\linewidth]{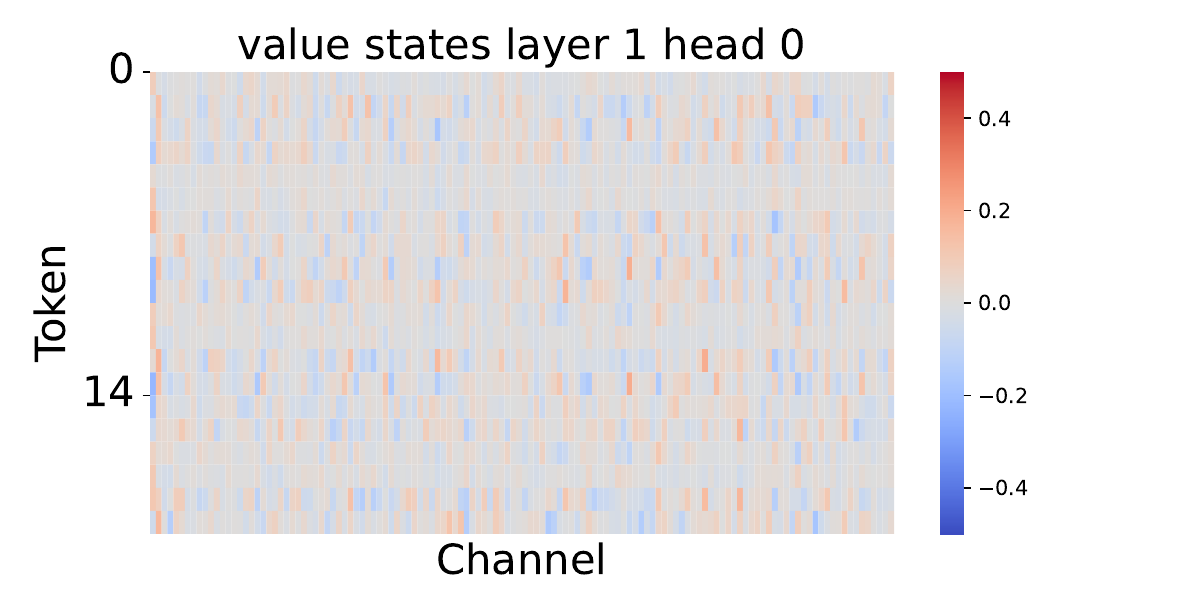}
    \end{subfigure}
    \begin{subfigure}{0.37\textwidth}
        \centering
    \includegraphics[width=\linewidth]{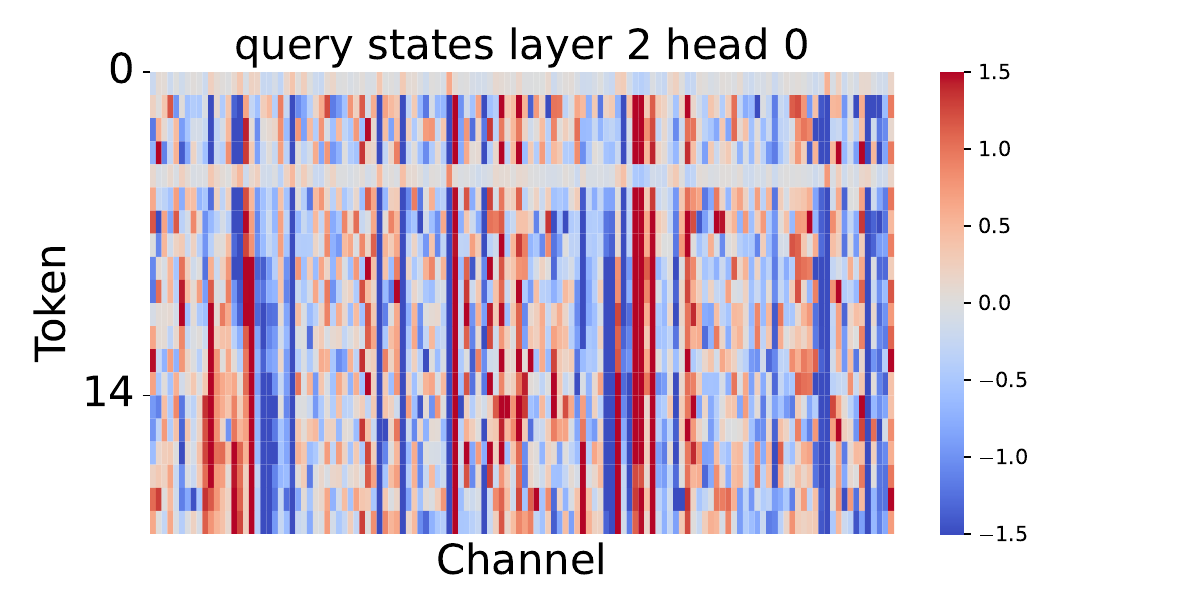}
    \end{subfigure}
    \hspace{-11mm} 
    \begin{subfigure}{0.37\textwidth}
        \centering
    \includegraphics[width=\linewidth]{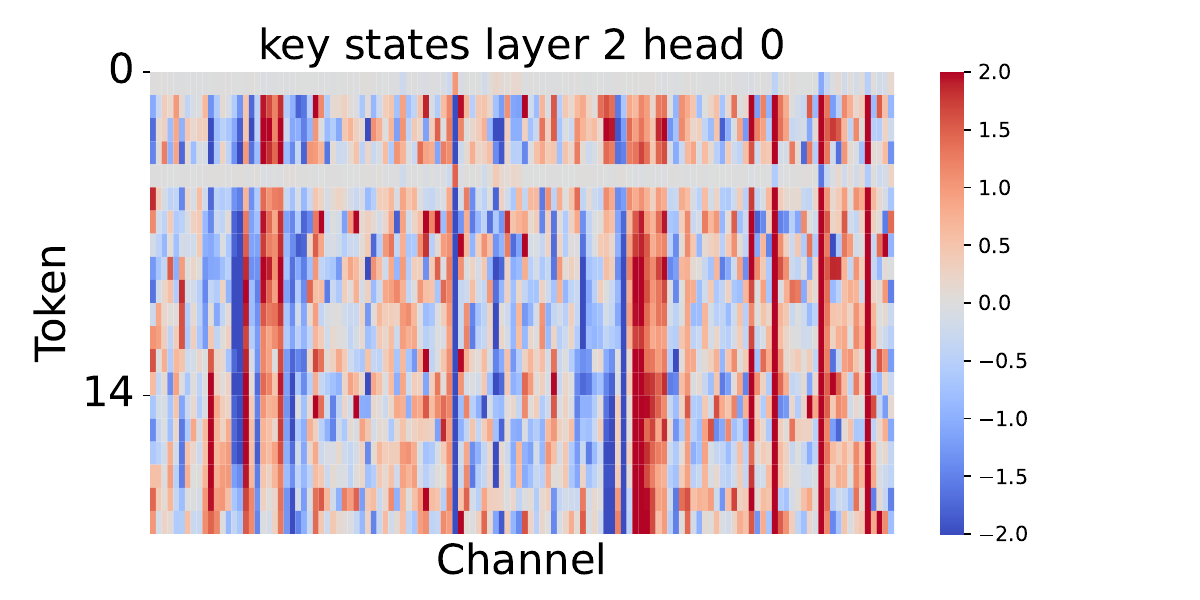}
    \end{subfigure}
    \hspace{-11mm} 
    \begin{subfigure}{0.37\textwidth}
        \centering
    \includegraphics[width=\linewidth]{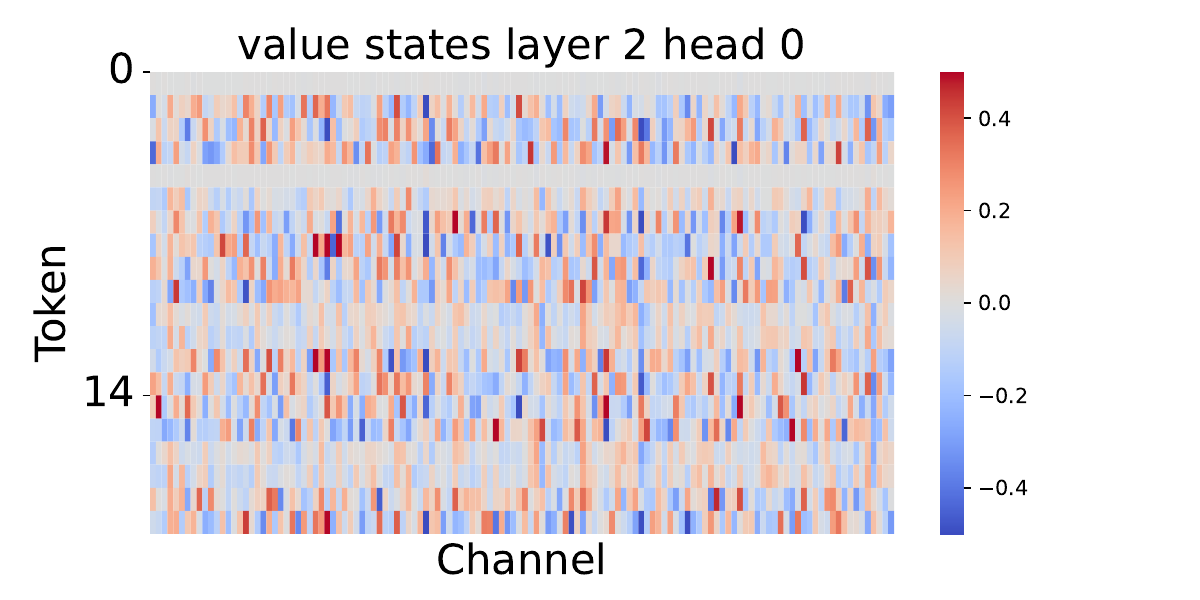}
    \end{subfigure}
    \begin{subfigure}{0.37\textwidth}
        \centering
    \includegraphics[width=\linewidth]{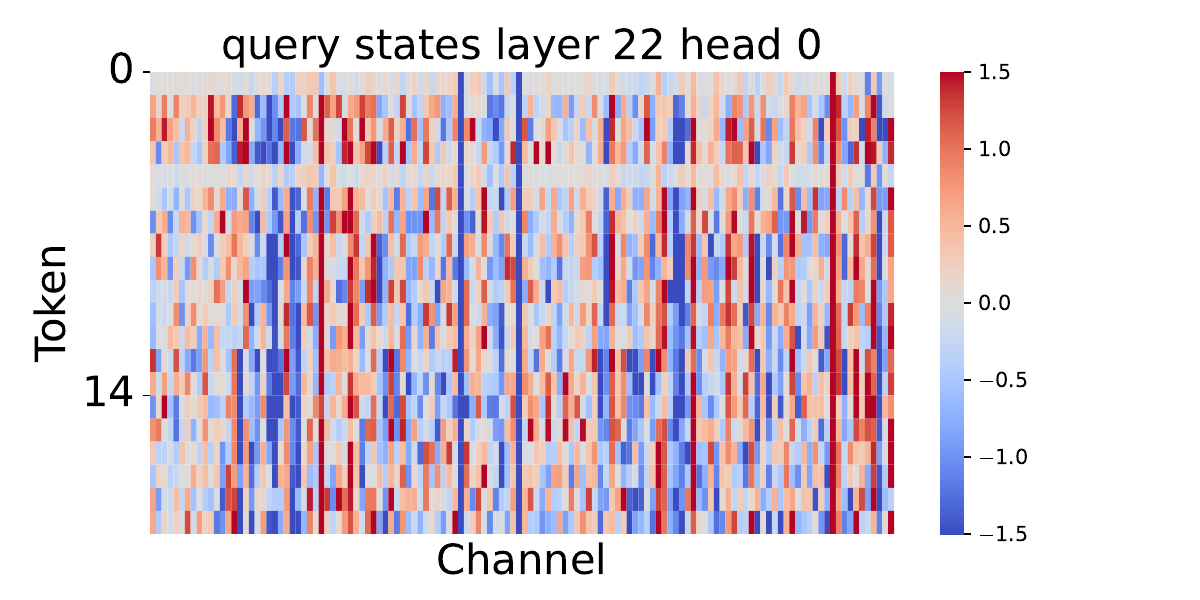}
    \end{subfigure}
    \hspace{-11mm} 
    \begin{subfigure}{0.37\textwidth}
        \centering
    \includegraphics[width=\linewidth]{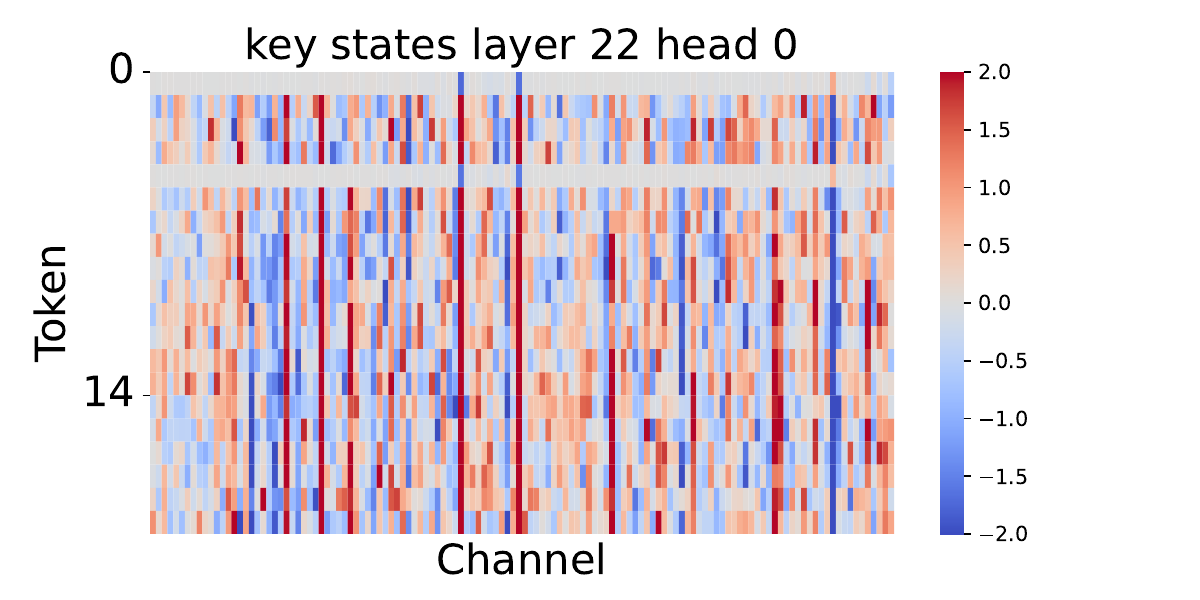}
    \end{subfigure}
    \hspace{-11mm} 
    \begin{subfigure}{0.37\textwidth}
        \centering
    \includegraphics[width=\linewidth]{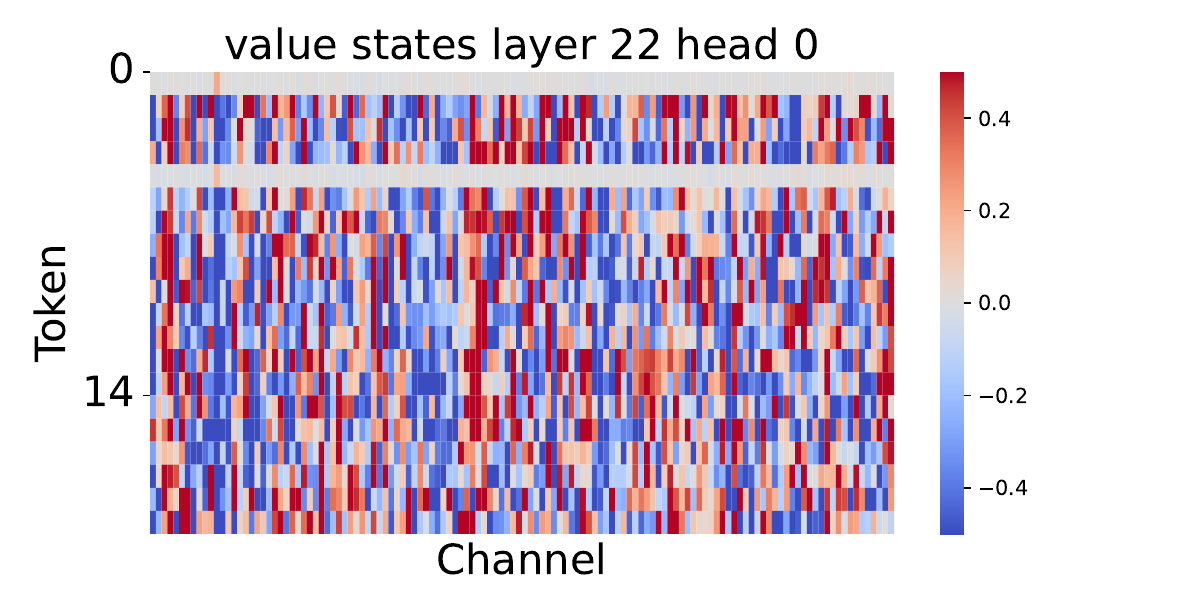}
    \end{subfigure}
    \begin{subfigure}{0.37\textwidth}
        \centering
    \includegraphics[width=\linewidth]{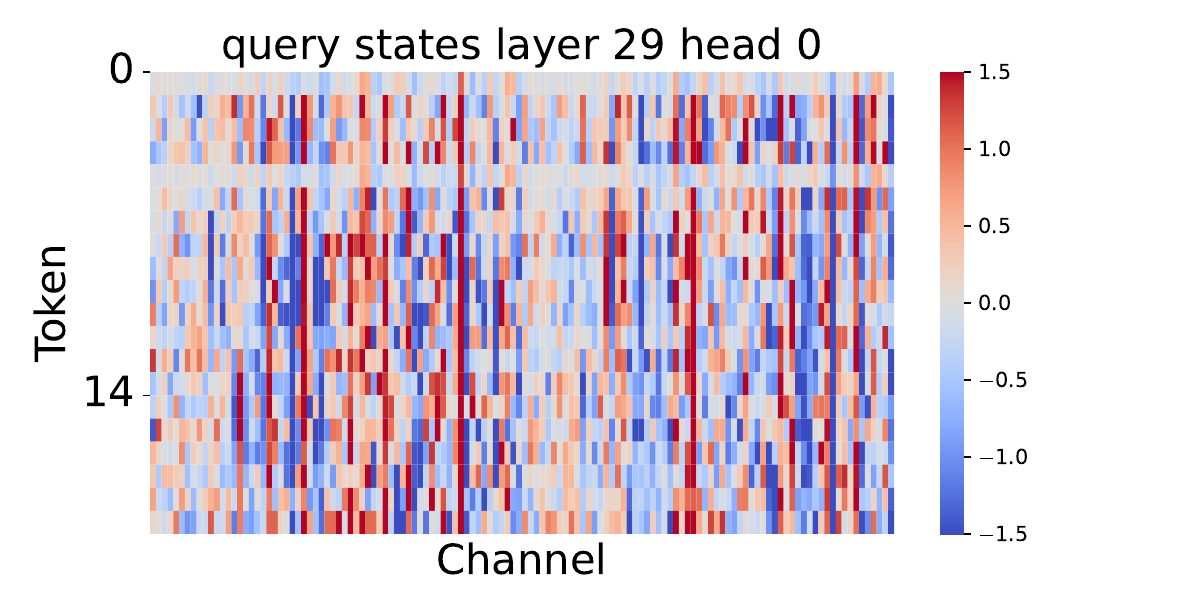}
    \end{subfigure} 
    \hspace{-11mm} 
    \begin{subfigure}{0.37\textwidth}
        \centering
    \includegraphics[width=\linewidth]{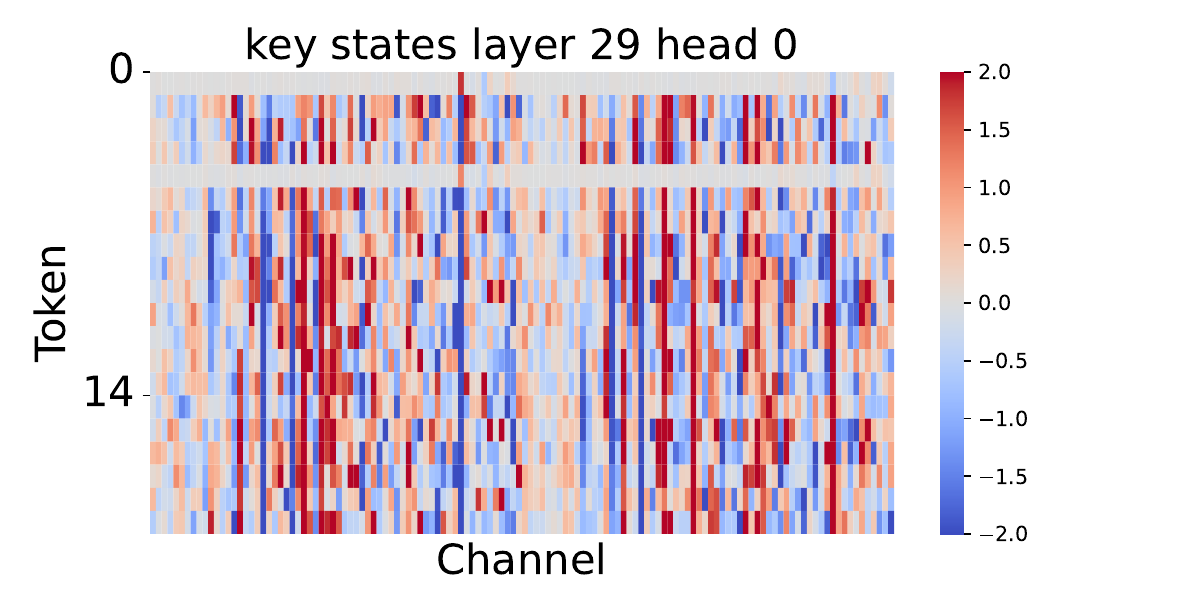}
    \end{subfigure}
    \hspace{-11mm} 
    \begin{subfigure}{0.37\textwidth}
        \centering
    \includegraphics[width=\linewidth]{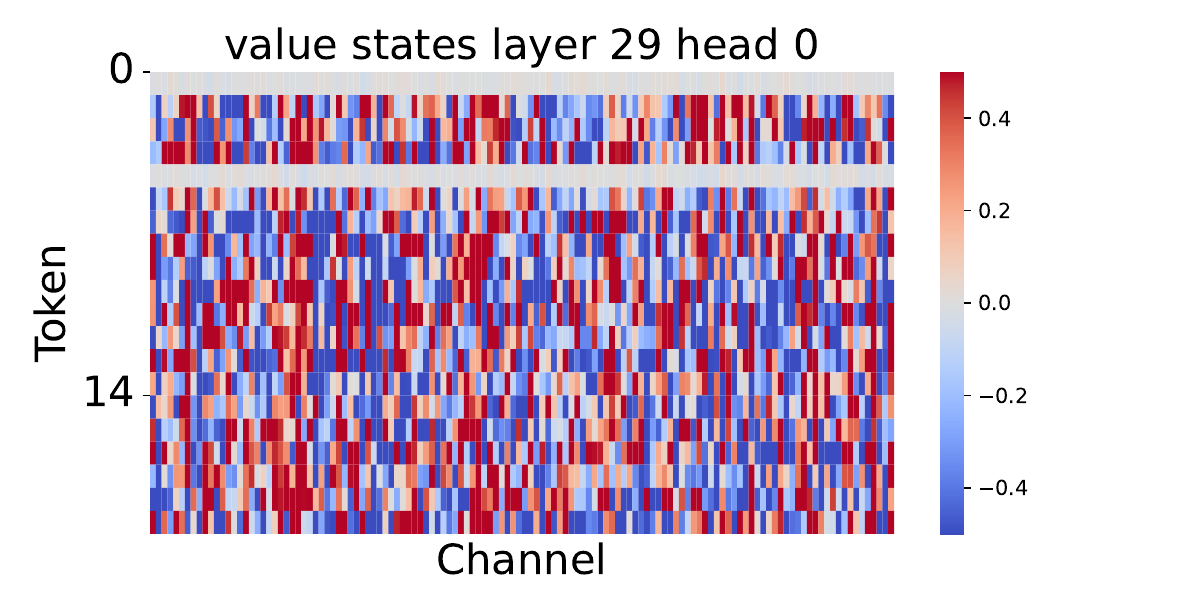}
    \end{subfigure}
    \begin{subfigure}{0.37\textwidth}
        \centering
    \includegraphics[width=\linewidth]{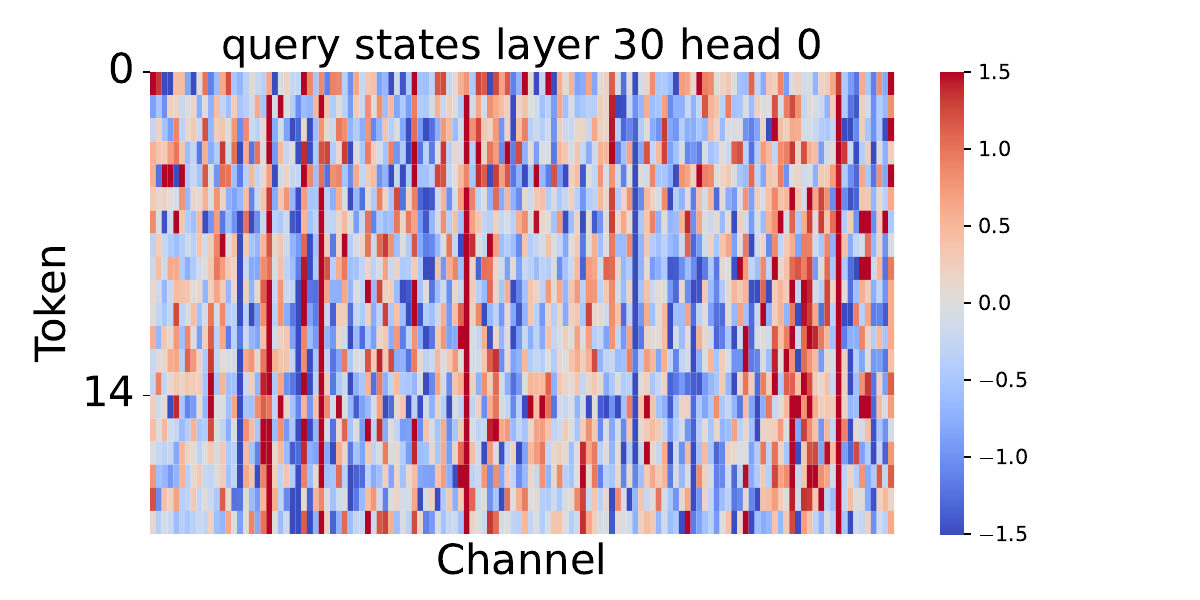}
    \end{subfigure}
    \hspace{-11mm} 
    \begin{subfigure}{0.37\textwidth}
        \centering
    \includegraphics[width=\linewidth]{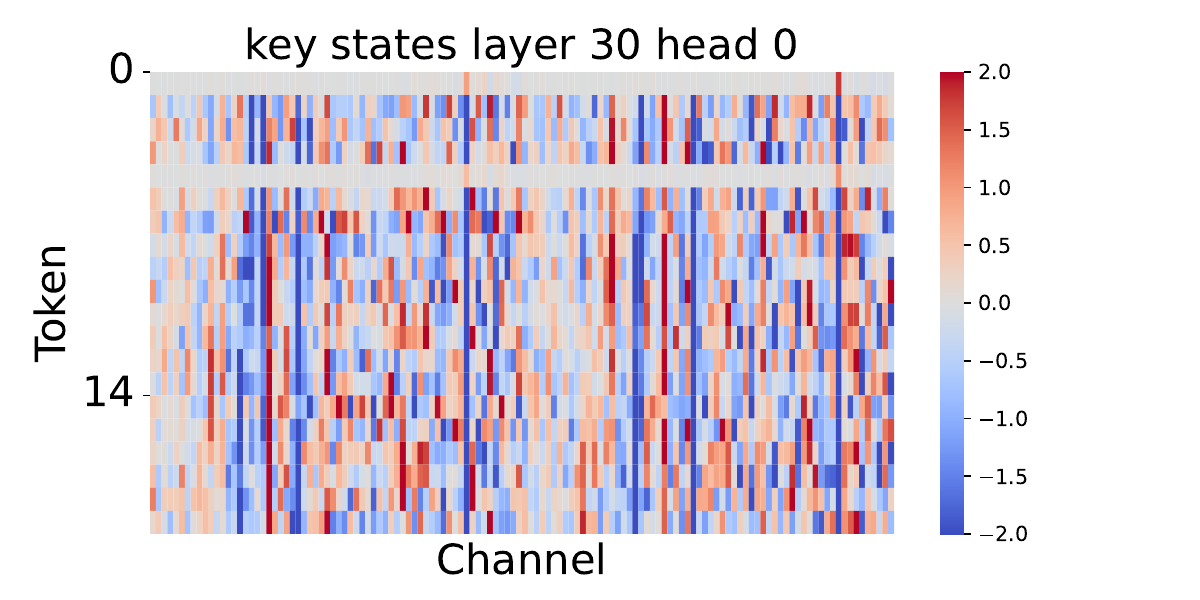}
    \end{subfigure}
    \hspace{-11mm} 
    \begin{subfigure}{0.37\textwidth}
        \centering
    \includegraphics[width=\linewidth]{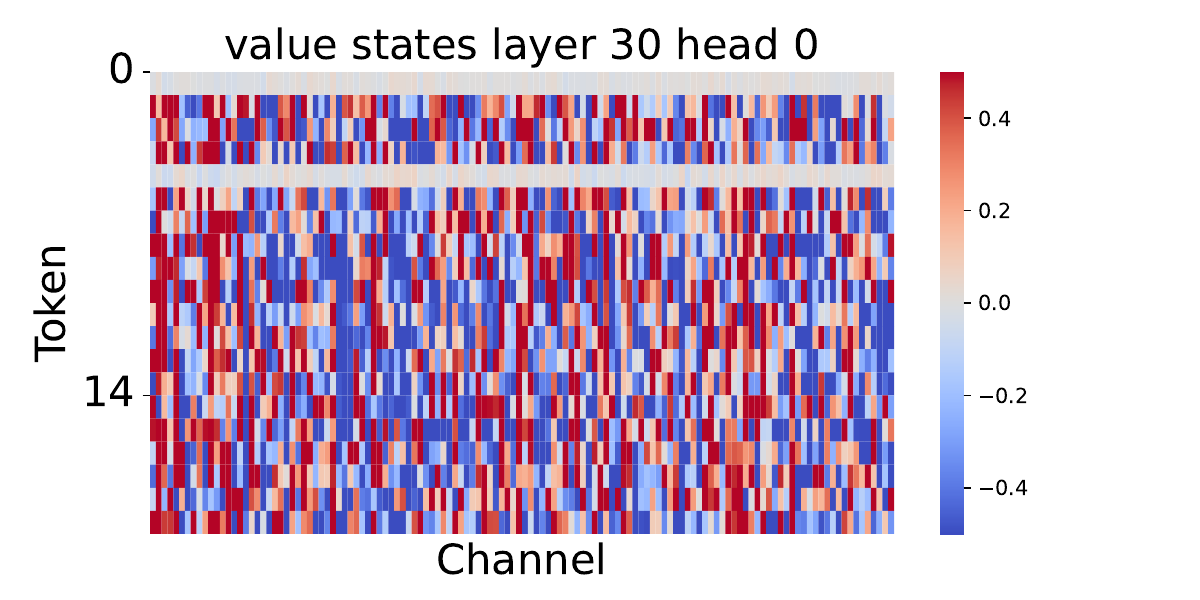}
    \end{subfigure}
    \begin{subfigure}{0.37\textwidth}
        \centering
    \includegraphics[width=\linewidth]{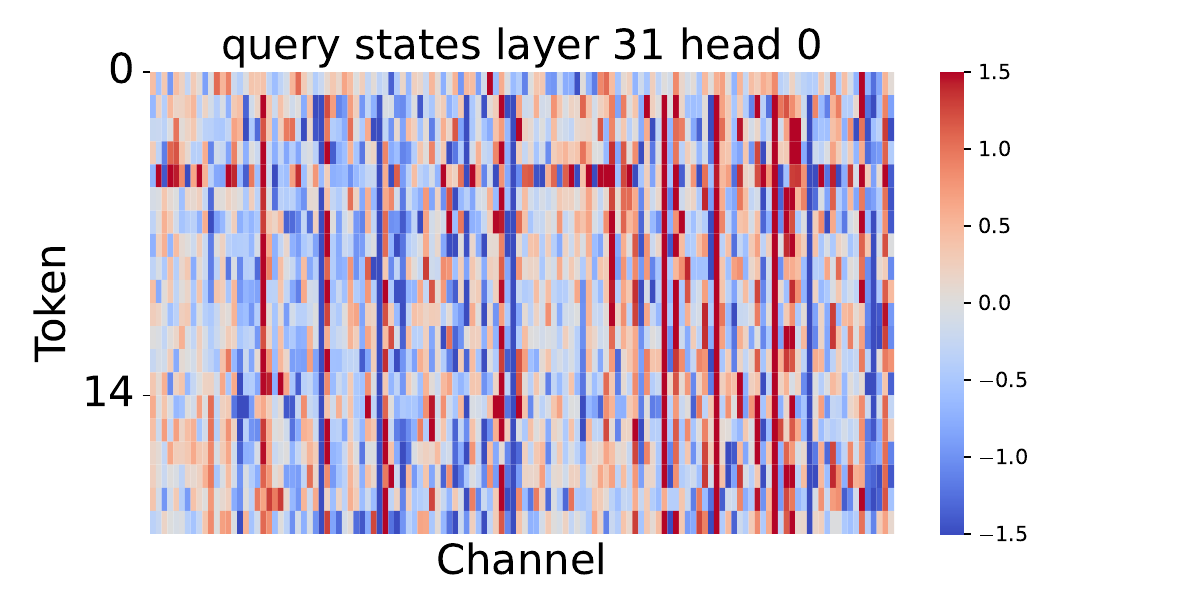}
    \end{subfigure}
    \hspace{-11mm} 
    \begin{subfigure}{0.37\textwidth}
        \centering
    \includegraphics[width=\linewidth]{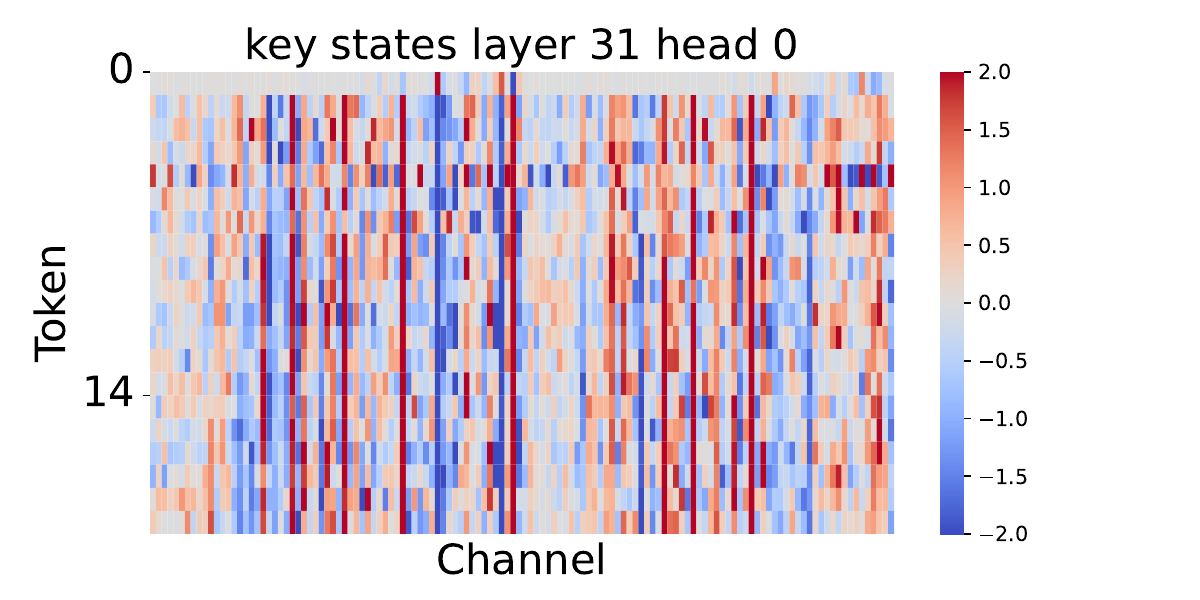}
    \end{subfigure}
    \hspace{-11mm} 
    \begin{subfigure}{0.37\textwidth}
        \centering
    \includegraphics[width=\linewidth]{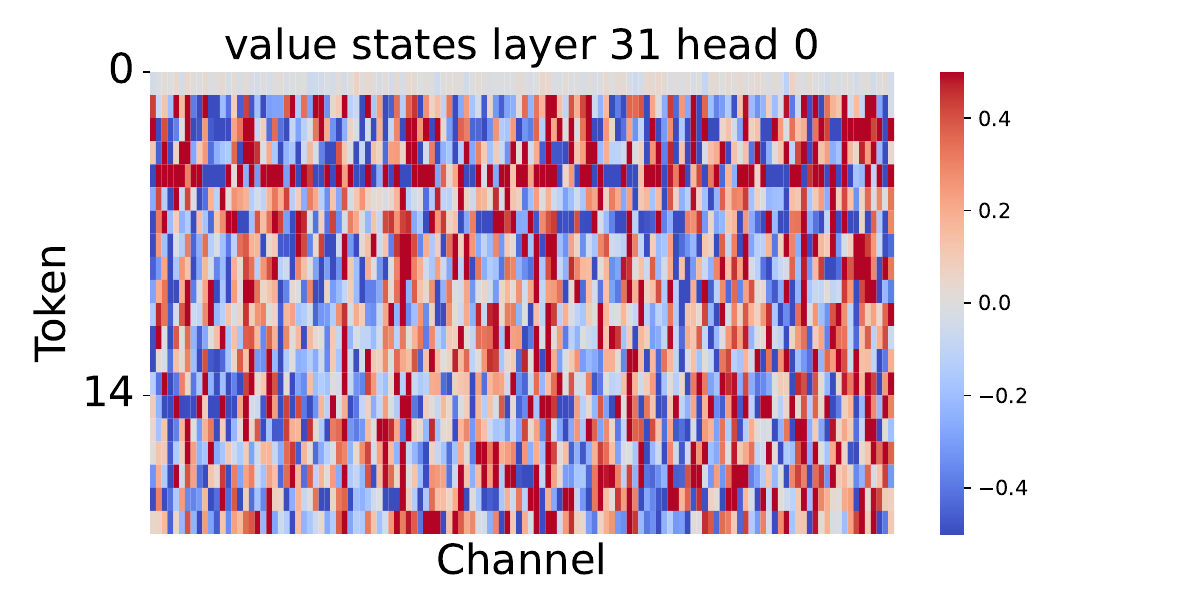}
    \end{subfigure}
    \caption{Queries, Keys, and Values for LLaMA2-7B using Prompt 2, with attention sinks occurring in layers beyond layer 0 and 1, at tokens 0 and 4.}
\label{QKV_appendix_state_2}
\end{figure}
\begin{figure}[t]
    \centering    
    \begin{subfigure}{0.32\textwidth}
        \centering
    \includegraphics[width=\linewidth]{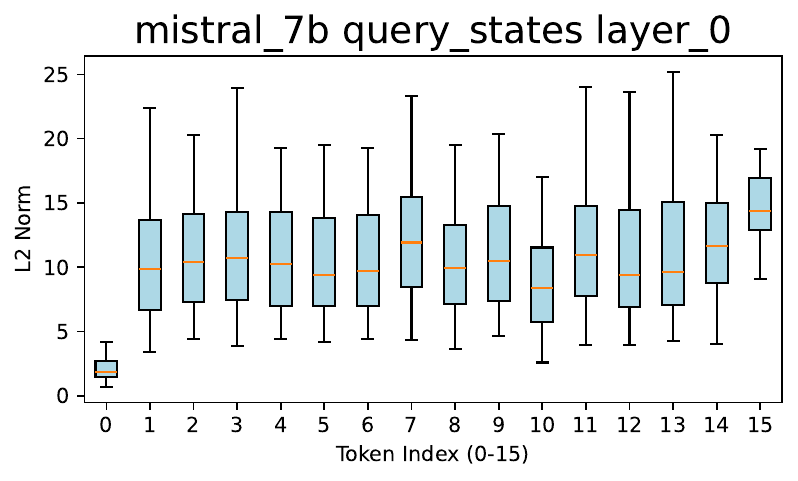}
    \end{subfigure}
    \begin{subfigure}{0.32\textwidth}
        \centering
    \includegraphics[width=\linewidth]{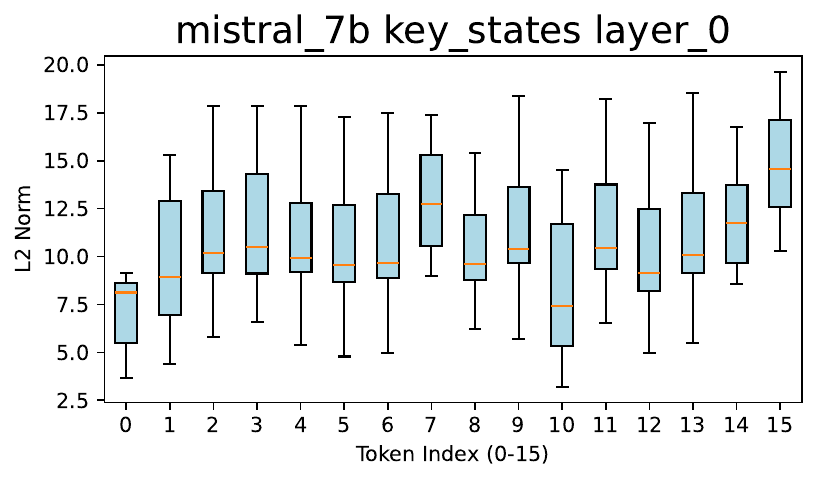}
    \end{subfigure}
    \begin{subfigure}{0.32\textwidth}
        \centering
    \includegraphics[width=\linewidth]{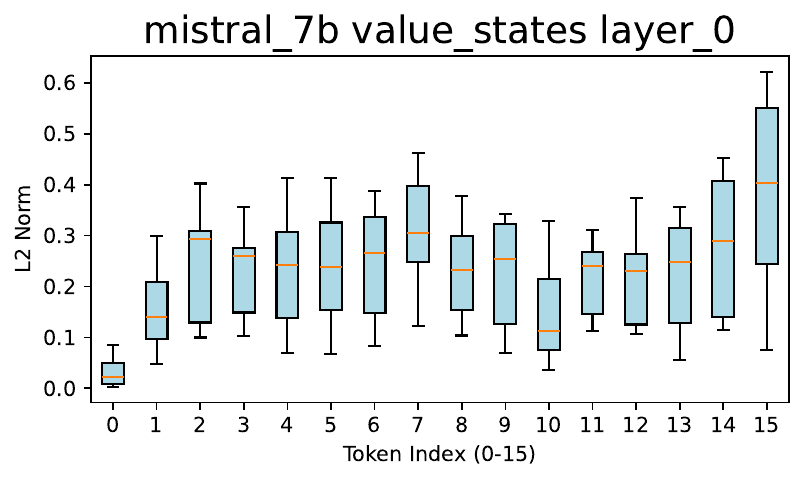}
    \end{subfigure}
    \begin{subfigure}{0.32\textwidth}
        \centering
    \includegraphics[width=\linewidth]{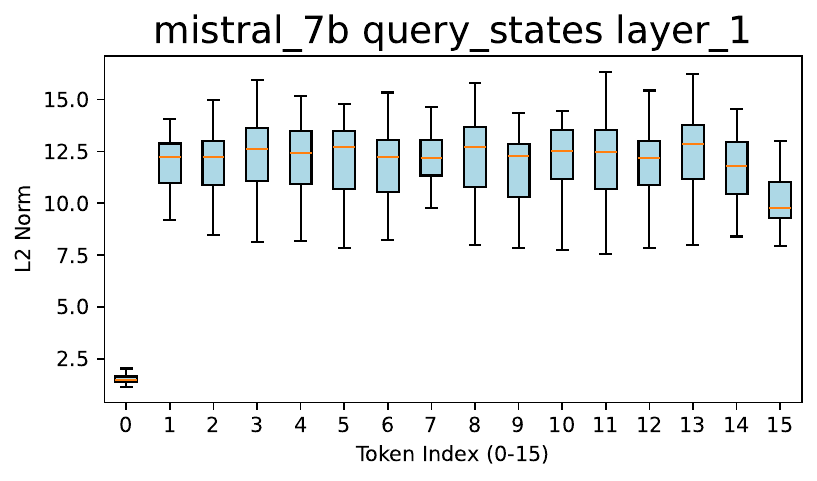}
    \end{subfigure}
    \begin{subfigure}{0.32\textwidth}
        \centering
    \includegraphics[width=\linewidth]{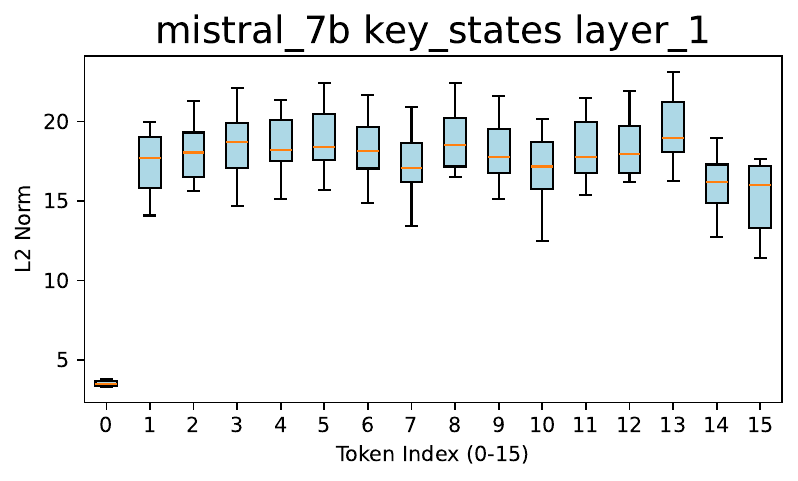}
    \end{subfigure}
    \begin{subfigure}{0.32\textwidth}
        \centering
    \includegraphics[width=\linewidth]{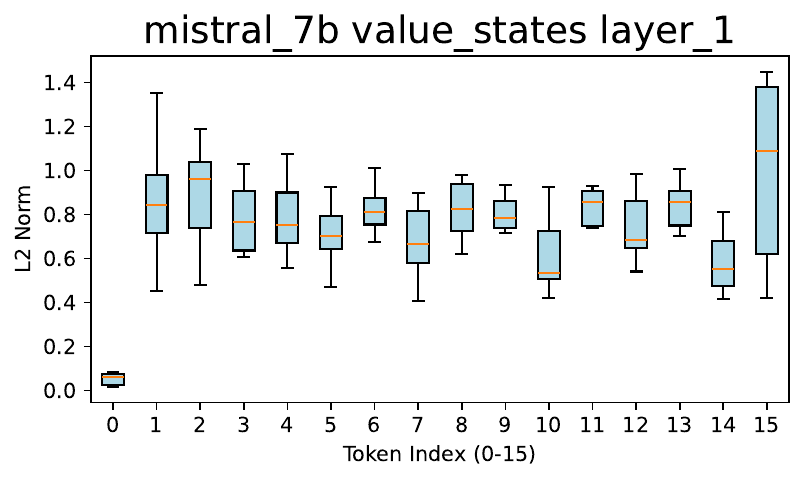}
    \end{subfigure}
    \begin{subfigure}{0.32\textwidth}
        \centering
    \includegraphics[width=\linewidth]{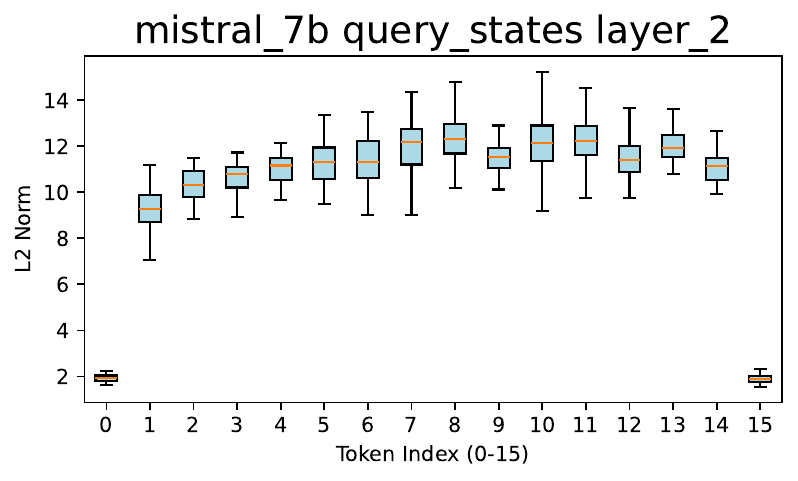}
    \end{subfigure}
    \begin{subfigure}{0.32\textwidth}
        \centering
    \includegraphics[width=\linewidth]{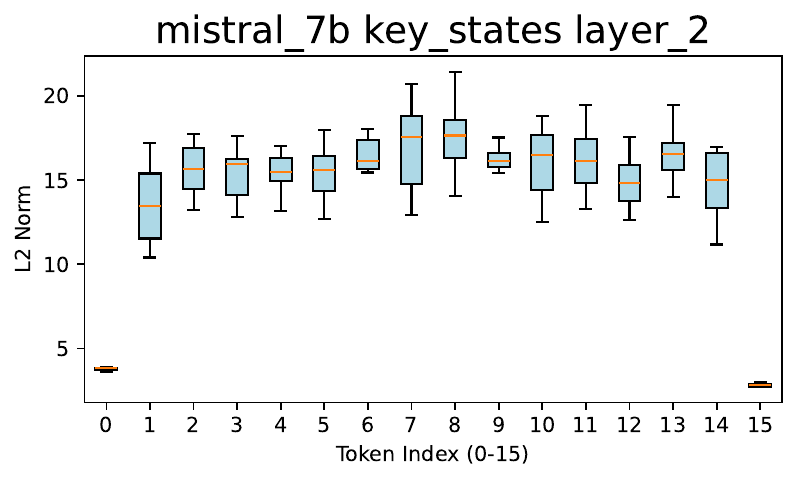}
    \end{subfigure}
    \begin{subfigure}{0.32\textwidth}
        \centering
    \includegraphics[width=\linewidth]{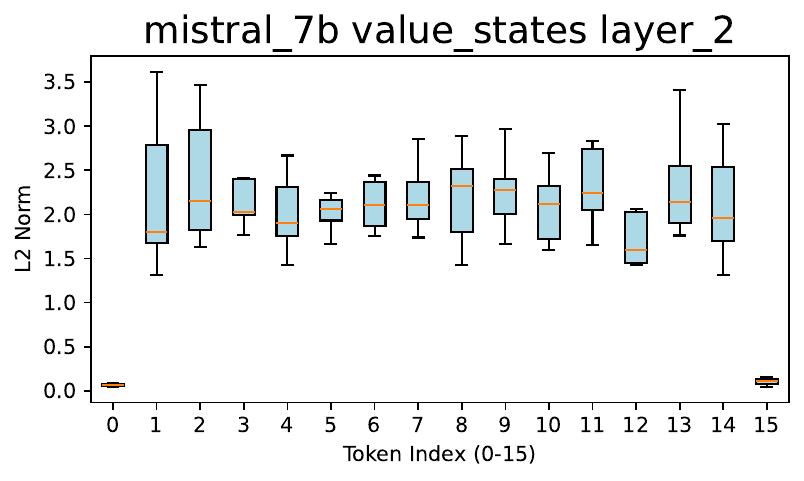}
    \end{subfigure}
    \begin{subfigure}{0.32\textwidth}
        \centering
    \includegraphics[width=\linewidth]{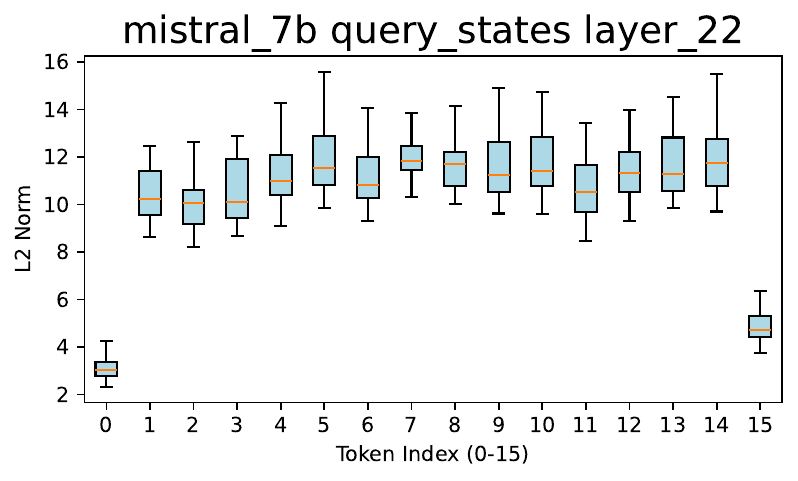}
    \end{subfigure}
    \begin{subfigure}{0.32\textwidth}
        \centering
    \includegraphics[width=\linewidth]{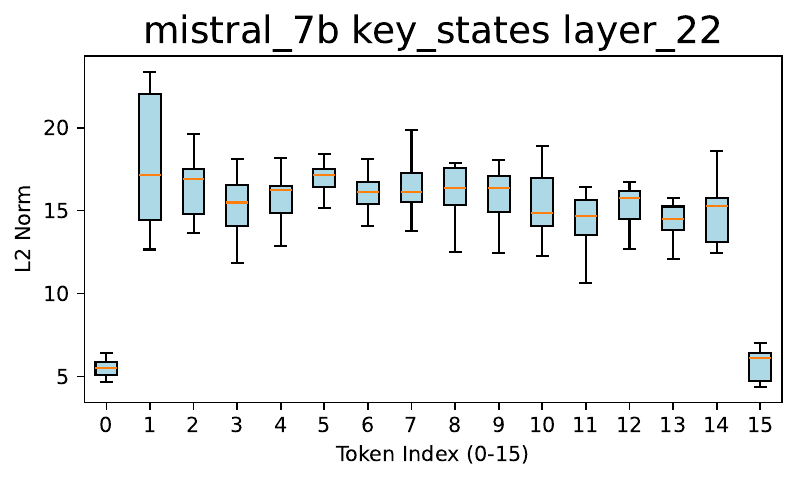}
    \end{subfigure}
    \begin{subfigure}{0.32\textwidth}
        \centering
    \includegraphics[width=\linewidth]{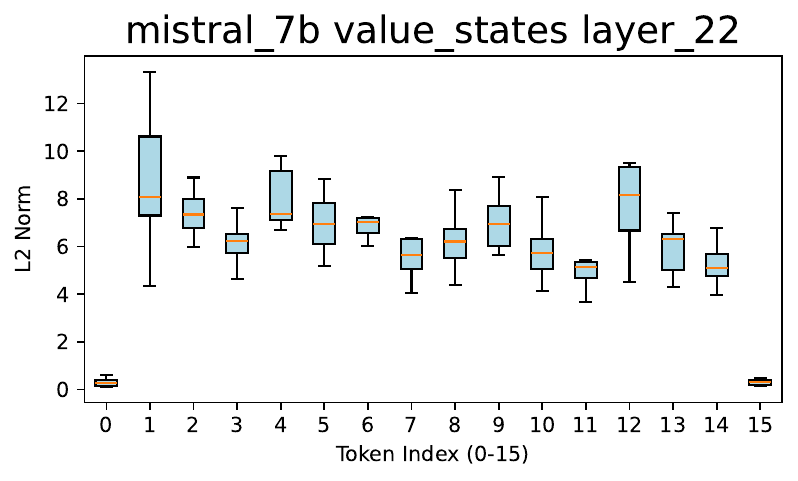}
    \end{subfigure}
    \begin{subfigure}{0.32\textwidth}
        \centering
    \includegraphics[width=\linewidth]{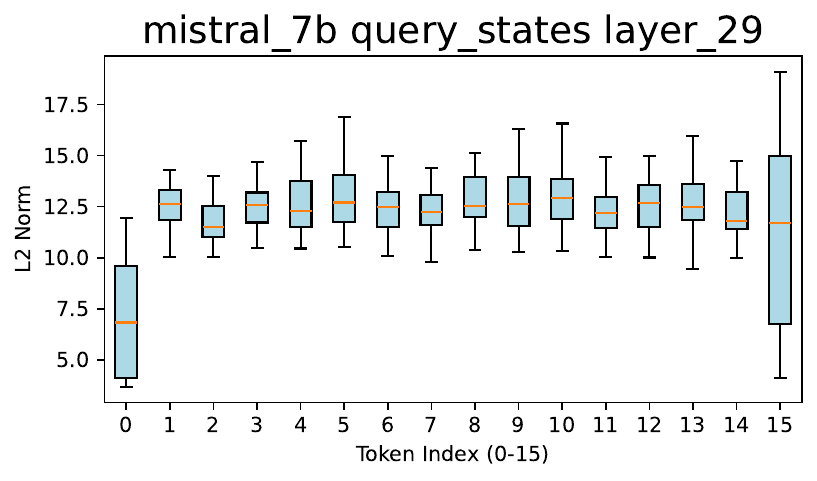}
    \end{subfigure}
    \begin{subfigure}{0.32\textwidth}
        \centering
    \includegraphics[width=\linewidth]{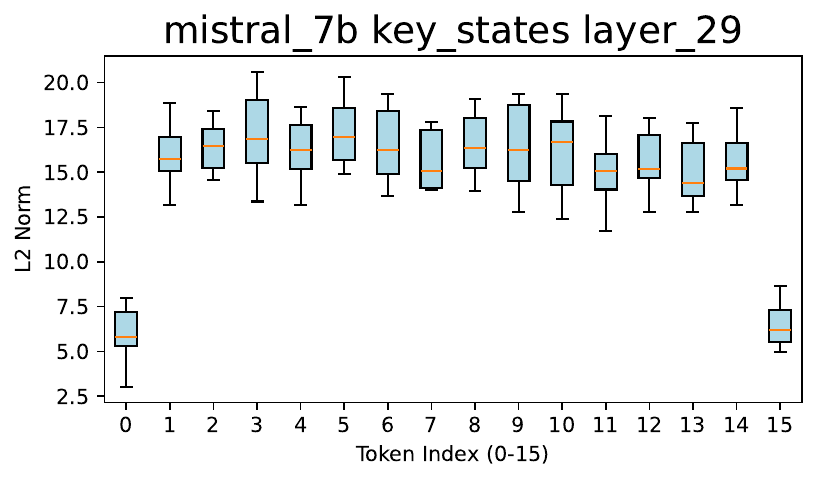}
    \end{subfigure}
    \begin{subfigure}{0.32\textwidth}
        \centering
    \includegraphics[width=\linewidth]{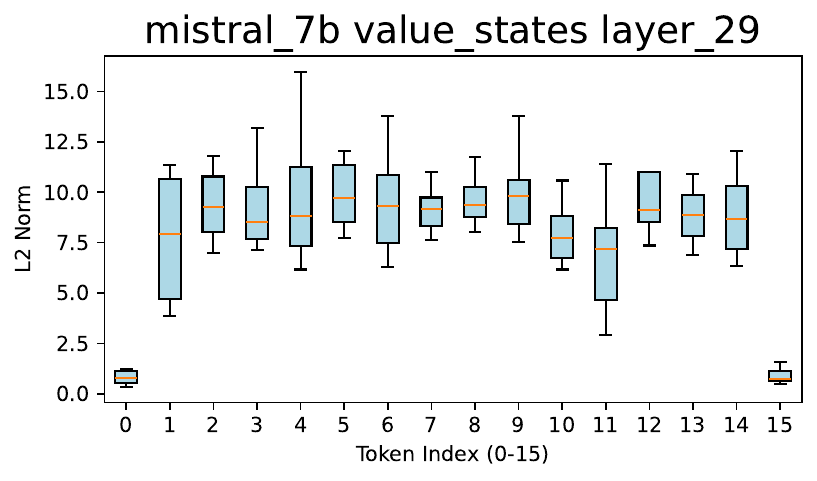}
    \end{subfigure}
    \begin{subfigure}{0.32\textwidth}
        \centering
    \includegraphics[width=\linewidth]{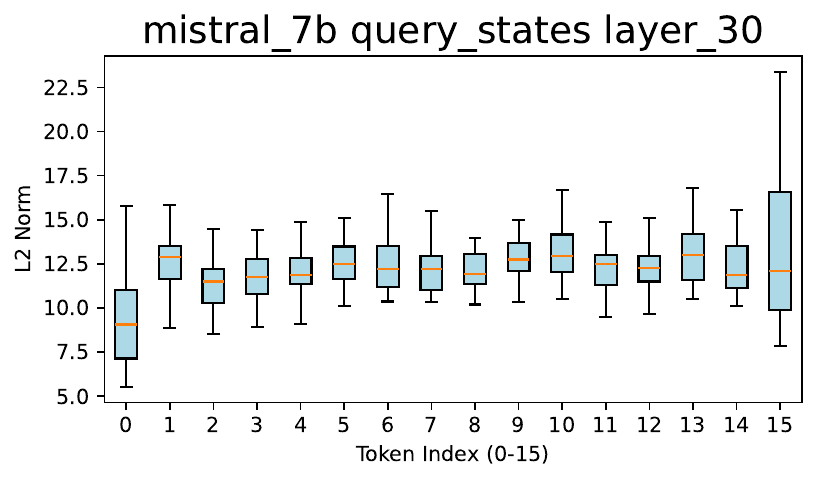}
    \end{subfigure}
    \begin{subfigure}{0.32\textwidth}
        \centering
    \includegraphics[width=\linewidth]{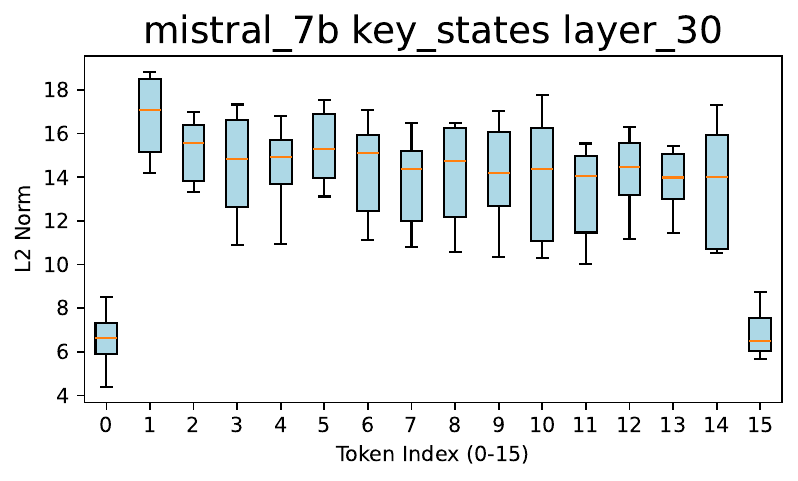}
    \end{subfigure}
    \begin{subfigure}{0.32\textwidth}
        \centering
    \includegraphics[width=\linewidth]{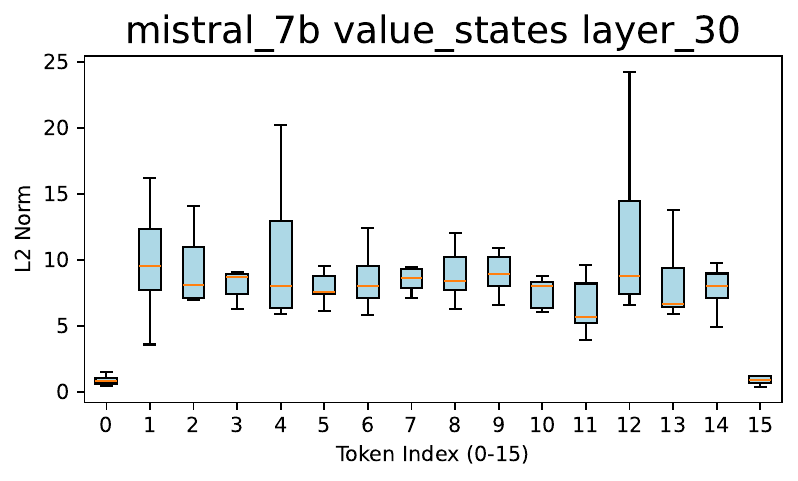}
    \end{subfigure}
    \begin{subfigure}{0.32\textwidth}
        \centering
    \includegraphics[width=\linewidth]{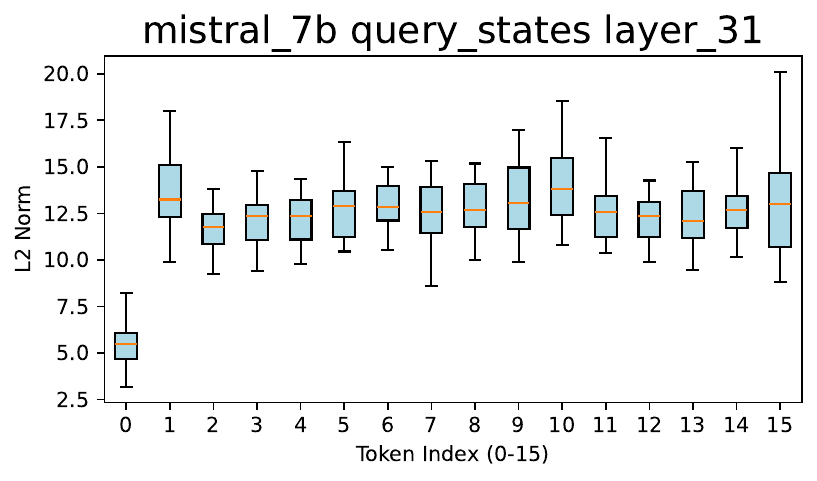}
    \end{subfigure}
    \begin{subfigure}{0.32\textwidth}
        \centering
    \includegraphics[width=\linewidth]{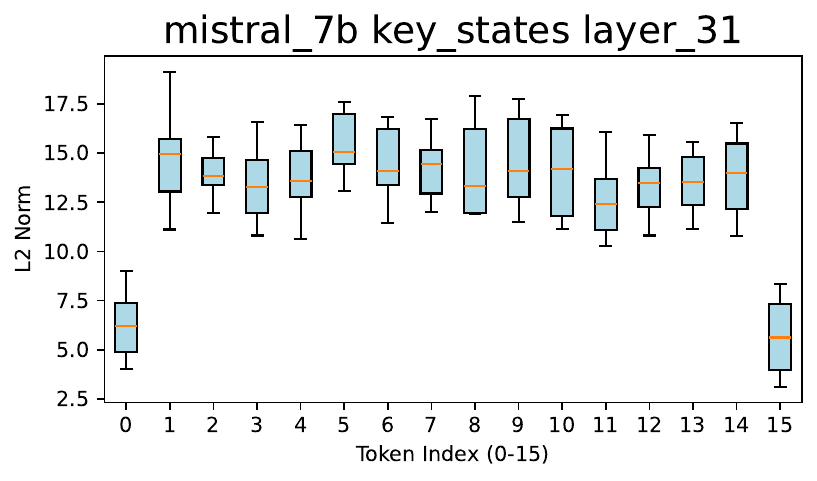}
    \end{subfigure}
    \begin{subfigure}{0.32\textwidth}
        \centering
    \includegraphics[width=\linewidth]{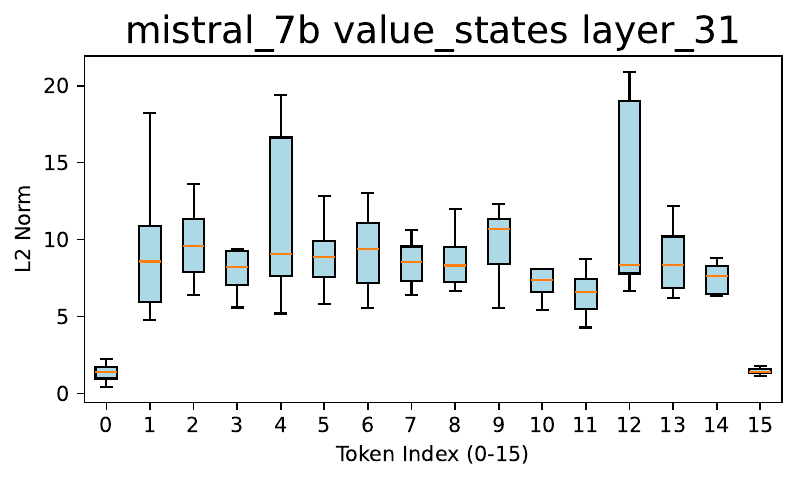}
    \end{subfigure}
    \caption{L2 norm distributions of Queries, Keys, and Values for Mistral-7B using Prompt 1, with attention sinks occurring in layers beyond layer 0 and 1, at tokens 0 and 15.}
\label{QKV_appendix_3}
\end{figure}
\begin{figure}[t]
    \centering    
    \begin{subfigure}{0.32\textwidth}
        \centering
    \includegraphics[width=\linewidth]{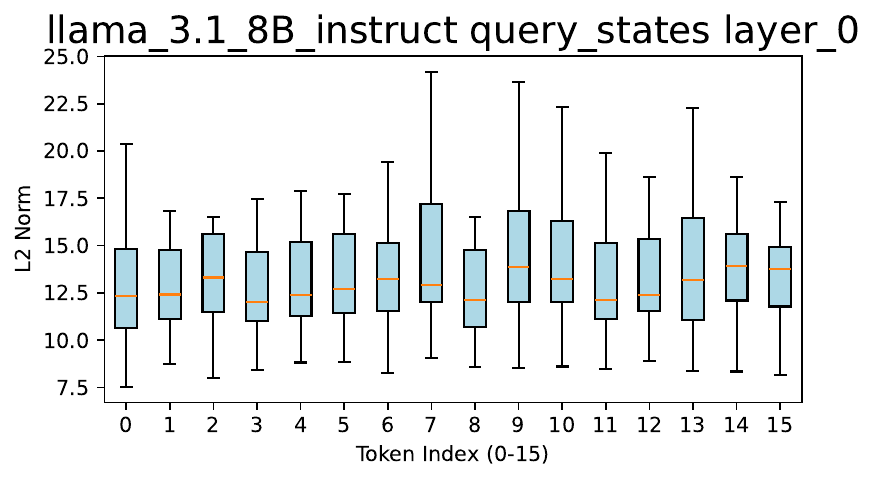}
    \end{subfigure}
    \begin{subfigure}{0.32\textwidth}
        \centering
    \includegraphics[width=\linewidth]{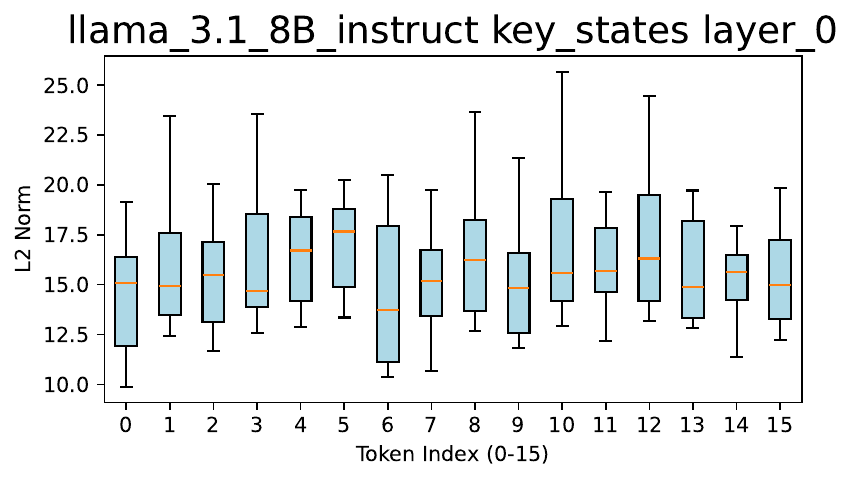}
    \end{subfigure}
    \begin{subfigure}{0.32\textwidth}
        \centering
    \includegraphics[width=\linewidth]{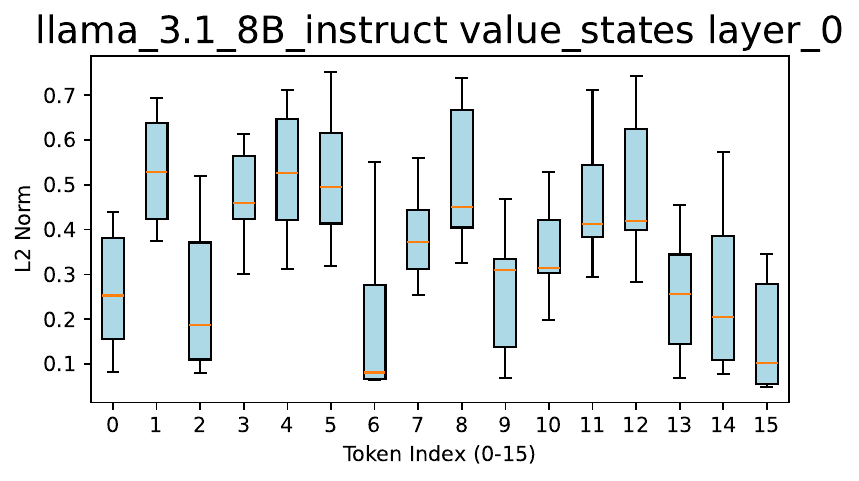}
    \end{subfigure}
    \begin{subfigure}{0.32\textwidth}
        \centering
    \includegraphics[width=\linewidth]{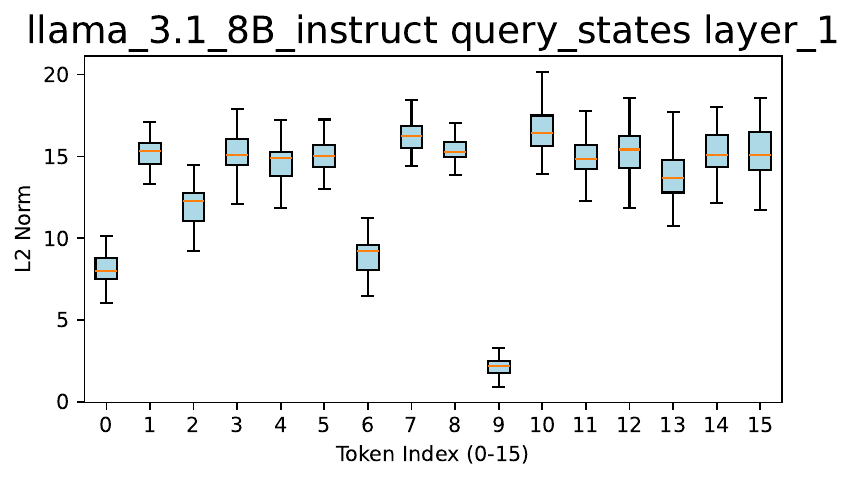}
    \end{subfigure}
    \begin{subfigure}{0.32\textwidth}
        \centering
    \includegraphics[width=\linewidth]{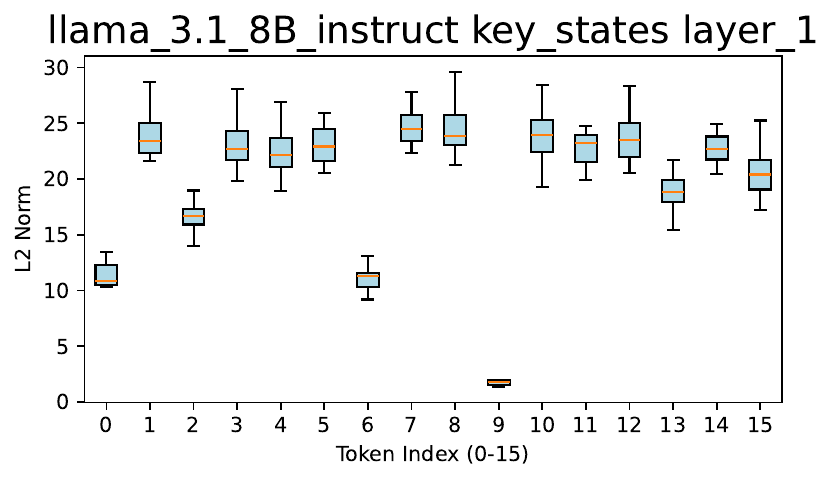}
    \end{subfigure}
    \begin{subfigure}{0.32\textwidth}
        \centering
    \includegraphics[width=\linewidth]{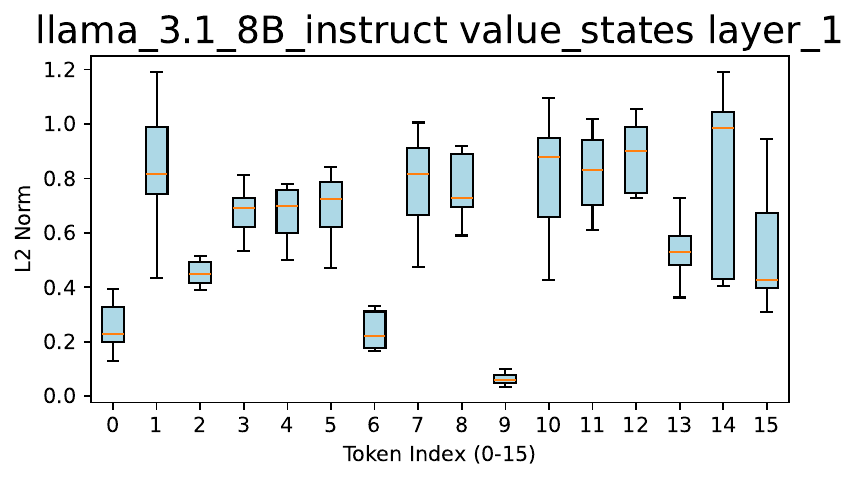}
    \end{subfigure}
    \begin{subfigure}{0.32\textwidth}
        \centering
    \includegraphics[width=\linewidth]{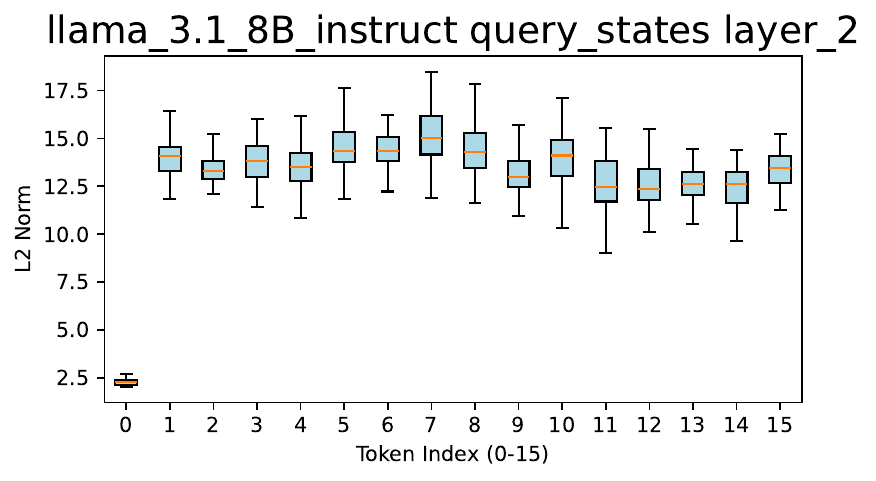}
    \end{subfigure}
    \begin{subfigure}{0.32\textwidth}
        \centering
    \includegraphics[width=\linewidth]{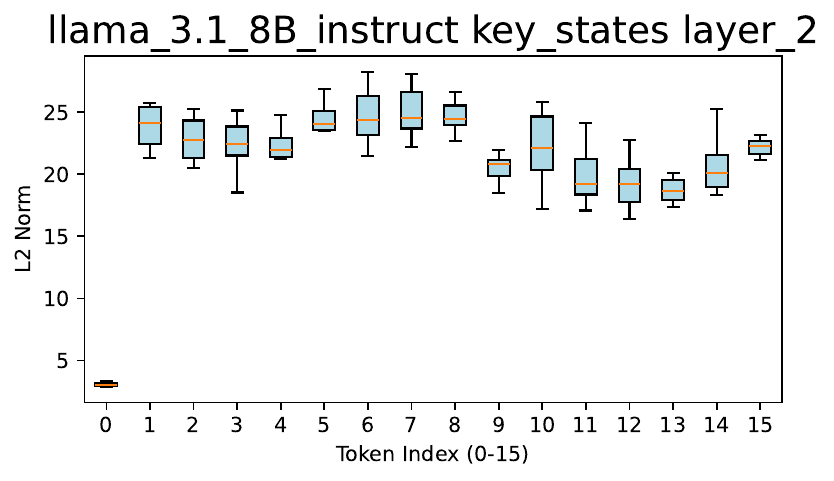}
    \end{subfigure}
    \begin{subfigure}{0.32\textwidth}
        \centering
    \includegraphics[width=\linewidth]{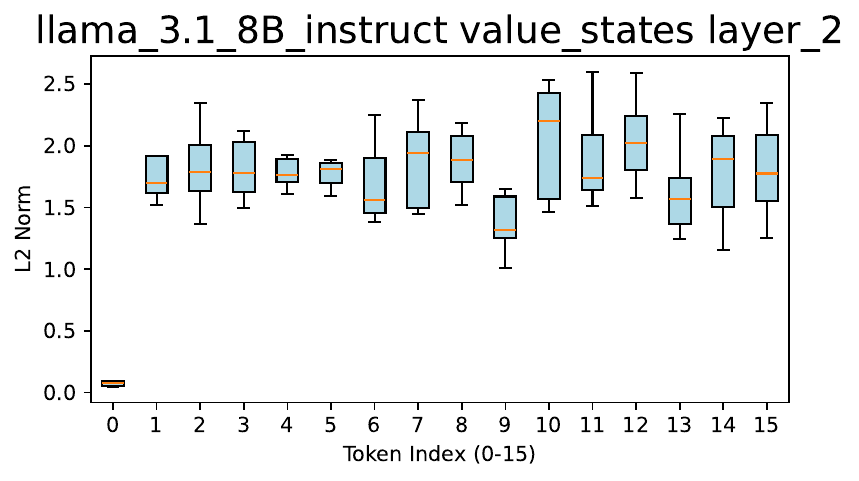}
    \end{subfigure}
    \begin{subfigure}{0.32\textwidth}
        \centering
    \includegraphics[width=\linewidth]{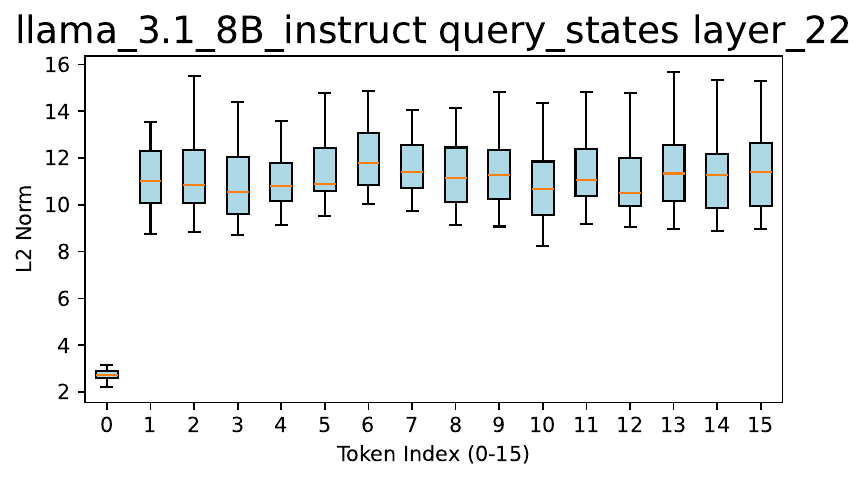}
    \end{subfigure}
    \begin{subfigure}{0.32\textwidth}
        \centering
    \includegraphics[width=\linewidth]{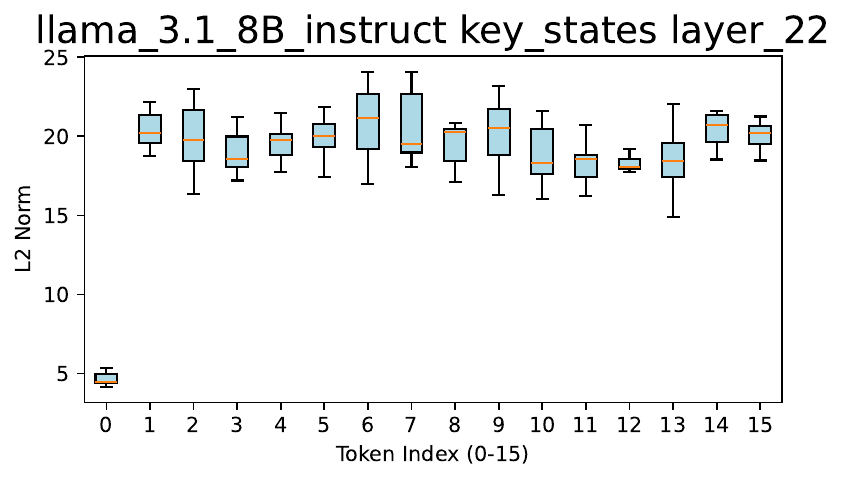}
    \end{subfigure}
    \begin{subfigure}{0.32\textwidth}
        \centering
    \includegraphics[width=\linewidth]{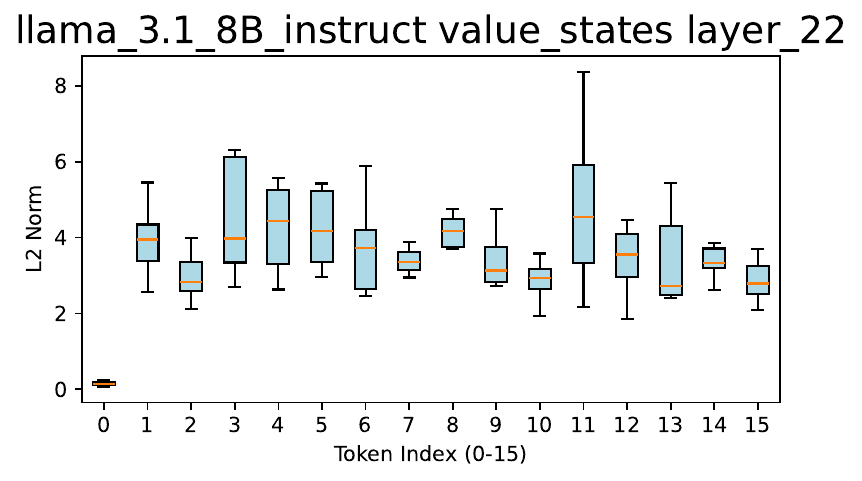}
    \end{subfigure}
    \begin{subfigure}{0.32\textwidth}
        \centering
    \includegraphics[width=\linewidth]{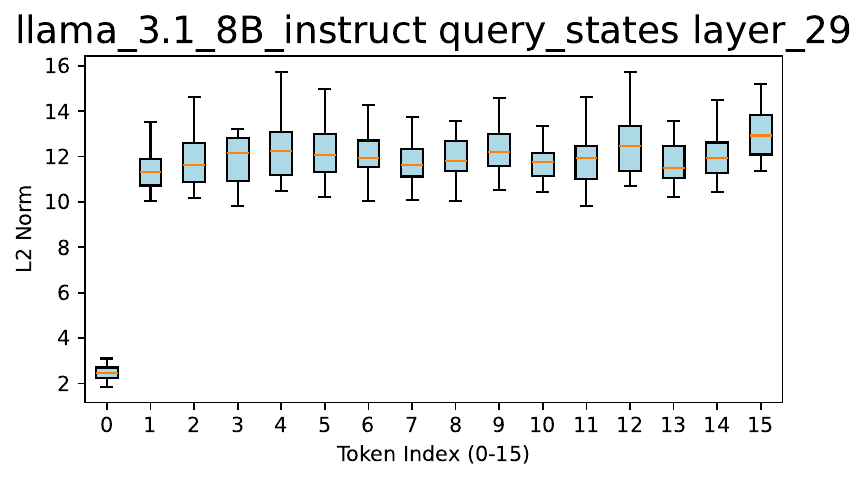}
    \end{subfigure}
    \begin{subfigure}{0.32\textwidth}
        \centering
    \includegraphics[width=\linewidth]{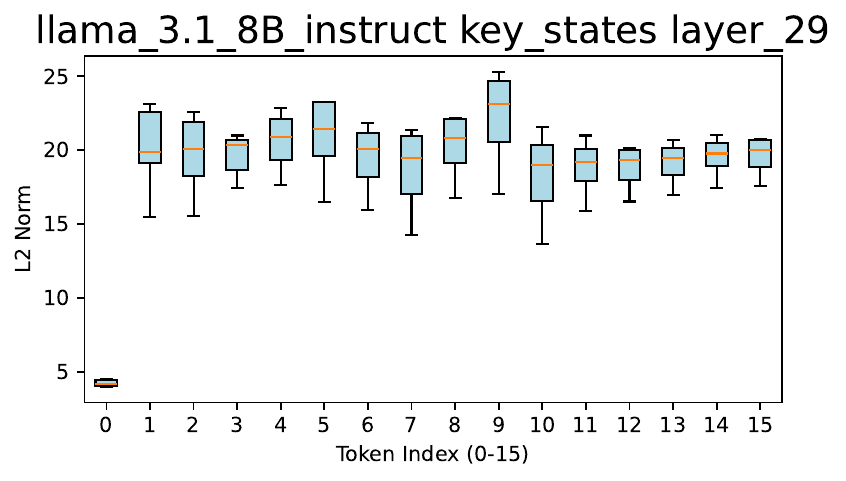}
    \end{subfigure}
    \begin{subfigure}{0.32\textwidth}
        \centering
    \includegraphics[width=\linewidth]{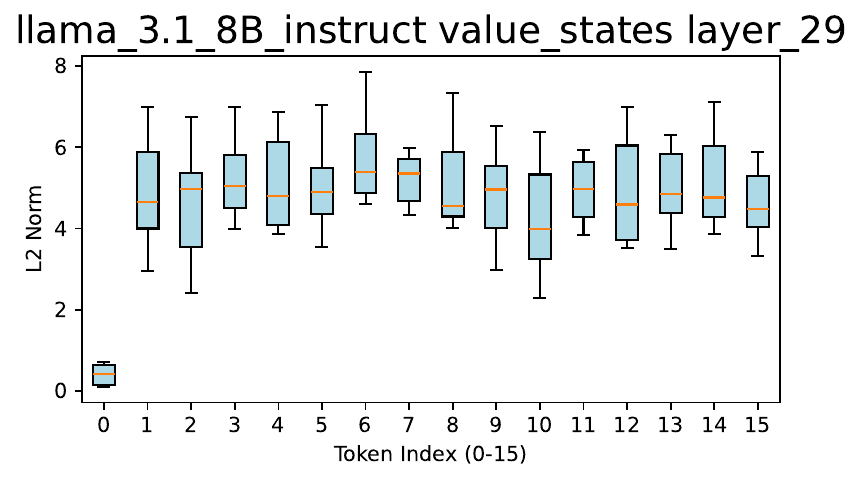}
    \end{subfigure}
    \begin{subfigure}{0.32\textwidth}
        \centering
    \includegraphics[width=\linewidth]{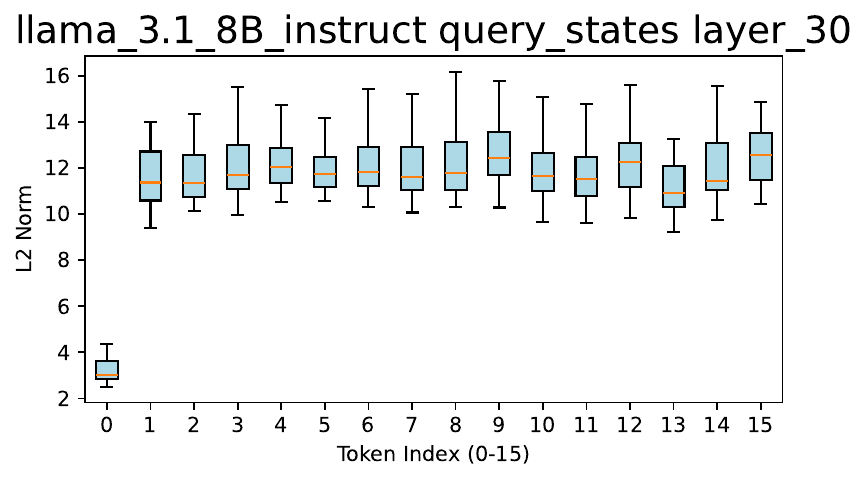}
    \end{subfigure}
    \begin{subfigure}{0.32\textwidth}
        \centering
    \includegraphics[width=\linewidth]{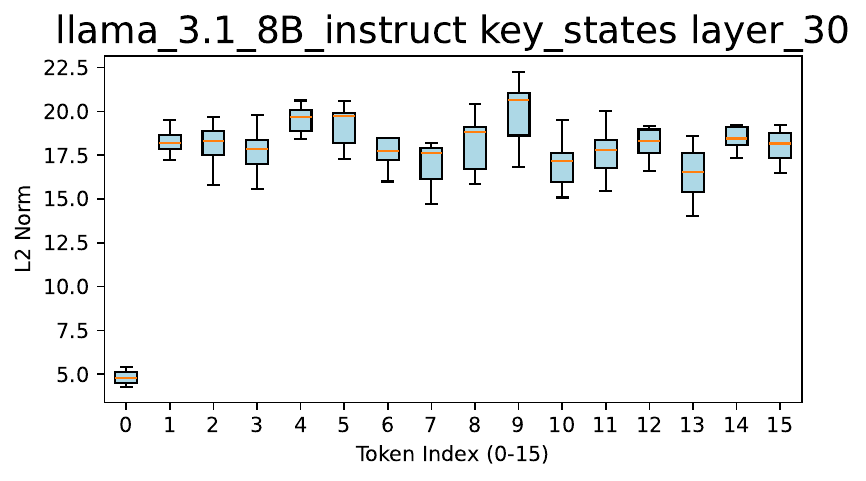}
    \end{subfigure}
    \begin{subfigure}{0.32\textwidth}
        \centering
    \includegraphics[width=\linewidth]{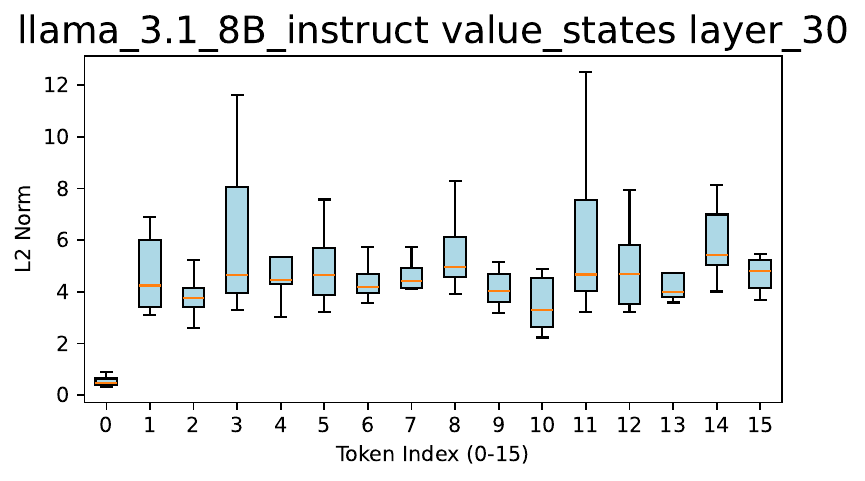}
    \end{subfigure}
    \begin{subfigure}{0.32\textwidth}
        \centering
    \includegraphics[width=\linewidth]{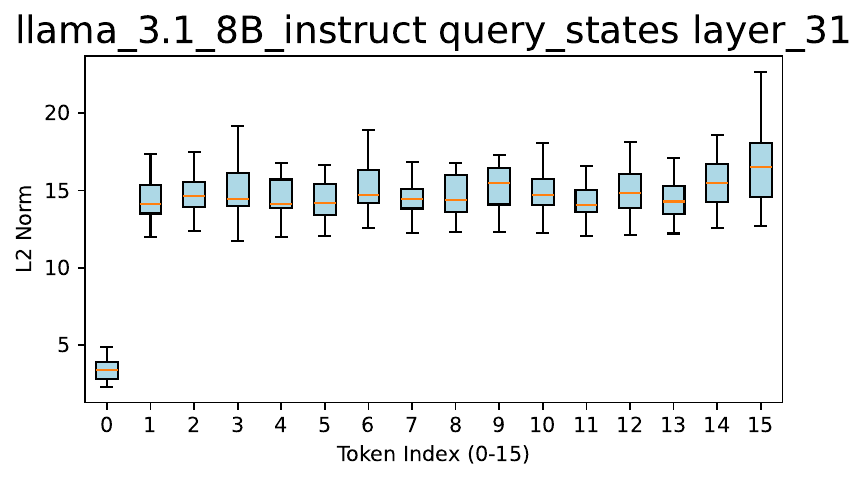}
    \end{subfigure}
    \begin{subfigure}{0.32\textwidth}
        \centering
    \includegraphics[width=\linewidth]{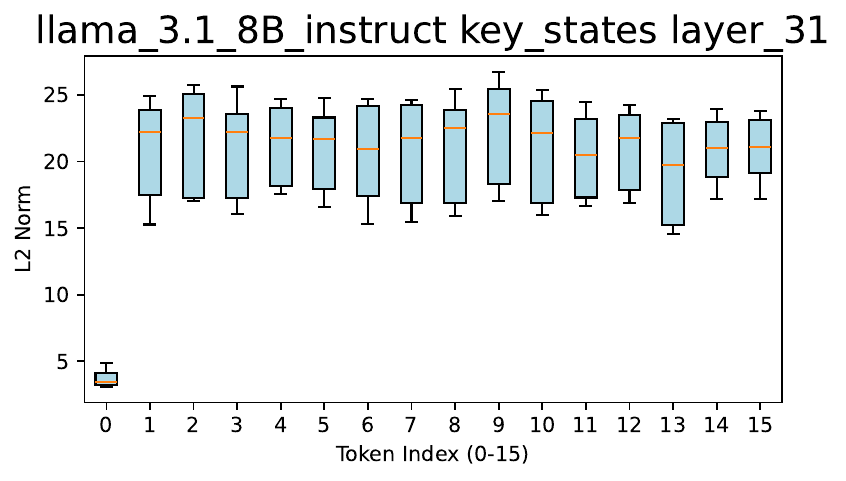}
    \end{subfigure}
    \begin{subfigure}{0.32\textwidth}
        \centering
    \includegraphics[width=\linewidth]{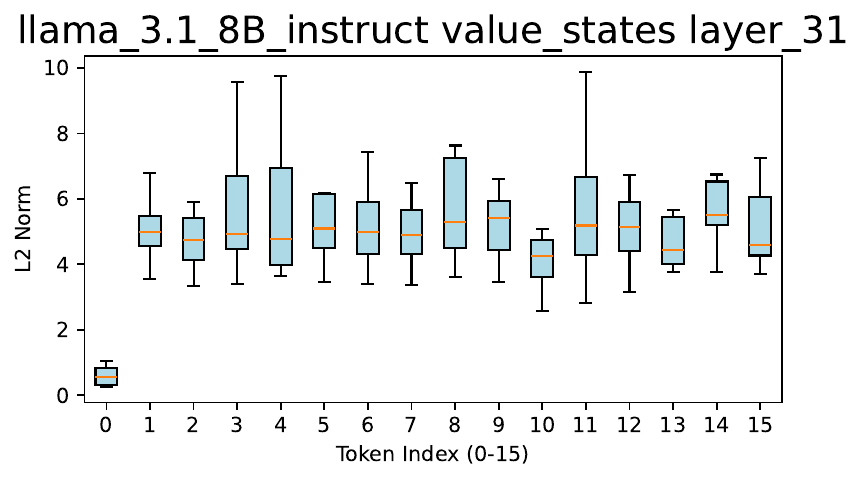}
    \end{subfigure}
    \caption{L2 norm distributions of Queries, Keys, and Values for LLaMA3.1-8B-instruct using Prompt 1, with attention sinks occurring in layers beyond layer 0 and 1, at token 0.}
\label{QKV_appendix_4}
\end{figure}
}

\begin{thebibliography}{51}
\providecommand{\natexlab}[1]{#1}
\providecommand{\url}[1]{\texttt{#1}}
\expandafter\ifx\csname urlstyle\endcsname\relax
  \providecommand{\doi}[1]{doi: #1}\else
  \providecommand{\doi}{doi: \begingroup \urlstyle{rm}\Url}\fi

\bibitem[Achiam et~al.(2023)Achiam, Adler, Agarwal, Ahmad, Akkaya, Aleman, Almeida, Altenschmidt, Altman, Anadkat, et~al.]{achiam2023gpt}
Josh Achiam, Steven Adler, Sandhini Agarwal, Lama Ahmad, Ilge Akkaya, Florencia~Leoni Aleman, Diogo Almeida, Janko Altenschmidt, Sam Altman, Shyamal Anadkat, et~al.
\newblock Gpt-4 technical report.
\newblock \emph{arXiv preprint arXiv:2303.08774}, 2023.

\bibitem[Ainslie et~al.(2023)Ainslie, Lee-Thorp, De~Jong, Zemlyanskiy, Lebr{\'o}n, and Sanghai]{ainslie2023gqa}
Joshua Ainslie, James Lee-Thorp, Michiel De~Jong, Yury Zemlyanskiy, Federico Lebr{\'o}n, and Sumit Sanghai.
\newblock Gqa: Training generalized multi-query transformer models from multi-head checkpoints.
\newblock \emph{arXiv preprint arXiv:2305.13245}, 2023.

\bibitem[An et~al.(2025)An, Zhao, Yu, Tang, and Wang]{an2025systematic}
Yongqi An, Xu~Zhao, Tao Yu, Ming Tang, and Jinqiao Wang.
\newblock Systematic outliers in large language models.
\newblock \emph{arXiv preprint arXiv:2502.06415}, 2025.

\bibitem[Bondarenko et~al.(2021)Bondarenko, Nagel, and Blankevoort]{bondarenko2021understanding}
Yelysei Bondarenko, Markus Nagel, and Tijmen Blankevoort.
\newblock Understanding and overcoming the challenges of efficient transformer quantization.
\newblock \emph{arXiv preprint arXiv:2109.12948}, 2021.

\bibitem[Bondarenko et~al.(2023)Bondarenko, Nagel, and Blankevoort]{bondarenko2023quantizable}
Yelysei Bondarenko, Markus Nagel, and Tijmen Blankevoort.
\newblock Quantizable transformers: Removing outliers by helping attention heads do nothing.
\newblock \emph{Advances in Neural Information Processing Systems}, 36:\penalty0 75067--75096, 2023.

\bibitem[Cai et~al.(2024)Cai, Zhang, Gao, Liu, Liu, Lu, Xiong, Dong, Chang, Hu, et~al.]{cai2024pyramidkv}
Zefan Cai, Yichi Zhang, Bofei Gao, Yuliang Liu, Tianyu Liu, Keming Lu, Wayne Xiong, Yue Dong, Baobao Chang, Junjie Hu, et~al.
\newblock Pyramidkv: Dynamic kv cache compression based on pyramidal information funneling.
\newblock \emph{arXiv preprint arXiv:2406.02069}, 2024.

\bibitem[Chang et~al.(2024)Chang, Lin, Lin, Chen, Hu, Wang, Huang, Ceze, Abdelfattah, and Wu]{chang2024palu}
Chi-Chih Chang, Wei-Cheng Lin, Chien-Yu Lin, Chong-Yan Chen, Yu-Fang Hu, Pei-Shuo Wang, Ning-Chi Huang, Luis Ceze, Mohamed~S Abdelfattah, and Kai-Chiang Wu.
\newblock Palu: Compressing kv-cache with low-rank projection.
\newblock \emph{arXiv preprint arXiv:2407.21118}, 2024.

\bibitem[Chaplot(2023)]{chaplot2023albert}
Devendra~Singh Chaplot.
\newblock Albert q. jiang, alexandre sablayrolles, arthur mensch, chris bamford, devendra singh chaplot, diego de las casas, florian bressand, gianna lengyel, guillaume lample, lucile saulnier, l{\'e}lio renard lavaud, marie-anne lachaux, pierre stock, teven le scao, thibaut lavril, thomas wang, timoth{\'e}e lacroix, william el sayed.
\newblock \emph{arXiv preprint arXiv:2310.06825}, 2023.

\bibitem[Clark et~al.(2019)Clark, Khandelwal, Levy, and Manning]{clark2019does}
Kevin Clark, Urvashi Khandelwal, Omer Levy, and Christopher~D Manning.
\newblock What does bert look at? an analysis of bert's attention.
\newblock \emph{arXiv preprint arXiv:1906.04341}, 2019.

\bibitem[Dao(2023)]{dao2023flashattention}
Tri Dao.
\newblock Flashattention-2: Faster attention with better parallelism and work partitioning.
\newblock \emph{arXiv preprint arXiv:2307.08691}, 2023.

\bibitem[Dao et~al.(2022)Dao, Fu, Ermon, Rudra, and R{\'e}]{dao2022flashattention}
Tri Dao, Dan Fu, Stefano Ermon, Atri Rudra, and Christopher R{\'e}.
\newblock Flashattention: Fast and memory-efficient exact attention with io-awareness.
\newblock \emph{Advances in neural information processing systems}, 35:\penalty0 16344--16359, 2022.

\bibitem[Darcet et~al.(2023)Darcet, Oquab, Mairal, and Bojanowski]{darcet2023vision}
Timoth{\'e}e Darcet, Maxime Oquab, Julien Mairal, and Piotr Bojanowski.
\newblock Vision transformers need registers.
\newblock \emph{arXiv preprint arXiv:2309.16588}, 2023.

\bibitem[Devlin et~al.(2019)Devlin, Chang, Lee, and Toutanova]{devlin2019bert}
Jacob Devlin, Ming-Wei Chang, Kenton Lee, and Kristina Toutanova.
\newblock Bert: Pre-training of deep bidirectional transformers for language understanding.
\newblock In \emph{Proceedings of the 2019 conference of the North American chapter of the association for computational linguistics: human language technologies, volume 1 (long and short papers)}, pp.\  4171--4186, 2019.

\bibitem[Dosovitskiy et~al.(2020)Dosovitskiy, Beyer, Kolesnikov, Weissenborn, Zhai, Unterthiner, Dehghani, Minderer, Heigold, Gelly, et~al.]{dosovitskiy2020image}
Alexey Dosovitskiy, Lucas Beyer, Alexander Kolesnikov, Dirk Weissenborn, Xiaohua Zhai, Thomas Unterthiner, Mostafa Dehghani, Matthias Minderer, Georg Heigold, Sylvain Gelly, et~al.
\newblock An image is worth 16x16 words: Transformers for image recognition at scale.
\newblock \emph{arXiv preprint arXiv:2010.11929}, 2020.

\bibitem[Duanmu et~al.(2024)Duanmu, Yuan, Li, Duan, Zhang, and Lin]{duanmu2024skvq}
Haojie Duanmu, Zhihang Yuan, Xiuhong Li, Jiangfei Duan, Xingcheng Zhang, and Dahua Lin.
\newblock Skvq: Sliding-window key and value cache quantization for large language models.
\newblock \emph{arXiv preprint arXiv:2405.06219}, 2024.

\bibitem[Dubey et~al.(2024)Dubey, Jauhri, Pandey, Kadian, Al-Dahle, Letman, Mathur, Schelten, Yang, Fan, et~al.]{dubey2024llama}
Abhimanyu Dubey, Abhinav Jauhri, Abhinav Pandey, Abhishek Kadian, Ahmad Al-Dahle, Aiesha Letman, Akhil Mathur, Alan Schelten, Amy Yang, Angela Fan, et~al.
\newblock The llama 3 herd of models.
\newblock \emph{arXiv preprint arXiv:2407.21783}, 2024.

\bibitem[Gu et~al.(2024)Gu, Pang, Du, Liu, Zhang, Du, Wang, and Lin]{gu2024attention}
Xiangming Gu, Tianyu Pang, Chao Du, Qian Liu, Fengzhuo Zhang, Cunxiao Du, Ye~Wang, and Min Lin.
\newblock When attention sink emerges in language models: An empirical view.
\newblock \emph{arXiv preprint arXiv:2410.10781}, 2024.

\bibitem[Guo et~al.(2025)Guo, Yang, Zhang, Song, Zhang, Xu, Zhu, Ma, Wang, Bi, et~al.]{guo2025deepseek}
Daya Guo, Dejian Yang, Haowei Zhang, Junxiao Song, Ruoyu Zhang, Runxin Xu, Qihao Zhu, Shirong Ma, Peiyi Wang, Xiao Bi, et~al.
\newblock Deepseek-r1: Incentivizing reasoning capability in llms via reinforcement learning.
\newblock \emph{arXiv preprint arXiv:2501.12948}, 2025.

\bibitem[Guo et~al.(2024)Guo, Pai, Bai, Jiao, Jordan, and Mei]{guo2024active}
Tianyu Guo, Druv Pai, Yu~Bai, Jiantao Jiao, Michael~I Jordan, and Song Mei.
\newblock Active-dormant attention heads: Mechanistically demystifying extreme-token phenomena in llms.
\newblock \emph{arXiv preprint arXiv:2410.13835}, 2024.

\bibitem[Hadi et~al.(2023)Hadi, Qureshi, Shah, Irfan, Zafar, Shaikh, Akhtar, Wu, Mirjalili, et~al.]{hadi2023survey}
Muhammad~Usman Hadi, Rizwan Qureshi, Abbas Shah, Muhammad Irfan, Anas Zafar, Muhammad~Bilal Shaikh, Naveed Akhtar, Jia Wu, Seyedali Mirjalili, et~al.
\newblock A survey on large language models: Applications, challenges, limitations, and practical usage.
\newblock \emph{Authorea Preprints}, 3, 2023.

\bibitem[He et~al.(2024)He, Zhang, Wu, Liu, Zhou, and Zhuang]{he2024zipcache}
Yefei He, Luoming Zhang, Weijia Wu, Jing Liu, Hong Zhou, and Bohan Zhuang.
\newblock Zipcache: Accurate and efficient kv cache quantization with salient token identification.
\newblock \emph{arXiv preprint arXiv:2405.14256}, 2024.

\bibitem[Hendrycks et~al.(2021)Hendrycks, Burns, Basart, Zou, Mazeika, Song, and Steinhardt]{hendryckstest2021}
Dan Hendrycks, Collin Burns, Steven Basart, Andy Zou, Mantas Mazeika, Dawn Song, and Jacob Steinhardt.
\newblock Measuring massive multitask language understanding.
\newblock \emph{Proceedings of the International Conference on Learning Representations (ICLR)}, 2021.

\bibitem[Hooper et~al.(2025)Hooper, Kim, Mohammadzadeh, Mahoney, Shao, Keutzer, and Gholami]{hooper2025kvquant}
Coleman Hooper, Sehoon Kim, Hiva Mohammadzadeh, Michael~W Mahoney, Sophia Shao, Kurt Keutzer, and Amir Gholami.
\newblock Kvquant: Towards 10 million context length llm inference with kv cache quantization.
\newblock \emph{Advances in Neural Information Processing Systems}, 37:\penalty0 1270--1303, 2025.

\bibitem[Kovaleva et~al.(2019)Kovaleva, Romanov, Rogers, and Rumshisky]{kovaleva2019revealing}
Olga Kovaleva, Alexey Romanov, Anna Rogers, and Anna Rumshisky.
\newblock Revealing the dark secrets of bert.
\newblock \emph{arXiv preprint arXiv:1908.08593}, 2019.

\bibitem[Li et~al.(2024{\natexlab{a}})Li, Li, Tian, Tang, Xu, Chen, Hu, Dong, Li, and Chen]{li2024survey}
Haoyang Li, Yiming Li, Anxin Tian, Tianhao Tang, Zhanchao Xu, Xuejia Chen, Nicole Hu, Wei Dong, Qing Li, and Lei Chen.
\newblock A survey on large language model acceleration based on kv cache management.
\newblock \emph{arXiv preprint arXiv:2412.19442}, 2024{\natexlab{a}}.

\bibitem[Li et~al.(2024{\natexlab{b}})Li, Zhang, Ye, Zhang, Wu, Sun, Ma, and Xie]{li2024flash}
Qingyuan Li, Bo~Zhang, Liang Ye, Yifan Zhang, Wei Wu, Yerui Sun, Lin Ma, and Yuchen Xie.
\newblock Flash communication: Reducing tensor parallelization bottleneck for fast large language model inference.
\newblock \emph{arXiv preprint arXiv:2412.04964}, 2024{\natexlab{b}}.

\bibitem[Liang et~al.(2024)Liang, Xu, Hong, Shang, Wang, Fu, and Liu]{liang2024survey}
Zijing Liang, Yanjie Xu, Yifan Hong, Penghui Shang, Qi~Wang, Qiang Fu, and Ke~Liu.
\newblock A survey of multimodel large language models.
\newblock In \emph{Proceedings of the 3rd International Conference on Computer, Artificial Intelligence and Control Engineering}, pp.\  405--409, 2024.

\bibitem[Liu et~al.(2024{\natexlab{a}})Liu, Feng, Xue, Wang, Wu, Lu, Zhao, Deng, Zhang, Ruan, et~al.]{liu2024deepseek}
Aixin Liu, Bei Feng, Bing Xue, Bingxuan Wang, Bochao Wu, Chengda Lu, Chenggang Zhao, Chengqi Deng, Chenyu Zhang, Chong Ruan, et~al.
\newblock Deepseek-v3 technical report.
\newblock \emph{arXiv preprint arXiv:2412.19437}, 2024{\natexlab{a}}.

\bibitem[Liu et~al.(2025)Liu, Liu, Pan, He, Haffari, and Zhuang]{liu2025minicache}
Akide Liu, Jing Liu, Zizheng Pan, Yefei He, Reza Haffari, and Bohan Zhuang.
\newblock Minicache: Kv cache compression in depth dimension for large language models.
\newblock \emph{Advances in Neural Information Processing Systems}, 37:\penalty0 139997--140031, 2025.

\bibitem[Liu et~al.(2024{\natexlab{b}})Liu, Bai, Lin, Li, Gao, Xu, Hou, Yao, and Yuan]{liu2024intactkv}
Ruikang Liu, Haoli Bai, Haokun Lin, Yuening Li, Han Gao, Zhengzhuo Xu, Lu~Hou, Jun Yao, and Chun Yuan.
\newblock Intactkv: Improving large language model quantization by keeping pivot tokens intact.
\newblock \emph{arXiv preprint arXiv:2403.01241}, 2024{\natexlab{b}}.

\bibitem[Liu et~al.(2024{\natexlab{c}})Liu, Yuan, Jin, Zhong, Xu, Braverman, Chen, and Hu]{liu2024kivi}
Zirui Liu, Jiayi Yuan, Hongye Jin, Shaochen Zhong, Zhaozhuo Xu, Vladimir Braverman, Beidi Chen, and Xia Hu.
\newblock Kivi: A tuning-free asymmetric 2bit quantization for kv cache.
\newblock \emph{arXiv preprint arXiv:2402.02750}, 2024{\natexlab{c}}.

\bibitem[Merity(2016)]{merity2016wikitext}
Stephen Merity.
\newblock The wikitext long term dependency language modeling dataset.
\newblock \emph{Salesforce Metamind}, 9, 2016.

\bibitem[Raffel et~al.(2020)Raffel, Shazeer, Roberts, Lee, Narang, Matena, Zhou, Li, and Liu]{2020t5}
Colin Raffel, Noam Shazeer, Adam Roberts, Katherine Lee, Sharan Narang, Michael Matena, Yanqi Zhou, Wei Li, and Peter~J. Liu.
\newblock Exploring the limits of transfer learning with a unified text-to-text transformer.
\newblock \emph{Journal of Machine Learning Research}, 21\penalty0 (140):\penalty0 1--67, 2020.
\newblock URL \url{http://jmlr.org/papers/v21/20-074.html}.

\bibitem[Saxena et~al.(2024)Saxena, Saha, Choudhary, and Roy]{saxena2024eigen}
Utkarsh Saxena, Gobinda Saha, Sakshi Choudhary, and Kaushik Roy.
\newblock Eigen attention: Attention in low-rank space for kv cache compression.
\newblock \emph{arXiv preprint arXiv:2408.05646}, 2024.

\bibitem[Shi et~al.(2024)Shi, Zhang, Yao, Li, and Zhao]{shi2024keep}
Luohe Shi, Hongyi Zhang, Yao Yao, Zuchao Li, and Hai Zhao.
\newblock Keep the cost down: A review on methods to optimize llm's kv-cache consumption.
\newblock \emph{arXiv preprint arXiv:2407.18003}, 2024.

\bibitem[Su et~al.(2024)Su, Ahmed, Lu, Pan, Bo, and Liu]{su2024roformer}
Jianlin Su, Murtadha Ahmed, Yu~Lu, Shengfeng Pan, Wen Bo, and Yunfeng Liu.
\newblock Roformer: Enhanced transformer with rotary position embedding.
\newblock \emph{Neurocomputing}, 568:\penalty0 127063, 2024.

\bibitem[Su et~al.(2025{\natexlab{a}})Su, Chen, Shen, Wei, Li, Yu, and Yuan]{su2025rotatekv}
Zunhai Su, Zhe Chen, Wang Shen, Hanyu Wei, Linge Li, Huangqi Yu, and Kehong Yuan.
\newblock Rotatekv: Accurate and robust 2-bit kv cache quantization for llms via outlier-aware adaptive rotations.
\newblock \emph{arXiv preprint arXiv:2501.16383}, 2025{\natexlab{a}}.

\bibitem[Su et~al.(2025{\natexlab{b}})Su, Shen, Li, Chen, Wei, Yu, and Yuan]{su2025akvq}
Zunhai Su, Wang Shen, Linge Li, Zhe Chen, Hanyu Wei, Huangqi Yu, and Kehong Yuan.
\newblock Akvq-vl: Attention-aware kv cache adaptive 2-bit quantization for vision-language models.
\newblock \emph{arXiv preprint arXiv:2501.15021}, 2025{\natexlab{b}}.

\bibitem[Sun et~al.(2024)Sun, Chen, Kolter, and Liu]{sun2024massive}
Mingjie Sun, Xinlei Chen, J~Zico Kolter, and Zhuang Liu.
\newblock Massive activations in large language models.
\newblock \emph{arXiv preprint arXiv:2402.17762}, 2024.

\bibitem[Touvron et~al.(2023)Touvron, Martin, Stone, Albert, Almahairi, Babaei, Bashlykov, Batra, Bhargava, Bhosale, et~al.]{touvron2023llama}
Hugo Touvron, Louis Martin, Kevin Stone, Peter Albert, Amjad Almahairi, Yasmine Babaei, Nikolay Bashlykov, Soumya Batra, Prajjwal Bhargava, Shruti Bhosale, et~al.
\newblock Llama 2: Open foundation and fine-tuned chat models.
\newblock \emph{arXiv preprint arXiv:2307.09288}, 2023.

\bibitem[Vaswani et~al.(2017)Vaswani, Shazeer, Parmar, Uszkoreit, Jones, Gomez, Kaiser, and Polosukhin]{vaswani2017attention}
Ashish Vaswani, Noam Shazeer, Niki Parmar, Jakob Uszkoreit, Llion Jones, Aidan~N Gomez, {\L}ukasz Kaiser, and Illia Polosukhin.
\newblock Attention is all you need.
\newblock \emph{Advances in neural information processing systems}, 30, 2017.

\bibitem[Wan et~al.(2024)Wan, Wu, Liu, Huang, Zhu, Jin, Wang, and Yuan]{wan2024look}
Zhongwei Wan, Ziang Wu, Che Liu, Jinfa Huang, Zhihong Zhu, Peng Jin, Longyue Wang, and Li~Yuan.
\newblock Look-m: Look-once optimization in kv cache for efficient multimodal long-context inference.
\newblock \emph{arXiv preprint arXiv:2406.18139}, 2024.

\bibitem[Xiao et~al.(2023)Xiao, Tian, Chen, Han, and Lewis]{xiao2023efficient}
Guangxuan Xiao, Yuandong Tian, Beidi Chen, Song Han, and Mike Lewis.
\newblock Efficient streaming language models with attention sinks.
\newblock \emph{arXiv preprint arXiv:2309.17453}, 2023.

\bibitem[Xiao et~al.(2024)Xiao, Tang, Zuo, Guo, Yang, Tang, Fu, and Han]{xiao2024duoattention}
Guangxuan Xiao, Jiaming Tang, Jingwei Zuo, Junxian Guo, Shang Yang, Haotian Tang, Yao Fu, and Song Han.
\newblock Duoattention: Efficient long-context llm inference with retrieval and streaming heads.
\newblock \emph{arXiv preprint arXiv:2410.10819}, 2024.

\bibitem[Yang et~al.(2024{\natexlab{a}})Yang, Kim, and Kim]{yang2024mitigating}
Jaewoo Yang, Hayun Kim, and Younghoon Kim.
\newblock Mitigating quantization errors due to activation spikes in glu-based llms.
\newblock \emph{arXiv preprint arXiv:2405.14428}, 2024{\natexlab{a}}.

\bibitem[Yang et~al.(2024{\natexlab{b}})Yang, Kim, Bae, Kwon, Park, Yang, Kwon, and Lee]{yang2024no}
June~Yong Yang, Byeongwook Kim, Jeongin Bae, Beomseok Kwon, Gunho Park, Eunho Yang, Se~Jung Kwon, and Dongsoo Lee.
\newblock No token left behind: Reliable kv cache compression via importance-aware mixed precision quantization.
\newblock \emph{arXiv preprint arXiv:2402.18096}, 2024{\natexlab{b}}.

\bibitem[Yu et~al.(2024)Yu, Wang, Fu, Shi, Shaikh, and Lin]{yu2024unveiling}
Zhongzhi Yu, Zheng Wang, Yonggan Fu, Huihong Shi, Khalid Shaikh, and Yingyan~Celine Lin.
\newblock Unveiling and harnessing hidden attention sinks: Enhancing large language models without training through attention calibration.
\newblock \emph{arXiv preprint arXiv:2406.15765}, 2024.

\bibitem[Zhang et~al.(2024)Zhang, Huang, Jin, and Lu]{zhang2024vision}
Jingyi Zhang, Jiaxing Huang, Sheng Jin, and Shijian Lu.
\newblock Vision-language models for vision tasks: A survey.
\newblock \emph{IEEE Transactions on Pattern Analysis and Machine Intelligence}, 2024.

\bibitem[Zhang et~al.(2023)Zhang, Sheng, Zhou, Chen, Zheng, Cai, Song, Tian, R{\'e}, Barrett, et~al.]{zhang2023h2o}
Zhenyu Zhang, Ying Sheng, Tianyi Zhou, Tianlong Chen, Lianmin Zheng, Ruisi Cai, Zhao Song, Yuandong Tian, Christopher R{\'e}, Clark Barrett, et~al.
\newblock H2o: Heavy-hitter oracle for efficient generative inference of large language models.
\newblock \emph{Advances in Neural Information Processing Systems}, 36:\penalty0 34661--34710, 2023.

\bibitem[Zhao et~al.(2023)Zhao, Zhou, Li, Tang, Wang, Hou, Min, Zhang, Zhang, Dong, et~al.]{zhao2023survey}
Wayne~Xin Zhao, Kun Zhou, Junyi Li, Tianyi Tang, Xiaolei Wang, Yupeng Hou, Yingqian Min, Beichen Zhang, Junjie Zhang, Zican Dong, et~al.
\newblock A survey of large language models.
\newblock \emph{arXiv preprint arXiv:2303.18223}, 1\penalty0 (2), 2023.

\bibitem[Zhu et~al.(2024)Zhu, Li, Liu, Ma, and Wang]{zhu2024survey}
Xunyu Zhu, Jian Li, Yong Liu, Can Ma, and Weiping Wang.
\newblock A survey on model compression for large language models.
\newblock \emph{Transactions of the Association for Computational Linguistics}, 12:\penalty0 1556--1577, 2024.

\end{thebibliography}
\end{document}